\documentclass[10pt,twocolumn,letterpaper]{article}

\PassOptionsToPackage{table}{xcolor}
\usepackage[pagenumbers]{cvpr} %

\usepackage{tabularx}
\usepackage{multirow}
\usepackage[per-mode=symbol, group-digits=integer, group-minimum-digits = 4, detect-all=true, input-symbols={()}]{siunitx}

\usepackage[percent]{overpic} %
\usepackage{fp} %
\usepackage{graphicx}
\usepackage{tikz}
\usetikzlibrary{positioning}
\usetikzlibrary{shapes.geometric, shapes, arrows, arrows.meta, bending}
\usetikzlibrary{backgrounds}
\usetikzlibrary{spy}
\usepackage{pgfplots}
\usepackage{bbding}
\usepackage{dsfont}
\usepackage{fontawesome5}
\usepackage{calc}
\pgfplotsset{compat=1.17}
\usepackage[outline]{contour} 
\contourlength{0.02em}
\usepackage{makecell}
\usepackage{pifont}%
\newcommand{\cmark}{\ding{51}}%
\newcommand{\xmark}{\ding{55}}%
\usepackage[scaled=0.97]{newtxtt}

\newcommand*{\inparagraph}[1]{\smallskip\noindent\textbf{#1}\hspace{0.4em}}
\newcommand*{\inparagraphnospace}[1]{\smallskip\noindent\textbf{#1}}

\newcolumntype{Z}{S[table-format=2.1]}
\newcolumntype{Y}{>{\centering\arraybackslash}X} %

\tikzset{
    double_trapezium/.pic={
        \pgfkeys{/double trapezium/width/.initial=1.15cm}
        \pgfkeys{/double trapezium/height/.initial=0.5cm}
        \pgfkeys{/double trapezium/midheight/.initial=0.25cm}
        \pgfkeys{/double trapezium/color/.initial=pole} %
        \pgfkeysgetvalue{/double trapezium/width}{\width}
        \pgfkeysgetvalue{/double trapezium/height}{\height}
        \pgfkeysgetvalue{/double trapezium/midheight}{\midheight}
        \pgfkeysgetvalue{/double trapezium/color}{\colorvar} %

        \filldraw[left color=\colorvar!50, right color=\colorvar!10, draw=\colorvar!10, rounded corners=1] 
            ({-\width}, \height) -- (0, {\midheight}) -- (0, {-\midheight}) -- ({-\width}, -\height) -- cycle;

        \filldraw[left color=\colorvar!10, right color=\colorvar!50, draw=\colorvar!10, rounded corners=1] 
            ({\width}, \height) -- (0, {\midheight}) -- (0, {-\midheight}) -- ({\width}, -\height) -- cycle;

        \filldraw[draw=black, fill=none, thick, rounded corners=1] 
            ({-\width}, \height) -- (0, {\midheight}) -- ({\width}, \height) -- 
            ({\width}, -\height) -- (0, {-\midheight}) -- ({-\width}, -\height) -- cycle;

    }
}

\tikzset{
    double_trapezium_small/.pic={
        \pgfkeys{/double trapezium/width/.initial=0.9625cm}
        \pgfkeys{/double trapezium/height/.initial=0.4375cm}
        \pgfkeys{/double trapezium/midheight/.initial=0.21875cm}
        \pgfkeys{/double trapezium/color/.initial=pole} %
        \pgfkeysgetvalue{/double trapezium/width}{\width}
        \pgfkeysgetvalue{/double trapezium/height}{\height}
        \pgfkeysgetvalue{/double trapezium/midheight}{\midheight}
        \pgfkeysgetvalue{/double trapezium/color}{\colorvar} %

        \filldraw[left color=\colorvar!50, right color=\colorvar!10, draw=\colorvar!10, rounded corners=1] 
            ({-\width}, \height) -- (0, {\midheight}) -- (0, {-\midheight}) -- ({-\width}, -\height) -- cycle;

        \filldraw[left color=\colorvar!10, right color=\colorvar!50, draw=\colorvar!10, rounded corners=1] 
            ({\width}, \height) -- (0, {\midheight}) -- (0, {-\midheight}) -- ({\width}, -\height) -- cycle;

        \filldraw[draw=black, fill=none, rounded corners=1] 
            ({-\width}, \height) -- (0, {\midheight}) -- ({\width}, \height) -- 
            ({\width}, -\height) -- (0, {-\midheight}) -- ({-\width}, -\height) -- cycle;

    }
}

\definecolor{cvprcolor}{RGB}{127,127,255}
\definecolor{tud0d}{RGB}{83,83,83}
\definecolor{tud0c}{RGB}{137,137,137}
\definecolor{tud0b}{RGB}{181,181,181}
\definecolor{tud0a}{RGB}{220,220,220}
\definecolor{tud1a}{RGB}{93,133,195}
\definecolor{tud2a}{RGB}{0,156,218}
\definecolor{tud3a}{RGB}{80,182,149}
\definecolor{tud4a}{RGB}{175,204,80}
\definecolor{tud5a}{RGB}{221,223,72}
\definecolor{tud6a}{RGB}{255,224,92}
\definecolor{tud7a}{RGB}{248,186,60}
\definecolor{tud8a}{RGB}{238,122,52}
\definecolor{tud9a}{RGB}{233,80,62}
\definecolor{tud10a}{RGB}{201,48,142}
\definecolor{tud11a}{RGB}{128,69,151}
\definecolor{tud1b}{RGB}{0,90,169}
\definecolor{tud2b}{RGB}{0,131,204}
\definecolor{tud3b}{RGB}{0,157,129}
\definecolor{tud4b}{RGB}{153,192,0}
\definecolor{tud5b}{RGB}{201,212,0}
\definecolor{tud6b}{RGB}{253,202,0}
\definecolor{tud7b}{RGB}{245,163,0}
\definecolor{tud8b}{RGB}{236,101,0}
\definecolor{tud9b}{RGB}{230,0,26}
\definecolor{tud10b}{RGB}{166,0,132}
\definecolor{tud11b}{RGB}{114,16,133}
\definecolor{tud1c}{RGB}{0,78,138}
\definecolor{tud2c}{RGB}{0,104,157}
\definecolor{tud3c}{RGB}{0,136,119}
\definecolor{tud4c}{RGB}{127,171,22}
\definecolor{tud5c}{RGB}{177,189,0}
\definecolor{tud6c}{RGB}{215,172,0}
\definecolor{tud7c}{RGB}{210,135,0}
\definecolor{tud8c}{RGB}{204,76,3}
\definecolor{tud9c}{RGB}{185,15,34}
\definecolor{tud10c}{RGB}{149,17,105}
\definecolor{tud11c}{RGB}{97,28,115}
\definecolor{tud1d}{RGB}{36,53,114}
\definecolor{tud2d}{RGB}{0,78,115}
\definecolor{tud3d}{RGB}{0,113,94}
\definecolor{tud4d}{RGB}{106,139,55}
\definecolor{tud5d}{RGB}{153,166,4}
\definecolor{tud6d}{RGB}{174,142,0}
\definecolor{tud7d}{RGB}{190,111,0}
\definecolor{tud8d}{RGB}{169,73,19}
\definecolor{tud9d}{RGB}{156,28,38}
\definecolor{tud10d}{RGB}{115,32,84}
\definecolor{tud11d}{RGB}{76,34,106}

\definecolor{our3}{RGB}{220, 38, 127}

\definecolor{our1}{RGB}{13, 146, 244}
\definecolor{our2}{RGB}{119, 205, 255}
\definecolor{our3}{RGB}{249, 84, 84}
\definecolor{our4}{RGB}{198, 46, 46}

\definecolor{unlabeled}{RGB}{0,0,0}
\definecolor{egovehicle}{RGB}{0,0,0}
\definecolor{rectification border}{RGB}{0,0,0}
\definecolor{outofroi}{RGB}{0,0,0}
\definecolor{static}{RGB}{0,0,0}
\definecolor{dynamic}{RGB}{111,74,0}
\definecolor{ground}{RGB}{81,0,81}
\definecolor{road}{RGB}{128,64,128}
\definecolor{sidewalk}{RGB}{244,35,232}
\definecolor{parking}{RGB}{250,170,160}
\definecolor{rail track}{RGB}{230,150,140}
\definecolor{building}{RGB}{70,70,70}
\definecolor{wall}{RGB}{102,102,156}
\definecolor{fence}{RGB}{190,153,153}
\definecolor{guard rail}{RGB}{180,165,180}
\definecolor{bridge}{RGB}{150,100,100}
\definecolor{tunnel}{RGB}{150,120,90}
\definecolor{pole}{RGB}{153,153,153}
\definecolor{polegroup}{RGB}{153,153,153}
\definecolor{trafficlight}{RGB}{250,170,30}
\definecolor{trafficsign}{RGB}{220,220,0}
\definecolor{vegetation}{RGB}{107,142,35}
\definecolor{terrain}{RGB}{152,251,152}
\definecolor{sky}{RGB}{70,130,180}
\definecolor{skylight}{RGB}{98,182,252}
\definecolor{person}{RGB}{220,20,60}
\definecolor{rider}{RGB}{255,0,0}
\definecolor{car}{RGB}{0,0,142}
\definecolor{truck}{RGB}{0,0,70}
\definecolor{bus}{RGB}{0,60,100}
\definecolor{caravan}{RGB}{0,0,90}
\definecolor{trailer}{RGB}{0,0,110}
\definecolor{train}{RGB}{0,80,100}
\definecolor{motorcycle}{RGB}{0,0,230}
\definecolor{bicycle}{RGB}{119,11,32}
\definecolor{licenseplate}{RGB}{0,0,142}

\usepackage{acronym}
\newacro{miou}[mIoU]{mean Intersection over Union}

\newcommand{\dlt}[1]{{\textcolor{tud3c}{#1}}}

\newcommand{\supp}{supp.\ material\@\xspace}

\newcommand{\MethodName}{CUPS\@\xspace} %

\definecolor{cvprblue}{rgb}{0.21,0.49,0.74}
\usepackage[pagebackref,breaklinks,colorlinks,allcolors=cvprblue]{hyperref}

\def\paperID{10582} %
\def\confName{CVPR}
\def\confYear{2025}

\title{Scene-Centric Unsupervised Panoptic Segmentation}

\newcommand{\authorstep}{\hspace{0.75cm}}
\newcommand{\affiliationstep}{\hspace{0.5cm}}

\author{
Oliver Hahn\textsuperscript{\normalfont{}* 1}
\authorstep Christoph Reich\textsuperscript{\normalfont{}* 1,2,4,5}
\authorstep Nikita Araslanov\textsuperscript{\normalfont{} 2,4}\\
Daniel Cremers\textsuperscript{\normalfont{} 2,4,5}
\authorstep Christian Rupprecht\textsuperscript{\normalfont{} 3}
\authorstep Stefan Roth\textsuperscript{\normalfont{} 1,5,6}\\[1pt]
\small{\textsuperscript{1}TU Darmstadt\affiliationstep \textsuperscript{2}TU Munich \affiliationstep \textsuperscript{3}University of Oxford \affiliationstep \textsuperscript{4}MCML\affiliationstep \textsuperscript{5}ELIZA\affiliationstep \textsuperscript{6}hessian.AI\affiliationstep
\textsuperscript{*}equal contribution}\\[-1pt]\small {\url{https://visinf.github.io/cups}}}

\usepackage{fancyhdr}
\usepackage{setspace}

\fancypagestyle{firststyle}
{

	\fancyhf{}
	\lfoot{{\footnotesize\begin{spacing}{.5}\parbox{\linewidth}{\vspace{2.5em}%
					To appear in \emph{Proceedings of the IEEE/CVF Conference on Computer Vision and Pattern Recognition}, Nashville, TN, USA, 2025.%
					\\\hrule\vspace{\baselineskip}
					\copyright~2025 IEEE. Personal use of this material is permitted. Permission from IEEE must be obtained for all other uses, in any current or future media, including reprinting/republishing this material for advertising or promotional purposes, creating new collective works, for resale or redistribution to servers or lists, or reuse of any copyrighted component of this work in other works. 
	}\end{spacing}}}
}

\begin{document}

\twocolumn[{%
\renewcommand\twocolumn[1][]{#1}%
\maketitle
\vspace{-1.75em}
\tikzset{every picture/.style={font=\sffamily\scriptsize}}
\pgfplotsset{
    teaser axis style/.style={
        width=3.5cm,
        height=3.5cm,
        ybar,
        bar width=17pt, 
        xtick=data, 
        ymin=14.5,     
        ytick=\empty,        
        axis x line=bottom,
        axis y line=none,
        axis line style={-},
        xtick style={draw=none},
        enlarge x limits=1.7,
        nodes near coords,
        nodes near coords={\pgfmathprintnumber[assume math mode=true]{\pgfplotspointmeta}},
        every node near coord/.append style={
            yshift=-10pt,
            color=white,
            font=\sffamily\scriptsize,
            /pgf/number format/.cd,
            fixed,
            fixed zerofill,
            precision=1,
        },
        legend style={
            at={(-0.0, 0.85)},
            anchor=north,
            draw=none,
            legend columns=1,
        },
        reverse legend,
        legend cell align={left},
        legend image code/.code={
            \draw[fill=#1, draw=none] (-0.08cm,-0.08cm) rectangle (0.12cm,0.12cm);
        },
    }
}

\newcommand{\teaserimgscale}{1.0} %
\FPeval{\csteaserimgscale}{round(\teaserimgscale*0.2,6)}
\FPeval{\kitticutoff}{round((((3.369565217391304/2.0)-1)/4),8)}

\newcommand{\imgoverlayopacity}{0.45}

\footnotesize
\sffamily
\setlength{\tabcolsep}{0pt}
\renewcommand{\arraystretch}{0.0}
\begin{tabular}{>{\centering\arraybackslash} m{0.0135\textwidth} 
                >{\centering\arraybackslash} m{0.1644\textwidth} 
                >{\centering\arraybackslash} m{0.1644\textwidth} 
                >{\centering\arraybackslash} m{0.1644\textwidth} 
                >{\centering\arraybackslash} m{0.1644\textwidth}
                >{\centering\arraybackslash} m{0.1644\textwidth}
                >{\centering\arraybackslash} m{0.1644\textwidth}}

 & {$\;\;\,$Cityscapes} & {$\;\;\,$KITTI} & {$\;\;\,$BDD} & {$\;\;\,$MUSES} & {$\;\;\,$Waymo} & {$\;\;\,$MOTS} \\[-1.5pt]

\rotatebox[origin=c]{90}{$\;\;\;\;\;\;\;\;\;$\MethodName} 
& \begin{tikzpicture}[spy using outlines={white, line width=0.5mm, dashed, dash pattern=on 1.25pt off 1.25pt, magnification=2.5, minimum width=0.65cm, minimum height=0.65cm, fill=white}]
    \node[anchor=south west] (img) at (0,0) {\includegraphics[width=\linewidth]{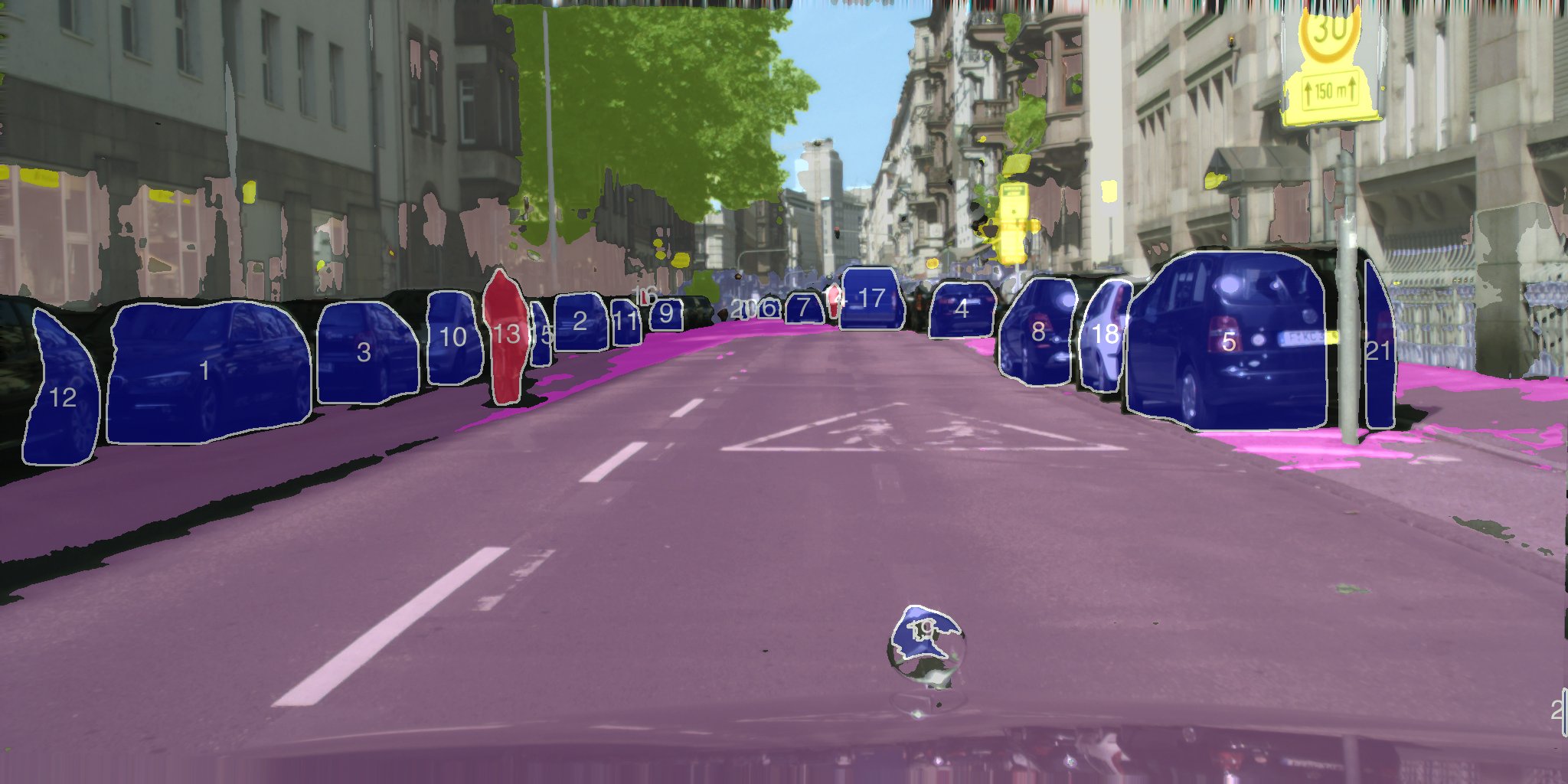}}; 
    \spy on (1.1, 0.925) in node [draw opacity=1.0, white, fill=white, anchor=south west] at (0.13, 0.13);
\end{tikzpicture} 
& \begin{tikzpicture}[spy using outlines={white, line width=0.5mm, dashed, dash pattern=on 1.5pt off 1.5pt, magnification=2.5, minimum width=0.65cm, minimum height=0.65cm, fill=white}]
    \node[anchor=south west] (img) at (0,0) {\includegraphics[width=\linewidth]{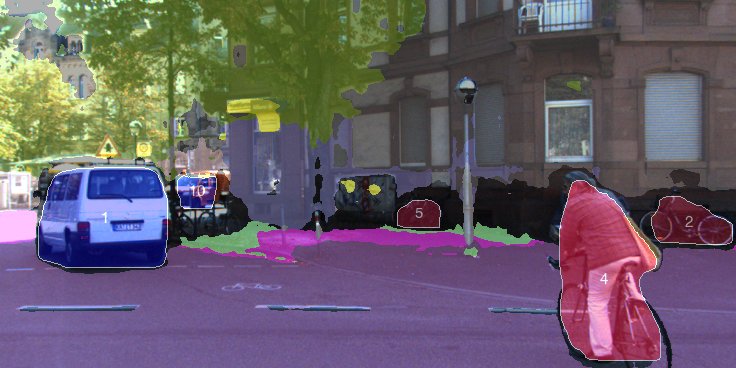}}; 
    \spy on (2.8, 0.655) in node [draw opacity=1.0, white, fill=white, anchor=south west] at (0.13, 0.13);
\end{tikzpicture}  
& \begin{tikzpicture}[spy using outlines={white, line width=0.5mm, dashed, dash pattern=on 1.5pt off 1.5pt, magnification=2.5, minimum width=0.65cm, minimum height=0.65cm, fill=white}]
    \node[anchor=south west] (img) at (0,0) {\includegraphics[width=\linewidth]{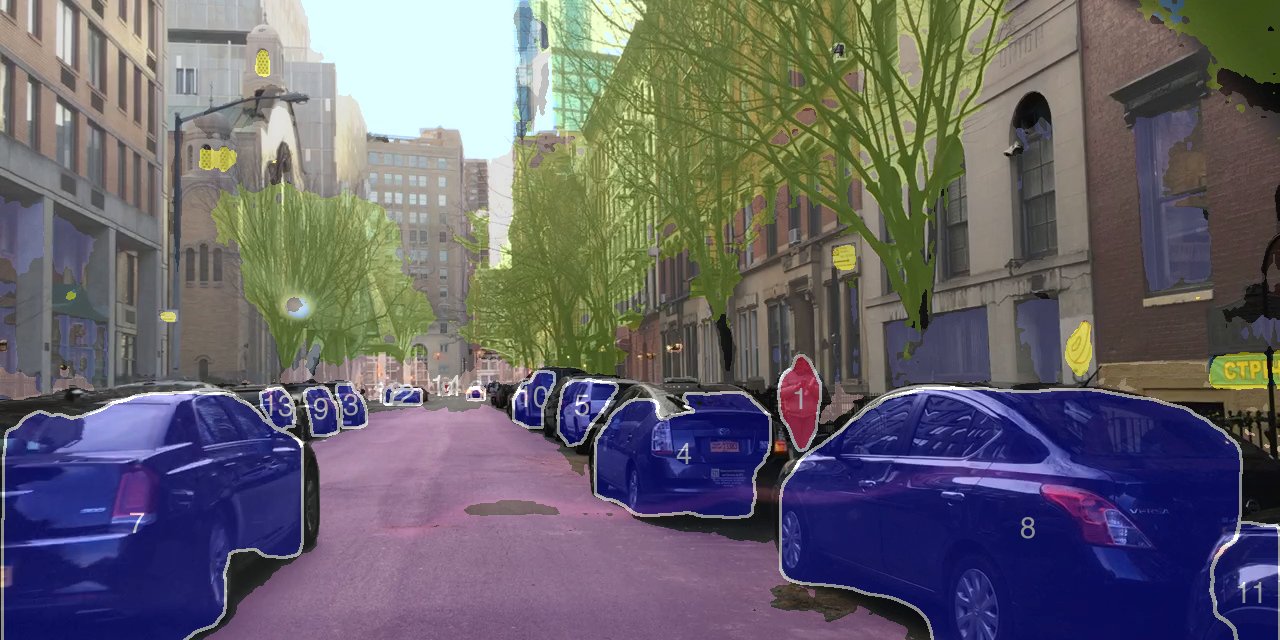}}; 
    \spy on (1.9, 0.65) in node [draw opacity=1.0, white, fill=white, anchor=south west] at (0.13, 0.13);
\end{tikzpicture}   
& \begin{tikzpicture}[spy using outlines={white, line width=0.5mm, dashed, dash pattern=on 1.5pt off 1.5pt, magnification=2.5, minimum width=0.65cm, minimum height=0.65cm, fill=white}]
    \node[anchor=south west] (img) at (0,0) {\includegraphics[width=\linewidth]{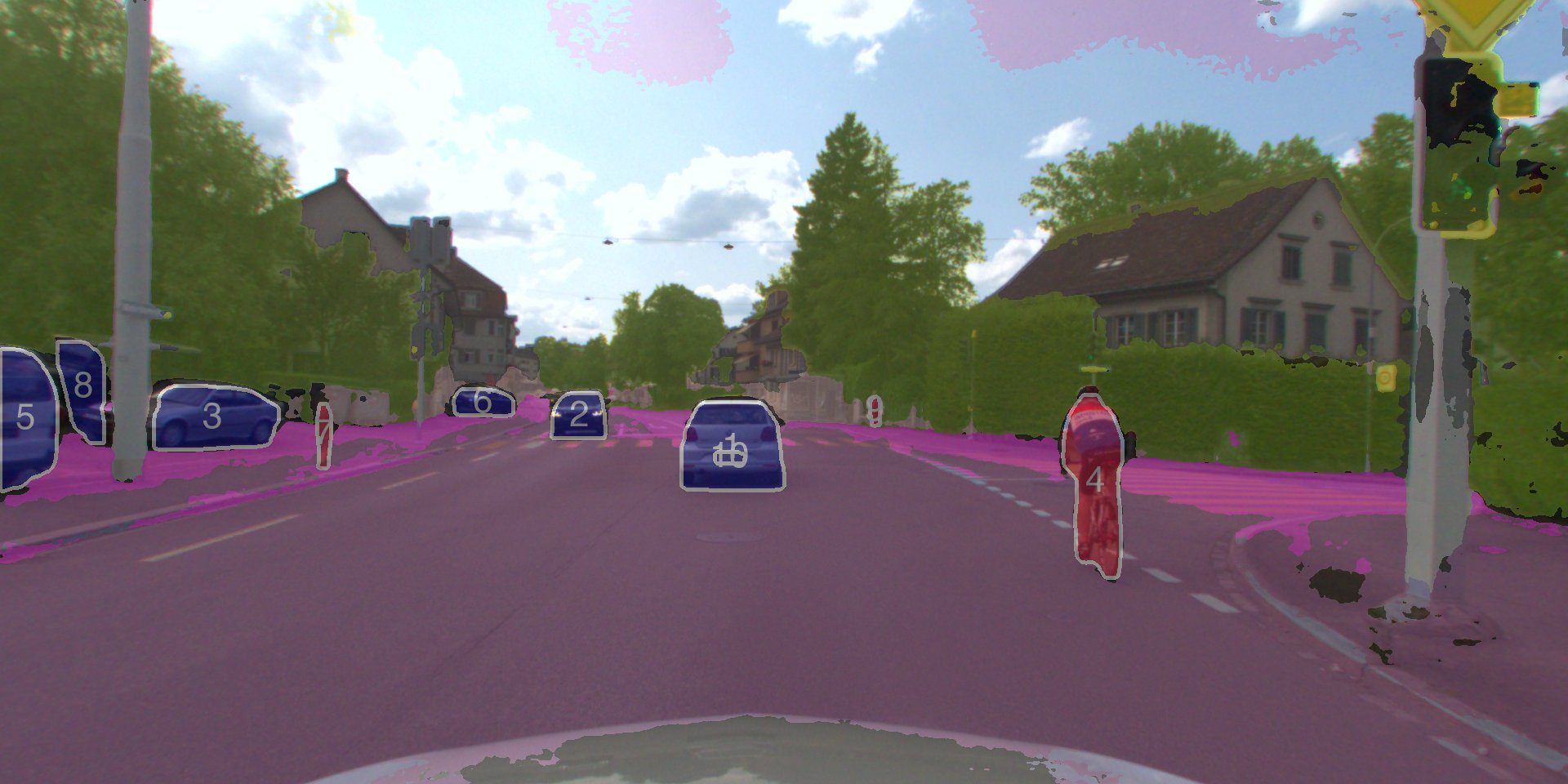}}; 
    \spy on (1.075, 0.835) in node [draw opacity=1.0, white, fill=white, anchor=south west] at (0.13, 0.13);
\end{tikzpicture}   
& \begin{tikzpicture}[spy using outlines={white, line width=0.5mm, dashed, dash pattern=on 1.5pt off 1.5pt, magnification=2.5, minimum width=0.65cm, minimum height=0.65cm, fill=white}]
    \node[anchor=south west] (img) at (0,0) {\includegraphics[width=\linewidth]{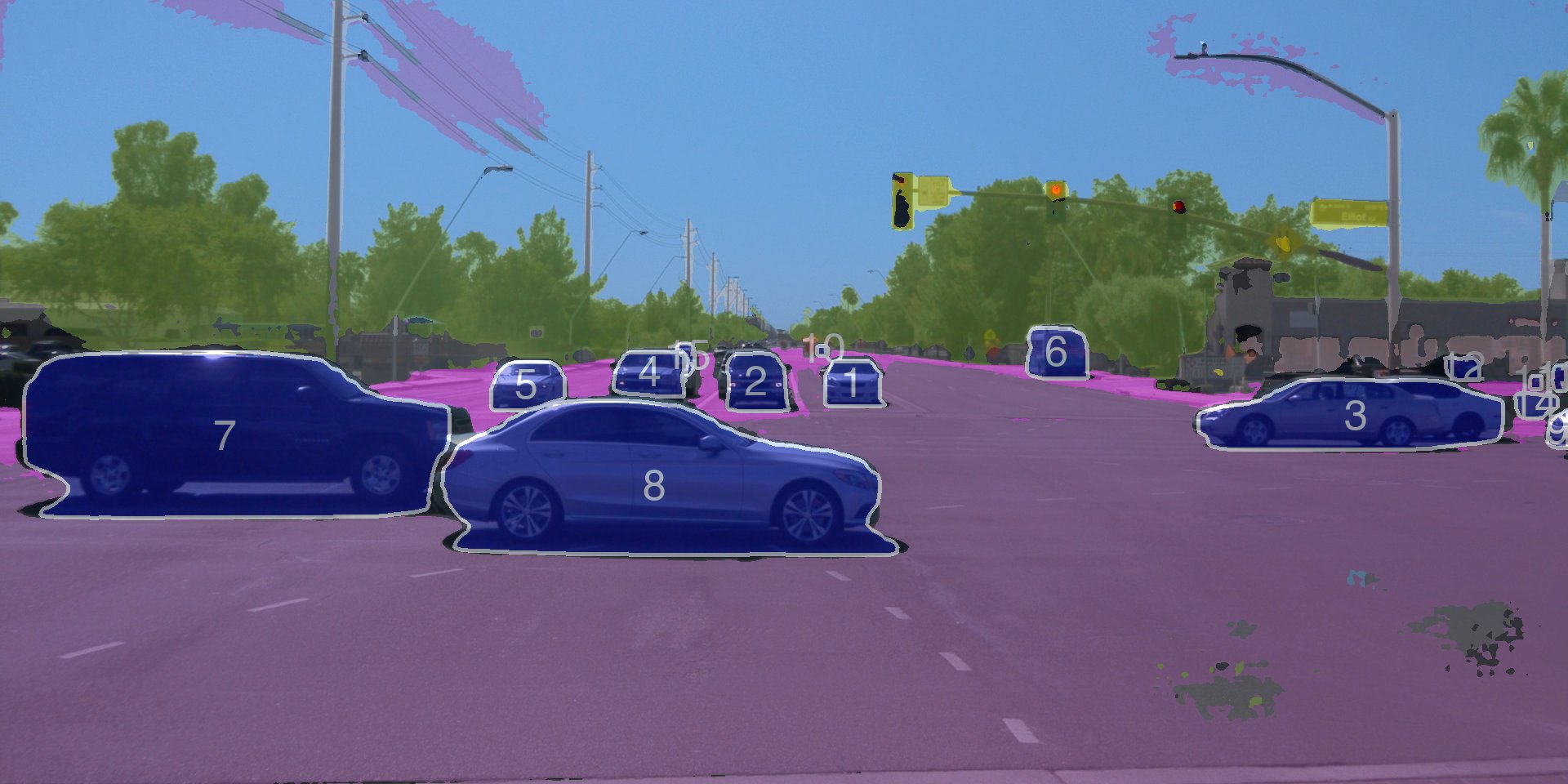}}; 
    \spy on (1.075, 0.835) in node [draw opacity=1.0, white, fill=white, anchor=south west] at (0.13, 0.13);
\end{tikzpicture}
& \begin{tikzpicture}[spy using outlines={white, line width=0.5mm, dashed, dash pattern=on 1.5pt off 1.5pt, magnification=2.5, minimum width=0.65cm, minimum height=0.65cm, fill=white}]
    \node[anchor=south west] (img) at (0,0) {\includegraphics[width=\linewidth]{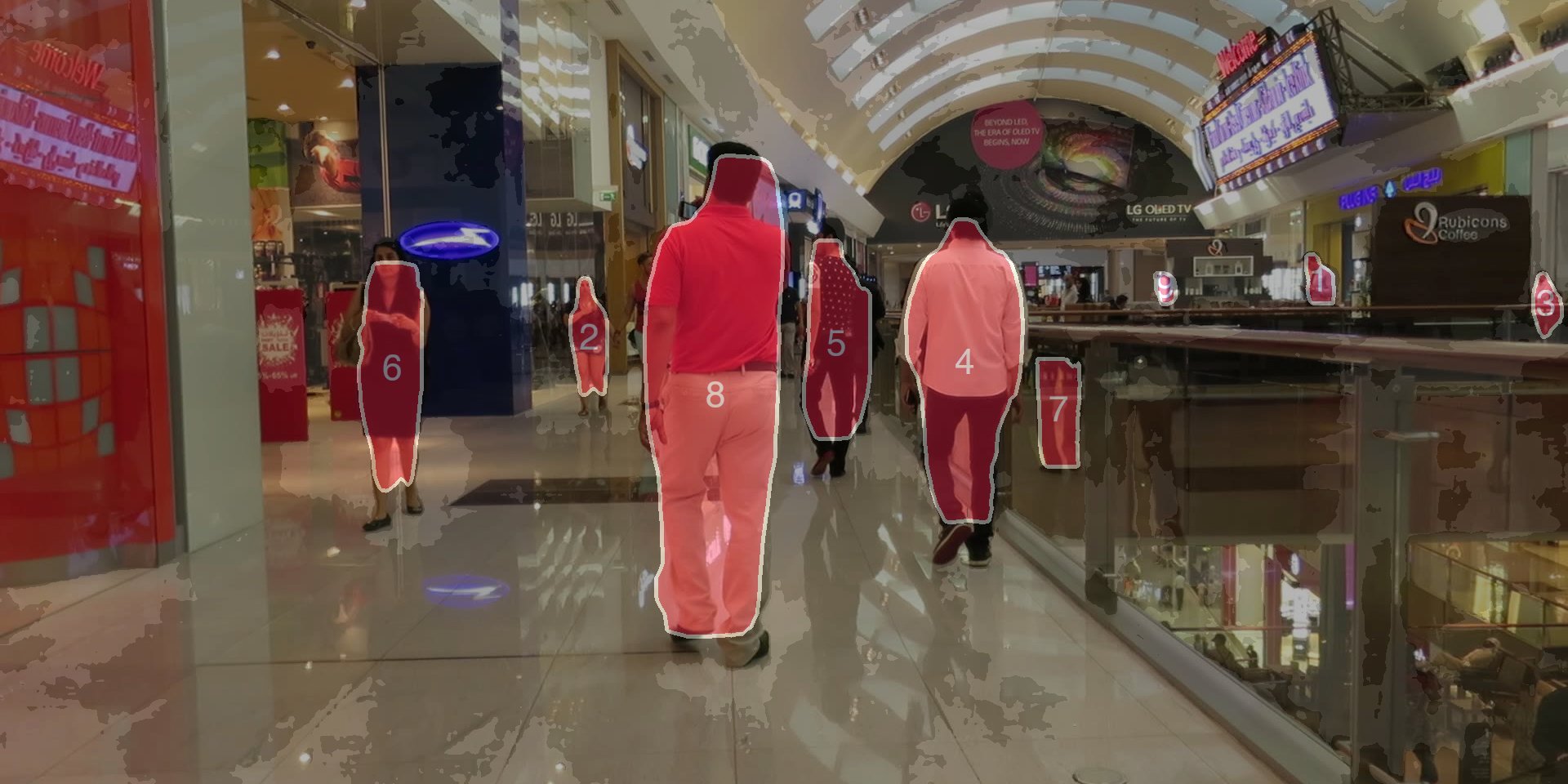}}; 
    \spy on (1.2, 0.925) in node [draw opacity=1.0, white, fill=white, anchor=south west] at (0.13, 0.13);
\end{tikzpicture} \\[-16.5pt]

\rotatebox[origin=l]{90}{\hspace{0.75em}U2Seg$\;\;\;\;$}
& \begin{tikzpicture}[spy using outlines={white, line width=0.5mm, dashed, dash pattern=on 1.5pt off 1.5pt, magnification=2.5, minimum width=0.65cm, minimum height=0.65cm, fill=white}]
    \node[anchor=south west] (img) at (0,0) {\includegraphics[width=\linewidth]{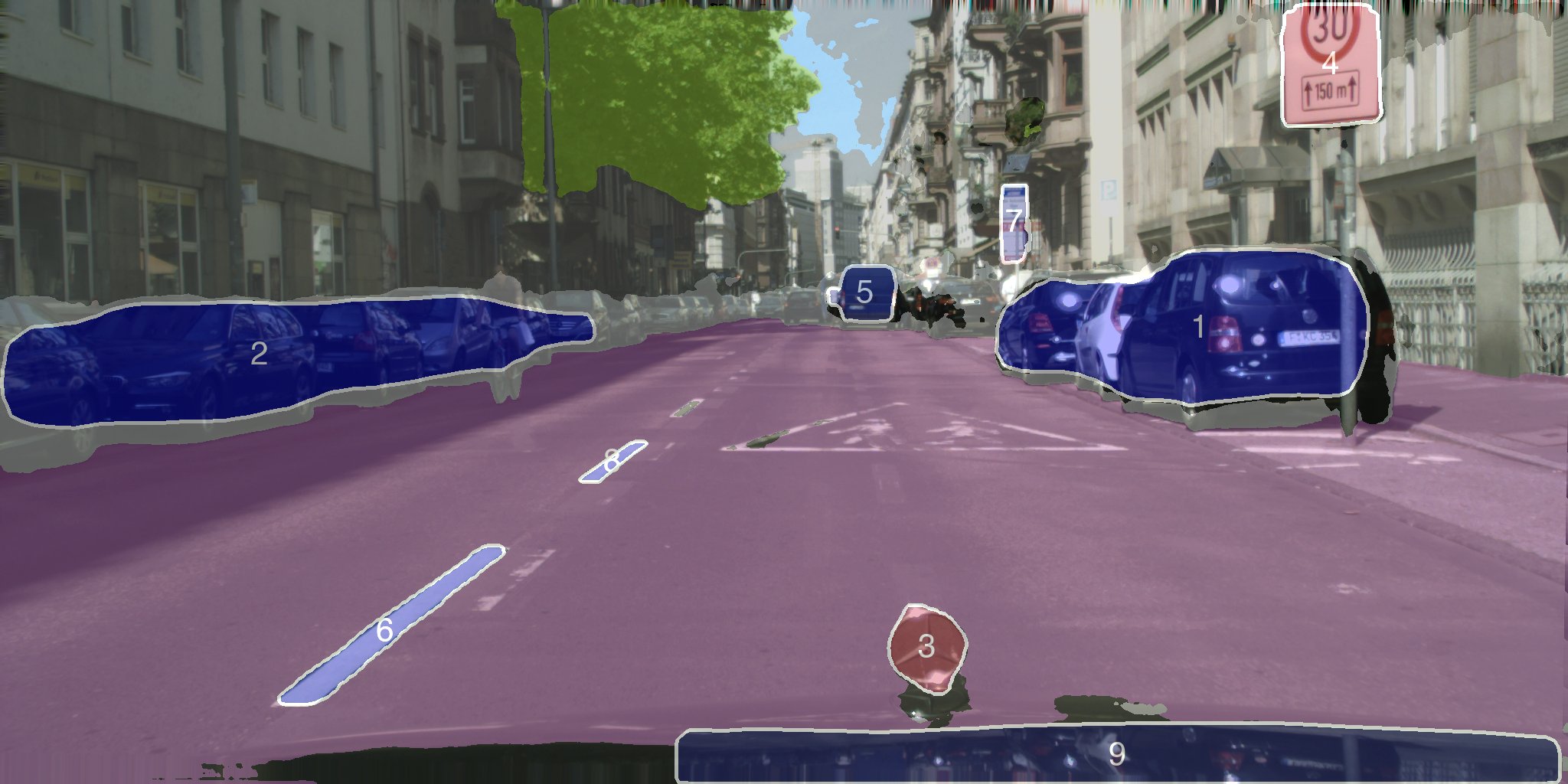}}; 
    \node[anchor=south, fill=white, fill opacity=0.8, text opacity=1, text=black, inner sep=1.5pt, rounded corners=1pt] at (0.54\linewidth, 1.375) {+\qty{9.4}{\percent} PQ};
    \spy on (1.1, 0.925) in node [draw opacity=1.0, white, fill=white, anchor=south west] at (0.13, 0.13);
\end{tikzpicture} 
& \begin{tikzpicture}[spy using outlines={white, line width=0.5mm, dashed, dash pattern=on 1.5pt off 1.5pt, magnification=2.5, minimum width=0.65cm, minimum height=0.65cm, fill=white}]
    \node[anchor=south west] (img) at (0,0) {\includegraphics[width=\linewidth]{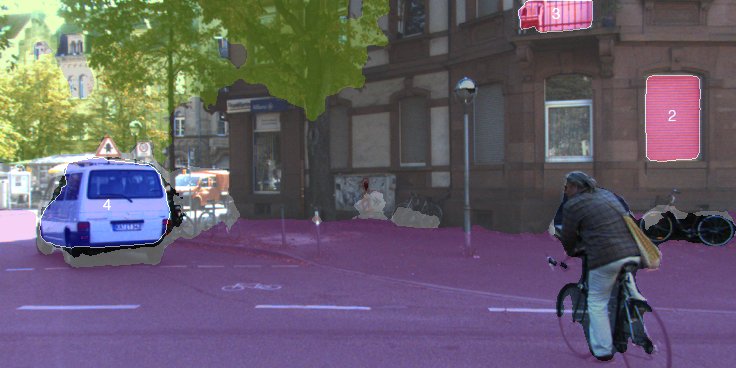}}; 
    \node[anchor=south, fill=white, text=black, fill opacity=0.8, text opacity=1, inner sep=1.5pt, rounded corners=1pt] at (0.54\linewidth, 1.375) {+\qty{4.9}{\percent} PQ};
    \spy on (2.8, 0.655) in node [draw opacity=1.0, white, fill=white, anchor=south west] at (0.13, 0.13);
\end{tikzpicture}  
& \begin{tikzpicture}[spy using outlines={white, line width=0.5mm, dashed, dash pattern=on 1.5pt off 1.5pt, magnification=2.5, minimum width=0.65cm, minimum height=0.65cm, fill=white}]
    \node[anchor=south west] (img) at (0,0) {\includegraphics[width=\linewidth]{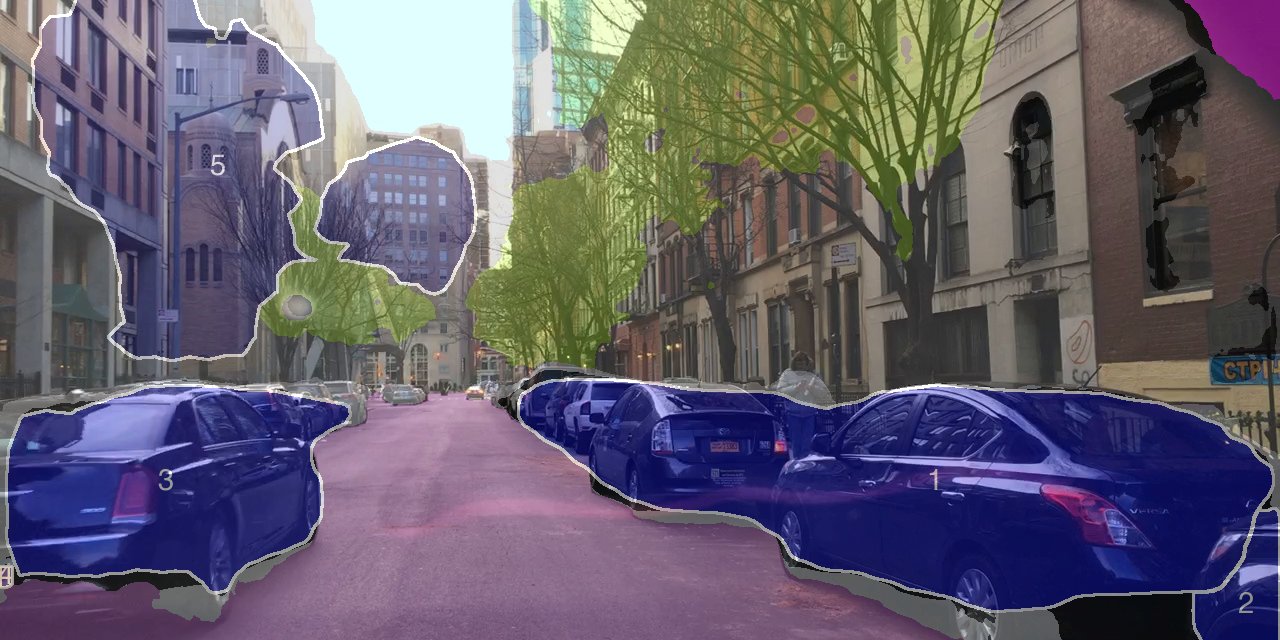}}; 
    \node[anchor=south, fill=white, text=black, fill opacity=0.8, text opacity=1, inner sep=1.5pt, rounded corners=1pt] at (0.54\linewidth, 1.375) {+\qty{4.1}{\percent} PQ};
    \spy on (1.9, 0.65) in node [draw opacity=1.0, white, fill=white, anchor=south west] at (0.13, 0.13);
\end{tikzpicture}   
& \begin{tikzpicture}[spy using outlines={white, line width=0.5mm, dashed, dash pattern=on 1.5pt off 1.5pt, magnification=2.5, minimum width=0.65cm, minimum height=0.65cm, fill=white}]
    \node[anchor=south west] (img) at (0,0) {\includegraphics[width=\linewidth]{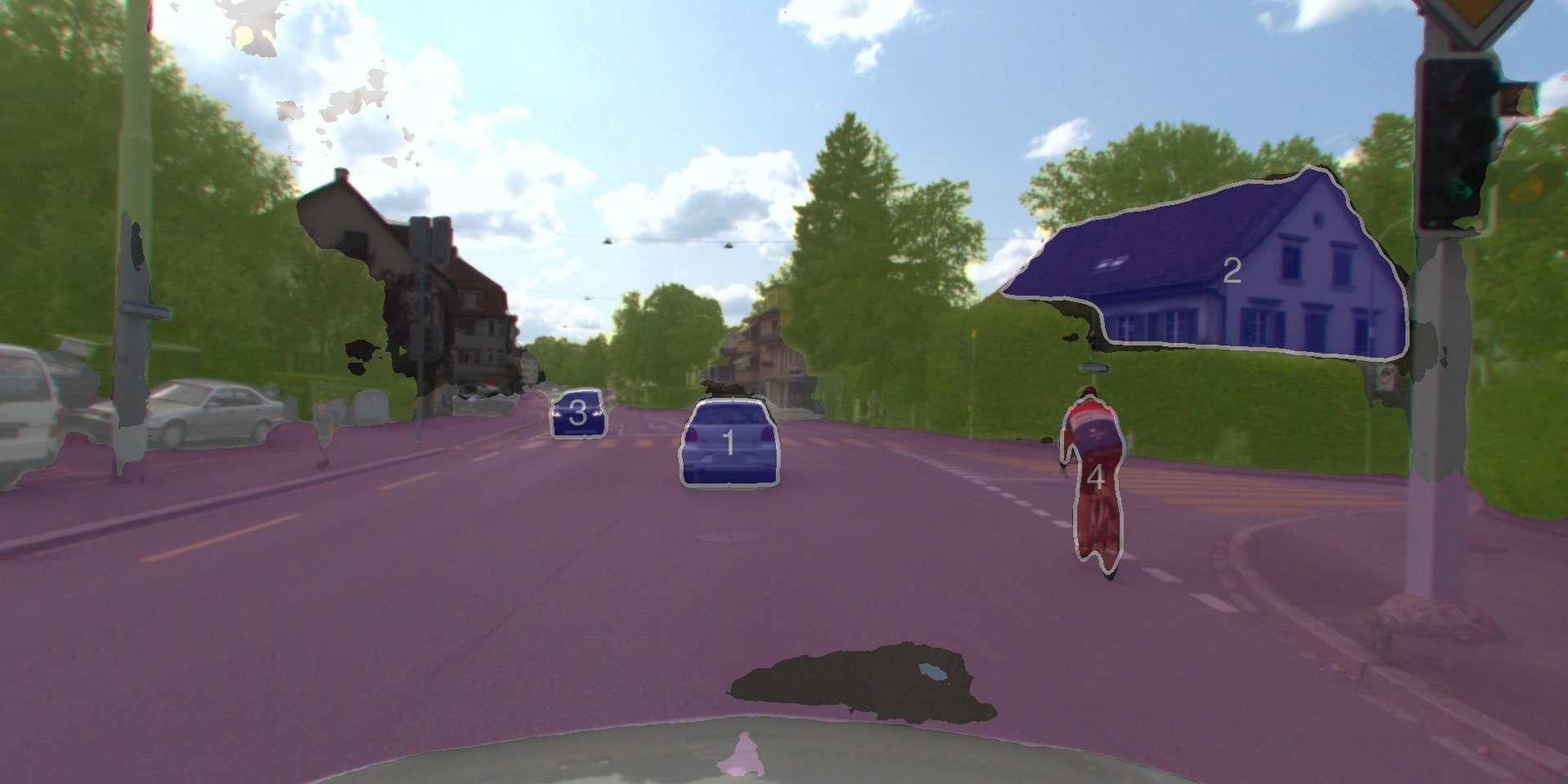}}; 
    \node[anchor=south, fill=white, text=black, fill opacity=0.8, text opacity=1, inner sep=1.5pt, rounded corners=1pt] at (0.54\linewidth, 1.375) {+\qty{4.1}{\percent} PQ};
    \spy on (1.075, 0.835) in node [draw opacity=1.0, white, fill=white, anchor=south west] at (0.13, 0.13);
\end{tikzpicture}   
& \begin{tikzpicture}[spy using outlines={white, line width=0.5mm, dashed, dash pattern=on 1.5pt off 1.5pt, magnification=2.5, minimum width=0.65cm, minimum height=0.65cm, fill=white}]
    \node[anchor=south west] (img) at (0,0) {\includegraphics[width=\linewidth]{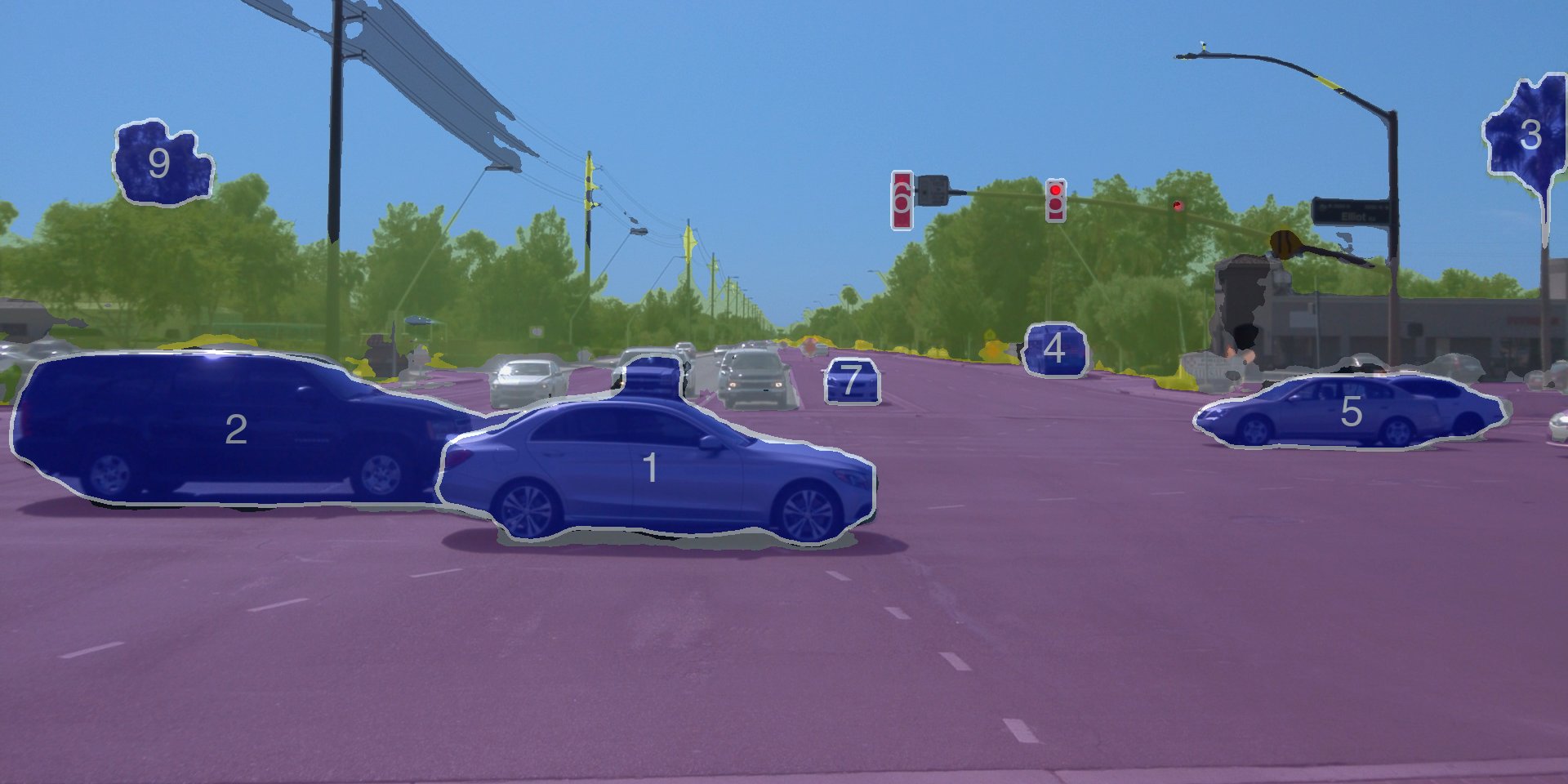}}; 
    \node[anchor=south, fill=white, text=black, fill opacity=0.8, text opacity=1, inner sep=1.5pt, rounded corners=1pt] at (0.54\linewidth, 1.375) {+\qty{6.6}{\percent} PQ};
    \spy on (1.075, 0.835) in node [draw opacity=1.0, white, fill=white, anchor=south west] at (0.13, 0.13);
\end{tikzpicture} 
& \begin{tikzpicture}[spy using outlines={white, line width=0.5mm, dashed, dash pattern=on 1.5pt off 1.5pt, magnification=2.5, minimum width=0.65cm, minimum height=0.65cm, fill=white}]
    \node[anchor=south west] (img) at (0,0) {\includegraphics[width=\linewidth]{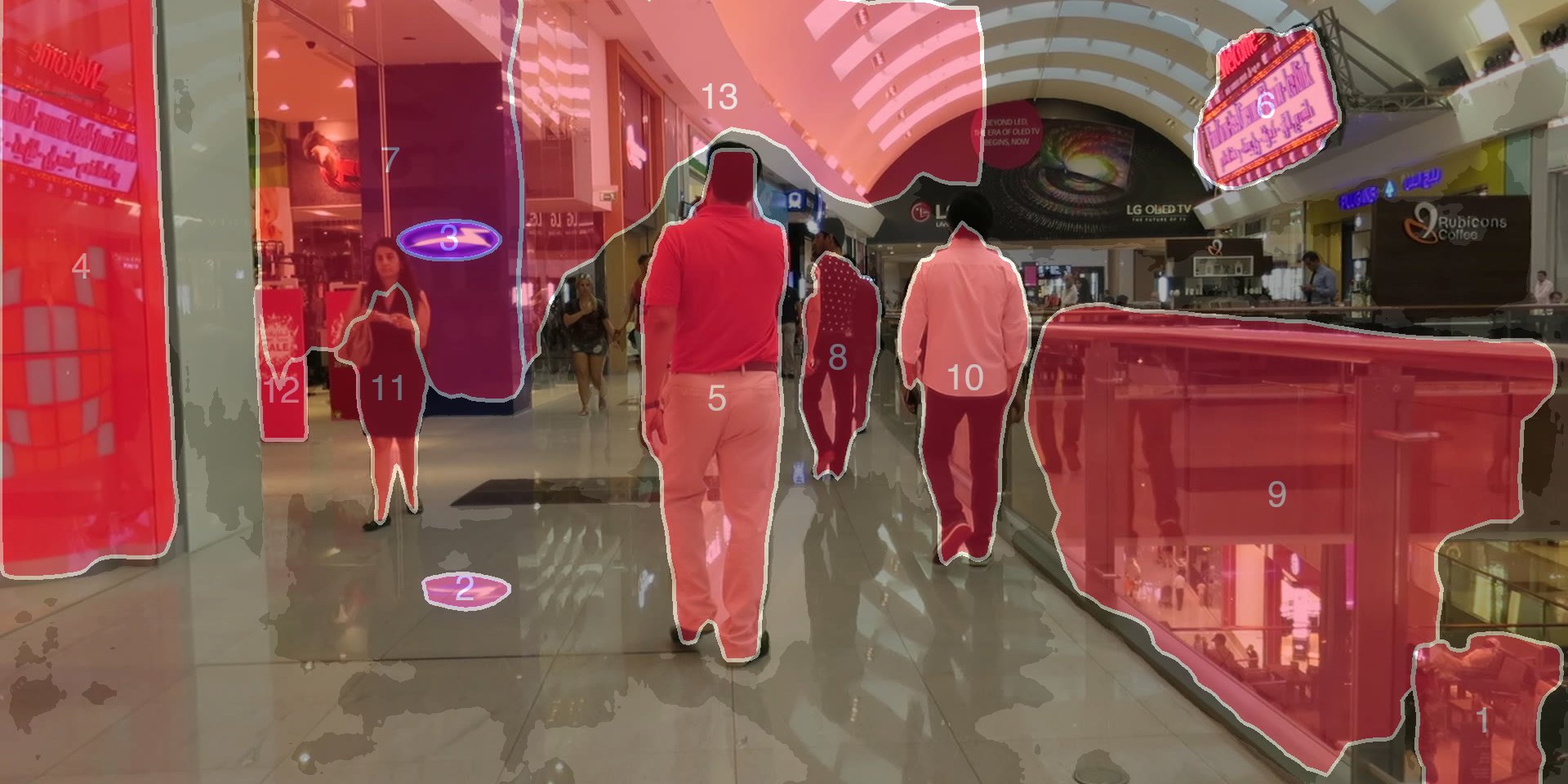}}; 
    \node[anchor=south, fill=white, text=black, fill opacity=0.8, text opacity=1, inner sep=1.5pt, rounded corners=1pt] at (0.54\linewidth, 1.375) {+\qty{17.1}{\percent} PQ};
    \spy on (1.2, 0.925) in node [draw opacity=1.0, white, fill=white, anchor=south west] at (0.13, 0.13);
\end{tikzpicture} \\

\end{tabular}%
\vspace{-0.45cm}%

\begin{center}
    \begin{tikzpicture}[>={Stealth[inset=0pt,length=5pt,angle'=45]}, scale=0.875]
        \tikzstyle{every node}=[font=\fontsize{5.5}{2} \selectfont]
        \node[white] at (15.25, 0.0) {.};
        \node[] at (3.45, 0.825) {\footnotesize \vphantom{p}Pseudo Label Generation};
        \begin{scope}[yshift=-0.075cm]
            \node at (-0.15, 0.15) {\includegraphics[width=1.4cm, decodearray={0.0 1.25 0 1.25 0.0 1.25}]{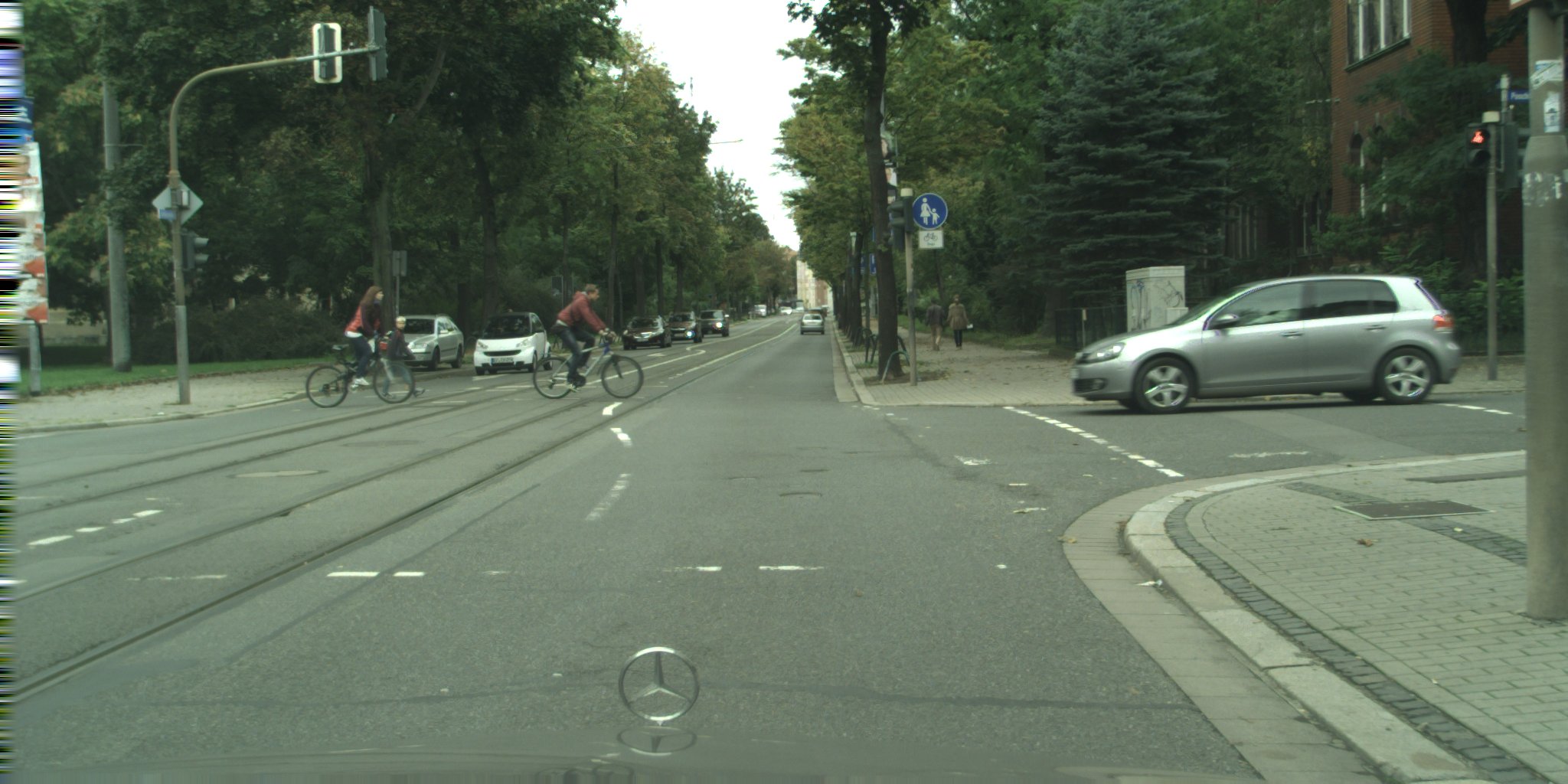}};
            \node at (-0.1, 0.1) {\includegraphics[width=1.4cm, decodearray={0.0 1.25 0 1.25 0.0 1.25}]{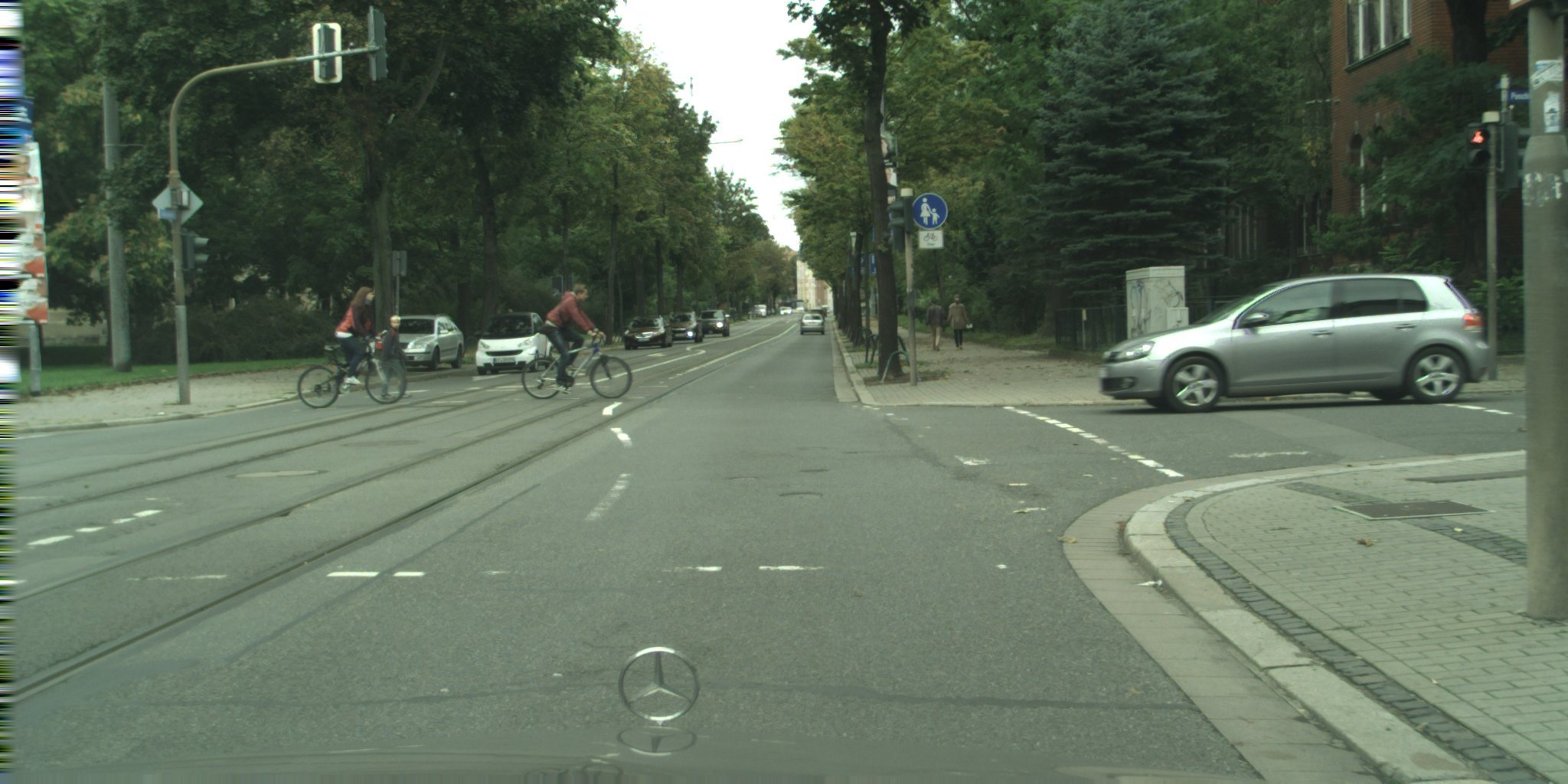}};
            \node at (-0.05, 0.05) {\includegraphics[width=1.4cm, decodearray={0.0 1.25 0 1.25 0.0 1.25}]{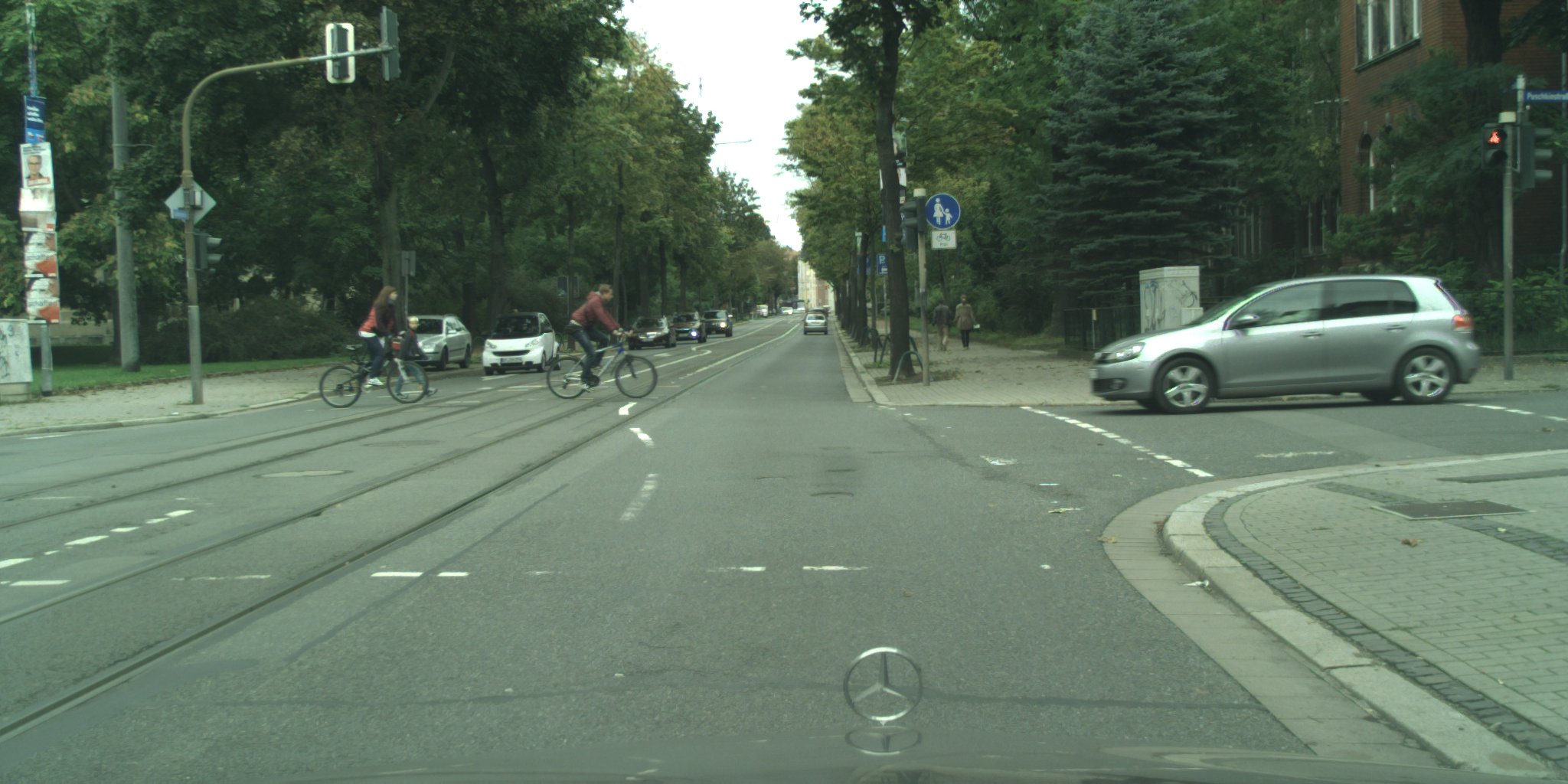}};
            \node at (0.0, 0.0) {\includegraphics[width=1.4cm, decodearray={0.0 1.25 0 1.25 0.0 1.25}]{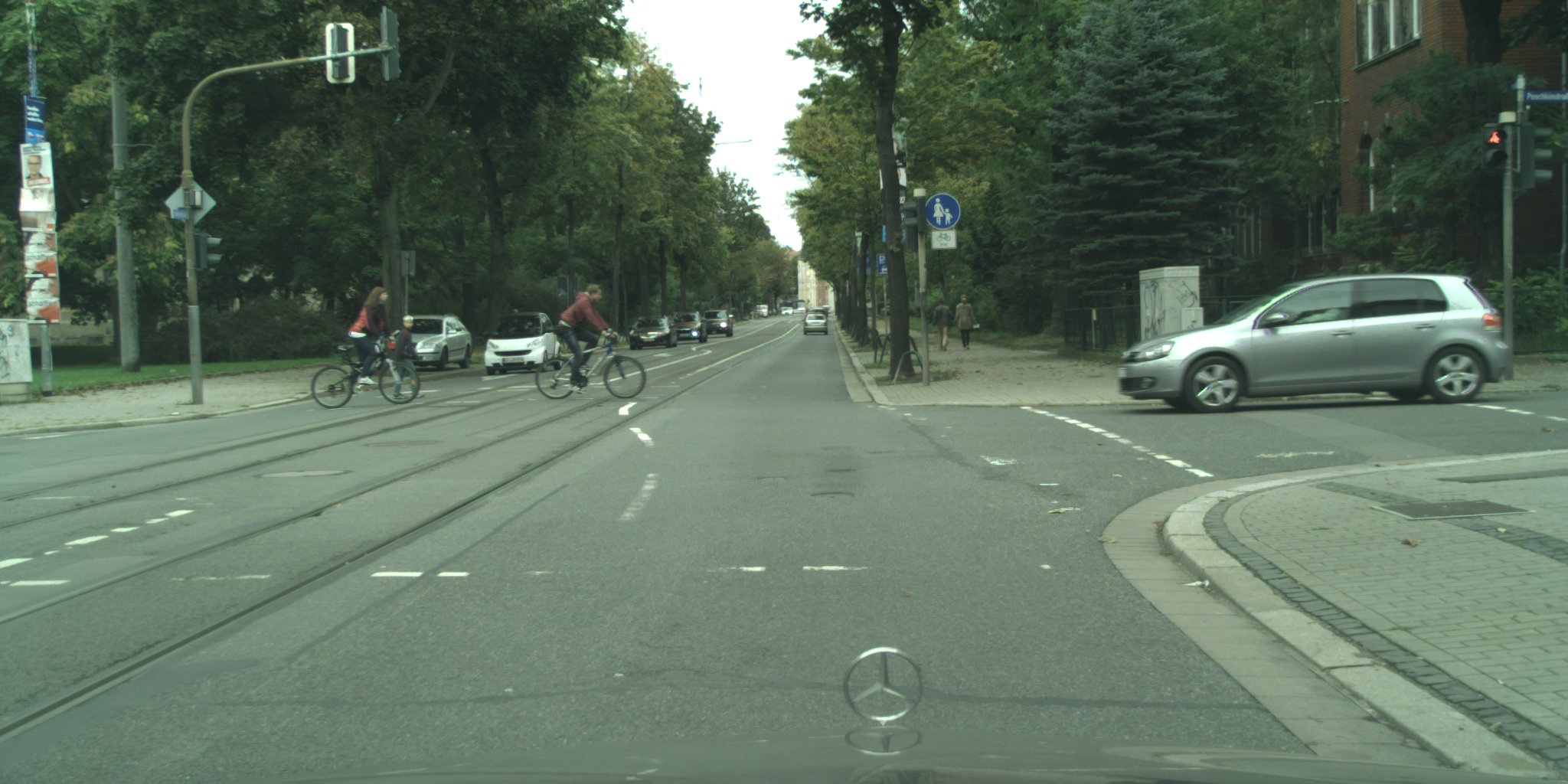}};
        \end{scope}
        \node[white] at (0.0, -0.3) {Stereo Frames\vphantom{p}};
        \node at (1.025, 0.0) {\small\textbf{+}};
        \node at (2.1, 0.075) {\frame{\includegraphics[width=1.4cm]{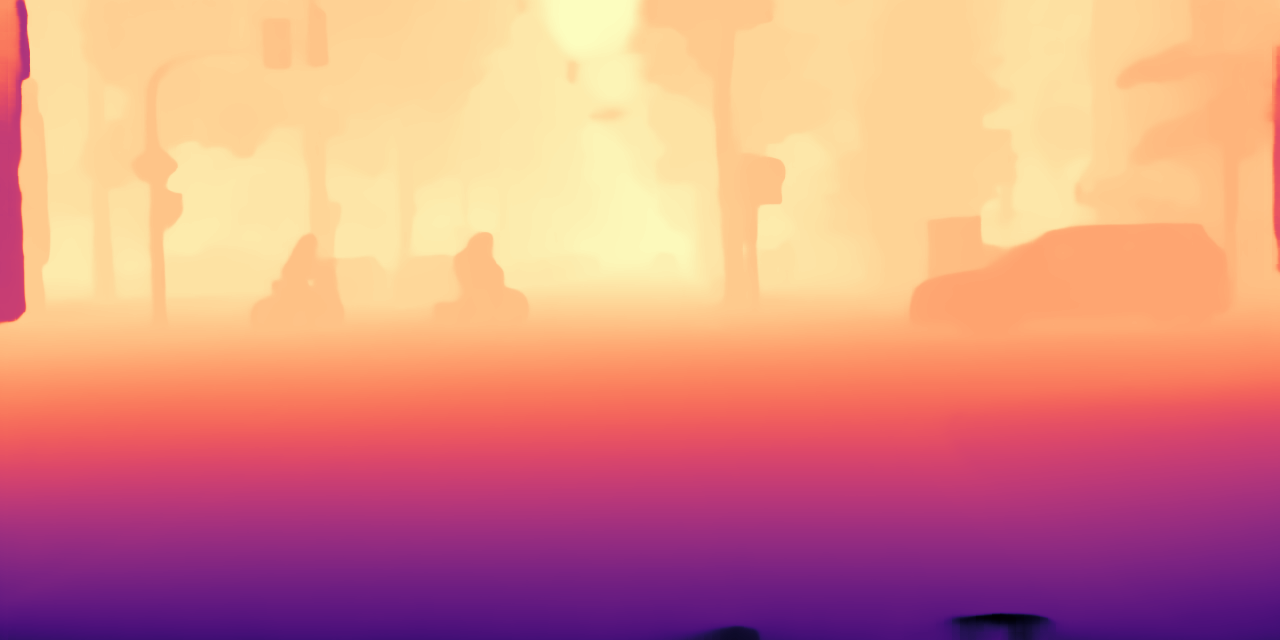}}};
        \node at (2.3, -0.075) {\frame{\includegraphics[width=1.4cm]{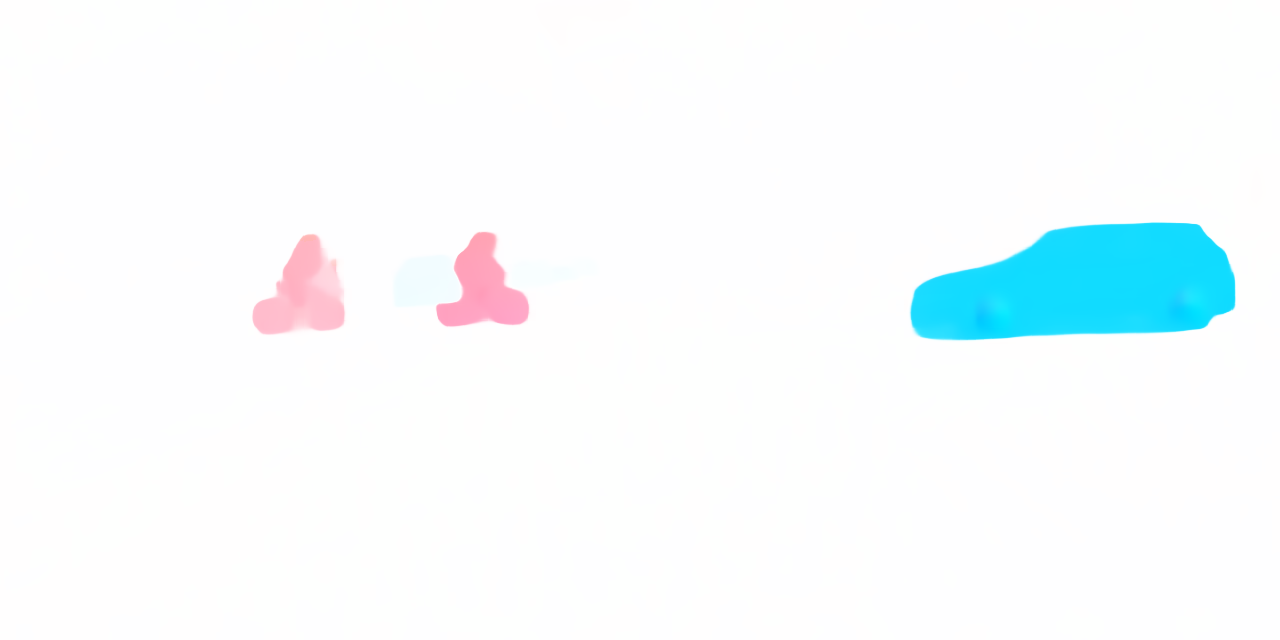}}};
        \node[black] at (2.3, -0.3) {Motion \& Depth};
        \draw[thick, ->] (3.25, 0.0) -- (3.65, 0.0);
        
        \draw[black, rounded corners=2, fill=gray!40] (3.7, 0.05) rectangle ++(2.0, 0.4565) node[pos=.5] {$\!\!$Instance Labeling};
        \draw[black, rounded corners=2, fill=gray!15] (3.7, -0.05) rectangle ++(2.0, -0.4565) node[pos=.5] {$\!\!$Semantic Labeling};
        \draw[thick, ->] (5.8, 0.0) -- (6.2, 0.0);
        \node at (7.05, 0.0) {\frame{\includegraphics[width=1.4cm]{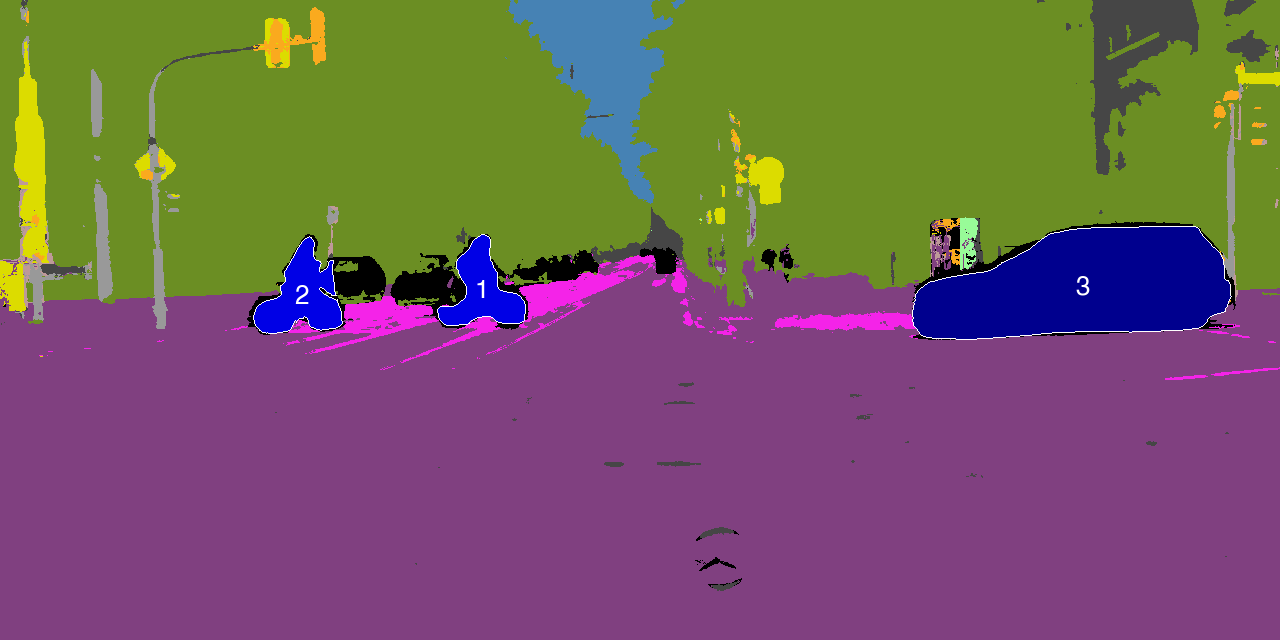}}};
        \node[white, align=center, text width=1.5cm] at (7.05, -0.0) {$\!\!\!$Panoptic Pseudo Label};
        \draw[thick, ->] (8.0, 0.0) -- (8.8, 0.0);
        \draw[-{>[flex=0.75]}, thick, shorten >=0.2cm, draw=pole] (10.2, 0.0) arc[start angle=0, end angle=-180, radius=0.55cm] node[midway, above=1pt] {\color{pole}Train$\,$};
        \draw[{[flex=0.75]<}-, thick, shorten <=0.2cm, draw=pole] (10.2, 0.0) arc[start angle=0, end angle=180, radius=0.55cm] node[midway, below=1pt] {\color{pole}Self$\,$};
        \draw[fill=white, draw=none] (9.1, 0.0) circle [radius=0.2cm] node {\scriptsize$\!\!\mathcal{L}$};
        \draw[fill=white, white] (10.2, 0.0) circle [radius=0.2cm];
        \draw[thick, <-] (9.4, 0.0) -- (10.2, 0.0);
        \pic [local bounding box=our_model] at (11.45, 0) {double_trapezium_small};
        \node[text=black, xshift=-0.025cm] at (our_model) {Panoptic\; Network};
        \draw[thick, <-] (12.65, 0.0) -- (13.05, 0.0);
        \node at (13.925, 0.0) {\includegraphics[width=1.4cm, decodearray={0.0 1.25 0 1.25 0.0 1.25}]{artwork/teaser/data/erfurt_000008_000019_leftImg8bit_enhanced.jpg}};
        \node[white] at (13.925, -0.225) {Input Image};
        \node[] at ((11.75, 0.825) {\footnotesize Unsupervised Panoptic Training};
    \end{tikzpicture}
\end{center}

\vspace{-2.0em}
\captionof{figure}{\textbf{Results and overview of our unsupervised panoptic segmentation approach \MethodName.} We visualize panoptic predictions \textit{(top)} of the current state of the art, U2Seg~\cite{Niu:2024:UUI}, and the proposed \MethodName on various scene-centric datasets. We utilize motion and depth cues from stereo frames \textit{(bottom left)} to generate scene-centric pseudo labels. Given a monocular image \textit{(bottom right)} we learn a panoptic network using our pseudo labels and self-training. \MethodName significantly outperforms U2Seg, indicated by the gains in panoptic quality (PQ).\label{fig:teaser}}
\vspace{1.5\baselineskip}
}]
\begin{abstract}Unsupervised panoptic segmentation aims to partition an image into semantically meaningful regions and distinct object instances without training on manually annotated data. In contrast to prior work on unsupervised panoptic scene understanding, we eliminate the need for object-centric training data, enabling the unsupervised understanding of complex scenes. To that end, we present the first unsupervised panoptic method that directly trains on scene-centric imagery. In particular, we propose an approach to obtain high-resolution panoptic pseudo labels on complex scene-centric data, combining visual representations, depth, and motion cues. Utilizing both pseudo-label training and a panoptic self-training strategy yields a novel approach that accurately predicts panoptic segmentation of complex scenes without requiring any human annotations. Our approach significantly improves panoptic quality, \eg,\ surpassing the recent state of the art in unsupervised panoptic segmentation on Cityscapes by \qty{9.4}{\percent} points in PQ.
\end{abstract}

\thispagestyle{firststyle}
\section{Introduction}\label{sec:introduction}

Panoptic image segmentation \cite{Kirillov:2019:PS} is a comprehensive scene understanding task that unifies semantic and instance segmentation. Semantic segmentation classifies each pixel into categories from a pre-defined semantic taxonomy, whereas instance segmentation aims to detect, segment, and classify each object instance \cite{Cordts:2016:CDS, Lin:2014:COC, Zhou:2024:ISF}. Achieving a semantic and an instance-level understanding of complex scenes are long-standing challenges with broad applications in robotics, autonomous driving, and medical image analysis \cite[see][for an overview]{Minaee:2022:ISU, Zhou:2024:ISF}. Recent progress in panoptic scene understanding \cite{Kirillov:2019:PS, Cheng:2020:PDL, Cheng:2022:M2F} has been primarily driven by supervised learning. However, obtaining the required pixel-level annotations for high-resolution imagery is time and resource-intensive~\cite {Cordts:2016:CDS, Brodermann:2024:MTM}. Although significant resources have been devoted to large-scale supervised %
models \cite{Kirillov:2023:SAM}, there remains a necessity to develop efficient approaches that overcome the need for annotated data. This is particularly relevant when training data is scarce or in ever-changing environments~\cite{Chen:2022:SSL, Verwimp:2024:CLS}.

\begin{figure}[t]
	\vspace{0.2em}
	\input{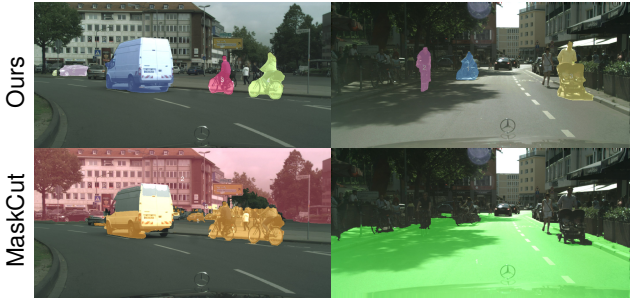}
	\vspace{-0.3em}
	\caption{\textbf{Comparing MaskCut~\cite{Wang:2023:CAL} to our instance labeling} on Cityscapes val. For scene-centric images, MaskCut attends to areas with high semantic correlation instead of instances, reflected in a mask precision (at a \qty{50}{\percent} IoU threshold) of \qty{6.5}{\percent} and \qty{59.6}{\percent} for MaskCut and our instance labels, respectively. \label{fig:maskcut}}
	\vspace{-1.2em}
\end{figure}

A highly promising opportunity lies in approaching panoptic segmentation without any manual supervision. \emph{Unsupervised panoptic segmentation} aims to automatically partition images into semantically meaningful regions and detect each object instance without annotations. This task is rather ambiguous due to the task-dependent and human-defined nature of semantic class boundaries and object notions. Despite these challenges, recent advances in self-supervised learning (SSL) of representations \cite{Caron:2021:EPS, He:2022:MAE, Oquab:2023:DLR} have led to remarkable successes in unsupervised scene understanding. 
Current unsupervised \emph{semantic} segmentation methods, \eg, STEGO~\cite{Hamilton:2022:USS}, leverage pre-trained DINO~\cite{Caron:2021:EPS} features to learn a lower-dimensional representation and clustering \cite{Hamilton:2022:USS, Seong:2023:LHP, Kim:2024:EAL, Sick:2024:USS}. Recent class-agnostic unsupervised \emph{instance} segmentation methods \cite{Simeoni:2021:LOS, Wang:2022:TSO, Wang:2022:FLS, Wang:2023:CAL, Arica:2024:CEU}, for example CutLER~\cite{Wang:2023:CAL}, use self-supervised representations to generate pseudo-instance masks for training detection models. 
In contrast, unsupervised \emph{panoptic} segmentation remains less explored. The only approach to date---U2Seg~\cite{Niu:2024:UUI}---demonstrates the feasibility of this task by building upon CutLER, combining distilled MaskCut~\cite{Wang:2023:CAL} with STEGO for panoptic pseudo labeling and training a panoptic segmentation network. Being only the first step, U2Seg has several limitations. \emph{First}, U2Seg relies on MaskCut, which assumes object-centric images. While highly effective in separating large foreground objects from background in object-centric data, MaskCut struggles with scene-centric images (\cf~\cref{fig:maskcut}).
\emph{Second}, due to the inability to train on scene-centric target datasets, U2Seg is compelled to bypass the classification into ``thing'' and ``stuff'' classes. Consequently, a large number of pseudo-classes are learned, letting this differentiation mainly arise from pseudo to ground-truth class matching during evaluation. \emph{Third}, U2Seg uses low-resolution semantic predictions from STEGO for pseudo supervision, which hampers results on high-resolution, scene-centric data. 

We present \textbf{\MethodName}: scene-\textbf{C}entric \textbf{U}nsupervised \textbf{P}anoptic \textbf{S}egmentation.
Drawing inspiration from Gestalt principles \cite{Wertheimer:1912:ESU, Koffka:1935:PGP} of perceptual organization---\eg, similarity, invariance, and common fate---we complement visual representations with depth and motion cues to extend unsupervised panoptic segmentation to scene-centric data. Gestalt psychology suggests that humans naturally group visual elements based on inherent perceptual cues.
Similarly, we argue that aside from the visual cues used by previous unsupervised scene understanding approaches \cite{Hamilton:2022:USS, Wang:2023:CAL, Niu:2024:UUI}, an additional signal capturing the spatial, three-dimensional properties of scenes is essential. Moreover, motion provides a cue for detecting object instances and achieving a distinction between ``thing'' and ``stuff'' categories, owing to the physical conception of objects being entities capable of moving or being moved.
Leveraging scene-centric data for learning is promising, as it provides an abundance of rich information and is predominantly present in real-world applications (\eg, autonomous driving)~\cite{Minaee:2022:ISU}.
We derive a semantic signal by utilizing depth-enhanced inference grounded in distilled SSL visual representations. Instance-level signals are obtained through SSL 3D motion estimation, leveraging an ensemble-based motion segmentation approach. By integrating these cues, we generate high-precision panoptic pseudo labels, which enable the bootstrapping of a panoptic segmentation network that is subsequently refined through self-training (\cf \cref{fig:teaser}). %
While previous work has utilized motion and depth for unsupervised semantic or instance segmentation, integrating these cues within a panoptic framework has remained unexplored. 

Specifically, we make the following contributions: \emph{(i)}~We derive high-quality panoptic pseudo labels of scene-centric images by leveraging SSL visual representations, SSL depth, and SSL motion. \emph{(ii)}~We effectively train a panoptic segmentation network on our pseudo labels using self-enhanced copy-paste augmentations and self-training. \emph{(iii)}~We demonstrate state-of-the-art unsupervised panoptic segmentation results from the proposed \MethodName across a wide range of scene-centric datasets. Additionally, our approach leads to impressive results on the sub-tasks of unsupervised semantic and instance segmentation. Finally, \MethodName reduces the gap to supervised panoptic segmentation, allowing for label-efficient learning on a fraction of the training data.

\begin{figure*}[t]
    \centering   
    \vspace{-0.1em} %
    \input{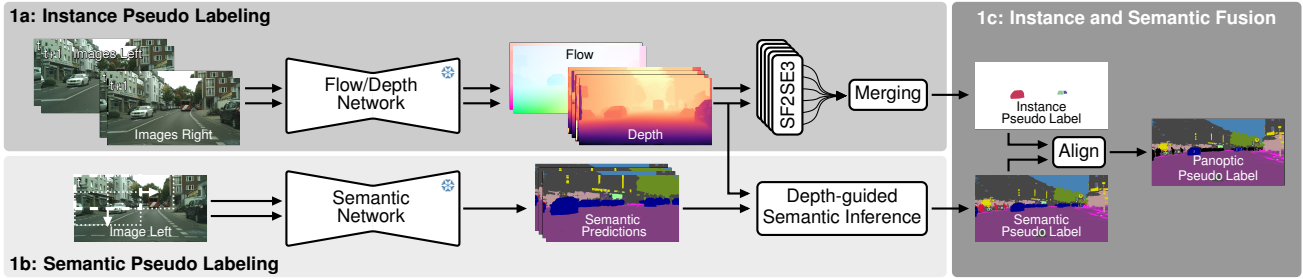}
    \vspace{-0.7em}
    \caption{\textbf{Stage 1: \MethodName pseudo-label generation.} \emph{Instance pseudo labeling} applies ensembling-based SF2SE3 motion segmentation~\cite{Sommer:2022:SCS} to scene flow extracted from flow and depth estimates. \emph{Semantic pseudo labeling} uses a semantic network, distilling and clustering DINO features \cite{Caron:2021:EPS}, combined with a depth-guided inference. \emph{Instance and semantic fusion} aligns the two signals into panoptic pseudo labels. \label{fig:pseudo_label_generation}}
    \vspace{-0.4em}
\end{figure*}

\section{Related Work \label{sec:related_work}}

Approaches for unsupervised segmentation tasks have been significantly influenced by the literature on self-supervised learning (SSL) and low-level vision tasks (\eg, optical flow estimation), which we review first.

\inparagraphnospace{Self-supervised representation learning} focuses on learning generic feature extractors from unlabeled data, aiming for expressive features that facilitate a broad range of downstream tasks~\cite{Ericsson:2022:SSL}. To that end, various self-supervised pretext tasks have been proposed~\cite{Albelwi:2022:SSL, Ericsson:2022:SSL}. The development of Vision Transformers (ViTs)~\cite{Dosovitskiy:2021:AIW} shaped current pretext tasks while allowing for data-scalable training~\cite{Caron:2021:EPS, He:2022:MAE}. Current approaches typically train ViTs on contrastive~\cite{Bachman:2019:LRM, Chen:2020:ISL, He:2020:MCU, Chen:2021:SSL}, negative-free~\cite{Bardes:2022:VIL, Caron:2021:EPS, Chen:2021:SSR, Grill:2020:BYL, Oquab:2023:DLR}, clustering-based~\cite{Asano:2020:SLC, Caron:2018:DCL, Caron:2020:ULV}, or masked modeling~\cite{He:2022:MAE, Gupta:2023:SMA, Nguyen:2024:RMA} pretext tasks. Recent state-of-the-art models (\eg, DINO~\cite{Caron:2021:EPS}) offer semantically rich and dense features suitable for unsupervised scene understanding~\cite{Hamilton:2022:USS, Wang:2023:CAL}.

\inparagraphnospace{Unsupervised optical flow} is concerned with learning optical flow estimation without the need for ground-truth data. While early deep networks relied on synthetic ground-truth flow for supervision~\cite{Dosovitskiy:2015:FN, Mayer:2016:OFD, Sun:2018:PWC, Teed:2020:RAF}, the domain gap to real videos, among other factors, has prompted the development of unsupervised deep optical flow pipelines~\cite{Yu:2016:UFL, Meister:2018:ULO, Jonschkowski:2020:UFL, Marsal:2023:BFL, Lifshitz:2024:UOP}. Current unsupervised optical flow methods (\eg, SMURF~\cite{Stone:2021:SST}) offer accurate flow estimates, fast inference, and generalization to various real-world domains.

\inparagraphnospace{Unsupervised instance segmentation} aims to discover and segment object instances in images~\cite{Simeoni:2024:UOL}. Recent work \cite{Simeoni:2021:LOS, Wang:2022:FLS, Gansbeke:2022:DOM, Wang:2023:CAL, Wang:2024:USA} bootstraps class-agnostic instance segmentation networks using pseudo labels extracted from SSL features on object-centric data. TokenCut~\cite{Wang:2022:TSO} applies normalized cuts \cite[N-Cut,][]{Shi:2000:NCI} to DINO features, providing a foreground pseudo mask. CutLER~\cite{Wang:2023:CAL} proposes MaskCut by iteratively applying N-Cuts, retrieving up to three pseudo masks per image. A second stream of works uses motion cues to obtain an unsupervised signal for object discovery~\cite{Yang:2021:DYS, Liu:2021:TEO, Choudhury:2022:GWM, Karazija:2022:UMS, Safadoust:2023:MOD, Sun:2024:MLM}. SF2SE3~\cite{Sommer:2022:SCS} clusters scene flow from consecutive stereo frames into independent rigid object motions in $\mathit{SE}(3)$ space, improving object segmentation and motion accuracy. MOD-UV~\cite{Sun:2024:MLM} uses motion segmentation for pseudo labeling and multi-stage training.

\inparagraphnospace{Unsupervised semantic segmentation} is approached by early deep learning methods via representation learning \cite{Ji:2019:IIC, Cho:2021:PUS, Harb:2021:IFS}. STEGO~\cite{Hamilton:2022:USS} leverages the self-supervised DINO features as an inductive prior and distills the features into a lower-dimensional space before unsupervised probing. Later, \cite{Seong:2023:LHP, Kim:2024:EAL, Sick:2024:USS} proposed improvements to the feature distillation or probing \cite{Hahn:2024:BUS}. DepthG~\cite{Sick:2024:USS} extends STEGO by spatially correlating the feature maps with depth maps and furthest point sampling in the contrastive loss. 
DiffSeg~\cite{Tian:2024:DAS} utilizes Stable Diffusion~\cite{Rombach:2022:LDM} and iterative attention merging for unsupervised semantic segmentation.

\inparagraphnospace{Unsupervised panoptic segmentation} is a nascent research avenue following recent advancements in unsupervised semantic and instance segmentation.
To the best of our knowledge, U2Seg~\cite{Niu:2024:UUI} is the only method to date to approach unsupervised panoptic segmentation. U2Seg leverages STEGO~\cite{Hamilton:2022:USS} and CutLER~\cite{Wang:2023:CAL} to create panoptic pseudo labels for training a panoptic network. However, its dependence on CutLER's MaskCut approach significantly limits its accuracy on scene-centric data. In contrast, we present the first unsupervised panoptic approach that learns directly from scene-centric data, addressing key limitations of U2Seg and MaskCut. 

\section{Method: \MethodName\label{sec:method}}
We aim to learn a panoptic segmentation network without any manual supervision. To that end, we leverage stereo video during pseudo labeling to incorporate depth and motion cues alongside visual features. Training and inference is done on single monocular images (\cf \cref{fig:teaser}). Our pipeline comprises three stages: \emph{(1)} Generating panoptic pseudo labels; \emph{(2)} bootstrapping a panoptic network with these pseudo labels; and \emph{(3)} self-training of the network.

\subsection{Stage 1: Pseudo-label generation\label{subsec:pseudo_label}}
We generate high-resolution panoptic pseudo labels directly on scene-centric data (\cf \cref{fig:pseudo_label_generation}). Our key insight is that motion and depth provide cues to disambiguate the object instances and semantics in complex scenes. Specifically, we leverage scene flow from an unsupervised framework and perform motion segmentation to obtain pseudo masks for moving objects. Semantic information is derived from our depth-enhanced inference based on pre-trained SSL features \cite{Caron:2021:EPS}.
Fusing these two signals---the semantic information and the instance masks---produces panoptic pseudo labels, which contain both ``thing'' and ``stuff'' classes. \cref{fig:pseudo_label_generation} depicts our pseudo-label generation pipeline.

\inparagraph{1a: Mining scene flow for precise object masks.}
Our first goal is to group scene flow---per-pixel displacement in 3D space and time---into coherently moving regions that likely correspond to object instances. Using two consecutive stereo video frames, $\{ (\mathbf{I}^{\rm l}_{t}, \mathbf{I}^{\rm r}_{t}), (\mathbf{I}^{\rm l}_{t + 1}, \mathbf{I}^{\rm r}_{t + 1})\}$ of dimension $\mathbb{R}^{3 \times \rm H \times \rm W}$, we estimate scene flow.
We use SMURF~\cite{Stone:2021:SST} to compute unsupervised optical flow and disparity from the pair of stereo images. Given the forward flow $\mathbf{f}^{\rm fw}$ and backward flow $\mathbf{f}^{\rm bw}$ of dimension $\mathbb{R}^{2\times \rm H \times \rm W}$, as well as the estimated disparity $\{ (\mathbf{d}^{\rm lr}_{t}$, $\mathbf{d}^{\rm rl}_{t}$), ($\mathbf{d}^{\rm lr}_{t + 1}, \mathbf{d}^{\rm rl}_{t + 1}) \}$ in $\mathbb{R}^{\rm H \times \rm W}$, we compute depth $\mathbf{D}\in\mathbb{R}^{\rm H \times \rm W}$ and scene flow $\mathbf{F}\in\mathbb{R}^{3\times \rm H \times \rm W}$ with respect to frame $\mathbf{I}^{\rm l}_{t}$ using the camera parameters of the dataset. We derive a consistency mask $\mathbf{O}\in\{0, 1\}^{\rm H \times \rm W}$ using forward-backward and left-right consistency~\cite{Sundaram:2010:OCC}.

Equipped with the scene flow $\mathbf{F}$ and the consistency mask $\mathbf{O}$, we perform motion segmentation to obtain object masks of moving objects. We employ SF2SE3~\cite{Sommer:2022:SCS}, a motion clustering algorithm that uses $\mathbf{F}$ and $\mathbf{O}$ to fit a variable number of rigid motions, defined in the Lie group $\mathit{SE}(3)$. This results in a set of $\mathit{SE}(3)$-motions with corresponding masks.
However, the original SF2SE3 algorithm is stochastic due to random initialization, which can produce inconsistent motion segmentation across multiple runs. %
To mitigate these inconsistencies, we perform sampling-based filtering and mask refinement.
While SF2SE3 ensures non-overlapping masks, running SF2SE3 $n$ times produces $m$ potentially overlapping masks $\mathbf{M}_{i,:,:}\in\{0, 1\}^{\rm H\times \rm W}$, $i \in \{1, ..., m\}$. To identify potentially incorrect masks, we compute a consistency score $c\in[0, 1]^{m}$ for each mask as
\begin{equation}
    c_{i}= \sum\limits_{h, w} \bigg(\mathbf{M}_{i,:,:}\odot\sum\limits_{j=1}^{m}\mathbf{M}_{j,:,:}\bigg) / \bigg( \sum\limits_{h', w'}\mathbf{M}_{i,:,:} \bigg),
\end{equation}
where $\odot$ is the Hardamard product.
We keep object masks that occur in at least \qty{80}{\percent} of SF2SE3 runs (\ie, $c_{i} \!\geq\! 0.8$).
This straightforward consistency filtering removes potentially erroneous predictions. Conflicts between overlapping masks are resolved using matrix non-maximum suppression~\cite{Wang:2020:NMS}. Finally, we isolate connected components of our object masks. Overall, this process results in $l$ high-precision moving object masks $\Tilde{\mathbf{M}}\in\{0, 1\}^{l\times \rm H\times \rm W}$. 

\begin{figure*}[t]

\begin{minipage}{0.48\textwidth}
    \centering
    \vspace{-0.1em} %
    \input{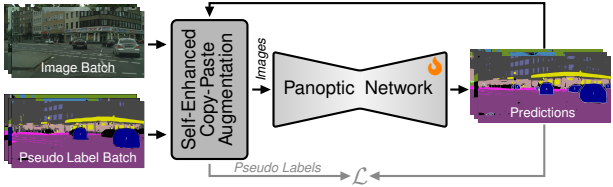}
    \subcaption{Stage 2: \MethodName panoptic bootstrapping. \label{fig:train}}
\end{minipage}
\hspace{0.04\textwidth}
\begin{minipage}{0.48\textwidth}
    \centering
    \vspace{-0.1em} %
    \input{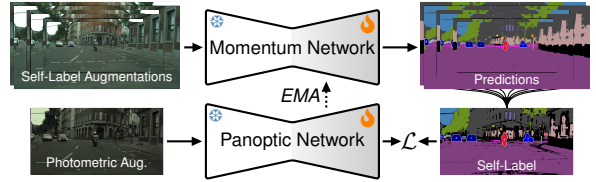}
    \vspace{3.575pt}
    \subcaption{Stage 3: \MethodName panoptic self-training. \label{fig:self_train}}
\end{minipage}
\vspace{-0.5em}
\caption{\textbf{Overview of our \MethodName training and self-training.} \emph{Panoptic bootstrapping \subref{fig:train}} optimizes the panoptic network based on the pseudo labels using self-enhanced copy-paste augmentation. 
\emph{Self-training of} \MethodName \emph{\subref{fig:self_train}} is performed by obtaining augmented predictions from a momentum network. We align, fuse, and filter the predictions to obtain refined 
self-labels. The panoptic network is trained on photometrically augmented images and pseudo labels. The momentum network prediction heads are updated using EMA weight updates.}
\label{fig:train-overview}
\vspace{-0.5em}
\end{figure*}

\inparagraph{1b: Depth-guided semantic pseudo labeling.\label{sec:depth_guided_semantic_inference}} Next, we proceed to extract the semantic pseudo labels. Thereby, our goal is to partition an image into semantically meaningful regions in an unsupervised manner while accounting for the high resolution and fine details present in scene-centric data. This is particularly relevant in the context of unsupervised semantic segmentation, where state-of-the-art methods typically operate at low resolutions (\eg, \(320 \times 320\)).\footnote{This is mainly due to the low resolution used in SSL pre-training \cite{Caron:2021:EPS}.} To address this limitation, we propose to enhance semantic inference by incorporating depth guidance.
We derive the semantic segmentation model by distilling a lower-dimensional representation of DINO features using a contrastive loss extended by depth as an auxiliary signal to guide feature correlation and sampling. The segmentation is obtained via stochastic cosine-distance $\rm K$\nobreakdash-means. The result is an unsupervised semantic segmentation model $\mathcal{S}: \mathbb{R}^{3\times \rm H \times \rm W} \rightarrow \mathbb{R}^{\rm K\times \rm H \times \rm W}$, which encodes an image $\mathbf{I}\in \mathbb{R}^{3\times \rm H \times \rm W}$ into a dense feature representation $\mathbf{P}\in \mathbb{R}^{\rm K\times \rm H \times \rm W}$ with $\rm K$ semantic pseudo classes per pixel.

The depth-guided semantic inference takes two semantic predictions at different resolutions, $\mathbf{P}^\text{low}$ and $\mathbf{P}^\text{high}$, and uses depth to compute their weighted average.
$\mathbf{P}^\text{low}$ comes from running $\mathcal{S}$ on a downscaled input resolution.
Conversely, $\mathbf{P}^\text{high}$ represents a semantic prediction at a higher resolution.
To compute $\mathbf{P}^\text{high}$, we slide a window over the image, spatially concatenate the output tensors, and average the soft predictions (see \supp for details).
Intuitively, $\mathbf{P}^\text{low}$ provides reliable feature representations for large-scale scene elements located at close range.
In contrast, $\mathbf{P}^\text{high}$ encapsulates a more fine-grained representation, which benefits small-scale objects at a distance.
Utilizing the inverse relationship between the projected object size and its distance to the camera in the pinhole model, we compute a depth-based weight $\boldsymbol{\alpha}$ at each pixel $(h,w)$ as 
\begin{equation}    
    \alpha_{h,w} = ({D}_{h,w}+1)^{-1}.
\end{equation}
Note that we add $1$ in the denominator for a bounded range, ensuring that $\alpha_{h,w} \in [0, 1]$. We use the depth estimate $\mathbf{D}$, obtained during instance pseudo labeling in \textbf{Step 1a} above.
The weighted semantic prediction $\mathbf{P}^{\ast}$ is then given as
\begin{equation}
    \mathbf{P}^{\ast}_{k,:,:} = \boldsymbol{\alpha} \odot \mathbf{P}^\text{low}_{k,:,:} + (1 - \boldsymbol{\alpha}) \odot \mathbf{P}^\text{high}_{k,:,:}.
\end{equation}
Observe that the pixels with small depth values ${D}_{h,w}$ will derive their feature representation predominantly from $\mathbf{P}^\text{low}$, whereas $\mathbf{P}^\text{high}$ will contribute to the semantic representation of the pixels with a large depth.
To further improve the alignment of $\mathbf{P}^{\ast}$ with the image structure, we perform post-processing with a fully-connected conditional random field \cite{Kraehenbuehl:2011:EIF}. 
For our unsupervised semantic segmentation model $\mathcal{S}$, we build upon DepthG~\cite{Sick:2024:USS}. To conform with our fully unsupervised setup, we re-train $\mathcal{S}$ with the stereo depth from the unsupervised SMURF model.

\inparagraph{1c: Instance and semantic fusion. \label{sec:label_fusion}}
We can finally obtain panoptic pseudo labels by fusing the instance and semantic pseudo labels.
A challenge here is distinguishing between the semantic classes belonging to ``stuff'' or ``thing'' categories in panoptic segmentation.
Aggregating pixel distributions across all images, we compute the ratio of each pseudo class’s frequency within the instance masks relative to its overall frequency. %
We designate semantic pseudo classes with a high ratio above a predefined threshold $\psi^{\rm ts}$ as ``thing'', and those below it as ``stuff''. Next, we assign a consistent semantic pseudo-label ID within each instance mask.
Given a semantic tensor $\mathbf{P}^{\ast}$, we assign the most frequent semantic pseudo-class ID within each pseudo instance mask.
Similarly, we assign image areas of pseudo-class IDs corresponding to pseudo-``thing'' classes that do not have an instance mask to ``ignore''. This results in the final pseudo labels, encompassing the training signal for all ``stuff'' regions as well as moving ``thing'' instances.

\subsection{Learning unsupervised panoptic segmentation\label{subsec:learn_ups}}
Using our panoptic pseudo labels, we train a panoptic network in two stages (\cf \cref{fig:train-overview}). First, we increase the coverage of the initial sparse set of pseudo labels in a bootstrapping stage. Then, we use self-training on self-labels from ensembled predictions to enhance the model's accuracy.

\inparagraph{Stage 2: Panoptic bootstrapping.} 
\Cref{fig:train} provides an overview of the bootstrapping stage. We utilize the pseudo labels from \cref{subsec:pseudo_label}, containing semantic information and a sparse set of instance masks for moving objects.
Despite the set sparsity, we can use this set of ``thing'' masks for network bootstrapping to accommodate the static objects as well. For this we employ the DropLoss~\cite{Wang:2023:CAL}, defined by
\begin{equation}\label{eq:droploss}
     \mathcal{L}_{\text{drop}}(\mathbf{R}_{j}, \hat{\mathbf{R}}_{i}) = \mathds{1}\big(\operatorname{IoU}_j^{\max} > \tau^{\operatorname{IoU}}\big)\,\mathcal{L}_{\text{Th}}(\mathbf{R}_{j}, \hat{\mathbf{R}}_{i}),
\end{equation}
where $\mathcal{L}_{\text{Th}}$ is the detection loss \cite{Kirillov:2019:PS}.
The DropLoss approach in \cref{eq:droploss} only supervises ``thing'' instances $\mathbf{R}_{j}$ whose maximum overlap $\operatorname{IoU}_j^{\max}$ with a pseudo mask $\hat{\mathbf{R}}_{i}$ exceeds $\tau^{\operatorname{IoU}}$. Importantly, the DropLoss does not penalize ``thing'' predictions that do not overlap with any pseudo mask from ``thing'' categories. %
This selective supervision allows the network to expand its prediction to potentially static objects not captured by the pseudo labels. For semantics, we pseudo-supervise using a standard cross-entropy loss while omitting ``ignore'' pixels in the pseudo semantics.

To enhance the accuracy of the panoptic network on small objects, we employ a variant of copy-paste augmentation~\cite{Dwibedi:2017:CPA}.
Instead of pasting masks from our pseudo labels, we copy-paste model predictions as they become confident during training.
This \textit{self-enhanced copy-paste augmentation} promotes the bootstrapping stage, since the network gradually discovers more potentially static objects.

\inparagraph{Stage 3: Panoptic self-training\label{subsec:self_train}.} 
In this final stage, illustrated in \cref{fig:self_train}, we further boost the panoptic segmentation accuracy of \MethodName. We self-train the network on self-labels by ensembling and confidence thresholding of augmented predictions.
Our self-training maintains a momentum network as an exponential moving average (EMA) of our panoptic network. 
This approach is much akin to the student-teacher framework \cite{Xie:2020:SST, He:2020:MCU, Araslanov:2021:SAC}, which we apply here for panoptic segmentation.
In detail, we create views of an input image using horizontal flipping and multiple image scales. The momentum network infers panoptic predictions for each view. We apply the inverse transform of each augmentation and merge the batch of predictions by averaging the soft semantic and instance predictions. 
Self-labels are retrieved by confidence thresholding of both the semantic and the instance signal. Given the averaged instance predictions $\tilde{\mathbf{R}}\in[0, 1]^{\rm J \times \rm H \times \rm W}$ and their confidence prediction $\kappa\in[0, 1]^{\rm J}$, we apply the threshold $\gamma \in [0,1]$ and only keep instance masks for which $\kappa_{j}>\gamma$ for the instance self-label.
For the semantic self-label $\mathbf{L}^{\text{sem}}$, we derive a class-dependent threshold $\zeta_{k} \in [0,1]$. 
Given the averaged semantic predictions $\tilde{\mathbf{P}}\in \mathbb{R}^{\rm K\times \rm H \times \rm W}$, we compute $\zeta_{k}$ from the semantic threshold $\hat{\zeta}$ and the maximum probability of every pseudo class as $\zeta_{k}=\hat{\zeta} \max (\tilde{\mathbf{P}}_{k,:,:})$.
Given the predicted pseudo class with the highest probability $k^*_{h,w} = \arg \max (\tilde{\mathbf{P}}_{:,h,w})$, we ignore low-confidence predictions using our class-dependent threshold $\zeta_{k}$:%
\begin{equation}
\mathbf{L}^{\text{sem}}_{h,w} = 
\begin{cases} 
      k^*_{h,w} & \text{if } \max (\tilde{\mathbf{P}}_{:,h,w}) \geq \zeta_{k^*_{h,w}} \\[-0.85pt]
      \text{ignore}, & \text{otherwise}.
\end{cases}
\end{equation}

In summary, for each optimization step, we apply augmentations to the image and generate a self-label by merging the augmentation-based predictions from the momentum network.
Obtaining a second prediction from a photometrically perturbed image from our panoptic network, we apply a standard panoptic loss \wrt the self-label.
We update the panoptic network with gradient descent and use EMA to update the momentum network.
We freeze all normalization layers and only train the heads of the network.

\begin{table*}[t]
    \centering
    \caption{\textbf{Unsupervised panoptic segmentation on Cityscapes val.} Comparing \MethodName to existing unsupervised panoptic methods, using PQ, SQ, and RQ, as well the PQ for ``thing'' and ``stuff'' classes (all in \%, $\uparrow$). $\dagger$ denotes results reported in \cite{Niu:2024:UUI}.\label{tab:panoptic_segmentation_cs}}
    \vspace{-0.3em}
    \footnotesize\sisetup{table-number-alignment=center}
\setlength{\tabcolsep}{8pt}
\renewcommand{\arraystretch}{0.885}
\begin{tabularx}{\textwidth}{>{\hspace{-\tabcolsep}\raggedright\columncolor{white}[\tabcolsep][\tabcolsep]}lYYZZZZZ}
    \toprule
    {\textbf{Method}} & {\textbf{Training data}} & {\textbf{Pseudo classes}} & {\textbf{PQ}} & {\textbf{SQ}} & {\textbf{RQ}} & {\textbf{PQ\textsuperscript{Th}}} & {\textbf{PQ\textsuperscript{St}}} \\
    \midrule
    \textcolor{tud0c}{Supervised~\cite{Kirillov:2019:PFP}} & \textcolor{tud0c}{Cityscapes} & \textcolor{tud0c}{--} & \color{tud0c}62.3 & \color{tud0c}81.8 & \color{tud0c}75.1 & \color{tud0c}62.4 & \color{tud0c}62.1 \\
    \midrule
    {DepthG~\cite{Sick:2024:USS} + CutLER~\cite{Wang:2023:CAL}} & Cityscapes \& ImageNet & 27 & 16.1 & 45.4	& 21.1	& 3.0	& 25.7  \\
    {U2Seg$^{\dagger}$~\cite{Niu:2024:UUI}} & COCO \& ImageNet & 800 + 27 & 17.6 & 52.7 & 21.7 & 8.4 & 24.2 \\
    {U2Seg~\cite{Niu:2024:UUI}} & COCO \& ImageNet & 800 + 27 & 18.4 & 55.8	& 22.7	& 10.2 & 24.3 \\
    \midrule
    \rowcolor{tud0c!20} 
    {\MethodName \textit{(Ours)}} & Cityscapes & 27 & \bfseries 27.8	& \bfseries 57.4	& \bfseries 35.2 & \bfseries 17.7	& \bfseries 35.1  \\
    {\textit{vs.\ prev.\ SOTA}} &  &  & \dlt{\hspace{-0.43pt}+9.4} & \dlt{\hspace{-0.43pt}+1.6} & \dlt{\hspace{-5pt}+12.5} & \dlt{\hspace{-0.43pt}+7.5} & \dlt{\hspace{-5pt}+10.8} \\  
    \bottomrule
\end{tabularx}

    \vspace{-0.5em}
\end{table*}

\section{Experiments \label{sec:experiments}}
We evaluate the proposed \MethodName on an extensive set of benchmarks and compare it to a simple baseline and the current state-of-the-art methods for unsupervised panoptic segmentation, semantic segmentation, and class-agnostic instance segmentation.
We refer to the supplemental material for additional quantitative and qualitative results.

\inparagraph{Datasets.} We train \MethodName on pseudo labels generated using Cityscapes training sequences and evaluate it on Cityscapes val~\cite{Cordts:2016:CDS}. We also evaluate the generalization of our method through cross-domain experiments on KITTI panoptic segmentation \cite{Mohan:2021:KIP, Geiger:2012:AWR}, BDD~\cite{Yu:2020:BAD}, MUSES~\cite{Brodermann:2024:MTM}, and Waymo~\cite{Sun:2020:SPA}. Following the evaluation by Niu \etal \cite{Niu:2024:UUI}, we use 19 semantic categories from Cityscapes for unsupervised panoptic segmentation.
For the cross-domain datasets, we ensured compatibility of their label space with the Cityscapes categories. Additionally, we test \MethodName in an out-of-domain setting by evaluating it on MOTS~\cite{Voigtlaender:2019:MMO}.
Our evaluation for unsupervised semantic segmentation follows the established protocol \cite{Ji:2019:IIC, Cho:2021:PUS, Hamilton:2022:USS, Sick:2024:USS} considering all 27 Cityscapes classes. For unsupervised class-agnostic instance segmentation, we follow previous work \cite{Cao:2023:HAS, Sun:2024:MLM} and evaluate on Waymo.
We additionally demonstrate the versatility of our method \wrt the training set choice.
Specifically, we alternatively use the raw KITTI data (instead of Cityscapes) for training, excluding all images used for evaluation. Please refer to our supplement for more details.

\inparagraph{Evaluation metrics.} For evaluating the panoptic segmentation accuracy, we utilize the panoptic quality (PQ)~\cite{Kirillov:2019:PS} metric. We also report the segmentation quality (SQ) and recognition quality (RQ), which are part of PQ. As we train without any supervision, the semantic pseudo IDs predicted by the model need to be aligned with the ground truth for evaluation.
To that end, we adapt the established evaluation in unsupervised semantic segmentation \cite{Ji:2019:IIC, Cho:2021:PUS, Hamilton:2022:USS, Sick:2024:USS} to unsupervised panoptic segmentation.
In particular, we match ``thing'' and ``stuff'' classes independently based on the pixel-wise overlap of the pseudo-label IDs to the ground-truth labels across the dataset using the Hungarian algorithm~\cite{Kuhn:1955:HUN}. After this one-to-one matching, we use maximum assignment for the remaining pseudo-label IDs. Notably, this matching solely depends on predicted pseudo semantics, avoiding extensive matching between segments, and does not introduce any hyperparameters, unlike the matching of U2Seg~\cite{Niu:2024:UUI}.
For unsupervised semantic segmentation, we report both the mean Intersection-over-Union (mIoU) and the all-pixel accuracy (Acc) following \cite{Ji:2019:IIC, Cho:2021:PUS, Hamilton:2022:USS, Sick:2024:USS}. Class-agnostic instance segmentation is evaluated using mask mean average precision (AP) \cite{Padilla:2020:SOD}, mask AP at an IoU threshold of \qty{50}{\percent} (AP\textsubscript{50}), and AP for small, medium, and large objects \cite{Lin:2014:COC}.

\inparagraph{Implementation details.} We utilize 27 pseudo classes for pseudo-label generation (stage 1) allowing for comparison with both existing unsupervised panoptic and semantic segmentation approaches. For a fair comparison, we follow Niu \etal \cite{Niu:2024:UUI} by using the same model architecture and initialization---Panoptic Cascade Mask R-CNN~\cite{Cai:2018:CRC, Kirillov:2019:PFP} with a self-supervised pre-trained DINO ResNet-50~\cite{He:2016:DRL} backbone. \MethodName pseudo-label training (stage 2) is using a thing-stuff threshold $\psi^{\rm ts}$ of \num{0.08}, applying DropLoss, and self-enhanced copy-paste augmentation for \num{4000} steps optimized with AdamW~\cite{Loshchilov:2018:ADW}. \MethodName self-training (stage 3) applies multi-scale, horizontal flipping, and photometric augmentations following Chen \etal \cite{Chen:2020:ISL}. We apply self-enhanced copy-paste augmentation and EMA for \num{1500} steps optimized with AdamW. We provide all implementation details in the supplement.

\inparagraph{Unsupervised panoptic segmentation baseline.}
To construct a strong baseline equivalent of our method, we integrate the unsupervised semantic segmentation method DepthG~\cite{Sick:2024:USS} with the unsupervised class-agnostic instance segmentation of CutLER~\cite{Wang:2023:CAL}. 
We obtain a panoptic prediction by fusing the semantic and instance predictions in the same fashion as we do for our pseudo labels (\cf~\cref{sec:label_fusion}).

\inparagraph{Supervised upper bound.}
To better assess the effectiveness of \MethodName, we train a supervised variant of our method and report its performance as an ``upper bound'' of our framework. In line with our experimental setup, we train on Cityscapes and test on the same datasets as \MethodName.

\begin{table*}[t]
    \centering
    \caption{\textbf{Generalization.} Comparing \MethodName with unsupervised panoptic segmentation methods, using PQ, SQ, and RQ (in \%, $\uparrow$) in terms of generalization to KITTI panoptic, BDD, MUSES, and Waymo. In addition, we analyze generalization to the OOD dataset MOTS.\label{tab:panoptic_segmentation_others}}
    \vspace{-0.3em}
    \footnotesize\sisetup{table-number-alignment=center}
\setlength{\tabcolsep}{5pt}
\renewcommand{\arraystretch}{0.885}
\begin{tabularx}{\textwidth}{>{\hspace{-\tabcolsep}\raggedright\columncolor{white}[\tabcolsep][\tabcolsep]}XZZZZZZZZZZZZZZZ}
    \toprule
    & \multicolumn{3}{c}{\textbf{KITTI}} & \multicolumn{3}{c}{\textbf{BDD}} & \multicolumn{3}{c}{\textbf{MUSES}} & \multicolumn{3}{c}{\textbf{Waymo}} & \multicolumn{3}{c}{\textbf{MOTS} \textit{(OOD)}}\\
    \cmidrule(l{0.2em}r{0.2em}){2-4} \cmidrule(l{0.2em}r{0.2em}){5-7} \cmidrule(l{0.2em}r{0.2em}){8-10} \cmidrule(l{0.2em}r{0.2em}){11-13} \cmidrule(l{0.2em}r{0.2em}){14-16}
    \multirow{-2}{*}{\vspace{0.5em}\textbf{Method}} & \textbf{PQ} & \textbf{SQ} & \textbf{RQ} & \textbf{PQ} & \textbf{SQ} & \textbf{RQ} & \textbf{PQ} & \textbf{SQ} & \textbf{RQ} & \textbf{PQ} & \textbf{SQ} & \textbf{RQ} & \textbf{PQ} & \textbf{SQ} & \textbf{RQ}\\
    \midrule
    \textcolor{tud0c}{Supervised~\cite{Kirillov:2019:PFP}} & \color{tud0c}31.9 & \color{tud0c}71.7 & \color{tud0c}40.4 & \color{tud0c}33.0 & \color{tud0c}76.3 & \color{tud0c}42.0 & \color{tud0c}38.1 & \color{tud0c}62.4 & \color{tud0c}49.6 & \color{tud0c}31.5 & \color{tud0c}70.1 & \color{tud0c}40.9 & \color{tud0c}73.8 & \color{tud0c}86.4 & \color{tud0c}84.6  \\
    \midrule
    {DepthG~\cite{Sick:2024:USS}~+~CutLER~\cite{Wang:2023:CAL}} & 11.0	& 34.5	& 13.8 & 14.4	& 41.9	& 19.2 & 10.1 & 30.1 & 13.1 & 13.4	& 37.3 & 17.0 & 49.6	& 78.4	& 60.6  \\
    {U2Seg~\cite{Niu:2024:UUI}} & 20.6 & 52.9 & 25.2 & 15.8	& 57.2 & 19.2 & 20.3 & 45.8 & 26.5 & 19.8 & 50.8 & 23.4 & 50.7	& 79.2	& 64.3 \\
    \midrule
    \rowcolor{tud0c!20} {\MethodName \textit{(Ours)}} & \bfseries 25.5 & \bfseries 58.1 & \bfseries 32.5 & \bfseries 19.9	& \bfseries 60.3	& \bfseries 25.9 & \bfseries 24.4	& \bfseries 48.5	& \bfseries 33.0 & \bfseries 26.4	& \bfseries 60.3	& \bfseries 33.0 & \bfseries 67.8	& \bfseries 86.4	& \bfseries 76.9  \\
    {\textit{vs. prev. SOTA}} & \dlt{\hspace{-0.43pt}+4.9} & \dlt{\hspace{-0.43pt}+5.2} & \dlt{\hspace{-0.43pt}+7.3} & \dlt{\hspace{-0.43pt}+4.1} & \dlt{\hspace{-0.43pt}+3.1} & \dlt{\hspace{-0.43pt}+6.7} & \dlt{\hspace{-0.43pt}+4.1} & \dlt{\hspace{-0.43pt}+2.7} & \dlt{\hspace{-0.43pt}+6.5} & \dlt{\hspace{-0.43pt}+6.6} & \dlt{\hspace{-0.43pt}+9.5} & \dlt{\hspace{-0.43pt}+9.6} & \dlt{\hspace{-5pt}+17.1} & \dlt{\hspace{-0.43pt}+7.2} & \dlt{\hspace{-5pt}+12.6} \\
    \bottomrule
\end{tabularx}

    \vspace{-0.6em}
\end{table*}%
\begin{table}[t]
    \centering
    \caption{\textbf{Unsupervised semantic segmentation.} Comparing \MethodName to existing unsupervised semantic segmentation methods on Cityscapes val, using Accuracy and mean IoU (in \%, $\uparrow$). \label{tab:semanticsegresults}}
    \vspace{-0.3em}
    \footnotesize\sisetup{table-number-alignment=center}
\setlength{\tabcolsep}{5pt}
\renewcommand{\arraystretch}{0.885}
\begin{tabularx}{\columnwidth}{>{\hspace{-\tabcolsep}\raggedright\columncolor{white}[\tabcolsep][\tabcolsep]}lYZZ}
	\toprule
    {\textbf{Method}} & {\textbf{Model}} & {\textbf{Acc}} & {\textbf{mIoU}}\\
    \midrule
    \textcolor{tud0c}{Supervised~\cite{Kirillov:2019:PFP}} & \textcolor{tud0c}{Pano. Cascade Mask R-CNN} & \color{tud0c} 94.7 & \color{tud0c} 76.7 \\
    \midrule
    {PiCIE~\cite{Cho:2021:PUS}} & ResNet-18 + FPN & 65.5 & 12.3 \\
    {DiffSeg~\cite{Tian:2024:DAS}} & Stable Diffusion V1.4 & 67.3 & 15.2 \\
    {HP~\cite{Seong:2023:LHP}} & DINO-S/8  & 80.1 & 18.4 \\
    {PriMaPs-EM~\cite{Hahn:2024:BUS}} & DINO-S/8  & 81.2 & 19.4 \\
    {STEGO~\cite{Hamilton:2022:USS}} & DINO-B/8  & 73.2 & 21.0 \\
    {EAGLE~\cite{Kim:2024:EAL}} & DINO-B/8  & 79.4 & 22.1 \\
    {DepthG~\cite{Sick:2024:USS}} & DINO-B/8 & 81.6 & 23.1 \\
    {U2Seg~\cite{Niu:2024:UUI}} & Pano. Cascade Mask R-CNN & 79.1 & 21.6  \\
    \midrule
    \rowcolor{tud0c!20} {\MethodName \textit{(Ours)}} & Pano. Cascade Mask R-CNN & \bfseries 83.2 & \bfseries 26.8 \\
	\bottomrule
\end{tabularx}

    \vspace{-0.9em}
\end{table}%

\subsection{Comparison to the state of the art \label{sec:results}}
Our experiments assess the unsupervised panoptic segmentation accuracy of \MethodName within its training domain and its generalization capabilities across diverse scene-centric datasets.
We further evaluate \MethodName on the two panoptic sub-tasks: unsupervised semantic and class-agnostic instance segmentation. Additionally, we analyze the impact of individual components of our method. Lastly, we investigate label-efficient learning.

\inparagraph{Unsupervised panoptic segmentation.}
\Cref{tab:panoptic_segmentation_cs} compares \MethodName with the state of the art in unsupervised panoptic segmentation, U2Seg~\cite{Niu:2024:UUI}, and our baseline, DepthG~\cite{Sick:2024:USS}\,+\,CutLER~\cite{Wang:2023:CAL}, evaluated on the Cityscapes validation dataset.
We also report a supervised upper bound.
Since the evaluation code of U2Seg for Cityscapes is not publicly available, we re-evaluate U2Seg using our pseudo-class-to-class matching and report the results of \cite{Niu:2024:UUI} for reference. 
\MethodName achieves a PQ of \qty{27.8}{\percent}, substantially improving over U2Seg (\qty{18.4}{\percent}). This demonstrates how \MethodName{} effectively utilizes scene-centric training data to improve panoptic quality.
Remarkably, \MethodName achieves superior panoptic quality across all ``thing'' classes (PQ$^{\text{Th}}$), despite not excessively over-clustering instance classes (contrary to U2Seg with 800 ``thing'' pseudo classes) and solving the additional thing-stuff assignment task.

In \cref{tab:panoptic_segmentation_others}, we explore the generalization of \MethodName across various scene-centric domains. \MethodName consistently surpasses the baseline as well as U2Seg on all datasets. These results indicate that \MethodName not only excels in the training domain, Cityscapes, but is also effective under a domain shift, underscoring its robustness.
By contrast, the supervised baseline suffers a significant drop in accuracy under the domain shift --- an effect not observed for the unsupervised approaches. In addition, we assess the generalization of \MethodName on MOTS, which represents an out-of-domain (OOD) testing scenario.
\MethodName achieves outstanding segmentation accuracy, substantially surpassing the baseline and U2Seg.
These results provide strong evidence that \MethodName excels in the OOD scenario as well.

\inparagraph{Unsupervised semantic segmentation.} As unsupervised panoptic segmentation implicitly solves the task of unsupervised semantic segmentation, we also assess the performance of \MethodName on this sub-task by comparing it to recent methods on Cityscapes. 
\Cref{tab:semanticsegresults} summarizes the results.
\MethodName achieves state-of-the-art semantic segmentation accuracy, improving over the previous best method, DepthG~\cite{Sick:2024:USS}, by a significant margin.
In comparison to U2Seg, \MethodName improves substantially by \qty{4.1}{\percent} in pixel accuracy and by \qty{5.2}{\percent} in mIoU.%

\begin{table}[t!]
    \centering
    \caption{\textbf{Unsupervised instance segmentation.} Comparing \MethodName, trained on Cityscapes, to unsupervised class-agnostic instance segmentation methods on Waymo using mask APs (\%, $\uparrow$).\label{tab:objectdetectionresults}}
    \vspace{-0.3em}
    \footnotesize\sisetup{table-number-alignment=center}
\setlength{\tabcolsep}{3.42pt}
\renewcommand{\arraystretch}{0.885}
\begin{tabularx}{\columnwidth}{>{\hspace{-\tabcolsep}\raggedright\columncolor{white}[\tabcolsep][\tabcolsep]}lcZS[table-format=2.1]S[table-format=2.1]S[table-format=2.1]S[table-format=2.1]S[table-format=2.1]}
	\toprule
    {\textbf{Method}} & {\textbf{Training data}} & {\textbf{AP\textsubscript{50}}} & {\textbf{AP}} & {\textbf{AP\textsubscript{S}}} & {\textbf{AP\textsubscript{M}}} & {\textbf{AP\textsubscript{L}}}\\
    \midrule
    \textcolor{tud0c}{Supervised~\cite{Kirillov:2019:PFP}} & \color{tud0c} Cityscapes & \color{tud0c} 44.6 & \color{tud0c} 27.6 & \color{tud0c} 10.3 & \color{tud0c} 45.0 & \color{tud0c} 73.3 \\
    \midrule
    {U2Seg~\cite{Niu:2024:UUI}}     & COCO \& ImageNet  & 4.3 & 2.3 & 0.0 & 1.5 & 17.0 \\
    {CutLER \cite{Wang:2023:CAL}}   & ImageNet          & 9.1 & 5.2 & 0.0 & 3.4 & 34.6 \\
    {HASSOD~\cite{Cao:2023:HAS}}    & COCO              & 3.9 & 2.0 & 0.0 & 0.9 & 18.3 \\
    {MOD-UV}~\cite{Sun:2024:MLM}    & Waymo             & 25.1 & 11.1 & \bfseries 4.5 & 15.6 & 36.3 \\
    \midrule
    \rowcolor{tud0c!20} {\MethodName \textit{(Ours)}}   & Cityscapes        & \bfseries 30.5 & \bfseries 12.4 & 2.6 & \bfseries 21.2 & \bfseries 45.3 \\
	\bottomrule
\end{tabularx}

    \vspace{-0.5em}
\end{table}

\inparagraph{Unsupervised instance segmentation.} Our unsupervised panoptic predictions include object instances, which we benchmark against current methods in class-agnostic instance segmentation.
\Cref{tab:objectdetectionresults} provides evaluation results on the Waymo dataset.
Our approach achieves excellent results and outperforms prior work.
Compared to the state of the art, MOD-UV~\cite{Sun:2024:MLM}, which directly trains on Waymo and does not provide panoptic segmentation, \MethodName outperforms it in terms of $\text{AP}_{50}$ and overall AP.
In more detail, our method shows stronger detection accuracy for large and medium instance sizes than MOD-UV, but performs slightly worse on small instances. 
Nevertheless, \MethodName achieves a state-of-the-art level of instance segmentation accuracy despite this being a side task in the overall framework.

\subsection{Analyzing \MethodName}

\textbf{\MethodName pseudo-label generation.} In \cref{tab:pseudolabel_ablation}, we analyze the contribution of individual pseudo-label generation sub-steps by gradually increasing the complexity. We start by simply combining the unsupervised semantic predictions and unsupervised instance predictions. As described in \cref{subsec:pseudo_label}, we add the alignment of the semantic pseudo IDs based on the instance masks (1c), our ensemble-based SF2SE3 extension (1a), and depth-guided semantic inference (1b). Every component contributes to the PQ of the labels. We also analyze the thing-stuff threshold $\psi^{\rm ts}$ in \cref{tab:thing_stuff_ablation} and observe highly stable behavior for different thresholds. The pseudo-label analysis is conducted on pseudo labels generated on Cityscapes val for better comparison.

\begin{table}[t]
    \centering
    \caption{\textbf{Pseudo-label generation ablation}, analyzing the contribution of individual generation components, using PQ, SQ, and RQ (in \%, $\uparrow$) for pseudo labels generated on Cityscapes val.\label{tab:pseudolabel_ablation}}
    \vspace{-0.3em}
    \footnotesize\sisetup{table-number-alignment=center}
\setlength{\tabcolsep}{5pt}
\renewcommand{\arraystretch}{0.885}
\begin{tabularx}{\columnwidth}{>{\hspace{-\tabcolsep}\raggedright\columncolor{white}[\tabcolsep][\tabcolsep]}XZZZ}
	\toprule
    \textbf{Pseudo-label configuration} & \textbf{PQ} & \textbf{SQ} & \textbf{RQ} \\
    \midrule
    Vanilla semantics + SF2SE3               & 14.3	& 35.5	& 17.8 \\
    $\;$+ Instance-aligned semantics (1c)                    & 14.9	& 43.8	& 18.2 \\
    $\;$+ SF2SE3-ensembling (1a)                           & 15.9	& 47.0	& 19.5 \\
    \rowcolor{tud0c!20} $\;$+ Depth-guided semantic inference (1b)   & \bfseries 18.1	& \bfseries 47.3	& \bfseries 22.6 \\
	\bottomrule
\end{tabularx}

    \vspace{-0.3em}
\end{table}

\begin{table}[t]
    \centering
    \caption{\textbf{Pseudo label thing-stuff threshold analysis} showing the impact of $\psi^{\rm ts}$ using PQ (in \%, $\uparrow$) on Cityscapes val pseudo labels. \label{tab:thing_stuff_ablation}}
    \vspace{-1.4em}
    \footnotesize\sisetup{table-number-alignment=center}
\setlength{\tabcolsep}{9.0pt}
\renewcommand{\arraystretch}{0.885}
\begin{tabularx}{\columnwidth}{lccccccc}
	\toprule
    $\mathbf{\psi^{\rm ts}}\;\to$ & 0.04 & 0.06 & \cellcolor{tud0c!20} 0.08 & 0.1 & 0.12 & 0.14 \\
    \midrule
    \bfseries PQ & 17.8  & \bfseries 18.1  & \bfseries 18.1 & \bfseries 18.1  & 17.7 & 16.7 \\
	\bottomrule
\end{tabularx}

    \vspace{-0.3em}
\end{table}

\begin{table}[t]
    \centering
    \caption{\textbf{\MethodName training analysis.} We study the role of \emph{\subref{tab:pseudo_label_train}} the pseudo labels, \emph{\subref{tab:overclustering}} the number of pseudo classes, and \emph{\subref{tab:training_components}} training components on the model's accuracy trained on Cityscapes and validated on Cityscapes val. We also ablate \emph{\subref{tab:kitti}} the training dataset by training on KITTI-raw and validating on KITTI panoptic. \label{tab:train_ablation}}
    \vspace{-0.4em}
    \renewcommand{\arraystretch}{0.885}
    \begin{minipage}{0.5\columnwidth}
        \vspace{-8.2pt}
        \subcaption{\textbf{Pseudo-label training analysis}\label{tab:pseudo_label_train}}
        \footnotesize
        \setlength{\tabcolsep}{4.41pt}
        \begin{tabularx}{\columnwidth}{>{\hspace{-\tabcolsep}\raggedright\columncolor{white}[\tabcolsep][\tabcolsep]}lZ}
        	\toprule
            {\textbf{Pseudo-label configuration}} & {\textbf{PQ}} \\
            \midrule
            Vanilla pseudo labels & 19.7 \\
            \rowcolor{tud0c!20} \MethodName pseudo labels & \bfseries 27.8 \\
        	\bottomrule
        \end{tabularx}%
    \end{minipage}%
    \hfill%
    \begin{minipage}{0.44\columnwidth}
        \subcaption{\textbf{Overclustering analysis}\label{tab:overclustering}}
        \footnotesize
        \setlength{\tabcolsep}{1.875pt}
        \begin{tabularx}{\columnwidth}{>{\hspace{-\tabcolsep}\raggedright\columncolor{white}[\tabcolsep][\tabcolsep]}lZZZ}
        	\toprule
            {\textbf{Pseudo classes}} & {\textbf{PQ}} & {\textbf{SQ}} & {\textbf{RQ}} \\
            \midrule
            \rowcolor{tud0c!20} 27 \textit{(default)} & 27.8 & 57.4 & 35.2 \\
            40 & 30.3 & 64.3 & 37.5 \\
            54 & \bfseries 30.6 & \bfseries 65.1 & \bfseries 37.8 \\
        	\bottomrule
        \end{tabularx}
    \end{minipage}
    \begin{minipage}{0.5\columnwidth}
        \vspace{-5pt}
        \subcaption{\textbf{Training components ablation}\label{tab:training_components}}
        \footnotesize
        \setlength{\tabcolsep}{5.75pt}
        \begin{tabularx}{\columnwidth}{>{\hspace{-\tabcolsep}\raggedright\columncolor{white}[\tabcolsep][\tabcolsep]}lZ}
            \toprule
            {\textbf{Training configuration}} & {\textbf{PQ}} \\
            \midrule
            Vanilla training & 24.1 \\
            $\;$+ DropLoss & 25.3 \\
            $\;$+ Copy-paste aug. & 26.3 \\
            $\;$+ Self-enhance copy-paste &  26.6 \\
            \rowcolor{tud0c!20} $\;$+ Self-training (\MethodName) & \bfseries 27.8 \\
            \bottomrule
        \end{tabularx}%
    \end{minipage}%
    \hfill%
    \begin{minipage}{0.44\columnwidth}
        \vspace{3.65pt}
        \subcaption{\textbf{Training dataset analysis}\label{tab:kitti}}
        \footnotesize
        \setlength{\tabcolsep}{2pt}
        \begin{tabularx}{\linewidth}{>{\hspace{-\tabcolsep}\raggedright\columncolor{white}[\tabcolsep][\tabcolsep]}lYZ}
            \toprule
             & {\textbf{Pseudo}} & \\
            \multirow{-2}{*}{\textbf{Method}} & {\textbf{classes}} & {\multirow{-2}{*}{\textbf{PQ}}} \\
            \midrule
            {DepthG+CutLER} & \hphantom{8}27 & 14.3  \\
            {U2Seg} & 827 & 20.6 \\
            \rowcolor{tud0c!20} {\MethodName (on KITTI)$\!\!\,$} & \hphantom{8}27 & \bfseries 22.0\\
            \bottomrule
        \end{tabularx}%
    \end{minipage}%
\vspace{-0.5em}
\end{table}

\inparagraph{\MethodName training analysis.} We analyze the contribution of different components and design choices of the \MethodName training in \cref{tab:train_ablation}. Building on \cref{tab:pseudolabel_ablation}, we show the significant performance metric differences between training on the simplest form of pseudo labels and our \MethodName pseudo labels in \cref{tab:pseudo_label_train}. \Cref{tab:training_components} reveals that each training component and stage complements the final panoptic quality. \Cref{tab:overclustering} demonstrates that increasing the number of pseudo classes for pseudo-label generation and training substantially improves unsupervised panoptic segmentation performance.
Finally, we train on KITTI without making any adjustments to our method and hyperparameters, see \cref{tab:kitti}. Our approach yields significantly better results than both the baseline and U2Seg \wrt the panoptic quality. Although the PQ is slightly inferior to that of the model trained on Cityscapes---likely due to the lower resolution and diversity of KITTI---we observe that \MethodName performs well, regardless of the training data.

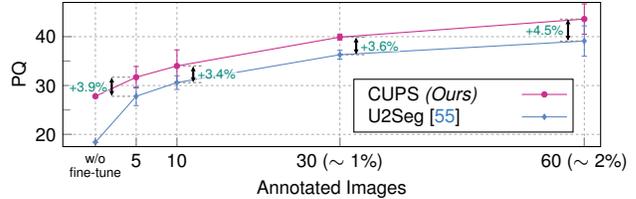
\begin{figure}[t]
    \centering
    \begin{tikzpicture}[clip, every node/.style={font=\sffamily\fontsize{7}{2}}, >={Stealth[inset=0pt,length=2pt,angle'=45]}]
    \tikzset{every picture/.style={/utils/exec={\sffamily\fontsize{7}{2}}}}
    \tikzset{every major tick/.append style={line width=.5pt, major tick length=3.5pt, gray!85}}
	\begin{axis}[
        enlargelimits=false,
        clip=true,
        height=3.5cm,
        xlabel shift=-3.65pt,
        width=0.5\textwidth,
        grid=both,
        xtick pos=bottom,
        ytick pos=left,
        grid style={line width=.1pt, draw=white, dash pattern=on 1pt off 1pt},
        major grid style={line width=.2pt,draw=gray!65},
        minor tick num=1,
        xmin=-4,
        xmax=62,
        ylabel shift=-1.5pt,
        xtick={
            0, 5, 10, 30, 60
        },
        xticklabels={
            {\tiny \shortstack{w/o\\[-1.5pt]fine-tune}}, 5\vphantom{/}, 10\vphantom{/}, 30 {($\sim$ 1\%)}, 60 {($\sim$ 2\%)}
        },
        ticklabel style = {font=\fontsize{7}{2}},
        ylabel=PQ,
        xlabel=Annotated Images,
        ymin=17.5,
        ymax=47,
        ytick={20, 30, 40},
        yticklabels={20, 30, 40},
        ]
    	\addplot[color=tud10a, mark=*, mark size=1.0pt, error bars/.cd, y dir=both, y explicit] table[x=n,y=pq, y error=pqstd] {cups.dat};\label{pgfplots:cups};
        \addplot[color=tud1a, mark=diamond*, mark size=1.0pt, error bars/.cd, y dir=both, y explicit] table[x=n,y=pq, y error=pqstd] {u2seg.dat};\label{pgfplots:u2seg};
        
        \draw[<->, thick] (axis cs:58, 39.1) -- (axis cs:58, 43.6) node[midway, left=1pt, fill=white, fill opacity=0.65, text opacity=1.0, font=\sffamily\tiny, inner sep=0.3pt] {\dlt{+4.5\%}};
        \draw[dashed, gray, dash pattern=on 1pt off 1pt] (axis cs:57, 39.1) -- (axis cs:60, 39.1);
        \draw[dashed, gray, dash pattern=on 1pt off 1pt] (axis cs:57, 43.6) -- (axis cs:60, 43.6);

        \draw[<->, thick] (axis cs:32, 36.3) -- (axis cs:32, 39.9) node[midway, right=1pt, fill=white, fill opacity=0.65, text opacity=1.0, font=\sffamily\tiny, inner sep=0.3pt] {\dlt{+3.6\%}};
        \draw[dashed, gray, dash pattern=on 1pt off 1pt] (axis cs:33, 36.3) -- (axis cs:30, 36.3);
        \draw[dashed, gray, dash pattern=on 1pt off 1pt] (axis cs:33, 39.9) -- (axis cs:30, 39.9);

        \draw[<->, thick] (axis cs:12, 30.6) -- (axis cs:12, 34.0) node[midway, right=1pt, fill=white, fill opacity=0.65, text opacity=1.0, font=\sffamily\tiny, inner sep=0.3pt] {\dlt{+3.4\%}};
        \draw[dashed, gray, dash pattern=on 1pt off 1pt] (axis cs:13, 30.6) -- (axis cs:10, 30.6);
        \draw[dashed, gray, dash pattern=on 1pt off 1pt] (axis cs:13, 34.0) -- (axis cs:10, 34.0);

        \draw[<->, thick] (axis cs:2, 27.8) -- (axis cs:2, 31.7) node[midway, left=1pt, fill=white, fill opacity=0.65, text opacity=1.0, font=\sffamily\tiny, inner sep=0.3pt] {\dlt{+3.9\%}};
        \draw[dashed, gray, dash pattern=on 1pt off 1pt] (axis cs:2, 27.8) -- (axis cs:5, 27.8);
        \draw[dashed, gray, dash pattern=on 1pt off 1pt] (axis cs:2, 31.7) -- (axis cs:5, 31.7);

	\end{axis}
    \node[draw,fill=white, inner sep=0.5pt, anchor=center] at (0.305\textwidth, 0.545) {\renewcommand{\arraystretch}{0.875}\setlength{\tabcolsep}{2.0pt}\fontsize{7}{2}
    \begin{tabular}{
    p{1.9cm}p{0.6cm}}
    \sffamily\MethodName \textit{(Ours)} & \ref*{pgfplots:cups} \\[-2pt]
    \sffamily U2Seg~\cite{Niu:2024:UUI} & \ref*{pgfplots:u2seg} \\
    \end{tabular}};
\end{tikzpicture}
    \vspace*{-1.8em}
    \caption{\textbf{Label-efficient learning results.} We fine-tune \MethodName and U2Seg on reduced amounts of annotated Cityscapes training images and compute PQ on Cityscapes val (in \%, $\uparrow$). We report the average and standard deviation across three different subsets. \label{fig:label_efficient_learning}}
    \vspace{-0.5em}
\end{figure}

\subsection{Label-efficient learning \label{subsec:label_eff_learning}} 
Ultimately, achieving high-quality, task-specific panoptic segmentation requires alignment with the desired taxonomy, which cannot be accomplished in a purely unsupervised fashion.
However, pre-training for unsupervised panoptic segmentation could be a promising modus operandi for label-efficient learning, where a minimal amount of labeled data is available to address the desired application.
Here, we explore this scenario by allocating a small share of annotated Cityscapes training images.
We fine-tune the segmentation heads of \MethodName and U2Seg, while keeping their backbones and feature pyramid networks frozen.
\cref{fig:label_efficient_learning} reports the average PQ for a varying amount of annotated training data.
Each data point averages the PQ across three runs with a random non-overlapping training subset of fixed size.
We observe that \MethodName scales reliably across all data regimes, maintaining a roughly constant and significant margin over U2Seg. Notably, using solely 60 annotated images, \ie, \qty{2}{\percent} of the total Cityscapes labels, achieves a PQ of \qty{43.6}{\percent} which amounts to \qty{70}{\percent} of the panoptic quality of the supervised upper bound.
This experiment suggests that \MethodName not only achieves outstanding accuracy in the unsupervised setting but also reduces the annotation effort in downstream tasks.

\section{Conclusion\label{sec:conclusion}}

We presented \MethodName, the first scene-centric unsupervised panoptic segmentation framework that trains directly on rich scene-centric images.
Integrating visual, depth, and motion cues, \MethodName overcomes the dependence on object-centric training data and achieves significant improvements on challenging scene-centric datasets where prior methods struggle. Our approach brings the quality of unsupervised panoptic, instance, and semantic segmentation to a new level and demonstrates highly promising results in label-efficient panoptic segmentation.

{\small \inparagraph{Acknowledgments.} This project was partially supported by the European Research Council (ERC) Advanced Grant SIMULACRON, DFG project CR 250/26-1 ``4D-YouTube'', and GNI Project ``AICC''. This project has also received funding from the ERC under the European Union’s Horizon 2020 research and innovation programme (grant agreement No.\ 866008). Additionally, this work has further been co-funded by the LOEWE initiative (Hesse, Germany) within the emergenCITY center [LOEWE/1/12/519/03/05.001(0016)/72] and by the State of Hesse through the cluster project ``The Adaptive Mind (TAM)''. Christoph Reich is supported by the Konrad Zuse School of Excellence in Learning and Intelligent Systems (ELIZA) through the DAAD programme Konrad Zuse Schools of Excellence in Artificial Intelligence, sponsored by the Federal Ministry of Education and Research. Finally, we acknowledge the support of the European Laboratory for Learning and Intelligent Systems (ELLIS) and thank Simone Schaub-Meyer and Leonhard Sommer for insightful discussions.}

{
    \small
    \bibliographystyle{ieeenat_fullname}
    \bibliography{bibtex/short, bibtex/new, bibtex/papers, bibtex/external}
}

\clearpage
\setcounter{section}{0}
\renewcommand\thesection{\Alph{section}}
\setcounter{page}{1}
\pagenumbering{roman}
\twocolumn[{%
\renewcommand\twocolumn[1][]{#1}%
\maketitlesupplementary
{\large Oliver Hahn\textsuperscript{\normalfont{}* 1}
\authorstep Christoph Reich\textsuperscript{\normalfont{}* 1,2,4,5}
\authorstep Nikita Araslanov\textsuperscript{\normalfont{} 2,4}\\[-3pt]
Daniel Cremers\textsuperscript{\normalfont{} 2,4,5}
\authorstep Christian Rupprecht\textsuperscript{\normalfont{} 3}
\authorstep Stefan Roth\textsuperscript{\normalfont{} 1,5,6}}\\[2pt]
\small{\textsuperscript{1}TU Darmstadt\affiliationstep \textsuperscript{2}TU Munich \affiliationstep \textsuperscript{3}University of Oxford \affiliationstep \textsuperscript{4}MCML\affiliationstep \textsuperscript{5}ELIZA\affiliationstep \textsuperscript{6}hessian.AI\affiliationstep
\textsuperscript{*}equal contribution}\\[1pt]\small {\url{https://visinf.github.io/cups}}
\vspace{0.75cm}
}]

In this appendix, we first highlight the conceptual features of our unsupervised panoptic segmentation method \MethodName{}. We elaborate on the training and validation approach as well as on implementation details to facilitate reproducibility. Next, we conduct further analyses of different design choices and the training stages. We then provide additional quantitative and qualitative results. Finally, we discuss the limitations of current unsupervised panoptic approaches as well as our \MethodName approach.

\section{\MethodName \textit{vs.}\! U2Seg:\! A Conceptual Comparison}
\Cref{tab:conceptual_comparison} conceptually compares \MethodName to U2Seg~\cite{Niu:2024:UUI}.
While both frameworks address the problem of unsupervised panoptic segmentation, \MethodName features novel distinctions:

\noindent \textit{(1) Scene-centric training.}
Object-centric images typically depict a center-aligned foreground object on a fairly homogeneous background.
The photographic bias inherent to this type of imagery also implies the need for manual curation in the data collection process. 
By contrast, scene-centric data encapsulates the complexity of real-world environments where multiple objects coexist and interact.
Furthermore, collecting scene-centric imagery is substantially cheaper, since it obviates the need for artificially isolating objects from their context. %
Training on scene-centric data is crucial to producing models that are capable of understanding real-world complexity and serving the needs of challenging applications, such as autonomous driving, robotic navigation, augmented reality, and assistive technologies for visually impaired individuals.
Although we are not the first to leverage motion for retrieving instance cues, accomplishing this in a self-supervised fashion is a novel aspect in the context of unsupervised panoptic segmentation.

\noindent \textit{(2) High-resolution pseudo labels.}
High-resolution training is crucial for capturing fine details in scene-centric data, which lower-resolution settings cannot address.
Our depth-guided semantic inference (\cf \cref{subsec:pseudo_label}) provides a semantic pseudo-labeling component with twice the resolution of previous methods. %
This enhancement allows \MethodName to learn semantic cues to a higher degree of detail, which can be observed in our qualitative results (\cf \cref{fig:panotic_qualitative}).

\noindent \textit{(3) Thing-stuff separation.}
Our integration of motion cues enables a precise distinction between semantic pseudo ``thing'' and pseudo ``stuff'' classes. 
This is because motion helps us identify ``thing'' classes as objects that move relative to the camera. %
In contrast, U2Seg cannot really distinguish between ``stuff'' and ``thing'' classes; this ambiguity is only resolved at test time via oracle matching of the pseudo labels with ground-truth semantic categories and object instances.
The capacity of \MethodName to discriminate between ``stuff'' and ``thing'' categories is an advancement toward solving unsupervised panoptic segmentation in a more principled way.

\begin{table}[t]
\footnotesize
\setlength{\tabcolsep}{5pt}
\renewcommand{\arraystretch}{0.885}
\caption{\textbf{A conceptual comparison of \MethodName and U2Seg.}\label{tab:conceptual_comparison}}
\vspace{-0.5em}
\begin{tabularx}{\linewidth}{>{\hspace{-\tabcolsep}\raggedright\columncolor{white}[\tabcolsep][\tabcolsep]}Xcc@{}}
    \toprule
    & \textbf{U2Seg~\cite{Niu:2024:UUI}} & \textbf{\MethodName} \textit{(Ours)}  \\[2pt]
    \midrule
    Unsupervised panoptic segmentation  & \dlt{\cmark} & \dlt{\cmark} \\
    Scene-centric training              & \color{tud9c}\xmark & \dlt{\cmark} \\
    High-resolution pseudo labels       & \color{tud9c}\xmark & \dlt{\cmark} \\
    Thing-stuff separation              & \color{tud8a}$\sim$ & \dlt{\cmark}  \\
    \bottomrule
\end{tabularx}
\vspace{-0.5em}
\end{table}

\section{Reproducibility}
To facilitate reproducibility, we elaborate on the technical and implementation details. 
Note that our code is available at\, \url{https://github.com/visinf/cups}. 

\subsection{Implementation details}

\MethodName is implemented using PyTorch~\cite{pytorch}, PyTorch Lightning~\cite{pytorchl}, and Kornia~\cite{kornia}. We partly build upon public codebases from previous work \cite{Stone:2021:SST, Hamilton:2022:USS, Sommer:2022:SCS, Sick:2024:USS, Niu:2024:UUI}.

\inparagraph{Stage 1.} \MethodName pseudo-label generation uses 27 pseudo classes and a thing-stuff threshold $\psi^{\rm ts}$ of \num{0.08}. This setting enables comparison against existing unsupervised panoptic and semantic segmentation approaches without relying on significant overclustering (\cf \cite{Niu:2024:UUI}). Instance pseudo labeling uses motion and depth estimates from a pre-trained SMURF model~\cite{Stone:2021:SST}. For our ensembling-based SF2SE3 clustering, we build upon the original implementation by Sommer \etal \cite{Sommer:2022:SCS}. Semantic pseudo labeling uses a pre-trained SMURF~\cite{Stone:2021:SST} to generate the depth to train the semantic segmentation network following Sick \etal \cite{Sick:2024:USS}. For depth-guided semantic inference, we first resize the input image so that its smaller side is 320 pixels, matching the standard resolution in unsupervised semantic segmentation. We then perform a second inference pass using sliding windows on an image scale of 640 pixels with a stride of half the window size. Depth-guided semantic inference uses the SMURF depth estimate to weight the two semantic segmentation predictions. The size of the sliding window is half the image width and height. Finally, we perform post-processing by further aligning the prediction to the image using a fully connected conditional random field (CRF)~\cite{Kraehenbuehl:2011:EIF, Gansbeke_2021_USS}. 

For a fair comparison and to demonstrate the impact of our pseudo labeling as well as the proposed training scheme, we use the same panoptic network as U2Seg. In particular, we follow Niu \etal \cite{Niu:2024:UUI} by employing the Panoptic Cascade Mask R-CNN~\cite{Cai:2018:CRC, Kirillov:2019:PFP} with a ResNet-50~\cite{He:2016:DRL} backbone pre-trained using self-supervised DINO~\cite{Caron:2021:EPS} for two epochs on ImageNet~\cite{imagenet}.

\inparagraph{Stage 2.} \MethodName pseudo-label bootstrapping proceeds by training for \num{4000} steps with AdamW~\cite{Loshchilov:2018:ADW}, using a learning rate of \num{e-4}, and a weight decay of \num{e-5}. The drop-loss overlap threshold $\tau^{\operatorname{IoU}}$ is set to \num{0.4}.
After \num{1000} steps, we start utilizing our self-enhanced copy-paste augmentation, randomly pasting between 1 and 8 objects into each image.

\inparagraph{Stage 3.} \MethodName self-training runs for \num{1500} steps using AdamW with a learning rate of \num{e-5} and no weight decay. The EMA decay for updating the momentum network is set to \num{0.9999}. We only update the detection heads, the mask head, and the semantic head, freezing all normalization layers.
For self-labeling augmentation, we use three different scales of the original image (\num{0.75}, \num{1.0}, and \num{1.25}), as well as horizontal flips at each scale, resulting in six views. We follow Chen \etal \cite{Chen:2020:ISL} to set up the photometric augmentation and employ our self-enhanced copy-paste augmentation also during self-training.

For both pseudo-label training and self-training, we utilize four NVIDIA A100 GPUs (\SI{40}{\giga\byte}) with a batch size of 16 per GPU. We evaluate \MethodName on the native resolution of each dataset, except for unsupervised semantic segmentation (\cf \cref{tab:semanticsegresults}) where we follow the common evaluation protocol \cite{Ji:2019:IIC, Cho:2021:PUS, Hamilton:2022:USS, Sick:2024:USS}.

\subsection{Computing panoptic quality}
As we train without any supervision, the semantic pseudo-class IDs are not aligned with the ground-truth semantic class IDs.
Therefore, to compute the panoptic quality (PQ)~\cite{Kirillov:2019:PS}, we need to align the pseudo-class IDs with the ground truth, distinguishing between ``thing'' and ``stuff'' semantic categories at the same time.

While U2Seg \cite{Niu:2024:UUI} also utilizes the panoptic quality and proposes an elaborate matching approach, significant limitations remain. Niu \etal \cite{Niu:2024:UUI} establish a semantic matching using three steps. First, predicted segments are matched with all ground-truth segments, ignoring the ``thing'' and ``stuff'' separation. Segments with an overlap of less than a pre-defined threshold (hyperparameter) are discarded. Second, using the set of matched segments and both the semantic pseudo-class IDs and the ground-truth class IDs, a cost matrix is constructed on a per-segment basis. Third, for each semantic pseudo class, the most frequent ground-truth class ID based on the cost matrix is matched. This matching approach entails two significant limitations. First, the overlap threshold is a crucial hyperparameter and can significantly impact the final PQ value. This is mainly due to the fact that the segment-wise cost matrix finds relatively few overlapping objects, and thresholding is required to consider only accurate predictions for matching.
Second, the matching approach does not consider the ``thing'' and ``stuff'' separation, leading to matches between both ``thing'' and ``stuff'' categories. This is highly undesired as ``thing'' segments entail object-level masks, whereas ``stuff'' segments only capture the semantic level. Finally, code for evaluation on the Cityscapes dataset has not been published by the authors of \cite{Niu:2024:UUI}.

\inparagraph{Principles.} We redefine the matching process in alignment with the following core principles:
\emph{Simplicity:} Introducing additional hyperparameters within the matching is undesirable, as it complicates evaluation. %
Semantic segmentation is a pixel-wise classification, hence we aim to perform matching of the pseudo classes to ground-truth classes on the pixel level as well. More specifically, every predicted pixel should be considered in the alignment between pseudo classes and ground-truth annotations. This resembles the simplest form of approaching the problem and is common in unsupervised semantic segmentation \cite{Ji:2019:IIC, Cho:2021:PUS, Hamilton:2022:USS, Sick:2024:USS}.
\emph{Clear thing and stuff separation:} The distinction between ``thing'' and ``stuff'' classes is a core aspect of panoptic segmentation. Consequently, it should be addressed by the method itself rather than the matching process. To ensure alignment, only pseudo classes labeled as ``stuff'' are matched with ``stuff'' ground-truth classes, and the same applies to ``thing'' classes.

\inparagraph{Approach.} To this end, we propose a simple but effective approach for matching. Taking inspiration from the established semantic matching for the task of unsupervised semantic segmentation \cite{Ji:2019:IIC, Cho:2021:PUS, Hamilton:2022:USS, Sick:2024:USS}, we perform matching purely utilizing semantics. In particular, we obtain the semantic segmentation prediction $\bar{\mathbf{P}}\in\{1, \ldots, \xi_{\text{p}}\}^{\rm H \times \rm W}$ from the unsupervised panoptic prediction, with $\xi_{\text{p}}$ denoting the number of pseudo classes. We use the ground-truth semantic segmentation $\hat{\mathbf{P}}\in\{1, \ldots, \xi_{\text{GT}}\}^{\rm H \times \rm W}$, with $\xi_{\text{GT}}$ indicating the number of ground-truth semantic classes, to construct a cost matrix $\mathbf{A}\in\mathbb{N}^{\xi_{\text{p}}\times \xi_{\text{GT}}}$. This cost matrix counts the number of overlapping pixels of each pseudo-class ID with all ground-truth class IDs. The full cost matrix is obtained using all validation samples. To ensure no ``thing'' class ID is matched to a ``stuff'' class ID or vice versa, we extract a ``thing'' and a ``stuff'' cost matrix from the full cost matrix $\mathbf{A}$. By using the ``thing'' and ``stuff'' splits of classes in the pseudo classes as well as the ground-truth classes, we construct a ``thing'' cost matrix $\mathbf{A}^{\text{Th}}\in\mathbb{N}^{\xi_{\text{p}}^{\text{Th}}\times \xi_{\text{GT}}^{\text{Th}}}$ and ``stuff'' cost matrix $\mathbf{A}^{\text{St}}\in\mathbb{N}^{\xi_{\text{p}}^{\text{St}}\times \xi_{\text{GT}}^{\text{St}}}$. Hungarian matching \cite{Kuhn:1955:HUN} is then applied to maximize overlap and establish a one-to-one matching between pseudo-class IDs and ground-truth class IDs by running matching on $\mathbf{A}^{\text{Th}}$ and $\mathbf{A}^{\text{St}}$, separately. As we can have more semantic pseudo-class IDs than ground-truth class IDs (\ie, $\xi_{\text{p}}^{\text{Th}}>\xi_{\text{GT}}^{\text{Th}}$ and/or $\xi_{\text{p}}^{\text{St}}>\xi_{\text{GT}}^{\text{St}}$), we assign all remaining pseudo classes, not assigned by Hungarian matching, to the respective ground-truth class ID with the maximum overlap. This process leads to a permutation of the pseudo-class IDs, maximizing the overlap with the ground-truth class IDs while adhering to the ``thing'' and ``stuff'' separation. 
Finally, we utilize the permuted (\ie, matched) semantics alongside the instance mask---the binary masks predicted for instances---to compute PQ. For evaluating on the task of unsupervised semantic segmentation, we skip the step of separating $\mathbf{A}$ into $\mathbf{A}^{\text{Th}}$ and $\mathbf{A}^{\text{St}}$ and perform a single matching on $\mathbf{A}$ as done by the related work in the field \cite{Ji:2019:IIC, Cho:2021:PUS, Hamilton:2022:USS, Sick:2024:USS}. 

To conclude, our class matching for unsupervised panoptic quality builds on established protocols, performs a straightforward and efficient matching, and adheres to the ``thing'' and ``stuff'' class split, while not introducing any hyperparameters. Interestingly, we observe that evaluating U2Seg with our matching leads to better panoptic quality than reported in the original paper (\cf \cref{tab:panoptic_segmentation_cs}). This suggests that we find a better correspondence between pseudo and ground-truth classes.
We make the evaluation code for all settings publicly available to facilitate future research.

\subsection{Datasets}
We provide further details about the datasets used to train and evaluate \MethodName.

\inparagraphnospace{Cityscapes}~\cite{Cordts:2016:CDS} is an ego-centric driving scene dataset, which contains 5\,000 high-resolution images with 2048$\times$1024 pixels. It is split into \num{2975} train, 500 val, and \num{1525} test images with pixel-level annotations provided for grouping into 27, 19, or 7 categories. Each of the training images stems from a short video sequence. We leverage all \num{86275} video frames of the training split for unsupervised training and evaluate on the validation split, in line with previous work. 

\smallskip\noindent The \textbf{KITTI}~\cite{Mohan:2021:KIP, Geiger:2012:AWR} vision benchmark suite is a comprehensive driving-scene dataset with ground truth for a variety of tasks, such as semantic segmentation, optical flow estimation, depth estimation.
Mohan \etal \cite{Mohan:2021:KIP} introduced the KITTI panoptic segmentation dataset for urban scene understanding by providing panoptic annotations for a subset of \num{1055} images. The images have a resolution of 1280$\times$384 pixels and adhere to the 19-class grouping of the Cityscapes taxonomy. We use the 200 validation images for evaluation. Furthermore, we use all \num{42150} rectified KITTI images excluding the validation split and calibration scenes for unsupervised training.

\inparagraphnospace{BDD}~\cite{Yu:2020:BAD} is a driving scene dataset, which also contains panoptic annotations with 19 class definitions identical to those in Cityscapes. The images have a resolution of 1280$\times$720 pixels. The validation set contains \num{1000} images.

\inparagraphnospace{MUSES}~\cite{Brodermann:2024:MTM} is a multi-modal dataset representing adverse conditions in driving scenes. The labels use the 19 class taxonomy of Cityscapes. For evaluation, we utilize the ``daytime clear'' validation split, containing 50 images with a resolution of 1920$\times$1080.

\inparagraphnospace{Waymo}~\cite{Sun:2020:SPA} is a another driving scene dataset. We use the ``front'' camera, providing a resolution of 1920$\times$1280 pixels and evaluate using the \num{1930} images of the 2D panoptic segmentation validation split. Waymo classes are remapped to ensure compatibility of its label space with the Cityscapes classes, resulting in 16 classes.  

\inparagraphnospace{MOTS}~\cite{Voigtlaender:2019:MMO} allows to assess scene-centric panoptic segmentation outside of driving scenarios.
Evaluation is performed using the MOTChallenge sequences for multi-object tracking and segmentation of humans in indoor and outdoor scenes. The annotations include two classes ``background'' and ``person'', where ``background'' is considered as a ``stuff'' class and ``person'' is a ``thing'' class. We evaluate on \num{2862} images of resolutions 640$\times$480 or 1920$\times$1080.

\begin{table}[t!]
    \centering
    \caption{\textbf{Comparison of motion networks for pseudo-label generation.} Investigating the contribution of the correspondence matching network, using PQ, SQ, and RQ (in \%, $\uparrow$) for pseudo labels generated on Cityscapes val. We use our full configuration and only change the motion network.\label{tab:pseudolabel_ablation_flow}}
    \vspace{-0.5em}\footnotesize\sisetup{table-number-alignment=center}
\setlength{\tabcolsep}{5pt}
\renewcommand{\arraystretch}{0.885}
\begin{tabularx}{\columnwidth}{>{\hspace{-\tabcolsep}\raggedright\columncolor{white}[\tabcolsep][\tabcolsep]}XZZZ}
	\toprule
    \textbf{Optical flow method} & \textbf{PQ} & \textbf{SQ} & \textbf{RQ} \\
    \midrule
    {BrightFlow~\cite{brightflow} (unsupervised)} & 17. 8 & 46.4 & 22.4 \\
    \rowcolor{tud0c!20} {SMURF~\cite{Stone:2021:SST} (unsupervised)} & 18.1	& 47.3	& 22.6 \\
    \midrule
    {SEA-RAFT~\cite{searaft} (supervised)} & 19.2	& 51.8	& 23.4 \\
    {RAFT~\cite{Teed:2020:RAF} (supervised)} & 20.4 & 52.6 & 24.7  \\ 
	\bottomrule
\end{tabularx}

    \vspace{-0.5em}
\end{table}

\section{Additional Results}
In the following, we analyze the results presented in the main paper in greater detail.

\begin{table}[t]
\caption{\textbf{Depth-guided semantic pseudo label analysis.} Semantic pseudo labels evaluated on Cityscapes val (for consistency both in 19 class setting). \emph{\subref{tab:depthguidedsemantic_single_combined}} comparing the resolutions and merging approaches. \emph{\subref{tab:depthguidedsemantic_depth_ranges}} decomposing depth-guided semantic segmentation accuracy for different depth ranges. All metrics in \%. \label{tab:dguided_analysis}}
\vspace{-0.75em}
    \begin{minipage}{\columnwidth}
        \subcaption{\textbf{Depth-guided semantic pseudo labeling components.}\label{tab:depthguidedsemantic_single_combined}}
        \footnotesize\sisetup{table-number-alignment=center}
        \setlength{\tabcolsep}{5pt}
        \renewcommand{\arraystretch}{0.885}
        \begin{tabularx}{\columnwidth}{>{\hspace{-\tabcolsep}\raggedright\columncolor{white}[\tabcolsep][\tabcolsep]}XZZZ}
        \toprule
        \textbf{Method} & \textbf{PQ} & \textbf{SQ} & \textbf{RQ} \\
        \midrule    
        Low Resolution ($\rm P\textsuperscript{low}$)        & 15.9	& 47.0	& 19.5 \\
        High Resolution ($\rm P\textsuperscript{high}$)      & 17.9   & 46.8  & 22.4 \\
        \midrule
        Mean           & 16.7   & 42.7  & 20.9 \\
        \rowcolor{tud0c!20} Depth-guided ($\rm P^{*}$)   &  18.1	&  47.3	&  22.6 \\
        \bottomrule
        \end{tabularx}
    \end{minipage}%
\\[-2pt]
    \begin{minipage}{\columnwidth}
        \vspace{0.5em}
        \subcaption{\textbf{Analyzing different depth ranges.}\label{tab:depthguidedsemantic_depth_ranges}}
        \footnotesize\sisetup{table-number-alignment=center}
        \setlength{\tabcolsep}{5pt}
        \renewcommand{\arraystretch}{0.885}
        \begin{tabularx}{\columnwidth}{>{\hspace{-\tabcolsep}\raggedright\columncolor{white}[\tabcolsep][\tabcolsep]}XZZZZ}
        \toprule
        & \multicolumn{4}{c}{\textbf{mIoU}\textsuperscript{19}}
        \\
        Distance (in m) & \textbf{0\,--\,10} & \textbf{10\,--\,30} & \textbf{$>$30} & \textbf{all} \\
        \midrule    
        Low Resolution ($\rm P\textsuperscript{low}$)        & 30.7 & 28.7 & 23.3 & 29.5 \\
        High Resolution ($\rm P\textsuperscript{high}$)       & 28.6 & 29.6 & 27.3 & 30.9 \\
        \midrule
        \rowcolor{tud0c!20} Depth-guided ($\rm P^{*}$)    & 29.2 & 29.7 & 27.3 & 31.1 \\
        \bottomrule
        \end{tabularx}
\end{minipage}%
\vspace{-0.5em}
\end{table}

\subsection{\MethodName pseudo-labels results}

\begin{figure*}[t]
    \centering
    \small
\sffamily
\setlength{\tabcolsep}{0pt}
\renewcommand{\arraystretch}{0.0}
\begin{tabular}{>{\centering\arraybackslash} m{0.2\textwidth} 
                >{\centering\arraybackslash} m{0.2\textwidth} 
                >{\centering\arraybackslash} m{0.2\textwidth}
                >{\centering\arraybackslash} m{0.2\textwidth}
                >{\centering\arraybackslash} m{0.2\textwidth}}

{Image} & {Ground Truth} & {Low Resolution} & {High Resolution} & {Depth Guided \textit{(Ours)}} \\[4pt]

\includegraphics[width=\linewidth]{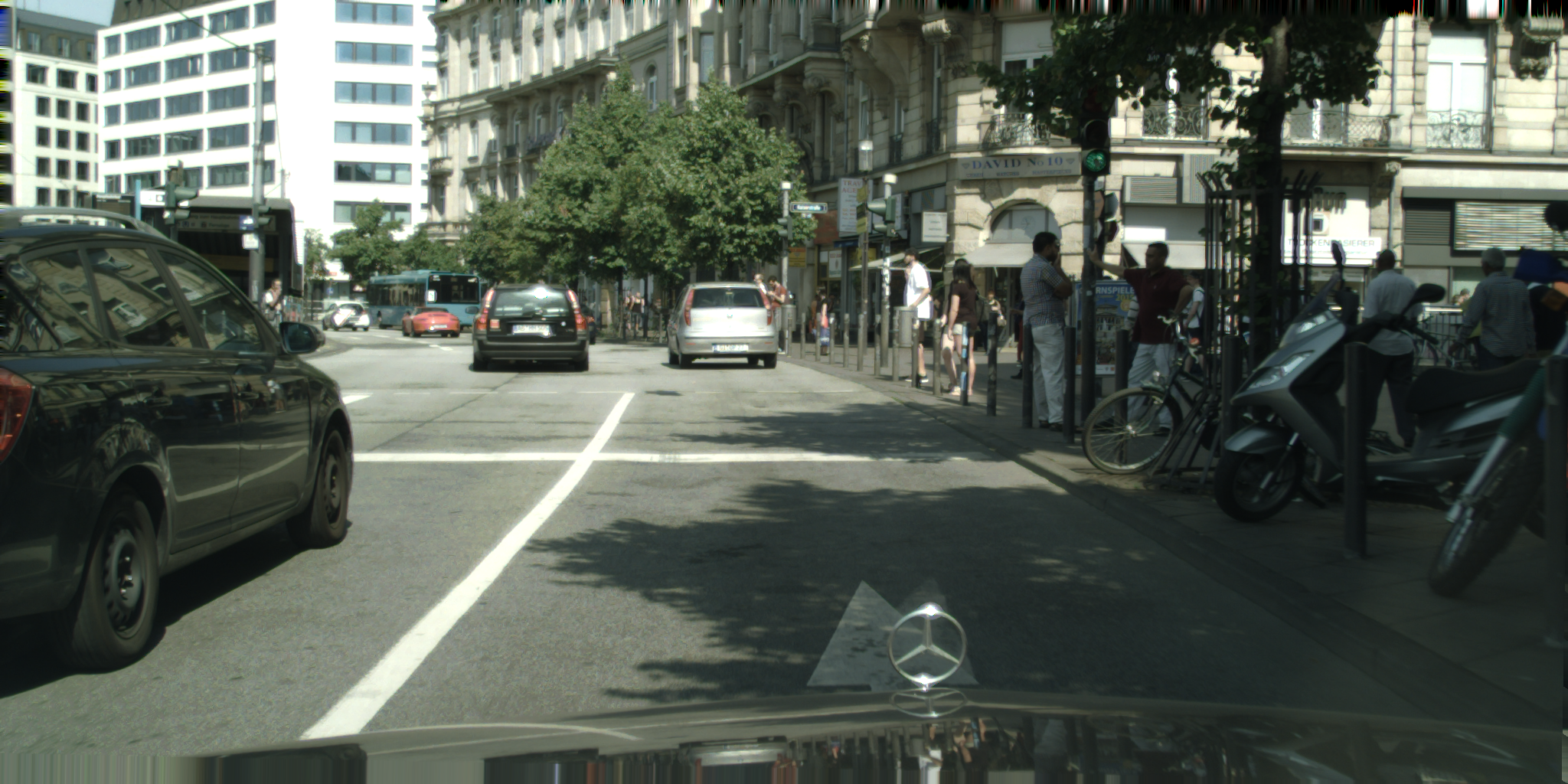} &
\includegraphics[width=\linewidth]{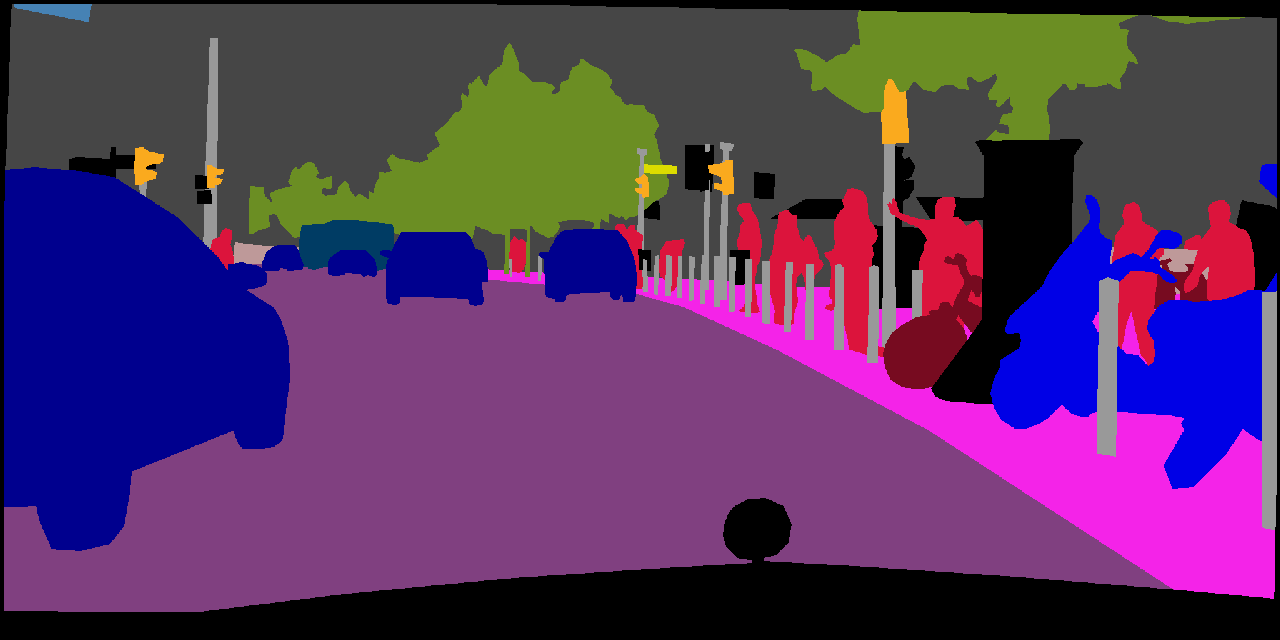} &
\includegraphics[width=\linewidth]{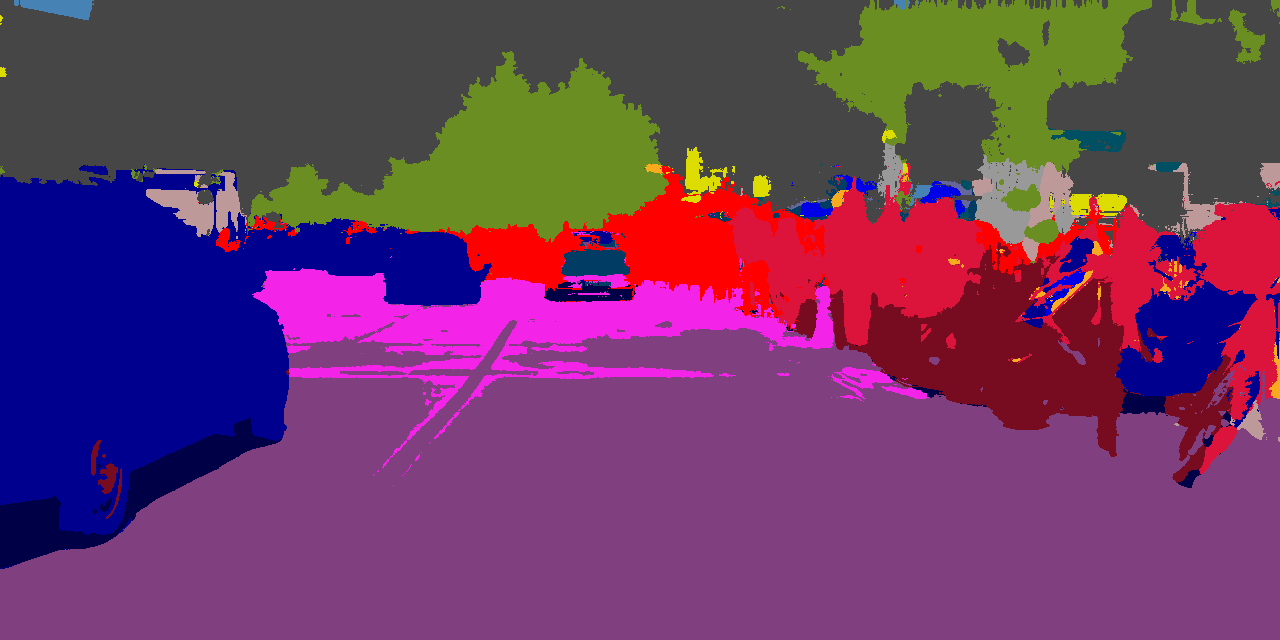} &
\includegraphics[width=\linewidth]{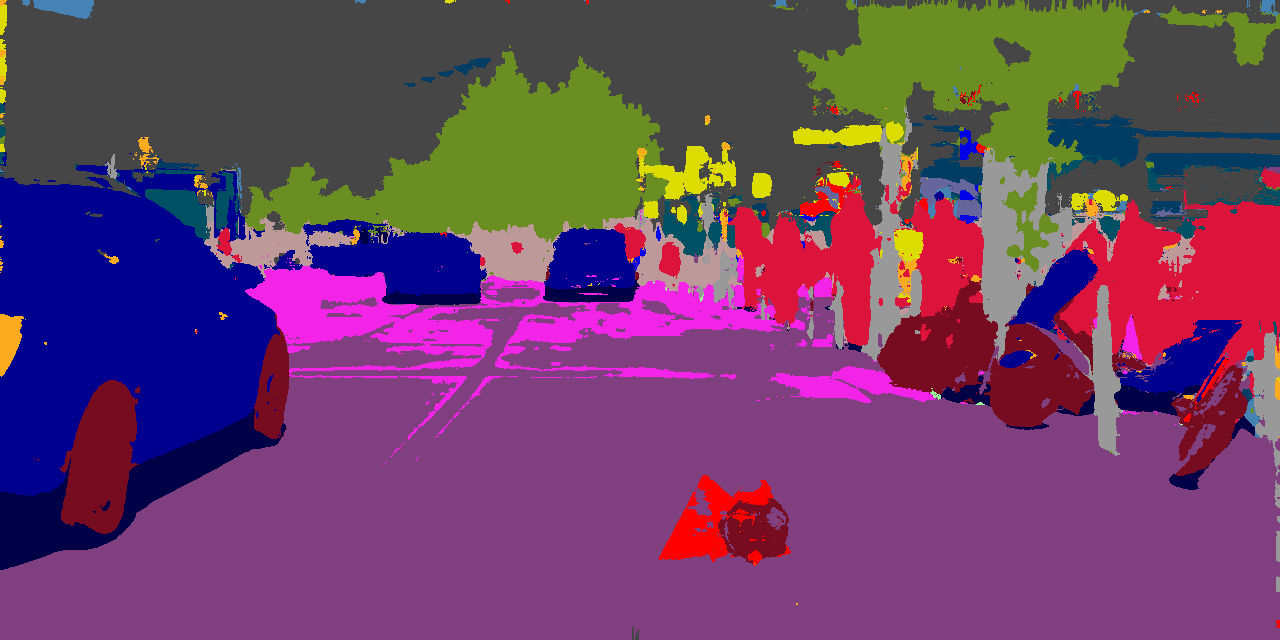} &
\includegraphics[width=\linewidth]{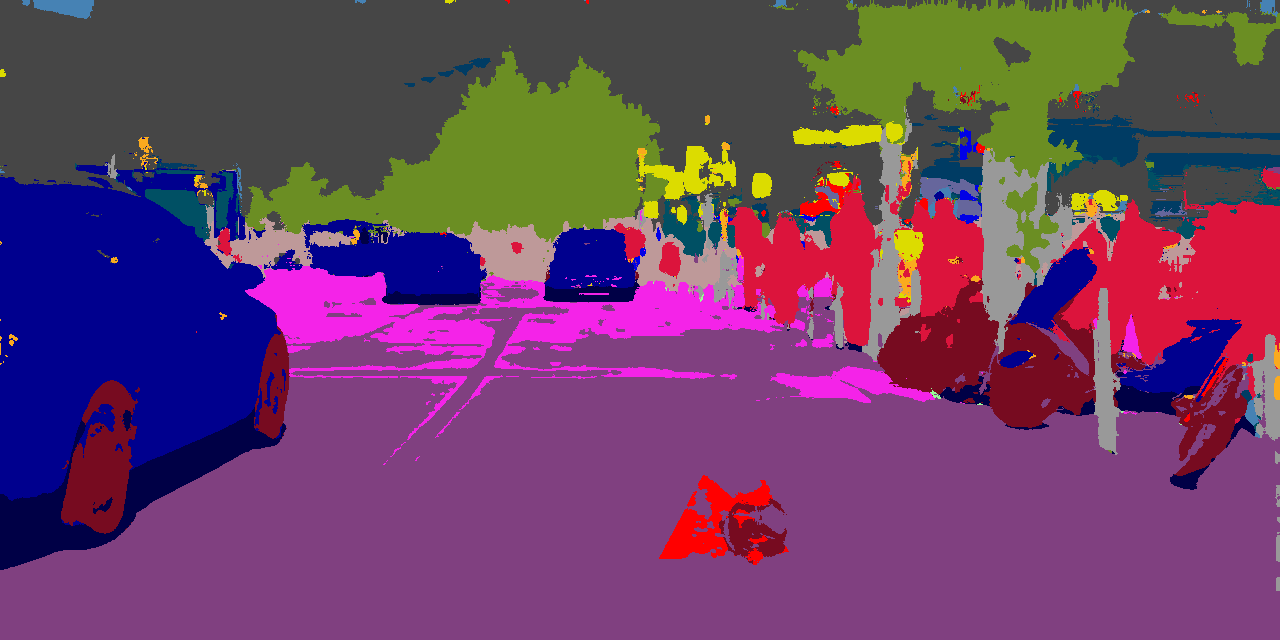} \\

\includegraphics[width=\linewidth]{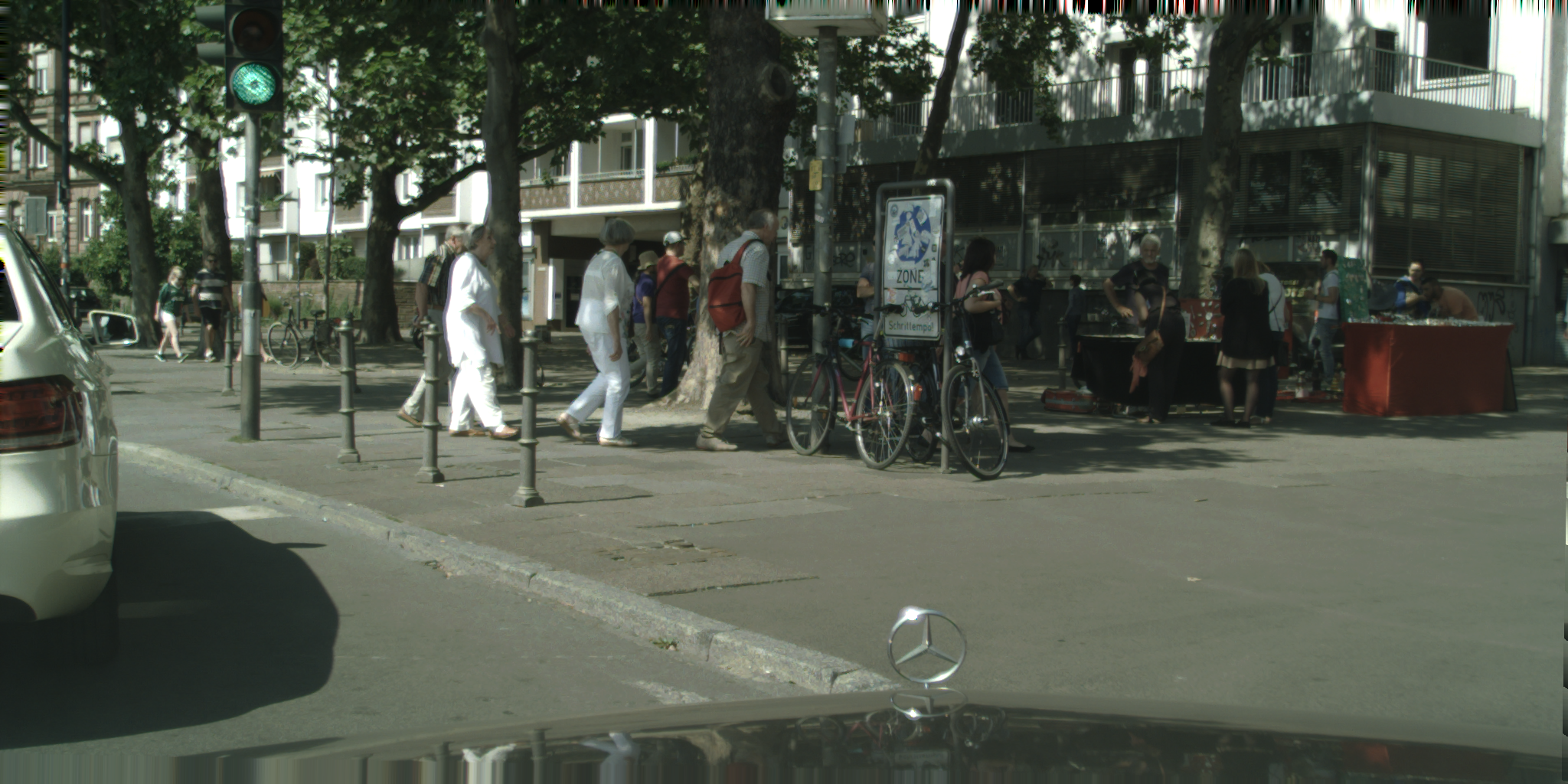} &
\includegraphics[width=\linewidth]{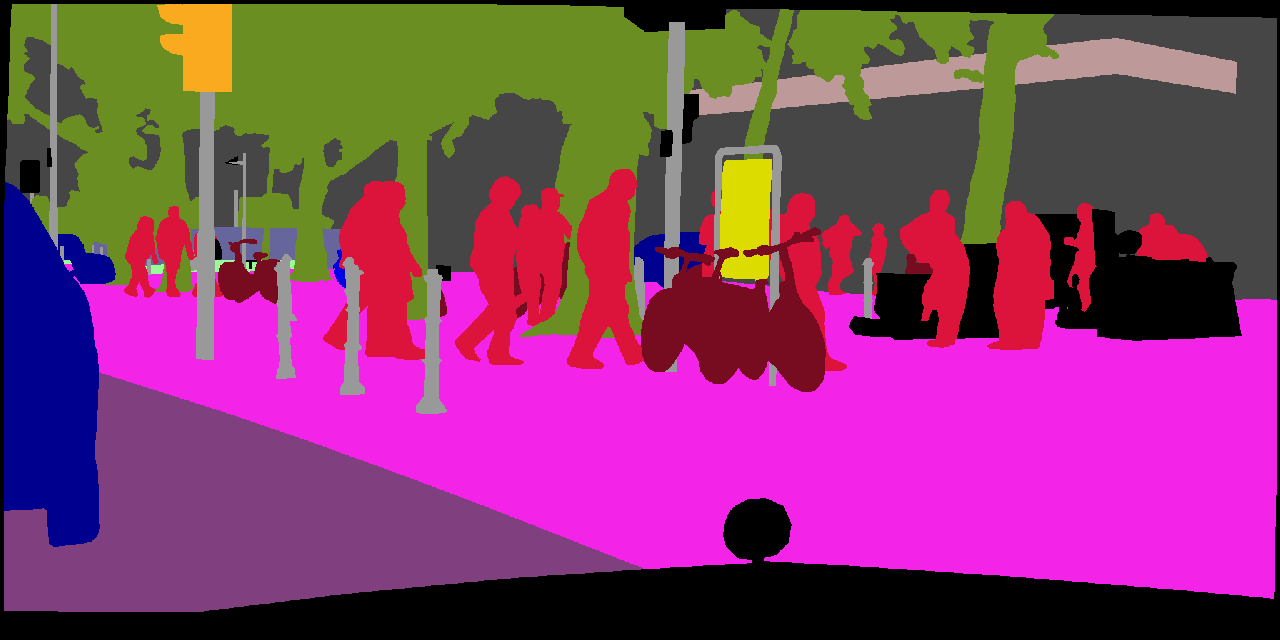} &
\includegraphics[width=\linewidth]{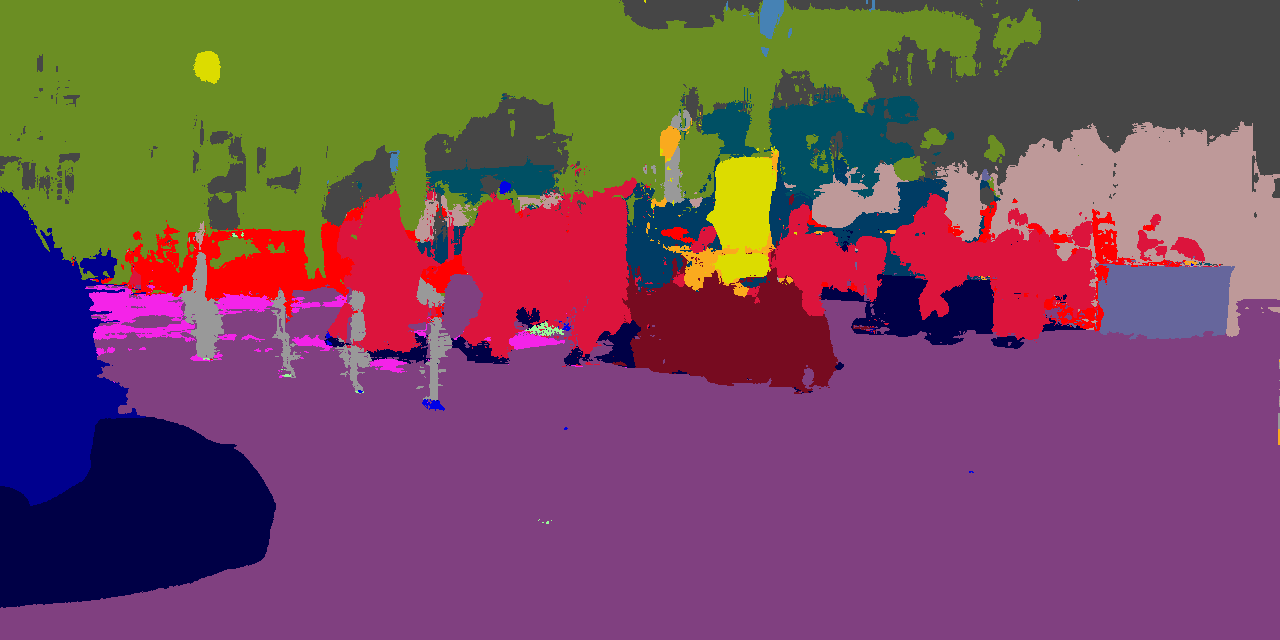} &
\includegraphics[width=\linewidth]{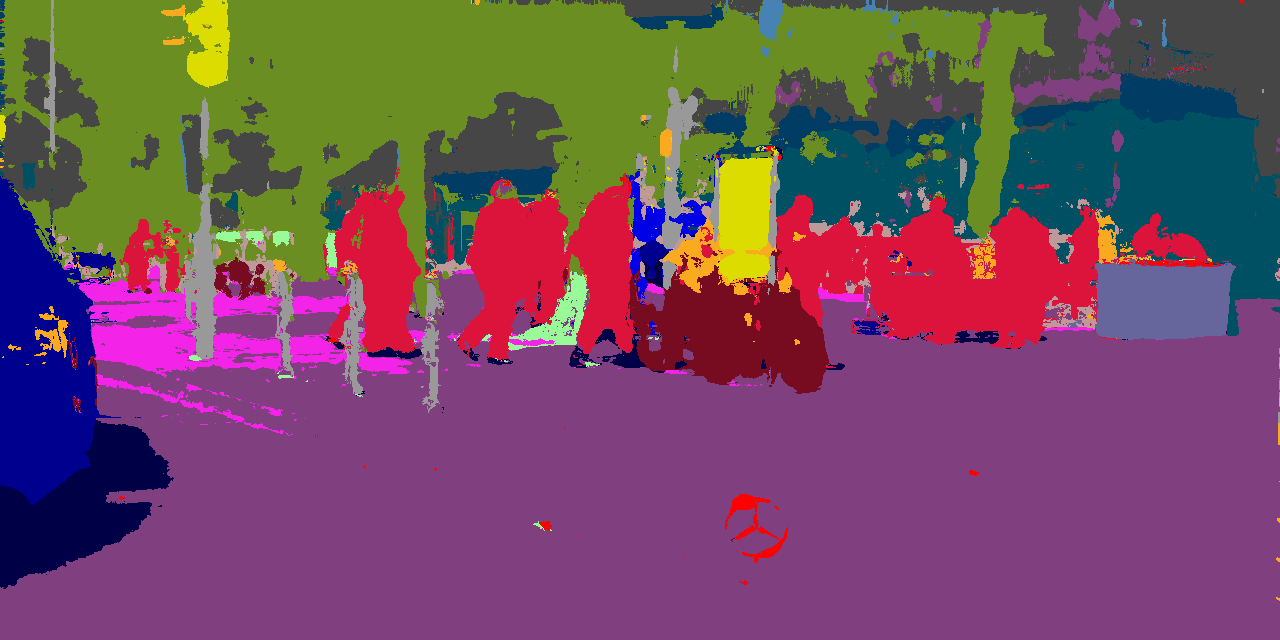} &
\includegraphics[width=\linewidth]{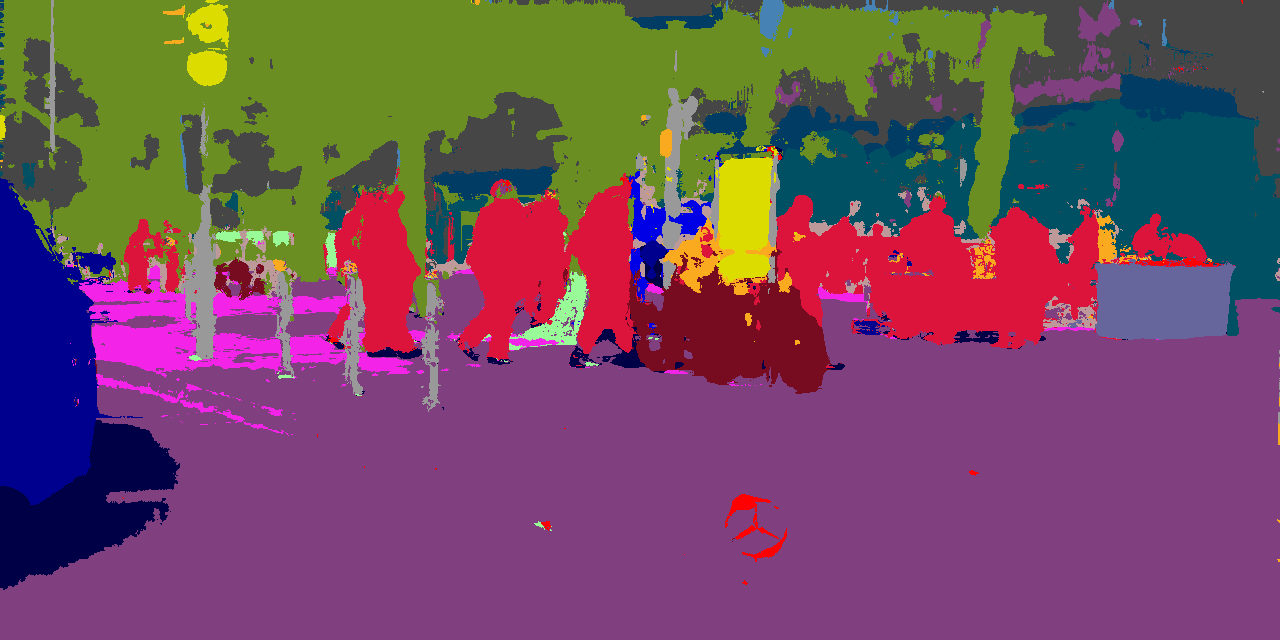} \\

\includegraphics[width=\linewidth]{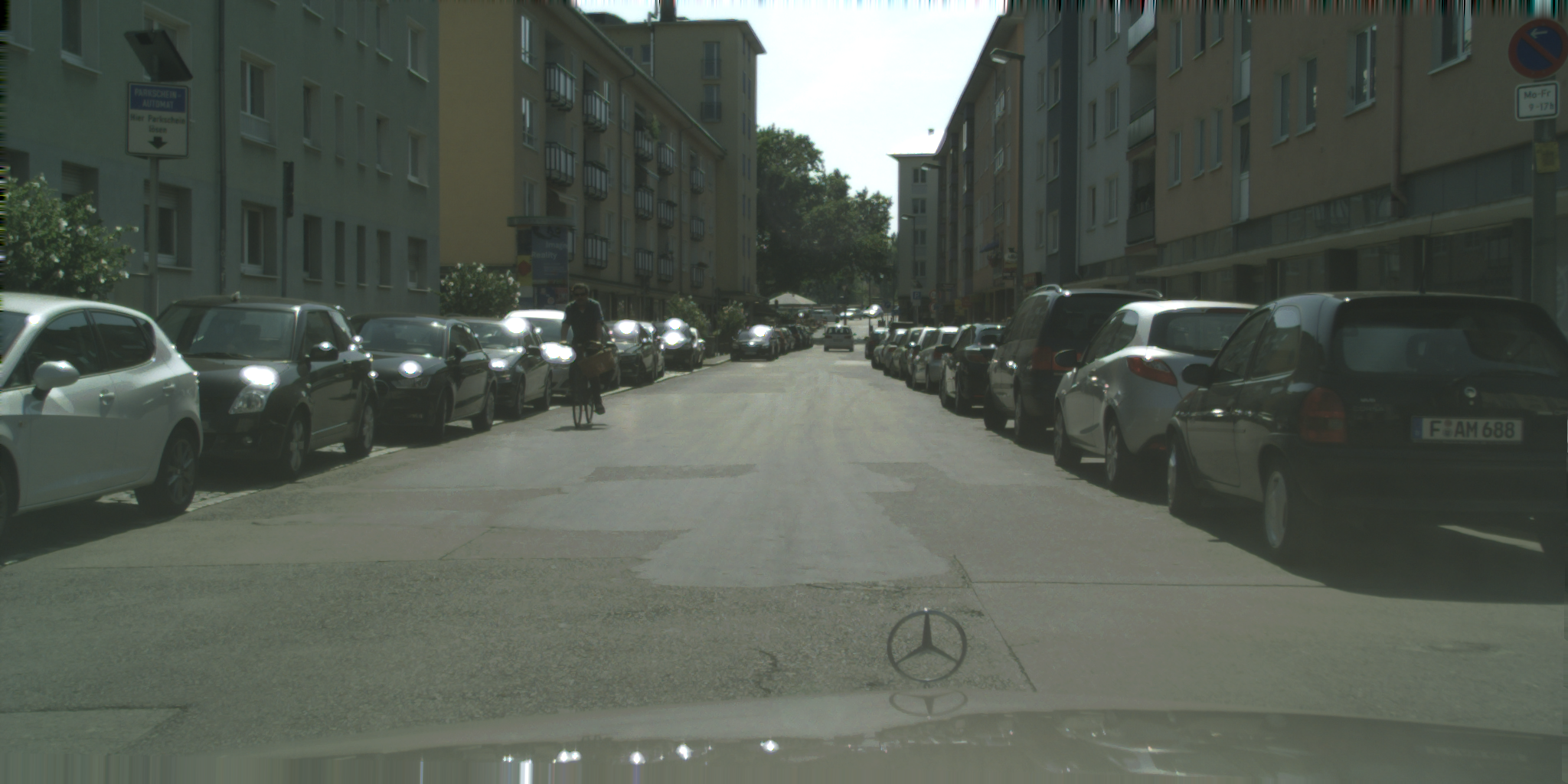} &
\includegraphics[width=\linewidth]{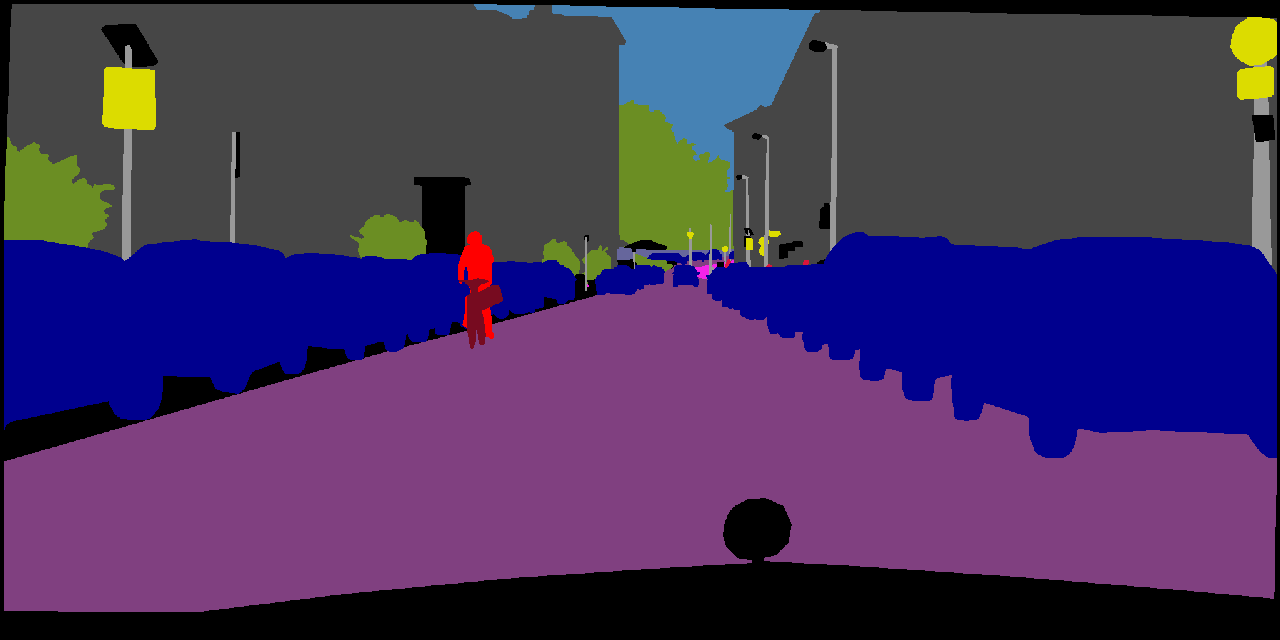} &
\includegraphics[width=\linewidth]{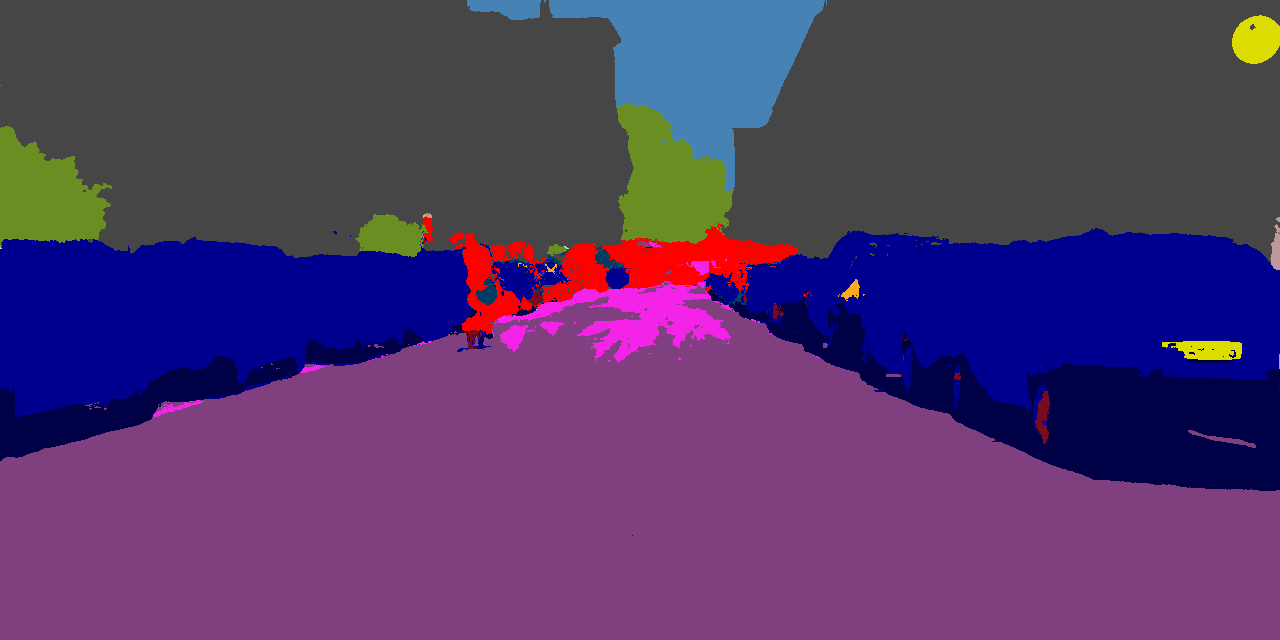} &
\includegraphics[width=\linewidth]{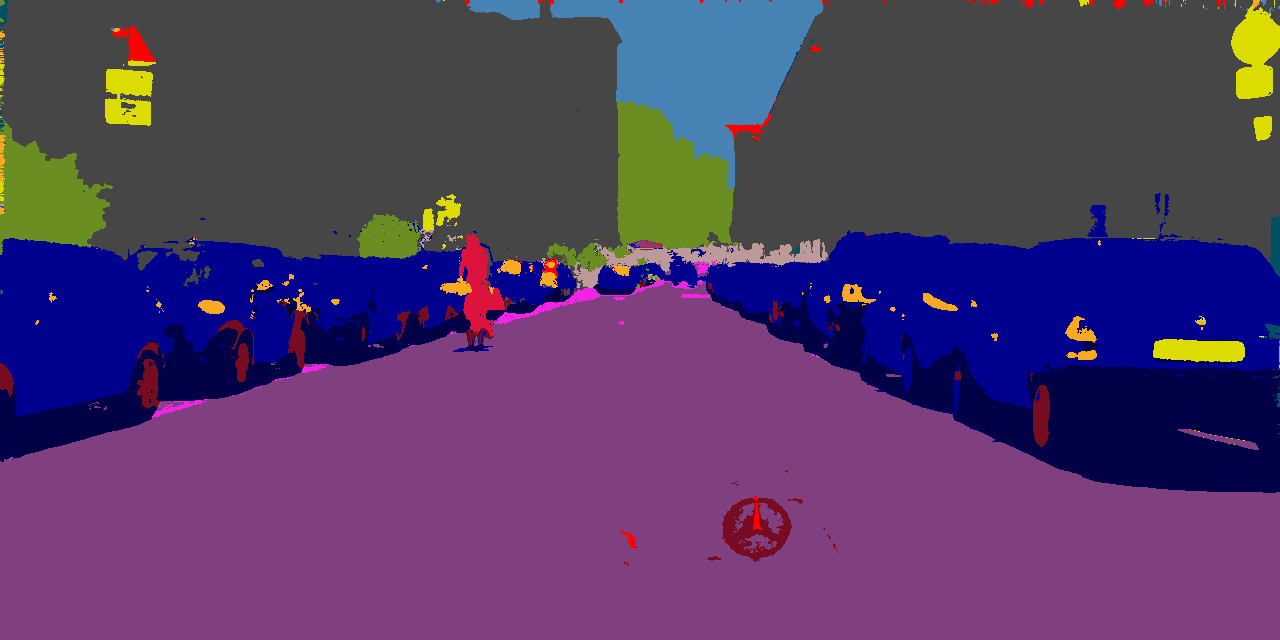} &
\includegraphics[width=\linewidth]{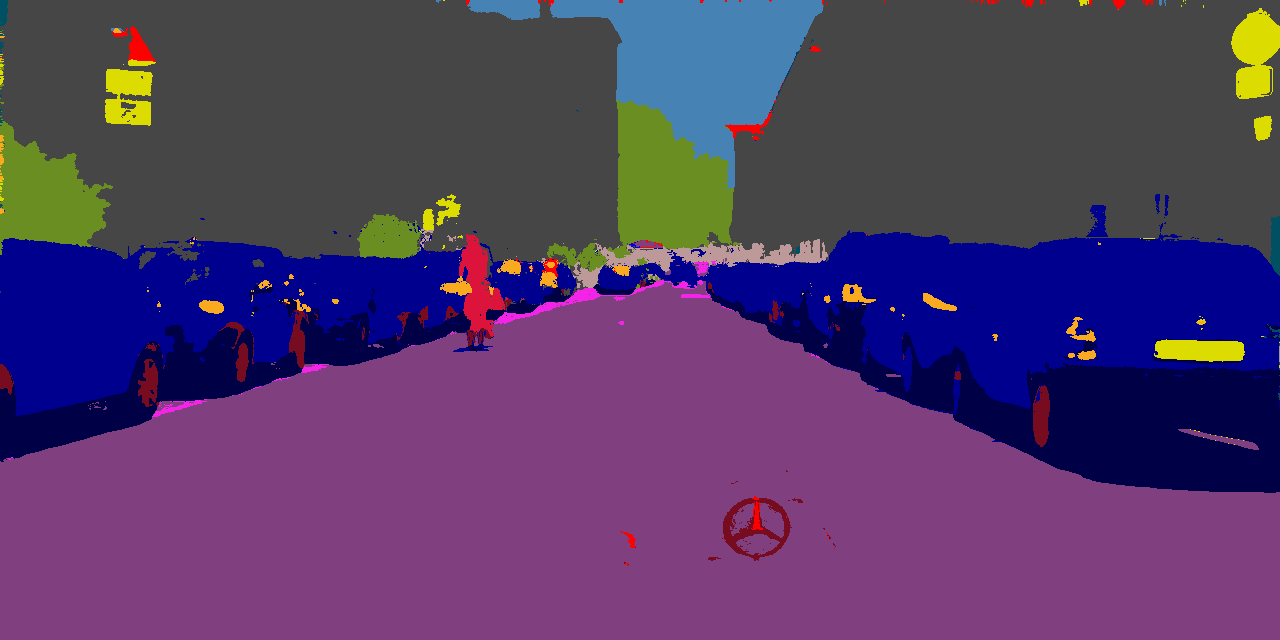} \\[-1pt]

\end{tabular}

\tiny
\renewcommand{\arraystretch}{1.3}
\begin{tabularx}{\textwidth}{*{19}{>{\centering\arraybackslash}X}}  
    \cellcolor{road}\textcolor{white}{Road}
    & \cellcolor{sidewalk}\textcolor{white}{Sidewalk}
    & \cellcolor{building}\textcolor{white}{Building}
    & \cellcolor{wall}\textcolor{white}{Wall}
    & \cellcolor{fence}\textcolor{white}{Fence}
    & \cellcolor{pole}\textcolor{white}{Pole}
    & \cellcolor{trafficlight}\textcolor{white}{Traffic~Light}
    & \cellcolor{trafficsign}\textcolor{white}{Traffic~Sign}
    & \cellcolor{vegetation}\textcolor{white}{Vegetation}
    & \cellcolor{terrain}\textcolor{white}{Terrain}
    & \cellcolor{sky}\textcolor{white}{Sky}
    & \cellcolor{person}\textcolor{white}{Person}
    & \cellcolor{rider}\textcolor{white}{Rider}
    & \cellcolor{car}\textcolor{white}{Car}
    & \cellcolor{truck}\textcolor{white}{Truck}
    & \cellcolor{bus}\textcolor{white}{Bus}
    & \cellcolor{train}\textcolor{white}{Train}
    & \cellcolor{motorcycle}\textcolor{white}{Motorcycle}
    & \cellcolor{bicycle}\textcolor{white}{Bicycle}
\end{tabularx}

    \vspace{-0.75em}
    \caption{\textbf{Depth-guided semantic pseudo-label examples.} Qualitative semantic pseudo-label examples comparing low resolution $\mathbf{P}\textsuperscript{low}$, high resolution $\mathbf{P}\textsuperscript{high}$, and depth-guided semantic fusion $\mathbf{P}^*$. \label{fig:qualitative_depthguided}}
    \vspace{-0.5em}
\end{figure*}

\inparagraph{Supervised \vs unsupervised optical flow.} In conjunction with the pseudo-label generation analysis presented in \cref{tab:pseudolabel_ablation}, we investigate the influence of different approaches for optical flow and two-frame disparity estimation on our pseudo labels in \cref{tab:pseudolabel_ablation_flow}. Identical to the analysis in the main paper, we generate pseudo labels on the validation split to ensure comparability with the \MethodName panoptic segmentation results and \MethodName analysis. %

\cref{tab:pseudolabel_ablation_flow} shows the direct quantitative evaluation of pseudo labels generated using different motion estimation methods against the ground truth (\ie, without the panoptic segmentation network). Alongside another unsupervised approach, BrightFlow~\cite{brightflow}, we include results obtained with supervised methods, RAFT-large~\cite{Teed:2020:RAF} (a supervised analog of SMURF~\cite{Stone:2021:SST}) and SEA-RAFT-large~\cite{searaft}.
We observe a rather consistent panoptic quality of the pseudo labels across different motion estimation networks.
As expected, the more accurate supervised optical flow methods can improve PQ further.
The slightly weaker panoptic quality with SEA-RAFT compared to RAFT might be due to SEA-RAFT being fine-tuned on multiple diverse datasets, whereas RAFT is fine-tuned specifically on KITTI.
To conclude, \MethodName is already effective with unsupervised flow and depth estimation methods, while exhibiting a notable margin for improvement in settings where some supervision of optical flow is available (and permissible).

\begin{table}[t]
    \centering
    \caption{\textbf{Instance pseudo label comparison.} Using MaskCut instance masks (U2Seg~\cite{Niu:2024:UUI}) in our CUPS pseudo-label generation. We compare using PQ, SQ, and RQ (in \%, $\uparrow$) for pseudo labels generated on Cityscapes val.}
    \vspace{-0.5em}
    \footnotesize\sisetup{table-number-alignment=center}
\setlength{\tabcolsep}{5pt}
\renewcommand{\arraystretch}{0.885}
\begin{tabularx}{\columnwidth}{>{\hspace{-\tabcolsep}\raggedright\columncolor{white}[\tabcolsep][\tabcolsep]}XZZZ}
	\toprule
    \textbf{Instance pseudo-label approach} & \textbf{PQ} & \textbf{SQ} & \textbf{RQ} \\
    \midrule
    MaskCut~\cite{Wang:2023:CAL} & 9.9 & 41.6 & 12.4  \\ 
    \rowcolor{tud0c!20} SF2SE3-ensembling \textit{(Ours)} & \bfseries 18.1	& \bfseries 47.3	& \bfseries 22.6 \\
	\bottomrule
\end{tabularx}

    \vspace{-0.5em}
    \label{tab:maskcuts}
\end{table}

\inparagraph{Analysis of depth-guided semantic pseudo labeling.} Following \cref{sec:depth_guided_semantic_inference}, we aim to analyze our proposed depth-guided semantic pseudo labeling in more detail.
\Cref{tab:dguided_analysis} shows that depth guidance fuses low- and high-resolution semantic predictions more effectively than an arithmetic mean. We use the identical experimental setting as in \cref{tab:pseudolabel_ablation}.
We further analyze pseudo labels by splitting images into depth ranges. Low resolution is best for pixels closer than 10\,m, both predictions perform similarly between 10\,--\,30\,m, and high resolution is superior beyond 30\,m.
These effects stem from DINO features trained on fixed-resolution, object-centric images, causing reduced representational quality at extreme scales. 
In short, low-resolution predictions produce blurry outputs for distant fine details, while high-resolution (sliding-window) predictions are more accurate at large distances but introduce errors near the camera. 
Overall, our depth-guided fusion yields the best metric performance.
We show qualitative examples in \cref{fig:qualitative_depthguided}.

\begin{table*}[t]
    \caption{\textbf{Per-class unsupervised panoptic segmentation on Cityscapes.} Comparing \MethodName to existing unsupervised panoptic methods, using PQ at the class level, as well as the mean PQ (in \%, $\uparrow$).\label{tab:classpq_analysis}}
    \vspace{-0.5em}
    \footnotesize\sisetup{table-number-alignment=center}
\setlength{\tabcolsep}{2.5pt}
\renewcommand{\arraystretch}{0.9}

\begin{tabularx}{\textwidth}{X *{20}{>{\centering\arraybackslash}Z}}
    \toprule
    {\textbf{Method}} 
    & \cellcolor{road}\rotatebox{90}{\textcolor{white}{\textbf{Road}}}
    & \cellcolor{sidewalk}\rotatebox{90}{\textcolor{white}{\textbf{Sidewalk}}}
    & \cellcolor{building}\rotatebox{90}{\textcolor{white}{\textbf{Building}}}
    & \cellcolor{wall}\rotatebox{90}{\textcolor{white}{\textbf{Wall}}}
    & \cellcolor{fence}\rotatebox{90}{\textcolor{white}{\textbf{Fence}}}
    & \cellcolor{pole}\rotatebox{90}{\textcolor{white}{\textbf{Pole}}}
    & \cellcolor{trafficlight}\rotatebox{90}{\textcolor{white}{\textbf{Traffic~Light$\;$}}}
    & \cellcolor{trafficsign}\rotatebox{90}{\textcolor{white}{\textbf{Traffic~Sign}}}
    & \cellcolor{vegetation}\rotatebox{90}{\textcolor{white}{\textbf{Vegetation}}}
    & \cellcolor{terrain}\rotatebox{90}{\textcolor{white}{\textbf{Terrain}}}
    & \cellcolor{sky}\rotatebox{90}{\textcolor{white}{\textbf{Sky}}}
    & \cellcolor{person}\rotatebox{90}{\textcolor{white}{\textbf{Person}}}
    & \cellcolor{rider}\rotatebox{90}{\textcolor{white}{\textbf{Rider}}}
    & \cellcolor{car}\rotatebox{90}{\textcolor{white}{\textbf{Car}}}
    & \cellcolor{truck}\rotatebox{90}{\textcolor{white}{\textbf{Truck}}}
    & \cellcolor{bus}\rotatebox{90}{\textcolor{white}{\textbf{Bus}}}
    & \cellcolor{train}\rotatebox{90}{\textcolor{white}{\textbf{Train}}}
    & \cellcolor{motorcycle}\rotatebox{90}{\textcolor{white}{\textbf{Motorcycle}}}
    & \cellcolor{bicycle}\rotatebox{90}{\textcolor{white}{\textbf{Bicycle}}}
    & \rotatebox{90}{\textbf{Mean (PQ)}}
    \\
    
    \midrule
    
    {\textcolor{tud0c}{Supervised~\cite{Kirillov:2019:PFP}}} & \color{tud0c}96.5 & \color{tud0c}72.4 & \color{tud0c}85.9 & \color{tud0c}16.4 & \color{tud0c}30.1 & \color{tud0c}48.6 & \color{tud0c}48.8 & \color{tud0c}67.2 & \color{tud0c}86.9 & \color{tud0c}34.8 & \color{tud0c}86.0 & \color{tud0c}65.0 & \color{tud0c}60.8 & \color{tud0c}79.2 & \color{tud0c}58.5 & \color{tud0c}77.2 & \color{tud0c}59.8 & \color{tud0c}54.9 & \color{tud0c}59.3 & \color{tud0c}62.3\\
    
    \midrule
    
    {DepthG~\cite{Sick:2024:USS}~+~CutLER~\cite{Wang:2023:CAL}} & 80.9 & 1.4 & 55.6 & \bfseries 3.0 & \bfseries 0.2 & 0.4 & 0.3 & 0.0 & 72.9 & 5.8 & 61.8 & 6.0 & 0.0 & 17.5 & 0.0 & 0.7 & 0.0 & 0.0 & 0.0 &16.1\\
    {U2Seg~\cite{Niu:2024:UUI}} & 82.5& 0.0& 42.4& 2.0& 0.0  & 0.0& 0.0& 0.0& 76.6& 1.5& 62.9& 8.3& \bfseries 2.2& 22.3& 10.2& \bfseries 27.0& \bfseries 4.7& \bfseries 0.7& 6.7 &18.4\\
    \midrule
    \rowcolor{tud0c!20} 
    {\MethodName \textit{(Ours)}} & \bfseries 85.8& \bfseries 6.0& \bfseries 64.4& 0.0& \bfseries 0.2& \bfseries 12.4& \bfseries 6.2& \bfseries 32.1& \bfseries 83.7& \bfseries 17.1& \bfseries 78.2& \bfseries 39.1& 0.0& \bfseries 62.9& \bfseries 16.3& 1.2& 0.0& 0.0& \bfseries 30.6 & \bfseries 27.8\\
    \bottomrule
\end{tabularx}

    \vspace{-0.5em}
\end{table*}

\begin{table}[t!]
    \centering
    \caption{\textbf{\MethodName self-training analysis}. Decomposing the self-training by analyzing the augmentation quality using PQ, SQ, and RQ (in \%, $\uparrow$) on Cityscapes val. \label{tab:tta_ablation}}
    \vspace{-0.5em}
    \footnotesize\sisetup{table-number-alignment=center}
\setlength{\tabcolsep}{5pt}
\renewcommand{\arraystretch}{0.885}
\begin{tabularx}{\columnwidth}{>{\hspace{-\tabcolsep}\raggedright\columncolor{white}[\tabcolsep][\tabcolsep]}XZZZZ}
	\toprule
    \textbf{Training configuration} & \textbf{PQ} & \textbf{SQ} & \textbf{RQ} & \textbf{Runtime (ms)} \\
    \midrule
    \MethodName w/o self-training   & 26.6 & \bfseries 57.5 & 33.5 & \hphantom{3}65.9 \\
    \MethodName w/o self-training + TTA   &  27.4 & 57.2 & 34.9 & 413.4 \\
    \rowcolor{tud0c!20} \MethodName \textit{(Ours)}   & \bfseries 27.8 & 57.4 & \bfseries 35.2 & \bfseries \hphantom{3}65.2\\ 
	\bottomrule
\end{tabularx}

    \vspace{-0.5em}
\end{table}

\inparagraph{Instance pseudo labeling analysis.} Supporting the qualitative results presented in \cref{fig:maskcut}, we further analyze the performance of our SF2SE3-ensembling approach against MaskCut~\cite{Wang:2023:CAL}. In particular, \cref{tab:maskcuts} presents pseudo-label evaluation results, replacing our SF2SE3-ensembling with MaskCut. All other pseudo-label generation components are kept the same. MaskCut fails to generate high-quality instance masks on scene-centric images, as PQ and RQ almost halved compared to our SF2SE3-ensembling.

\subsection{Unsupervised panoptic segmentation results}

\inparagraph{Class-level PQ.} \Cref{tab:classpq_analysis} expands \cref{tab:panoptic_segmentation_cs} by detailing class-wise PQ.
\MethodName demonstrates substantial improvements on most categories, particularly excelling on ``Car'' (\qty{62.9}{\percent}), ``Person'' (\qty{39.1}{\percent}), ``Traffic Sign'' (\qty{32.1}{\percent}), and ``Sky'' (\qty{78.2}{\percent}). 
Although \MethodName has difficulties with a few classes, \eg , ``Wall'', ``Fence'', and ``Rider'', our baseline DepthG~\cite{Sick:2024:USS}~+~CutLER~\cite{Wang:2023:CAL} and U2Seg~\cite{Niu:2024:UUI} also struggle with segmenting these classes. The only exception is ``Bus'', on which \MethodName exhibits lower PQ than U2Seg. In the case of ``Rider'', CUPS does not learn this as a separate class, which is probably due to the motion cue used for instance pseudo labeling, which cannot easily separate a ``Rider'' from their means of transportation. Accordingly, \MethodName usually predicts person instead of rider or the entire unit of ``Bicycle'' and ``Rider'' is predicted as ``Bicycle'' (\cf \cref{fig:qualitative_cs}, second example).
Nevertheless, \MethodName significantly improves the panoptic quality for the majority of classes and narrows the gap to the supervised upper bound. %

\inparagraph{Panoptic self-training \vs test-time augmentation.} Following up on the ablation in \cref{tab:pseudo_label_train}, we provide a finer-grained analysis of the self-training process  in \cref{tab:tta_ablation} by comparing against using the self-labeling augmentations as test-time augmentation (TTA) at inference time directly after Stage 2 instead of the self-training.
Recall that the self-labeling augmentations involve resizing the input image to three different scales and applying horizontal flipping, followed by aggregating the predictions.
Self-labeling augmentations, combined with confidence thresholding and self-enhanced copy-paste augmentations, provide self-labels for self-training (Stage~3). Note that we report TTA without thresholding in \cref{tab:tta_ablation}.
While the results in \cref{tab:tta_ablation} show that TTA improves the panoptic quality, it is not a practical approach due to the significantly increased inference time.
By contrast, panoptic self-training retains the original runtime of the network and even surpasses TTA in panoptic quality.

\begin{table}[t]
\caption{\textbf{DepthG~\cite{Sick:2024:USS} + VideoCutLER~\cite{videocutler} baseline.} We compare \MethodName to a baseline using VideoCutLER on the Cityscapes val dataset (all metrics in \%, $\uparrow$).\label{tab:videocutler}}
\vspace{-0.5em}
        \footnotesize\sisetup{table-number-alignment=center}
        \setlength{\tabcolsep}{5pt}
        \renewcommand{\arraystretch}{0.885}
        \begin{tabularx}{\columnwidth}{>{\hspace{-\tabcolsep}\raggedright\columncolor{white}[\tabcolsep][\tabcolsep]}XZZZ}
        \toprule
        \textbf{Method} & \textbf{PQ} & \textbf{SQ} & \textbf{RQ} \\
        \midrule    
        {DepthG~\cite{Sick:2024:USS} + CutLER~\cite{Wang:2023:CAL}} & 16.1 & 45.4 & 21.1 \\
        {DepthG~\cite{Sick:2024:USS} + VideoCutLER~\cite{videocutler}} & 16.6 & 42.6 & 20.5 \\
        \rowcolor{tud0c!20} \MethodName \textit{(Ours)}   & \bfseries 27.8 & \bfseries 57.4 & \bfseries 35.2\\ 
        \bottomrule
        \end{tabularx}
        \vspace{-0.5em}
\end{table}%

\begin{table*}[t]
    \caption{\textbf{Unsupervised panoptic segmentation for \MethodName on Cityscapes, KITTI, BDD, MUSES, Waymo, and MOTS.} Comparing \MethodName to existing unsupervised panoptic methods, using PQ, SQ, and RQ (in \%, $\uparrow$) for different numbers of pseudo classes. By default, \MethodName uses 27 pseudo classes to facilitate the comparison against both unsupervised panoptic and unsupervised semantic segmentation approaches. We also test 40 (\qty{150}{\percent} of the default) and 54 pseudo classes (\qty{200}{\percent} of the default), showcasing the impact of overclustering. \label{tab:overclustering_analysis}}
    \vspace{-0.5em}
    \footnotesize\sisetup{table-number-alignment=center}
\setlength{\tabcolsep}{2.5pt}
\renewcommand{\arraystretch}{0.885}
\begin{tabularx}{\textwidth}{>{\hspace{-\tabcolsep}\raggedright\columncolor{white}[\tabcolsep][\tabcolsep]}lYZZZZZZZZZZZZZZZZZZ}
    \toprule
    & & \multicolumn{3}{c}{\textbf{Cityscapes}} & \multicolumn{3}{c}{\textbf{KITTI}} & \multicolumn{3}{c}{\textbf{BDD}} & \multicolumn{3}{c}{\textbf{MUSES}} & \multicolumn{3}{c}{\textbf{Waymo}} & \multicolumn{3}{c}{\textbf{MOTS}}\\
    \cmidrule(l{0.2em}r{0.2em}){3-5} \cmidrule(l{0.2em}r{0.2em}){6-8} \cmidrule(l{0.2em}r{0.2em}){9-11} \cmidrule(l{0.2em}r{0.2em}){12-14} \cmidrule(l{0.2em}r{0.2em}){15-17} \cmidrule(l{0.2em}r{0.2em}){18-20}

    \multirow{-2}{*}{\vspace{0.5em}\textbf{Method}} & \multirow{-2}{*}{\vspace{0.5em}\textbf{$\!$Pseudo classes$\!$}} & {\textbf{PQ}} & {\textbf{SQ}} & {\textbf{RQ}} & \textbf{PQ} & \textbf{SQ} & \textbf{RQ} & \textbf{PQ} & \textbf{SQ} & \textbf{RQ} & \textbf{PQ} & \textbf{SQ} & \textbf{RQ} & \textbf{PQ} & \textbf{SQ} & \textbf{RQ} & \textbf{PQ} & \textbf{SQ} & \textbf{RQ}\\
    \midrule
    \textcolor{tud0c}{Supervised~\cite{Kirillov:2019:PFP}} & \textcolor{tud0c}{--} & \color{tud0c}62.3 & \color{tud0c}81.8 & \color{tud0c}75.1 & \color{tud0c}31.9 & \color{tud0c}71.7 & \color{tud0c}40.4 & \color{tud0c}33.0 & \color{tud0c}76.3 & \color{tud0c}42.0 & \color{tud0c}38.1 & \color{tud0c}62.4 & \color{tud0c}49.6 & \color{tud0c}31.5 & \color{tud0c}70.1 & \color{tud0c}40.9 & \color{tud0c}73.8 & \color{tud0c}86.4 & \color{tud0c}84.6  \\
    \midrule
    {DepthG~\cite{Sick:2024:USS}~+~CutLER~\cite{Wang:2023:CAL}} & 27 & 16.1 & 45.4	& 21.1 & 11.0	& 34.5	& 13.8 & 14.4	& 41.9	& 19.2 & 10.1 & 30.1 & 13.1 & 13.4	& 37.3 & 17.0 & 49.6	& 78.4	& 60.6  \\
    {U2Seg~\cite{Niu:2024:UUI}} & 800 + 27 & 18.4 & 55.8	& 22.7 & 20.6 & 52.9 & 25.2 & 15.8	& 57.2 & 19.2 & 20.3 & 45.8 & 26.5 & 19.8 & 50.8 & 23.4 & 50.7	& 79.2	& 64.3 \\
    \midrule
    \rowcolor{tud0c!20} {\MethodName \textit{(Ours)}} & 27 \textit{(default)} & 27.8 & 57.4	& 35.2 & 25.5 & 58.1 & 32.5 & 19.9	& 60.3	& 25.9 & 24.4	& 48.5	& 33.0 & 26.4	& 60.3	& 33.0 & 67.8	& 86.4	& 76.9  \\
    {\MethodName \textit{(Ours)}} & 40 & 30.3 & 64.3 & 37.5 & 28.1 & \bfseries 63.1 & 35.3 & \bfseries 21.9 & 57.3 & \bfseries 28.1 & \bfseries 28.2 & \bfseries 52.9 & \bfseries 35.4 &  27.2 &  62.4 &  \bfseries 33.6 & 74.0 & 88.4 & 82.8 \\
    {\MethodName \textit{(Ours)}} & 54 & \bfseries 30.6 & \bfseries 65.1 & \bfseries 37.8 & \bfseries 28.5 & 60.6  & \bfseries 36.0 & 21.8 & \bfseries 62.5 & 27.6 & 22.8 & 45.4 & 29.3 & \bfseries 27.3 & \bfseries 65.3 & 32.5 & \bfseries 78.7 & \bfseries 89.3 & \bfseries 87.4 \\
    \bottomrule
\end{tabularx}

\end{table*}
\begin{table*}[t]
    \caption{\textbf{Hierarchical unsupervised panoptic segmentation on Cityscapes.} Comparing \MethodName to existing unsupervised panoptic methods, using PQ (all in \%, $\uparrow$) on different class hierarchies. All datasets are analyzed on 19 and 7 ground truth classes. The number of ground-truth classes is indicated by the superscript of the metric. \label{tab:hierarchical}}
    \vspace{-0.5em}
    \footnotesize\sisetup{table-number-alignment=center}
\setlength{\tabcolsep}{5pt}
\renewcommand{\arraystretch}{0.885}
\begin{tabularx}{\textwidth}{>{\hspace{-\tabcolsep}\raggedright\columncolor{white}[\tabcolsep][\tabcolsep]}lY *{11}{>{\centering\arraybackslash}Z}}
    \toprule
    & & \multicolumn{2}{c}{\textbf{Cityscapes}} & \multicolumn{2}{c}{\textbf{KITTI}} & \multicolumn{2}{c}{\textbf{BDD}} & \multicolumn{2}{c}{\textbf{MUSES}} & \multicolumn{2}{c}{\textbf{Waymo}} \\
    \cmidrule(l{0.2em}r{0.2em}){3-4} \cmidrule(l{0.2em}r{0.2em}){5-6} \cmidrule(l{0.2em}r{0.2em}){7-8} \cmidrule(l{0.2em}r{0.2em}){9-10} \cmidrule(l{0.2em}r{0.2em}){11-12}

    \multirow{-2}{*}{\vspace{0.5em}\textbf{Method}} & \multirow{-2}{*}{\vspace{0.5em}\textbf{Pseudo classes}} & \textbf{PQ}\textsuperscript{19} & \textbf{PQ}\textsuperscript{7} & \textbf{PQ}\textsuperscript{19} & \textbf{PQ}$\textsuperscript{7}$ & \textbf{PQ}\textsuperscript{19} & \textbf{PQ}\textsuperscript{7} & \textbf{PQ}\textsuperscript{19} & \textbf{PQ}\textsuperscript{7} & \textbf{PQ}\textsuperscript{16} & \textbf{PQ}\textsuperscript{7}\\
    \midrule
    \textcolor{tud0c}{Supervised~\cite{Kirillov:2019:PFP}} & \textcolor{tud0c}{--} & \color{tud0c} 62.3 & \color{tud0c} 79.8 & \color{tud0c} 31.9 & \color{tud0c} 57.9 & \color{tud0c} 33.0 & \color{tud0c} 54.6 & \color{tud0c} 38.1 & \color{tud0c} 69.4 & \color{tud0c} 31.5 & \color{tud0c} 62.3 \\
    \midrule
    {DepthG~\cite{Sick:2024:USS}~+~CutLER~\cite{Wang:2023:CAL}} & 27 & 16.1 & 44.1 & 10.9 & 27.6 & 14.4 & 38.5 & 10.1 & 22.1 & 13.4 & 37.7 \\
    {U2Seg~\cite{Niu:2024:UUI}}  & 800 + 27 & 18.4 & 43.5 & 20.6 & 44.4 & 15.8 & 37.3 & 20.3 & 41.4 & 19.8 & 39.6 \\
    \midrule
    \rowcolor{tud0c!20} {\MethodName \textit{(Ours)}} & 27 & \bfseries 27.8 & \bfseries 63.9 & \bfseries 25.4 & \bfseries 57.4 & \bfseries 19.9 & \bfseries 49.3 & \bfseries 24.4 & \bfseries 53.5 & \bfseries 26.4 & \bfseries 54.7\\
    \bottomrule
\end{tabularx}

    \vspace{-0.5em}
\end{table*}

\inparagraph{DepthG+VideoCutLER baseline.}
Since CUPS leverages two consecutive frames to generate instance pseudo labels, it inherently exploits temporal consistency. Consequently, we combine VideoCutLER~\cite{videocutler}, an unsupervised method for video instance segmentation, with DepthG as an additional baseline.
We performed the experiment using five consecutive frames as the video input to VideoCutLER.
The semantic and instance predictions of DepthG and VideoCutLER are combined identically to the DepthG+CutLER baseline. 
As shown in \cref{tab:videocutler}, DepthG+VideoCutLER is slightly worse for SQ and RQ, yet better in PQ. We attribute this to the improved temporal consistency. %
Our CUPS approach strongly outperforms this video baseline as well.

\inparagraph{Overclustering analysis.} Overclustering refers to setting the number of pseudo labels significantly higher than the number of ground-truth categories.
Extending the analysis presented in \cref{tab:overclustering}, we analyze the impact of overclustering along two dimensions.
First, we test \MethodName with an increased number of pseudo classes.
Second, we run \MethodName in the default setting, but evaluate it on the group-level class hierarchies defined by Cityscapes. Here, the 19-class taxonomy is mapped  down to 7 broader groups of classes.

When training \MethodName with a larger number of pseudo classes---specifically, 40 (\qty{150}{\percent} of the default number of pseudo classes)---we observe a significant improvement in the panoptic segmentation metrics (\cf \cref{tab:overclustering_analysis}). Further increasing the number of pseudo classes to 54 (\qty{200}{\percent} of the default number of pseudo classes) yields additional improvements but also exhibits a saturation trend. However, significantly increasing the number of pseudo-classes can impede generalization, as visible on MUSES when using 54 pseudo classes. In general, we use 27 pseudo classes for fair comparison, as it is the lowest number of pseudo classes that allows for a comparison to both unsupervised panoptic and unsupervised semantic segmentation.

In \cref{tab:hierarchical}, we evaluate \MethodName on different class hierarchies. While the main paper demonstrates substantial gains in the standard 19-class evaluation, we show that the gains extrapolate to the setting with a coarser grouping of 7 Cityscapes classes: ``Flat'' (\eg, ``Road'', ``Sidewalk''), ``Human'' (\eg, ``Person'', ``Rider''), ``Vehicle'' (\eg, ``Car'', ``Truck''), ``Construction'' (\eg, ``Building'', ``Wall''), ``Object'' (\eg, ``Pole'', ``Traffic Sign''), ``Nature'' (\eg, ``Vegetation'', ``Terrain''), and ``Sky''.
Although the accuracy improvement on the coarser label set is expected, this experiment empirically demonstrates that our analysis and conclusions hold for different granularities of the semantic taxonomy.
As another remark, we follow up on our observation from \cref{tab:panoptic_segmentation_others} in the main text, where the supervised model (trained on Cityscapes) suffers a noticeable drop in segmentation performance outside the training domain. 
In the coarser setting here, this observation applies to a more striking extent: \MethodName nearly approaches the supervised bound and achieves competitive panoptic quality with the supervised model (\eg, only \qty{0.3}{\percent} worse on KITTI).

\begin{table}[t]
\caption{\textbf{Panoptic segmentation architecture analysis.} We evaluate CUPS after stage-1 training on the Cityscapes val datasets (all metrics in \%, $\uparrow$). \label{tab:segmodel}}
\vspace{-0.75em}
        \footnotesize\sisetup{table-number-alignment=center}
        \setlength{\tabcolsep}{5pt}
        \renewcommand{\arraystretch}{0.885}
        \begin{tabularx}{\columnwidth}{>{\hspace{-\tabcolsep}\raggedright\columncolor{white}[\tabcolsep][\tabcolsep]}XZZZ}
        \toprule
        \textbf{Segmentation model} & \textbf{PQ} & \textbf{SQ} & \textbf{RQ} \\
        \midrule    
        {Mask2Former~\cite{Cheng:2022:M2F}} & 25.1 & \bfseries 57.7 & 31.7 \\
        \rowcolor{tud0c!20} {Panoptic Cascade Mask R-CNN~\cite{Cai:2018:CRC, Kirillov:2019:PFP}} & \bfseries 26.6 & 57.5 & \bfseries 33.5 \\
        \bottomrule
        \end{tabularx}
        \vspace{-0.5em}
\end{table}%

\begin{figure*}[t]
    \caption{\textbf{Qualitative unsupervised panoptic segmentation examples} across all datasets after Hungarian matching. We compare \MethodName \textit{(Ours)} to the DepthG+CutLER baseline and U2Seg. \MethodName produces more consistent and accurate panoptic segmentations. \label{fig:panotic_qualitative}}
    
    \begin{subfigure}[t]{\textwidth}
        \centering
        \small
\sffamily
\setlength{\tabcolsep}{0pt}
\renewcommand{\arraystretch}{0.0}
\begin{tabular}{>{\centering\arraybackslash} m{0.2\textwidth} 
                >{\centering\arraybackslash} m{0.2\textwidth} 
                >{\centering\arraybackslash} m{0.2\textwidth}
                >{\centering\arraybackslash} m{0.2\textwidth}
                >{\centering\arraybackslash} m{0.2\textwidth}}

{Image} & {Ground Truth} & {Baseline} & {U2Seg~\cite{Niu:2024:UUI}} & {\MethodName \textit{(Ours)}} \\[4pt]

\includegraphics[width=\linewidth]{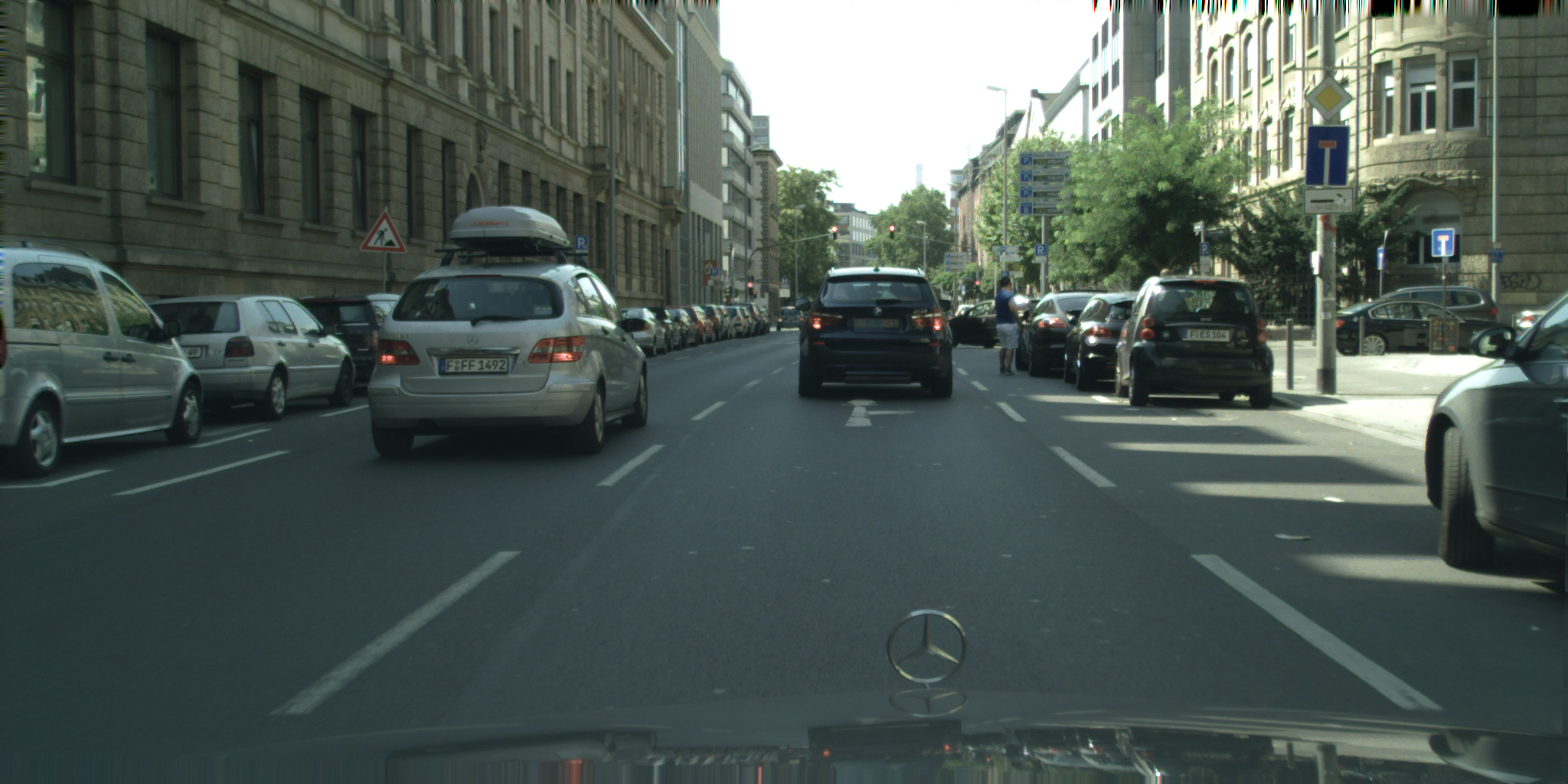} &
\includegraphics[width=\linewidth]{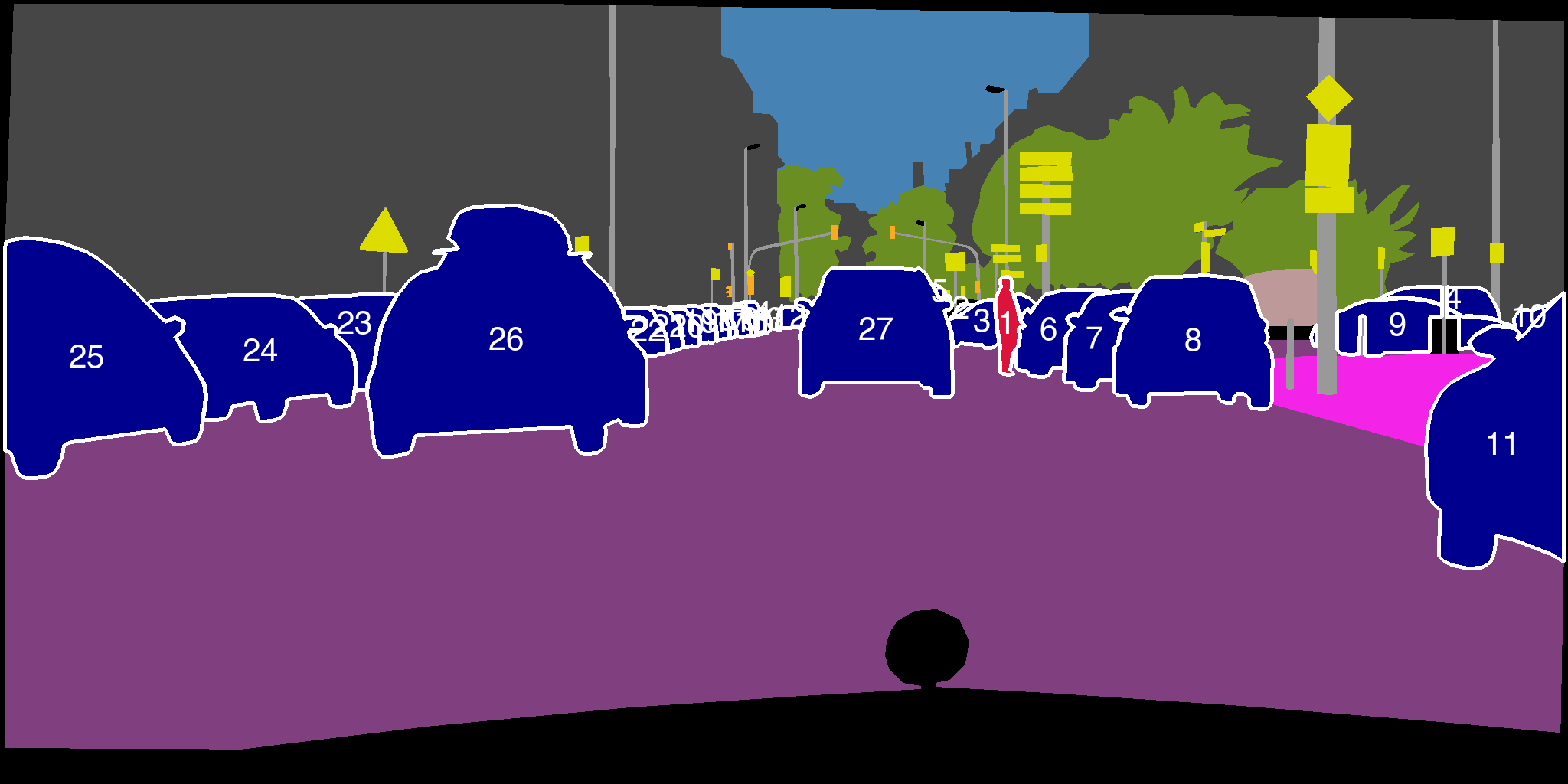} &
\includegraphics[width=\linewidth]{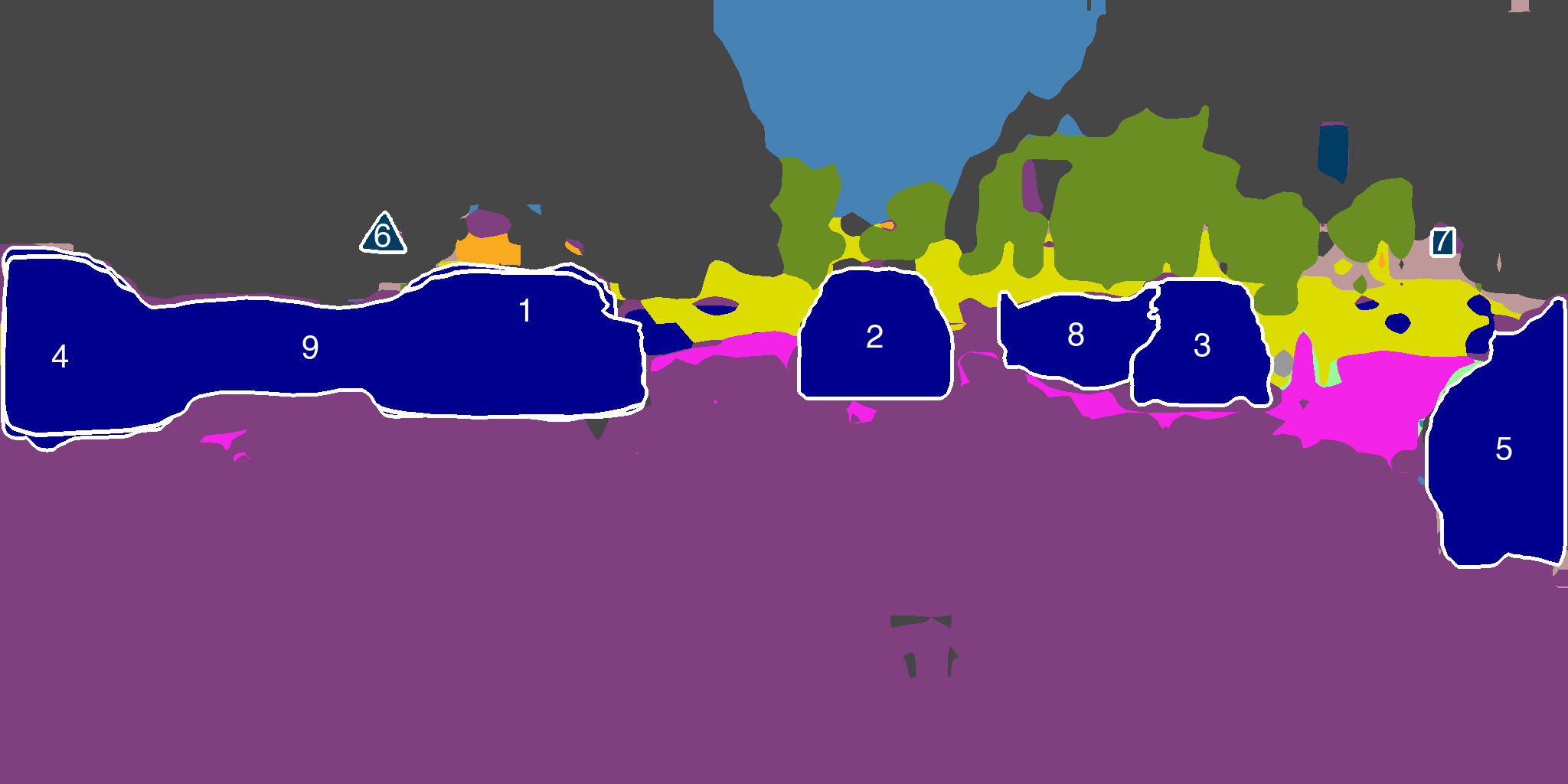} &
\includegraphics[width=\linewidth]{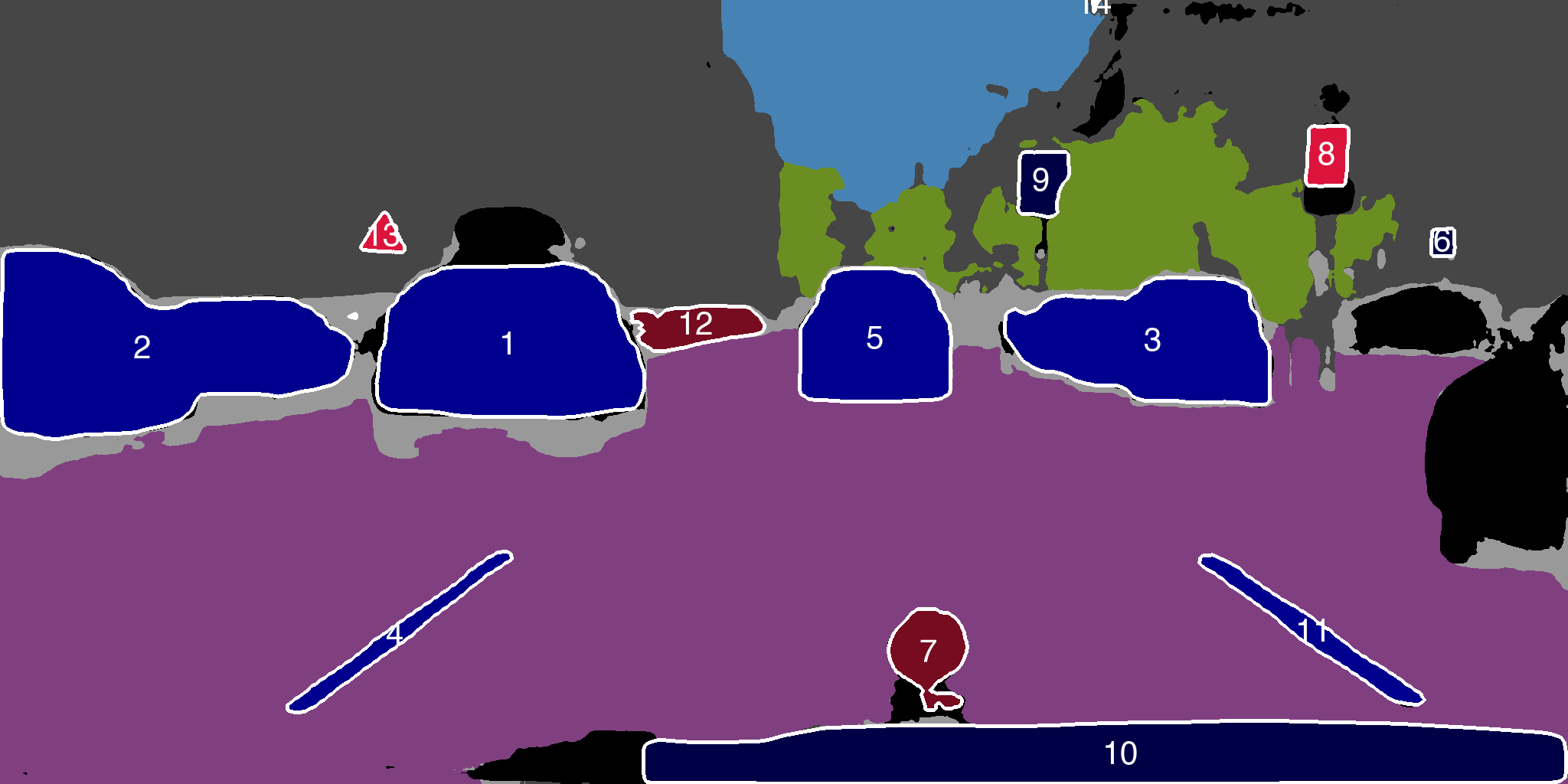} &
\includegraphics[width=\linewidth]{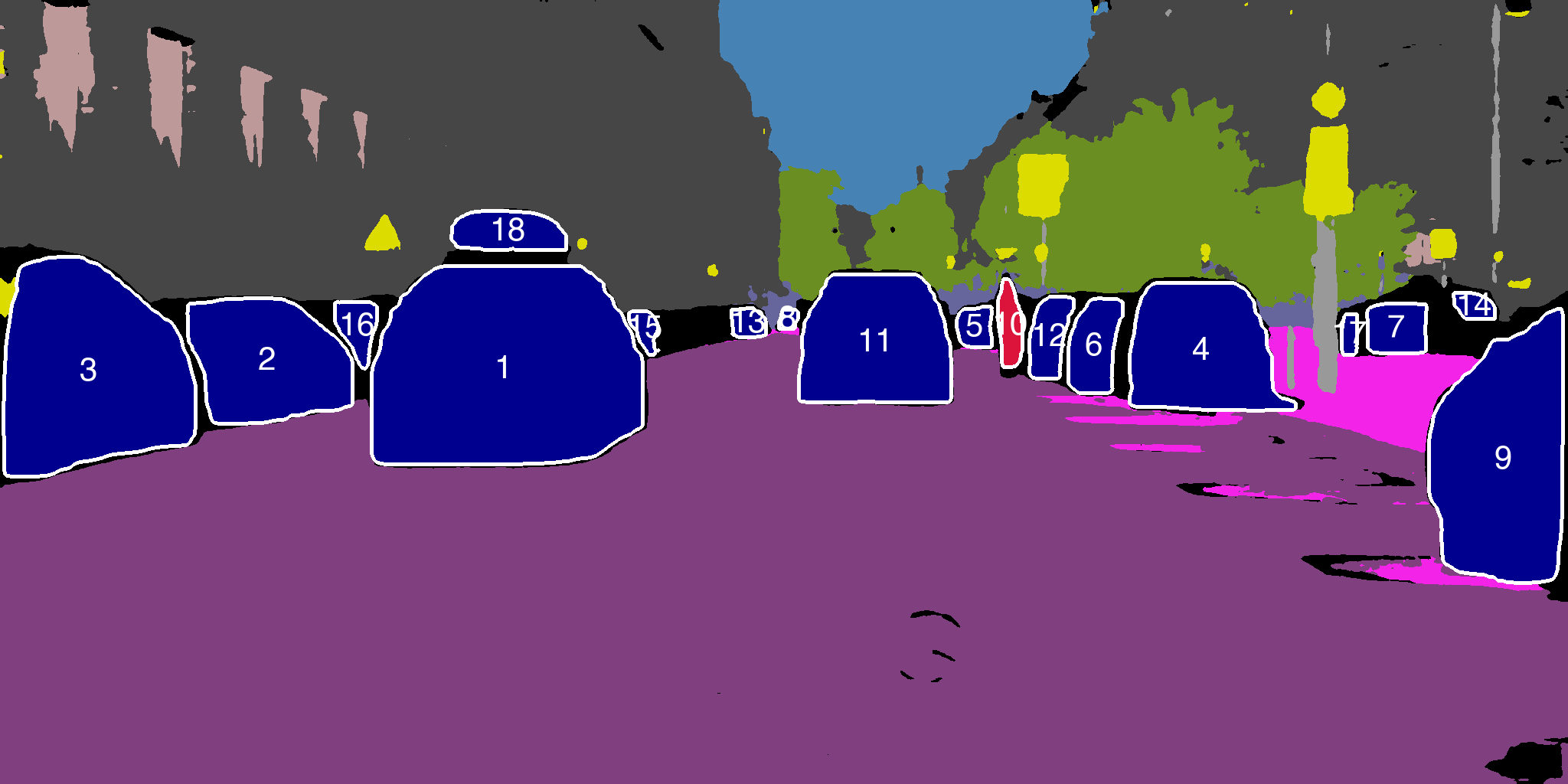} \\

\includegraphics[width=\linewidth]{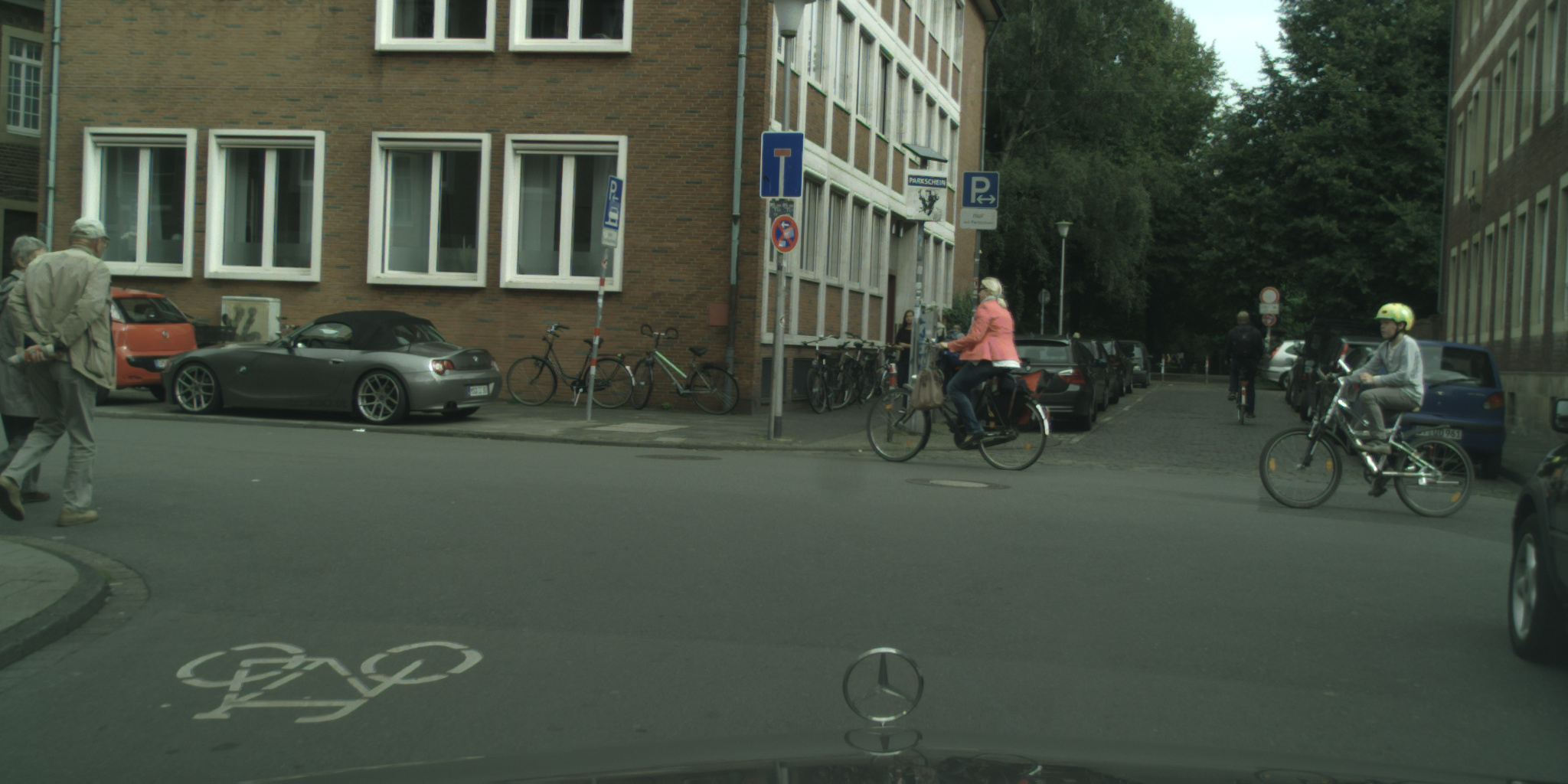} &
\includegraphics[width=\linewidth]{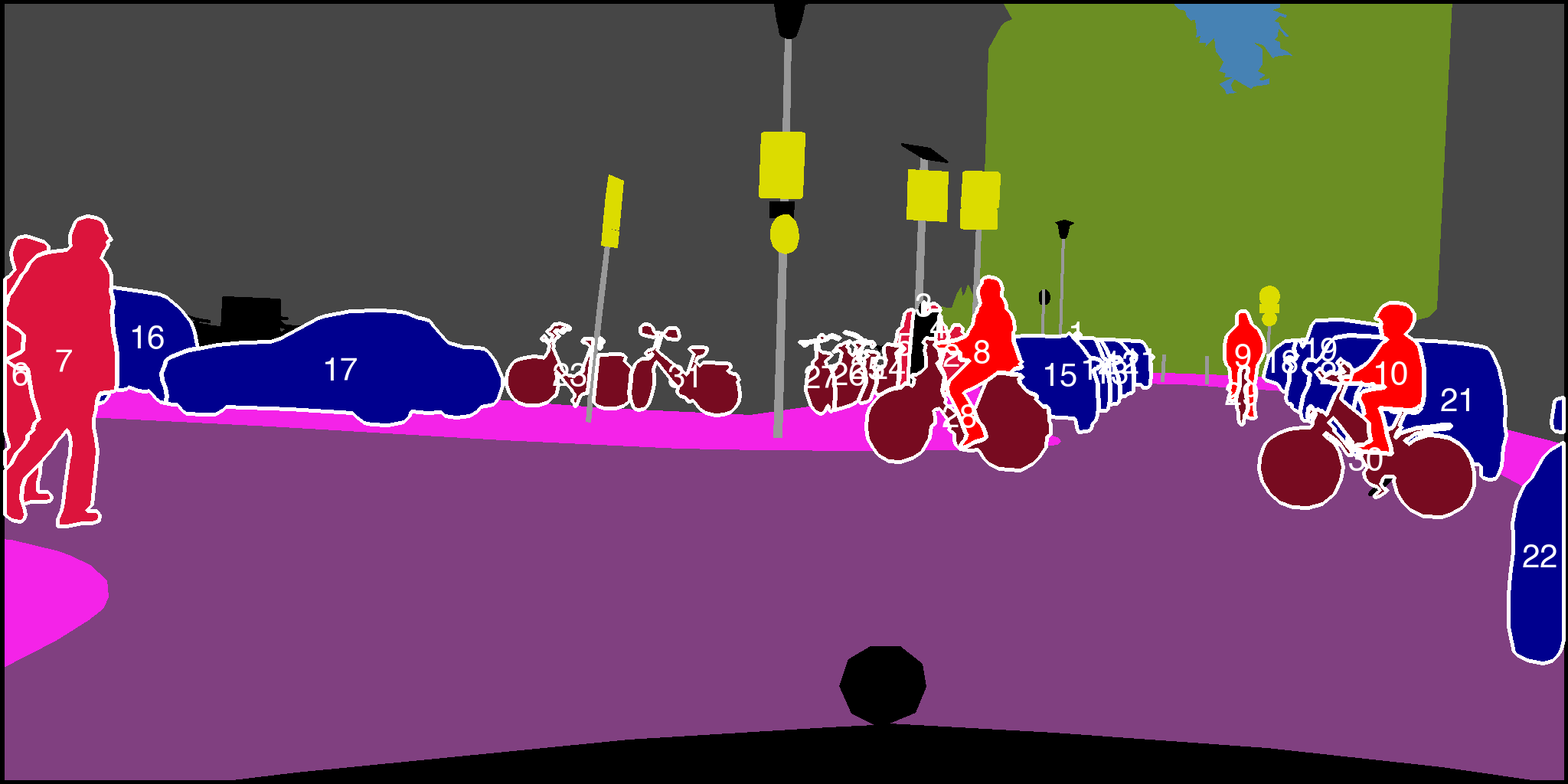} &
\includegraphics[width=\linewidth]{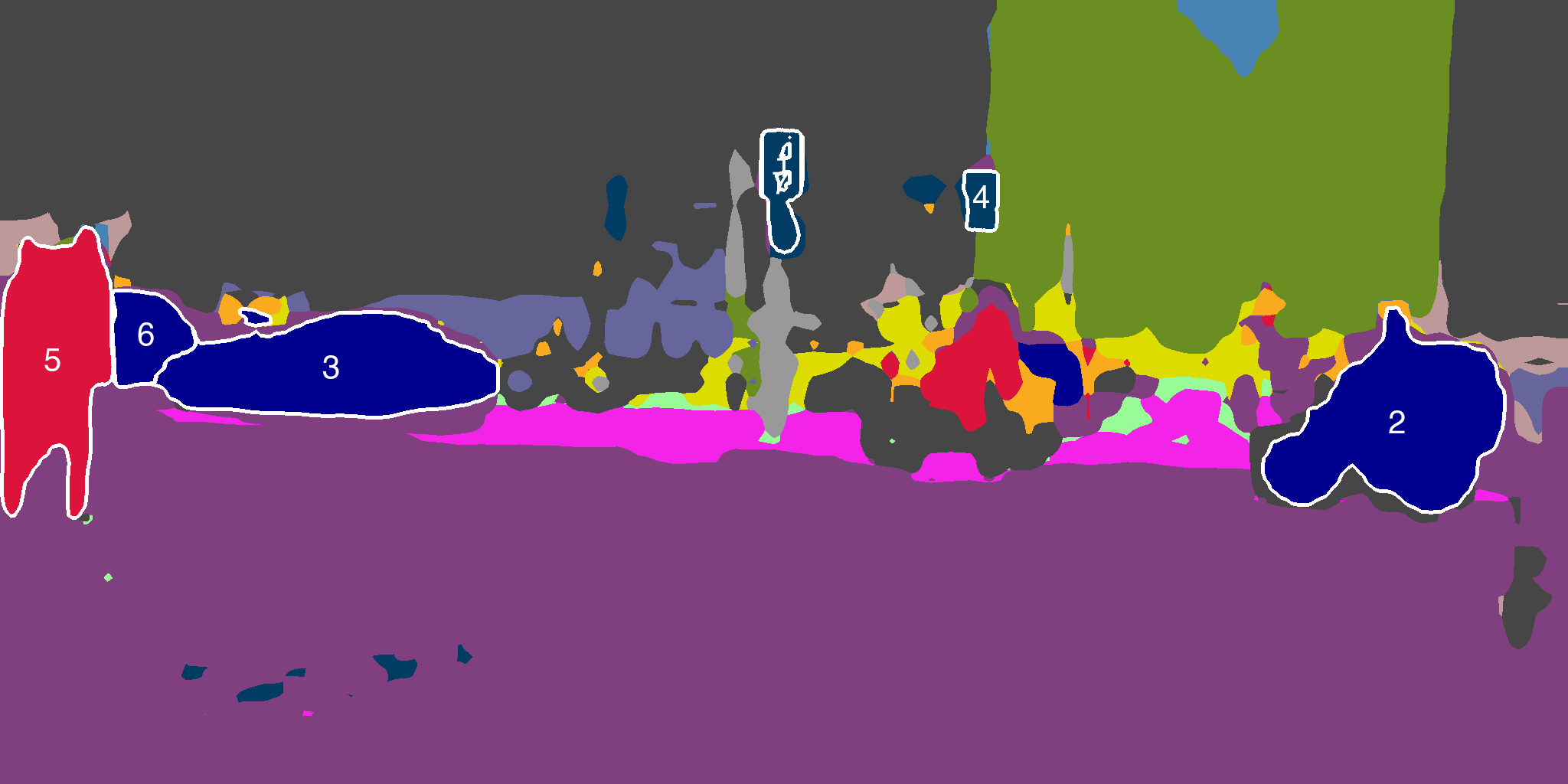} &
\includegraphics[width=\linewidth]{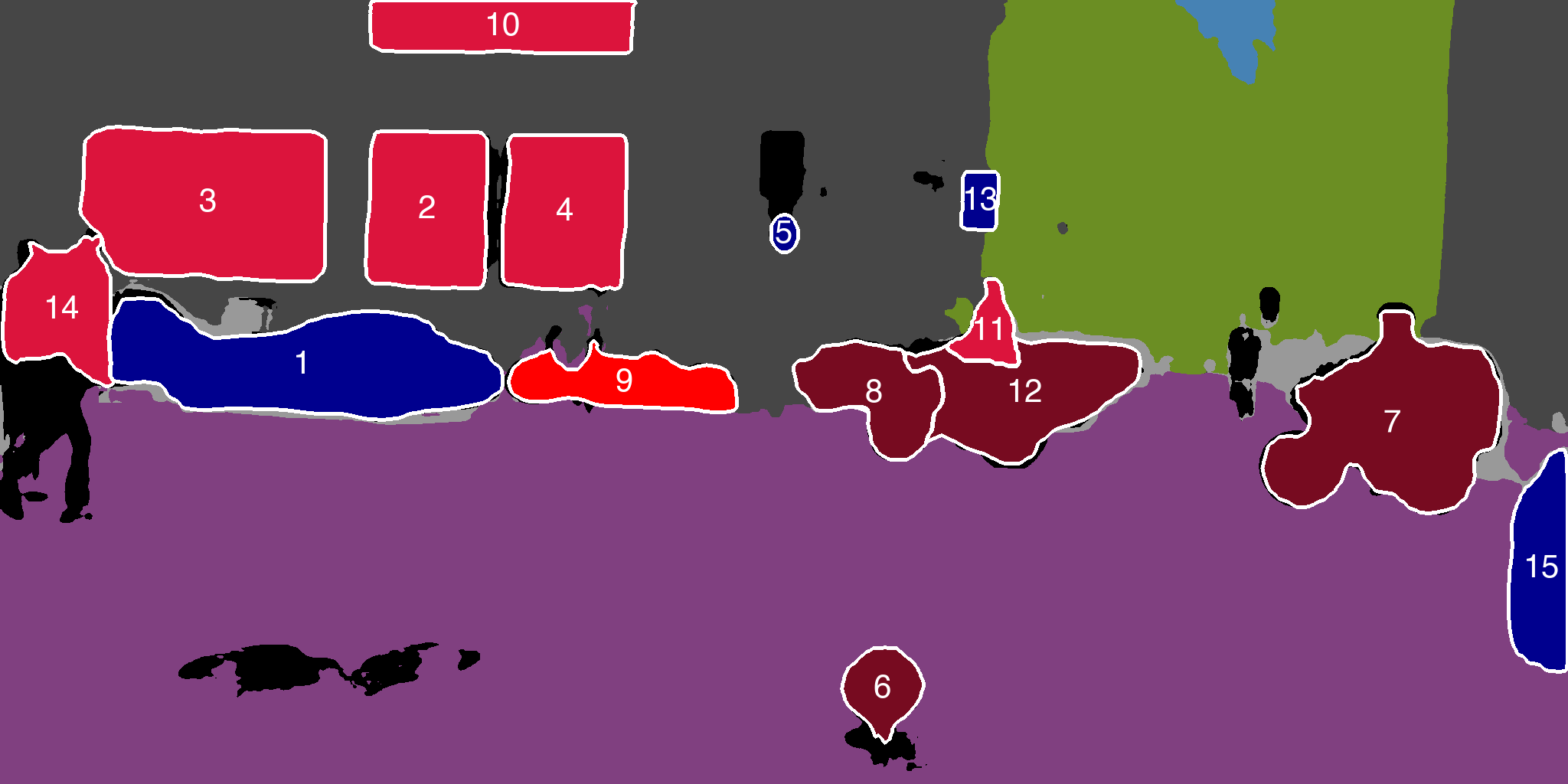} &
\includegraphics[width=\linewidth]{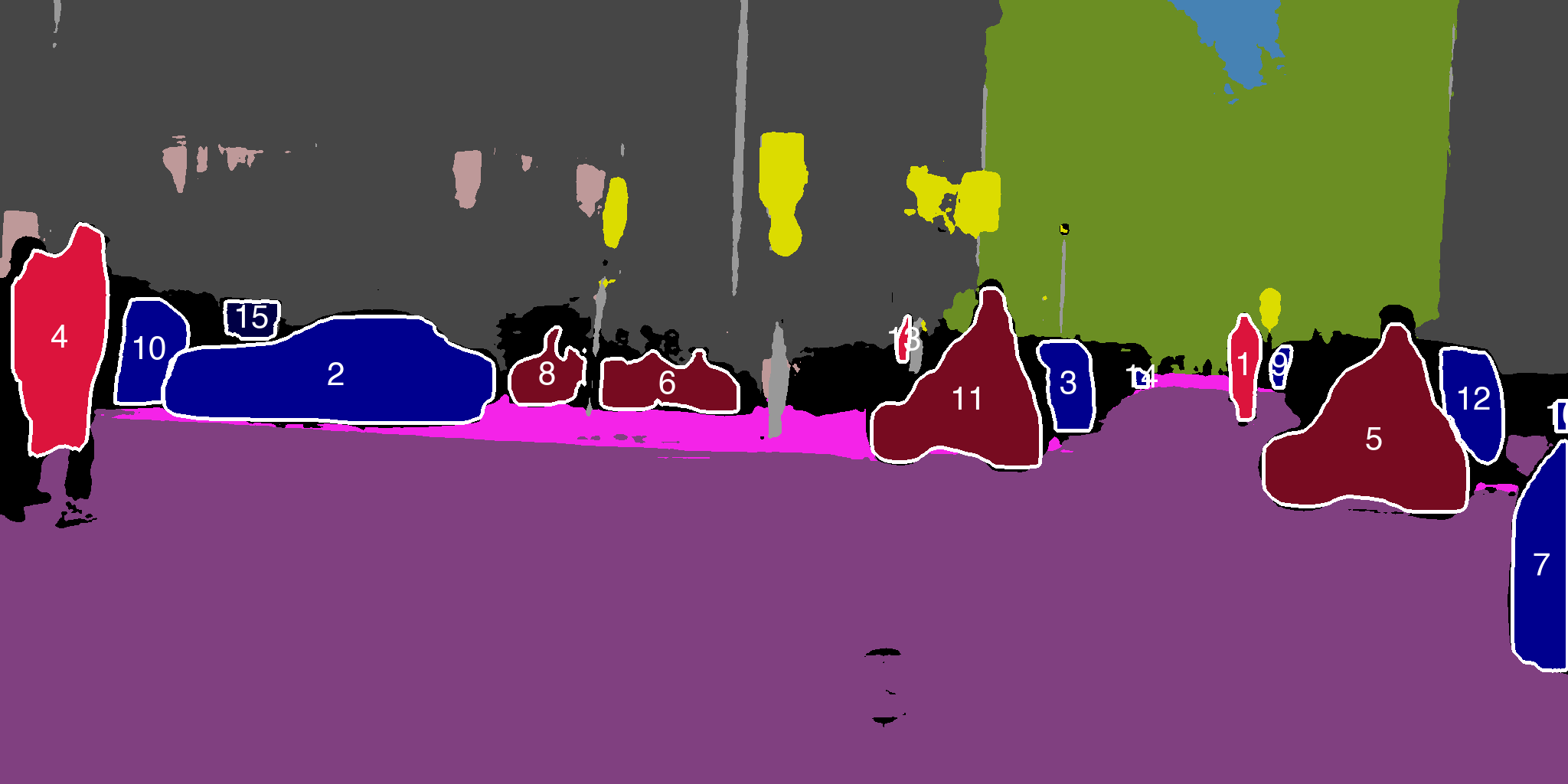} \\

\includegraphics[width=\linewidth]{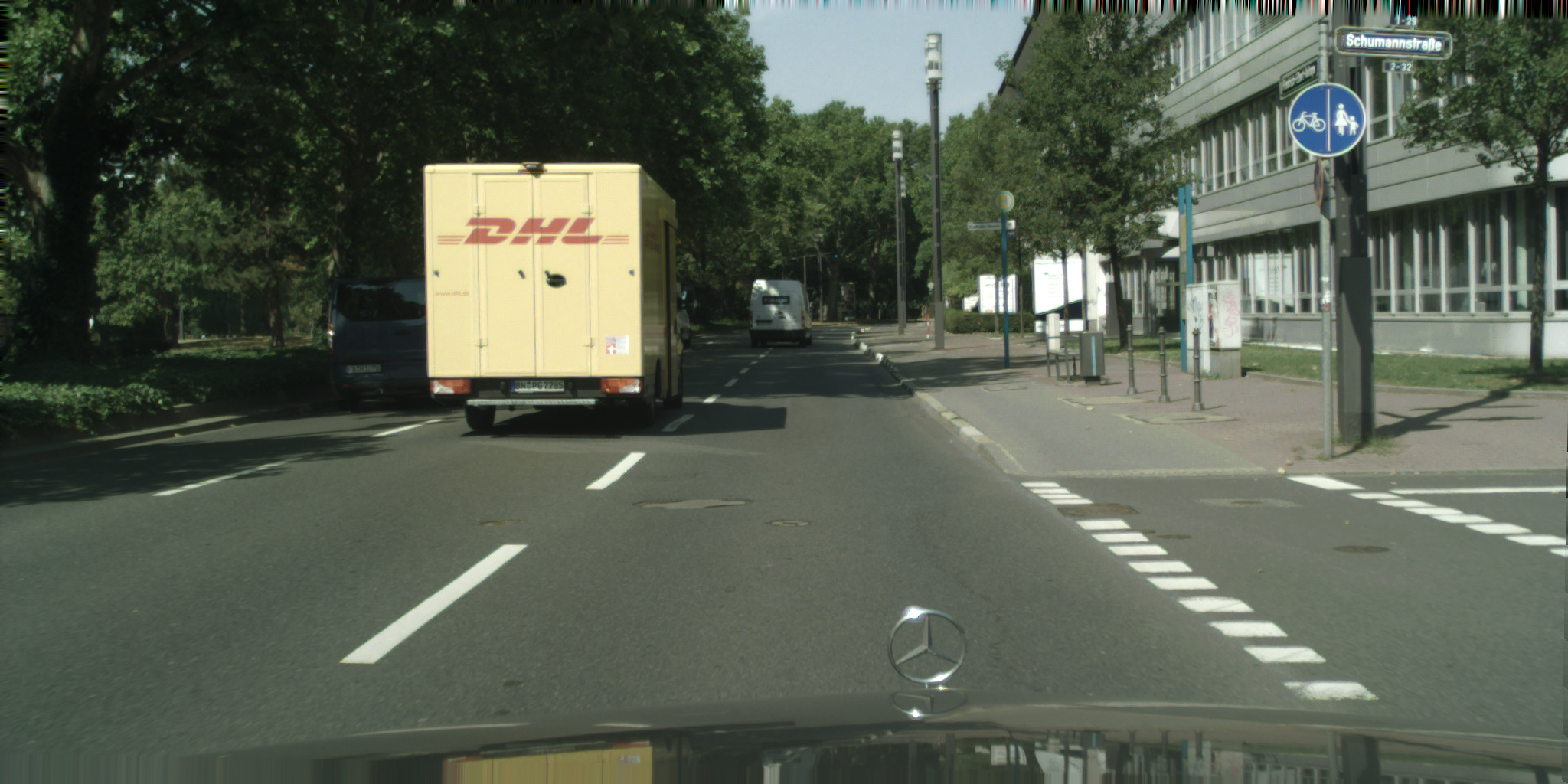} &
\includegraphics[width=\linewidth]{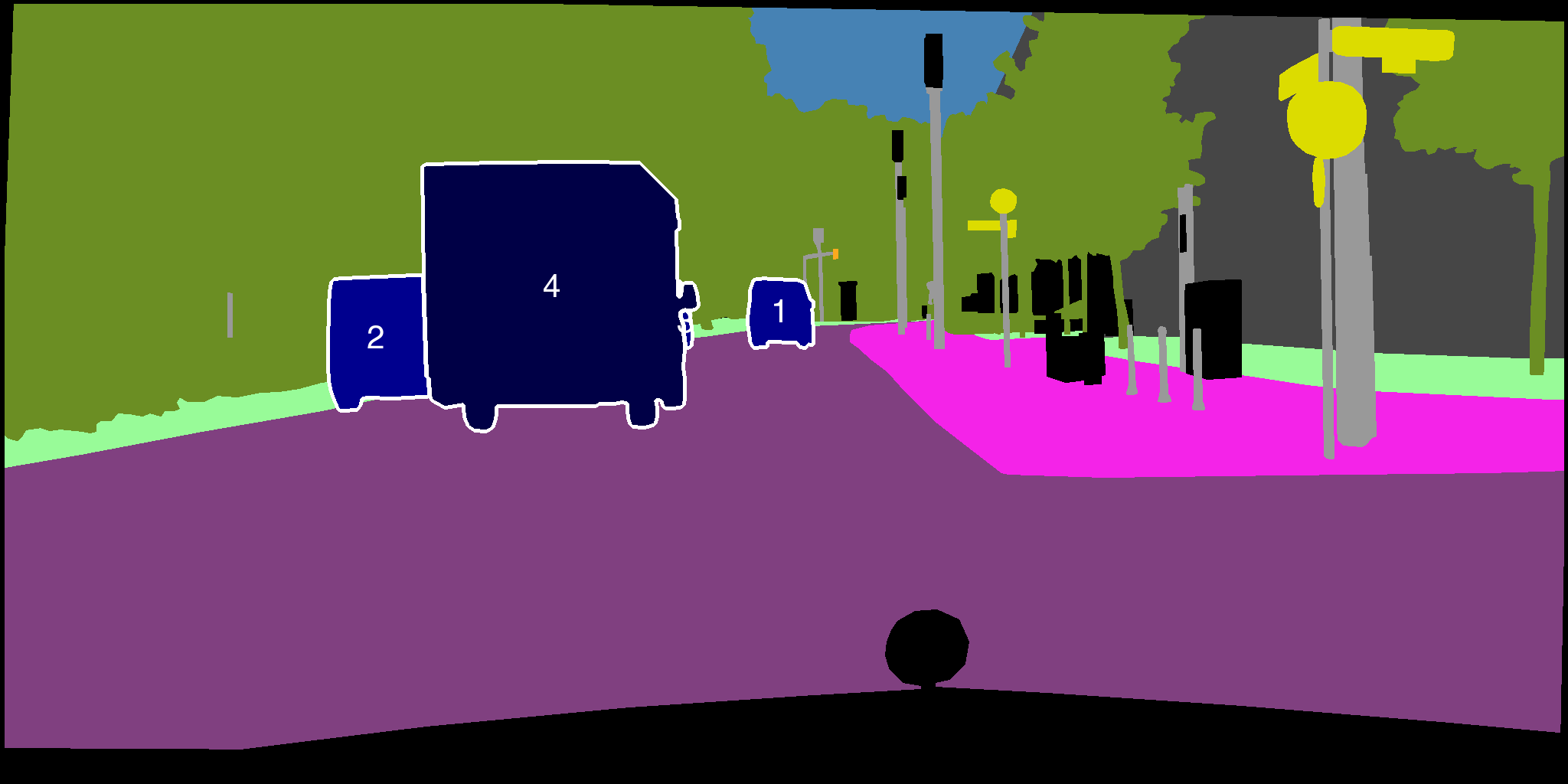} &
\includegraphics[width=\linewidth]{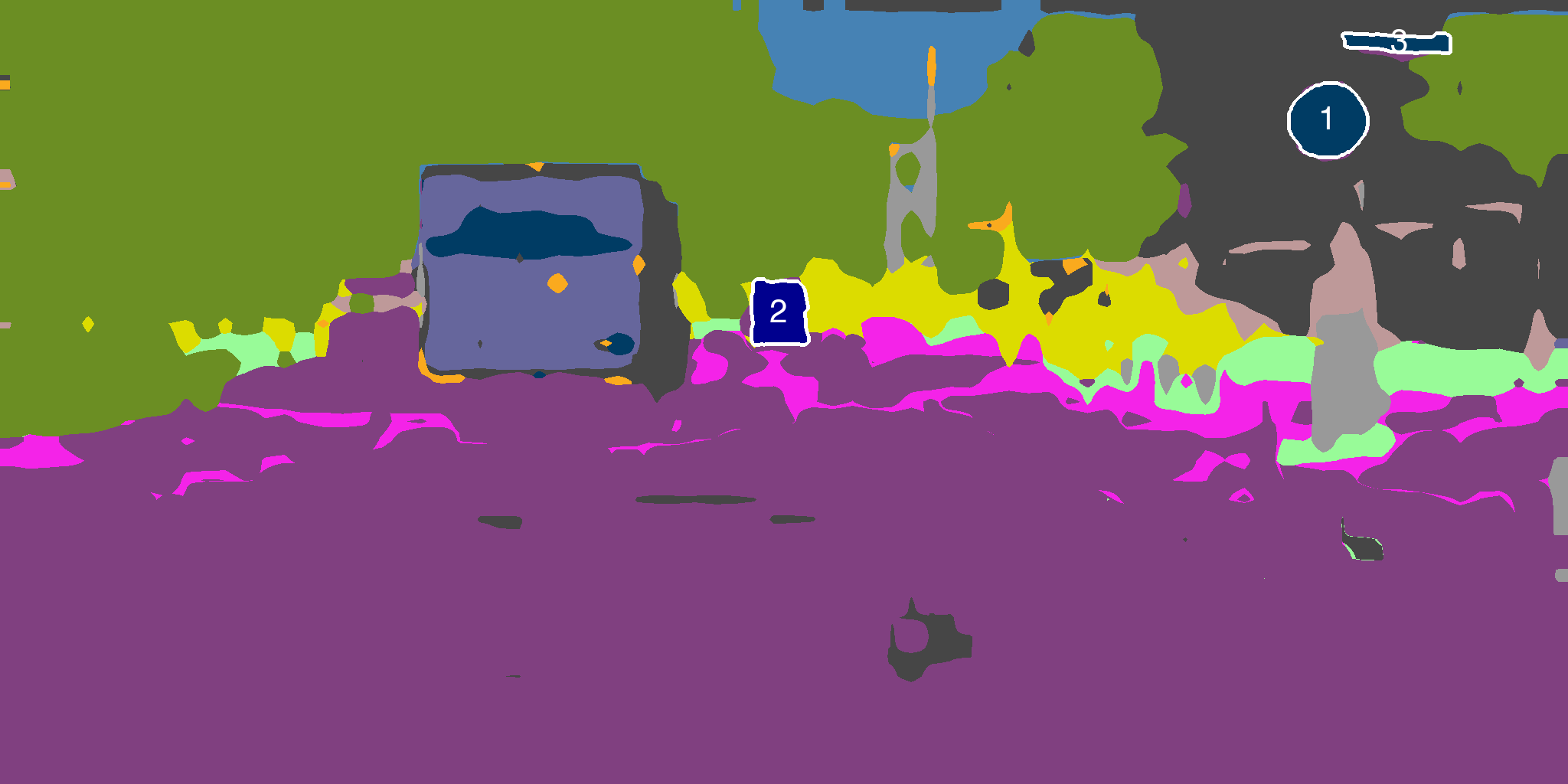} &
\includegraphics[width=\linewidth]{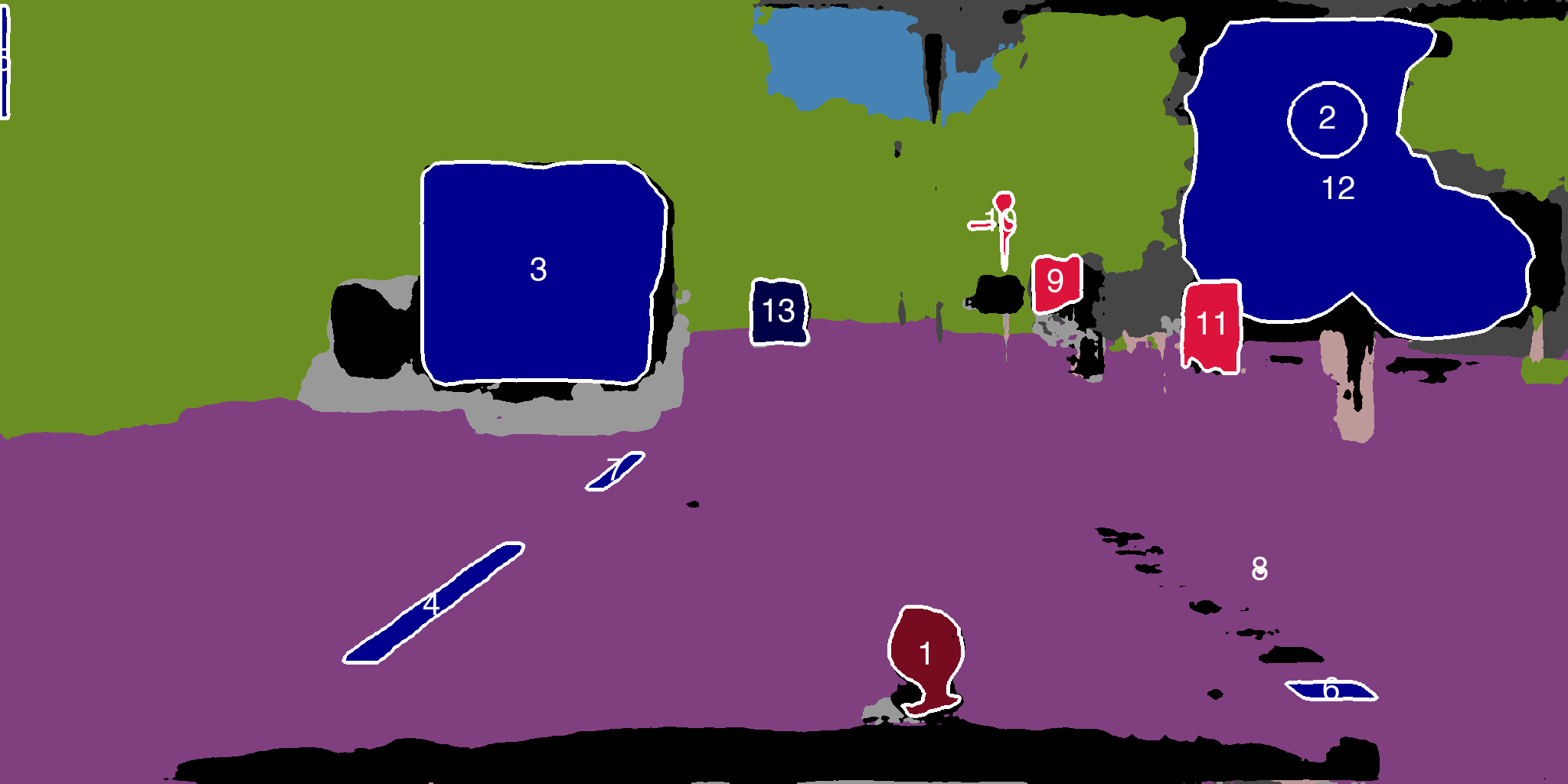} &
\includegraphics[width=\linewidth]{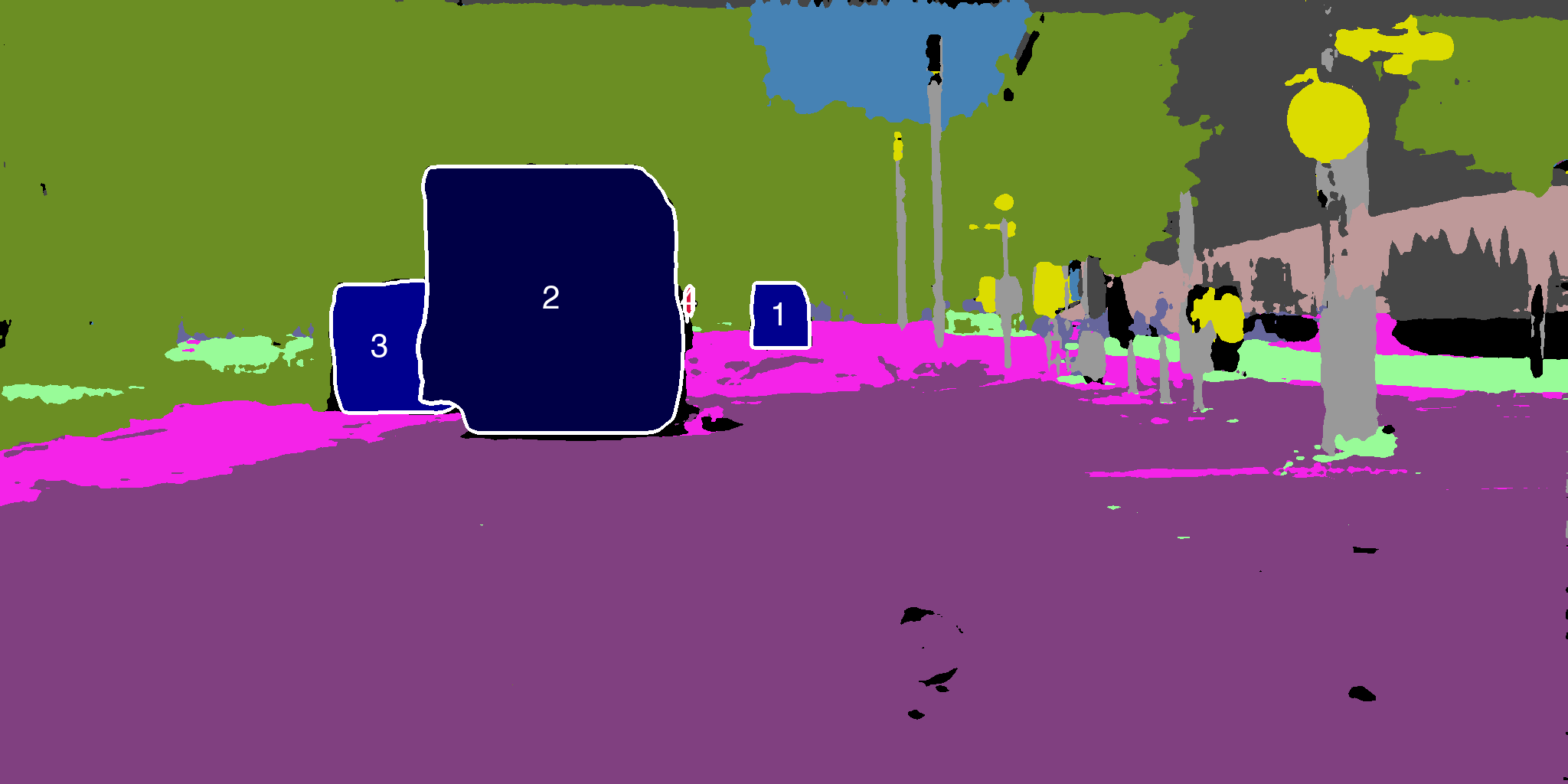} \\[-1pt]

\end{tabular}

\tiny
\renewcommand{\arraystretch}{1.3}
\begin{tabularx}{\textwidth}{*{19}{>{\centering\arraybackslash}X}}  
    \cellcolor{road}\textcolor{white}{Road}
    & \cellcolor{sidewalk}\textcolor{white}{Sidewalk}
    & \cellcolor{building}\textcolor{white}{Building}
    & \cellcolor{wall}\textcolor{white}{Wall}
    & \cellcolor{fence}\textcolor{white}{Fence}
    & \cellcolor{pole}\textcolor{white}{Pole}
    & \cellcolor{trafficlight}\textcolor{white}{Traffic~Light}
    & \cellcolor{trafficsign}\textcolor{white}{Traffic~Sign}
    & \cellcolor{vegetation}\textcolor{white}{Vegetation}
    & \cellcolor{terrain}\textcolor{white}{Terrain}
    & \cellcolor{sky}\textcolor{white}{Sky}
    & \cellcolor{person}\textcolor{white}{Person}
    & \cellcolor{rider}\textcolor{white}{Rider}
    & \cellcolor{car}\textcolor{white}{Car}
    & \cellcolor{truck}\textcolor{white}{Truck}
    & \cellcolor{bus}\textcolor{white}{Bus}
    & \cellcolor{train}\textcolor{white}{Train}
    & \cellcolor{motorcycle}\textcolor{white}{Motorcycle}
    & \cellcolor{bicycle}\textcolor{white}{Bicycle}
\end{tabularx}

        \vspace{-0.5em}
        \caption{\textbf{Cityscapes} --- Qualitative unsupervised panoptic segmentation examples. \label{fig:qualitative_cs}}
    \end{subfigure}\vspace{3em}\\

    \begin{subfigure}[t]{\textwidth}
        \centering
        \small
\sffamily
\setlength{\tabcolsep}{0pt}
\renewcommand{\arraystretch}{0.0}
\begin{tabular}{>{\centering\arraybackslash} m{0.2\textwidth} 
                >{\centering\arraybackslash} m{0.2\textwidth} 
                >{\centering\arraybackslash} m{0.2\textwidth}
                >{\centering\arraybackslash} m{0.2\textwidth}
                >{\centering\arraybackslash} m{0.2\textwidth}}

{Image} & {Ground Truth} & {Baseline} & {U2Seg~\cite{Niu:2024:UUI}} & {\MethodName \textit{(Ours)}} \\[4pt]

\includegraphics[width=\linewidth]{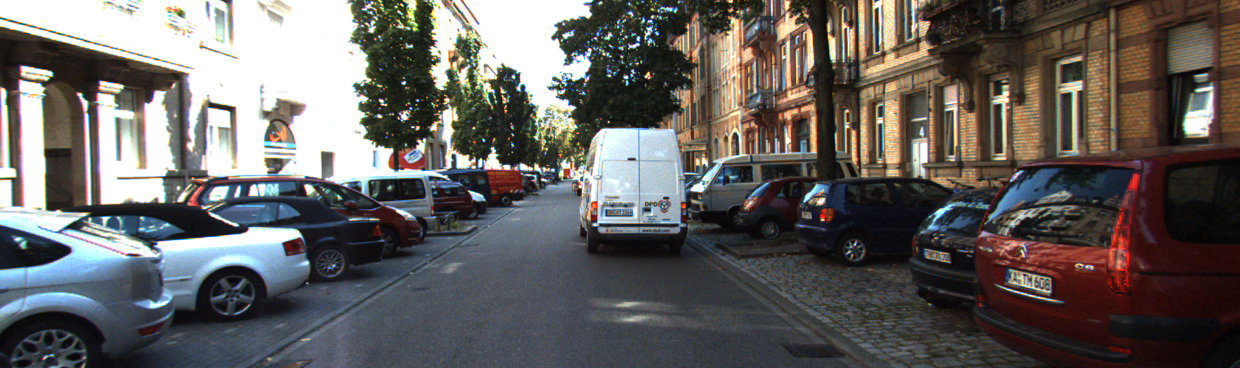} &
\includegraphics[width=\linewidth]{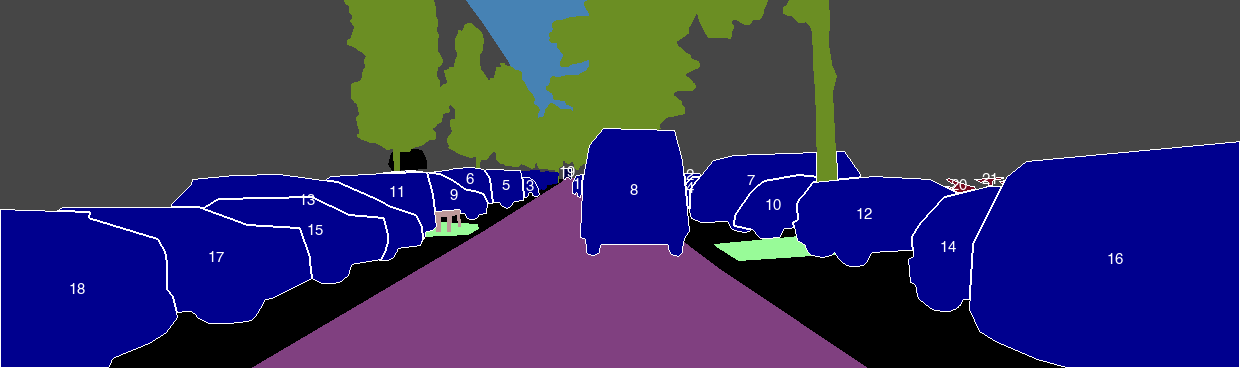} &
\includegraphics[width=\linewidth]{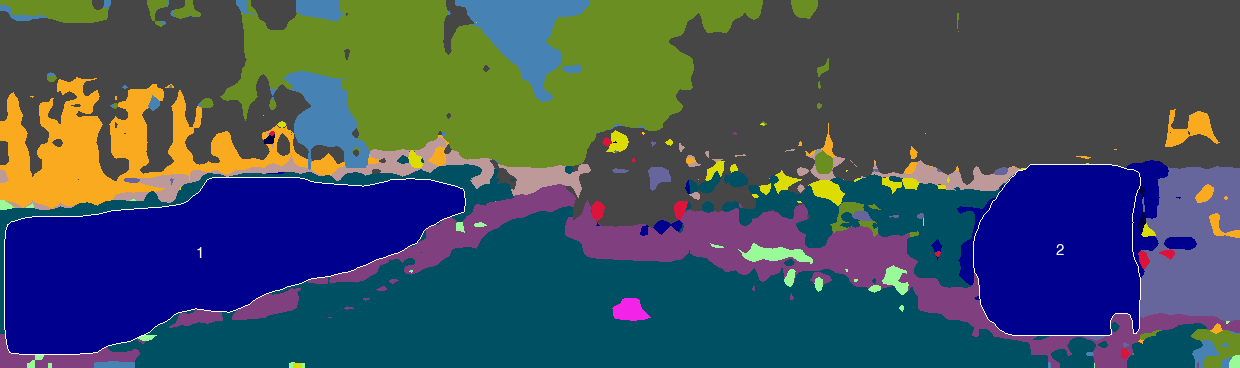} &
\includegraphics[width=\linewidth]{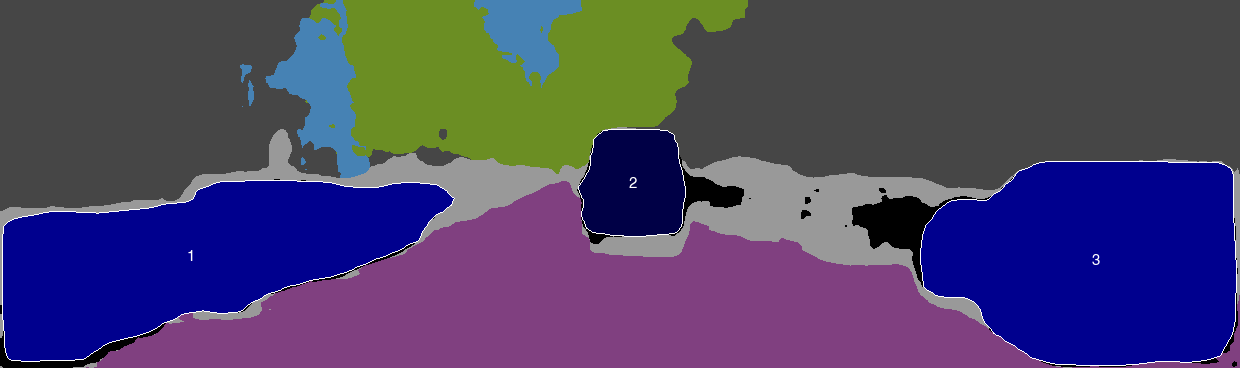} &
\includegraphics[width=\linewidth]{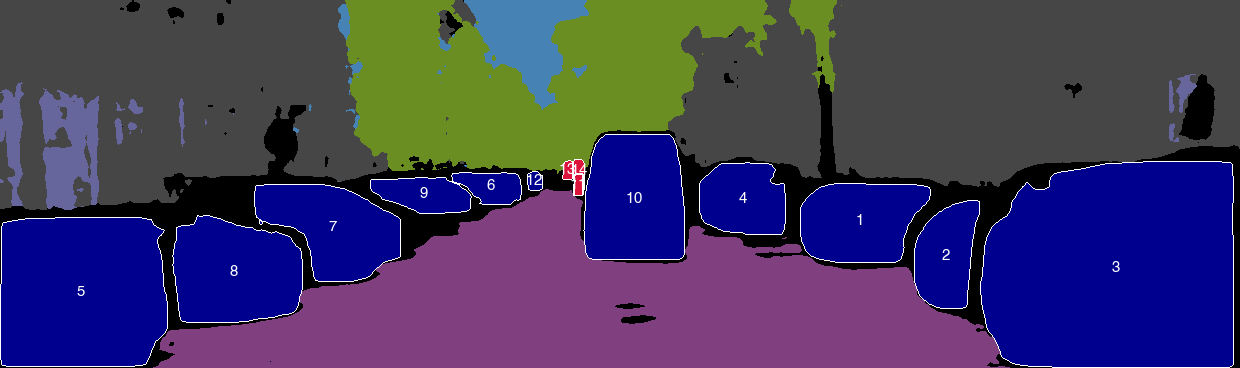} \\

\includegraphics[width=\linewidth]{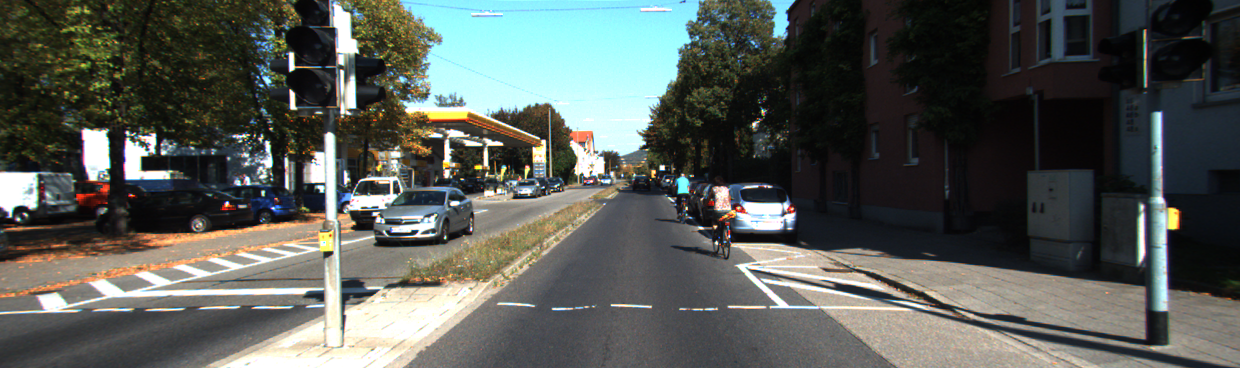} &
\includegraphics[width=\linewidth]{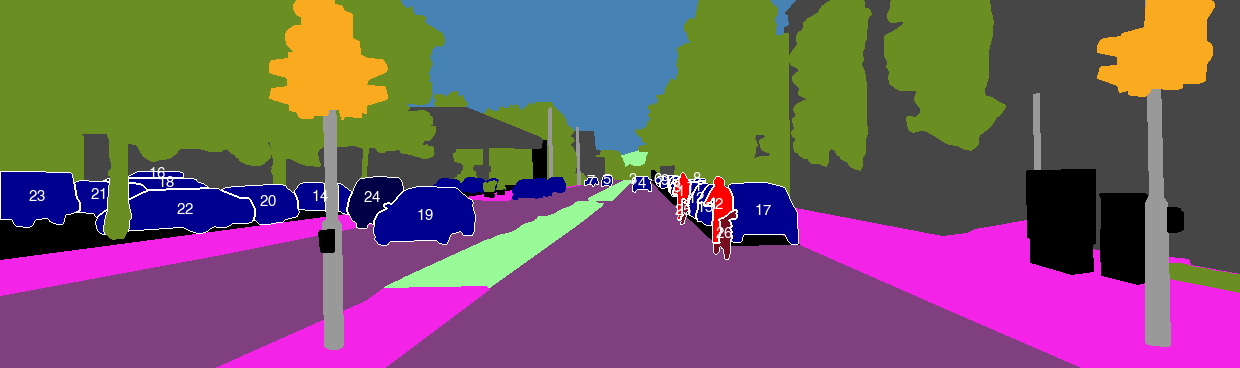} &
\includegraphics[width=\linewidth]{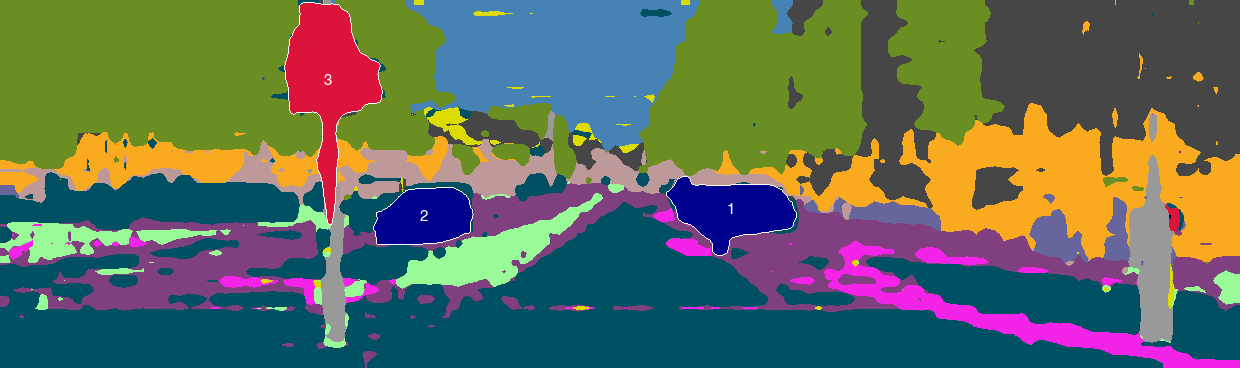} &
\includegraphics[width=\linewidth]{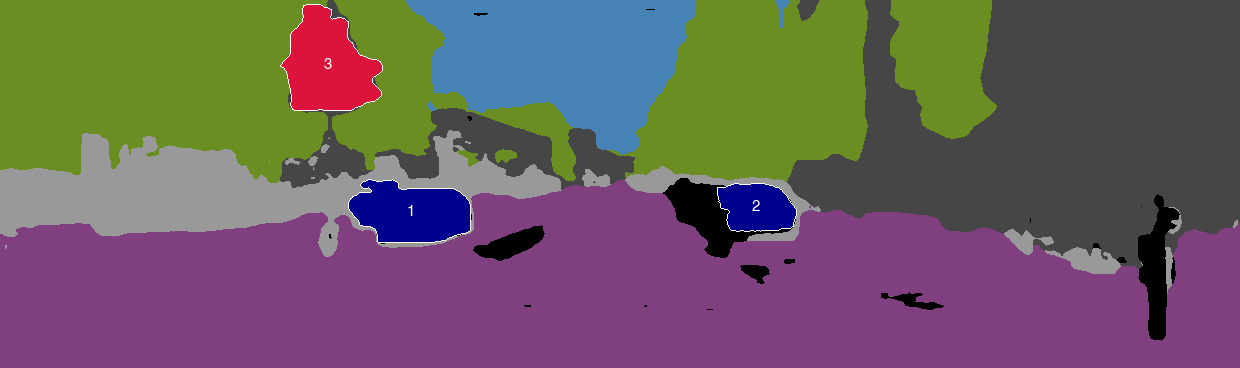} &
\includegraphics[width=\linewidth]{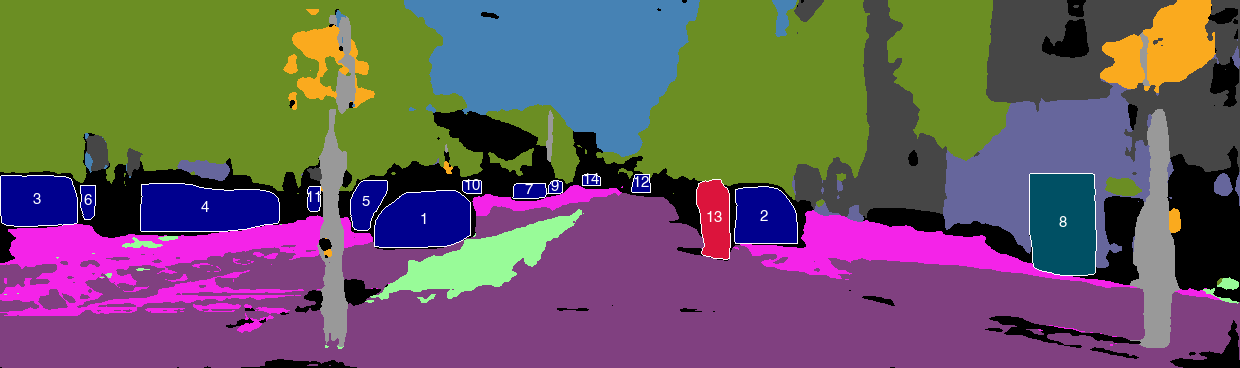} \\

\includegraphics[width=\linewidth]{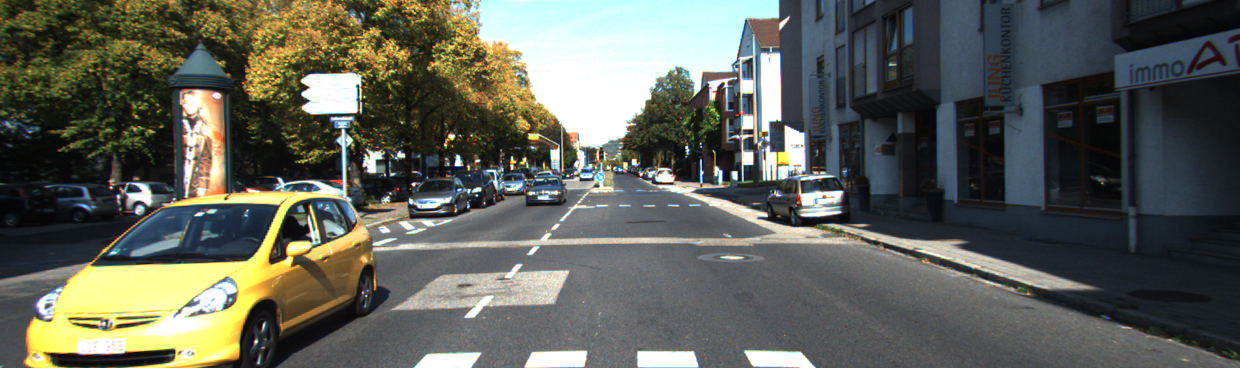} &
\includegraphics[width=\linewidth]{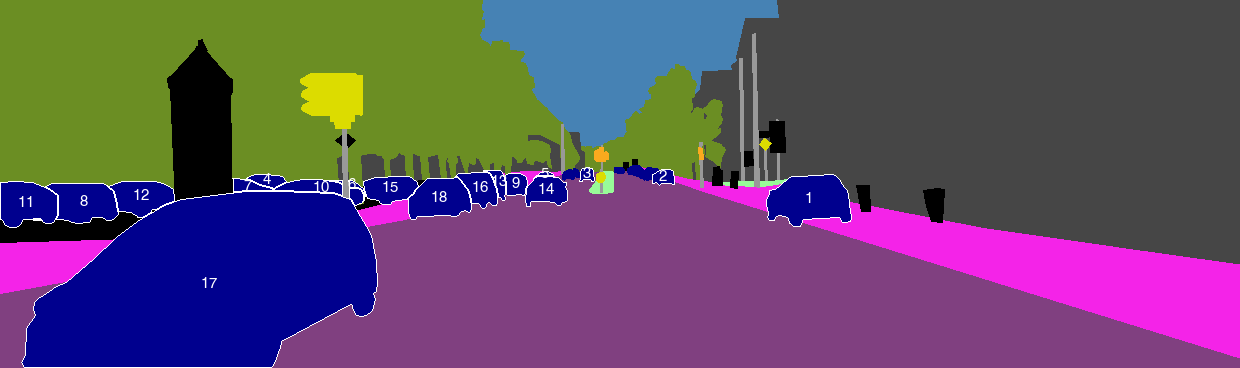} &
\includegraphics[width=\linewidth]{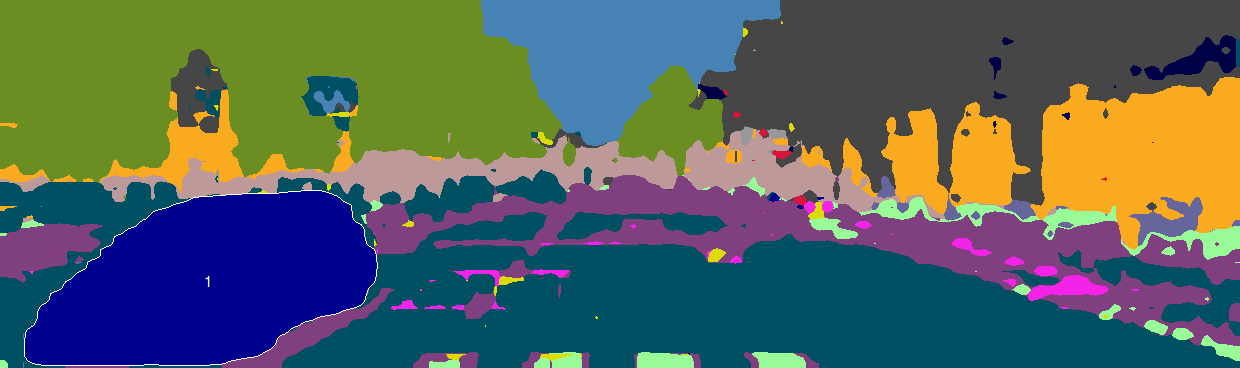} &
\includegraphics[width=\linewidth]{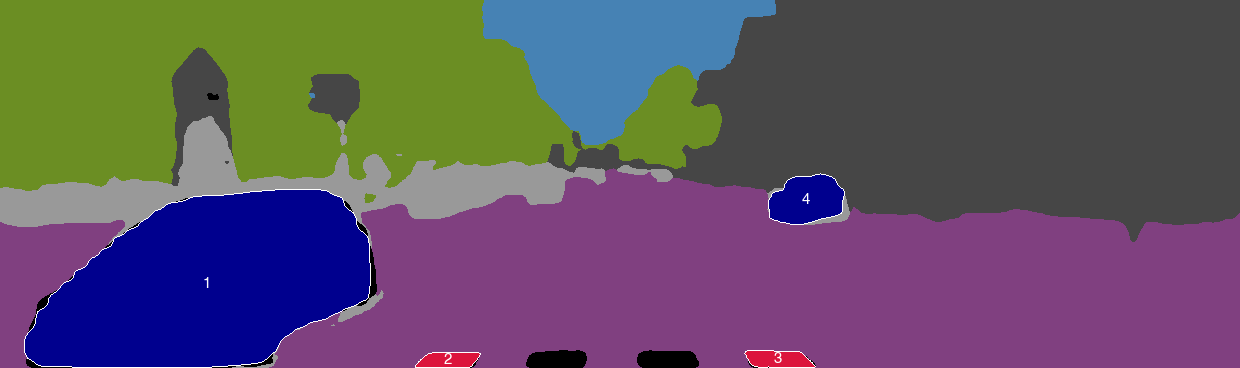} &
\includegraphics[width=\linewidth]{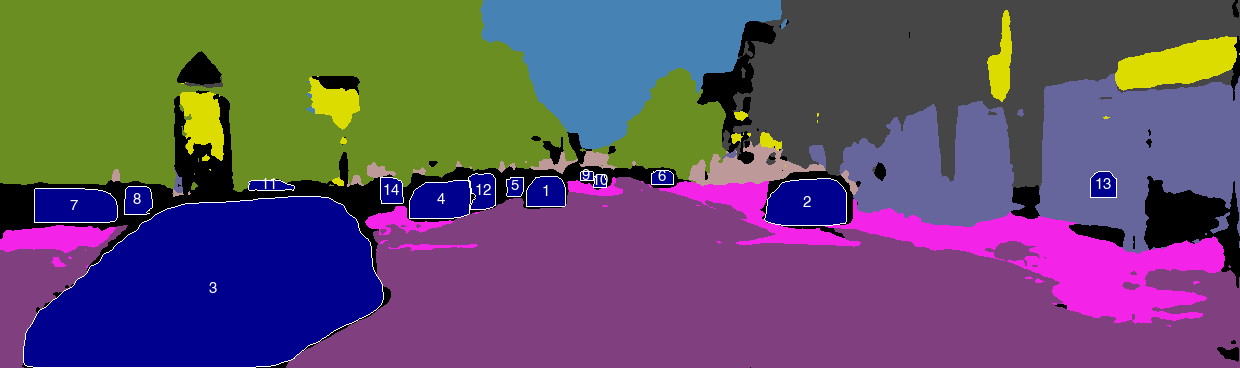} \\[-1pt]

\end{tabular}

\tiny
\renewcommand{\arraystretch}{1.3}
\begin{tabularx}{\textwidth}{*{19}{>{\centering\arraybackslash}X}}  
    \cellcolor{road}\textcolor{white}{Road}
    & \cellcolor{sidewalk}\textcolor{white}{Sidewalk}
    & \cellcolor{building}\textcolor{white}{Building}
    & \cellcolor{wall}\textcolor{white}{Wall}
    & \cellcolor{fence}\textcolor{white}{Fence}
    & \cellcolor{pole}\textcolor{white}{Pole}
    & \cellcolor{trafficlight}\textcolor{white}{Traffic~Light}
    & \cellcolor{trafficsign}\textcolor{white}{Traffic~Sign}
    & \cellcolor{vegetation}\textcolor{white}{Vegetation}
    & \cellcolor{terrain}\textcolor{white}{Terrain}
    & \cellcolor{sky}\textcolor{white}{Sky}
    & \cellcolor{person}\textcolor{white}{Person}
    & \cellcolor{rider}\textcolor{white}{Rider}
    & \cellcolor{car}\textcolor{white}{Car}
    & \cellcolor{truck}\textcolor{white}{Truck}
    & \cellcolor{bus}\textcolor{white}{Bus}
    & \cellcolor{train}\textcolor{white}{Train}
    & \cellcolor{motorcycle}\textcolor{white}{Motorcycle}
    & \cellcolor{bicycle}\textcolor{white}{Bicycle}
\end{tabularx}

        \vspace{-0.5em}
        \caption{\textbf{KITTI} --- Qualitative unsupervised panoptic segmentation examples.\label{fig:qualitative_kitti}}
    \end{subfigure}\vspace{3em}\\

    \ContinuedFloat
    \begin{subfigure}[t]{\textwidth}
        \centering
        \small
\sffamily
\setlength{\tabcolsep}{0pt}
\renewcommand{\arraystretch}{0.0}
\begin{tabular}{>{\centering\arraybackslash} m{0.2\textwidth} 
                >{\centering\arraybackslash} m{0.2\textwidth} 
                >{\centering\arraybackslash} m{0.2\textwidth}
                >{\centering\arraybackslash} m{0.2\textwidth}
                >{\centering\arraybackslash} m{0.2\textwidth}}

{Image} & {Ground Truth} & {Baseline} & {U2Seg~\cite{Niu:2024:UUI}} & {\MethodName \textit{(Ours)}} \\[4pt]

\includegraphics[width=\linewidth]{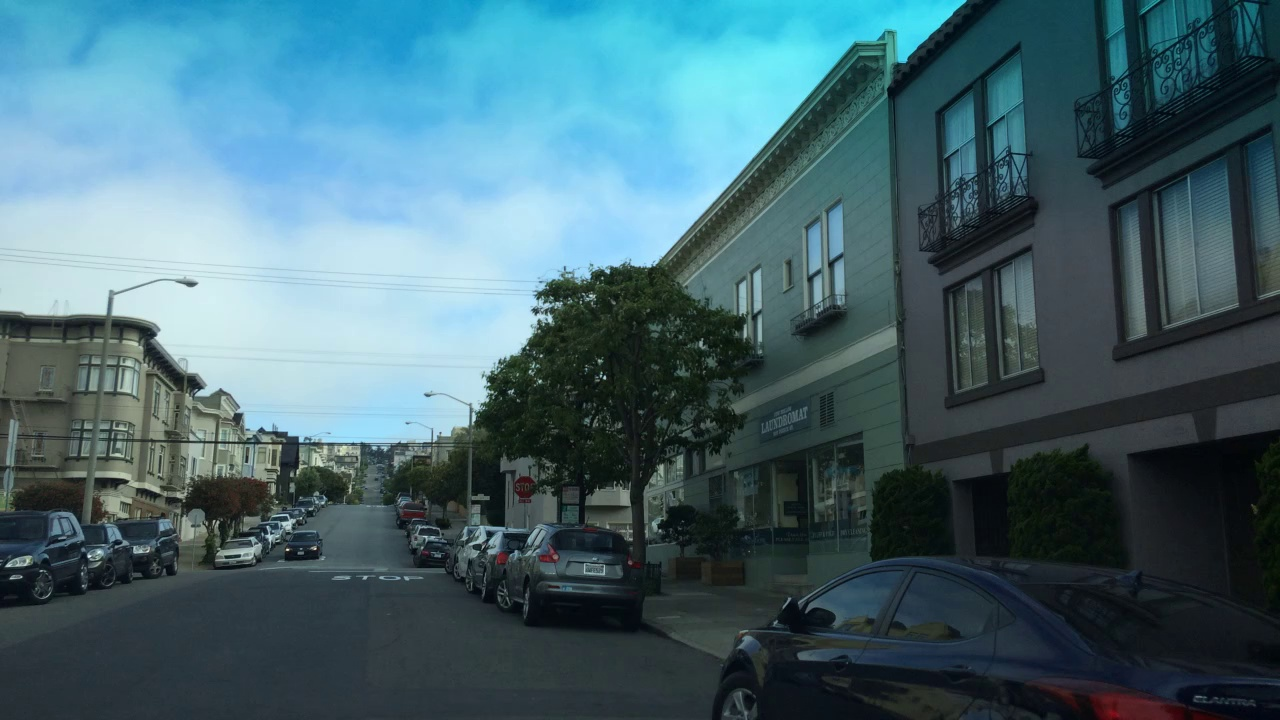} &
\includegraphics[width=\linewidth]{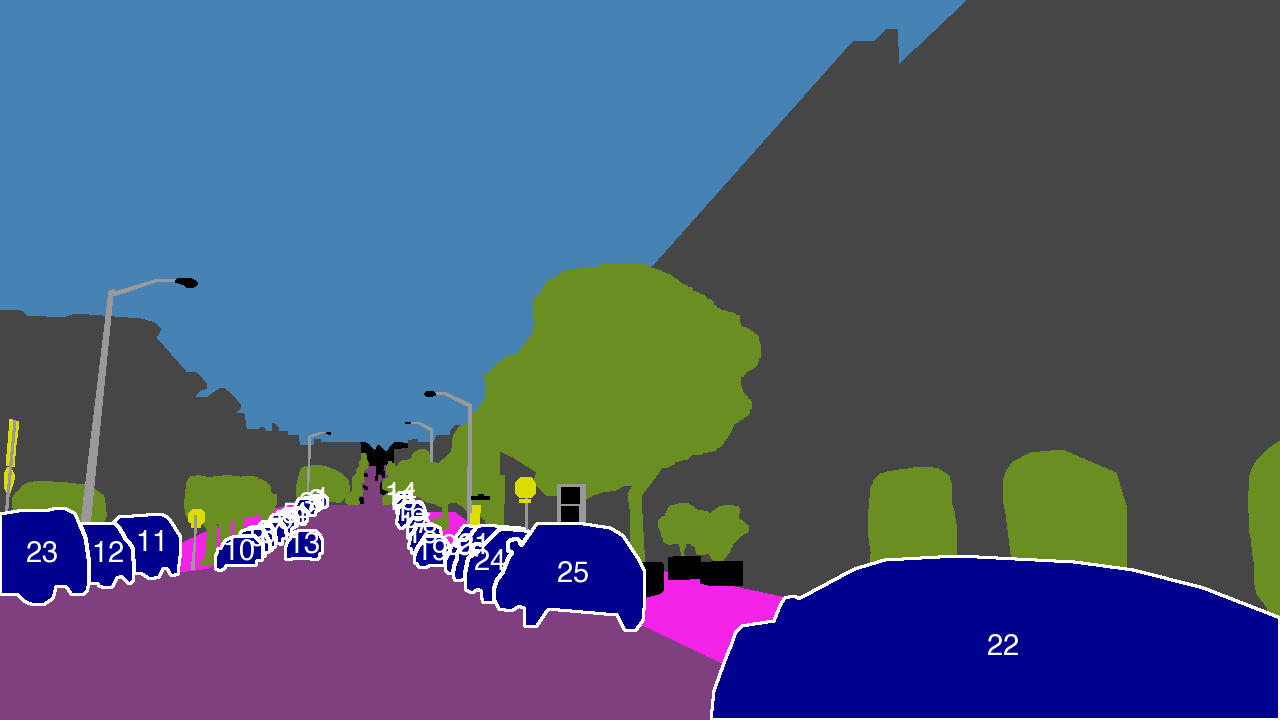} &
\includegraphics[width=\linewidth]{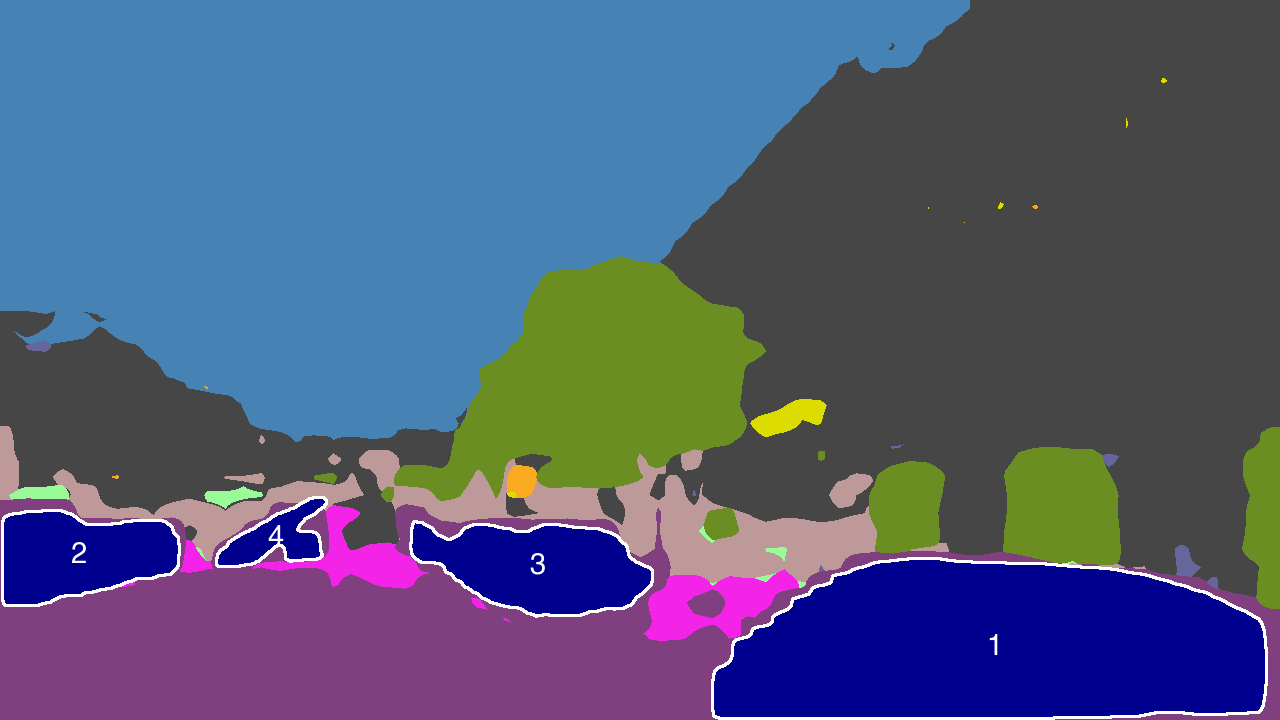} &
\includegraphics[width=\linewidth]{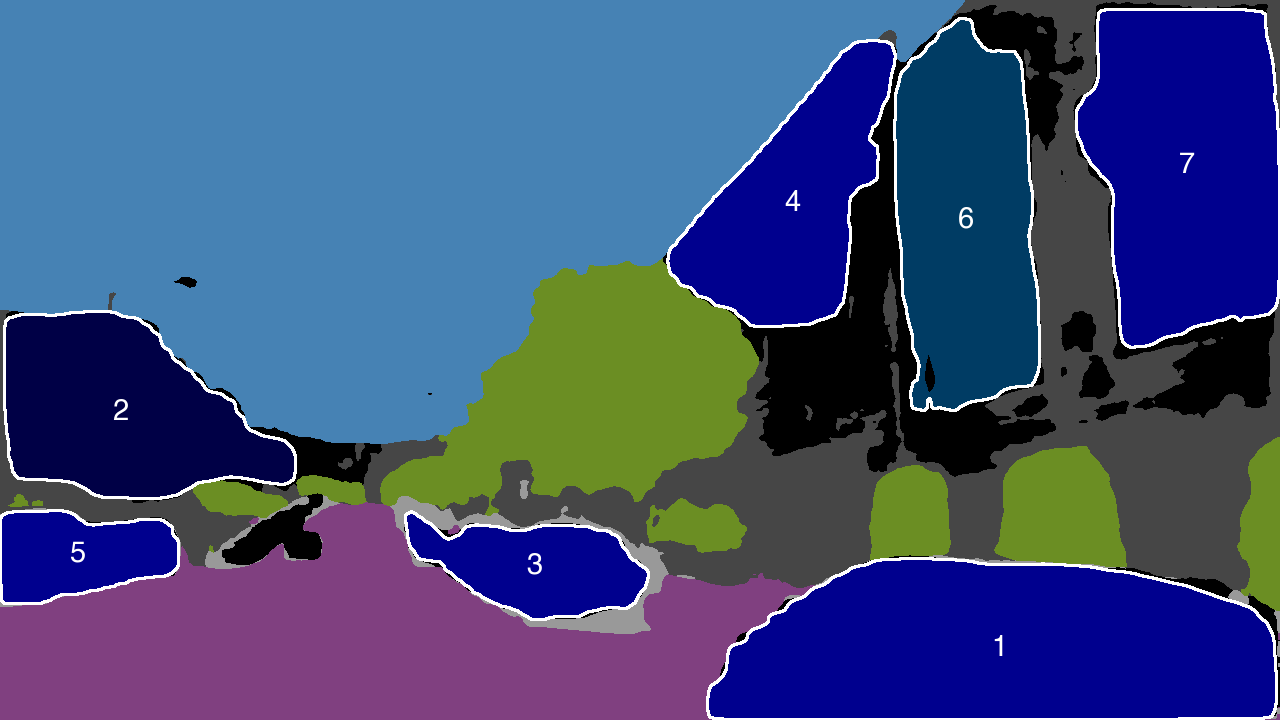} &
\includegraphics[width=\linewidth]{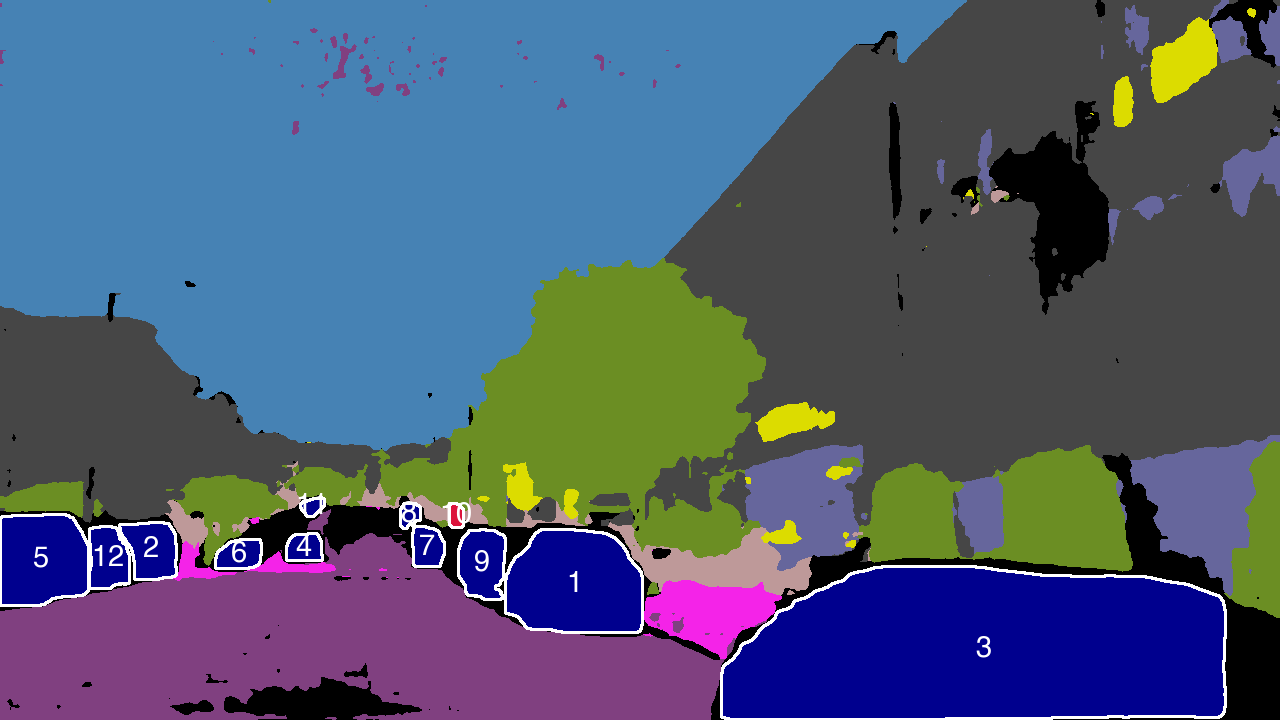} \\

\includegraphics[width=\linewidth]{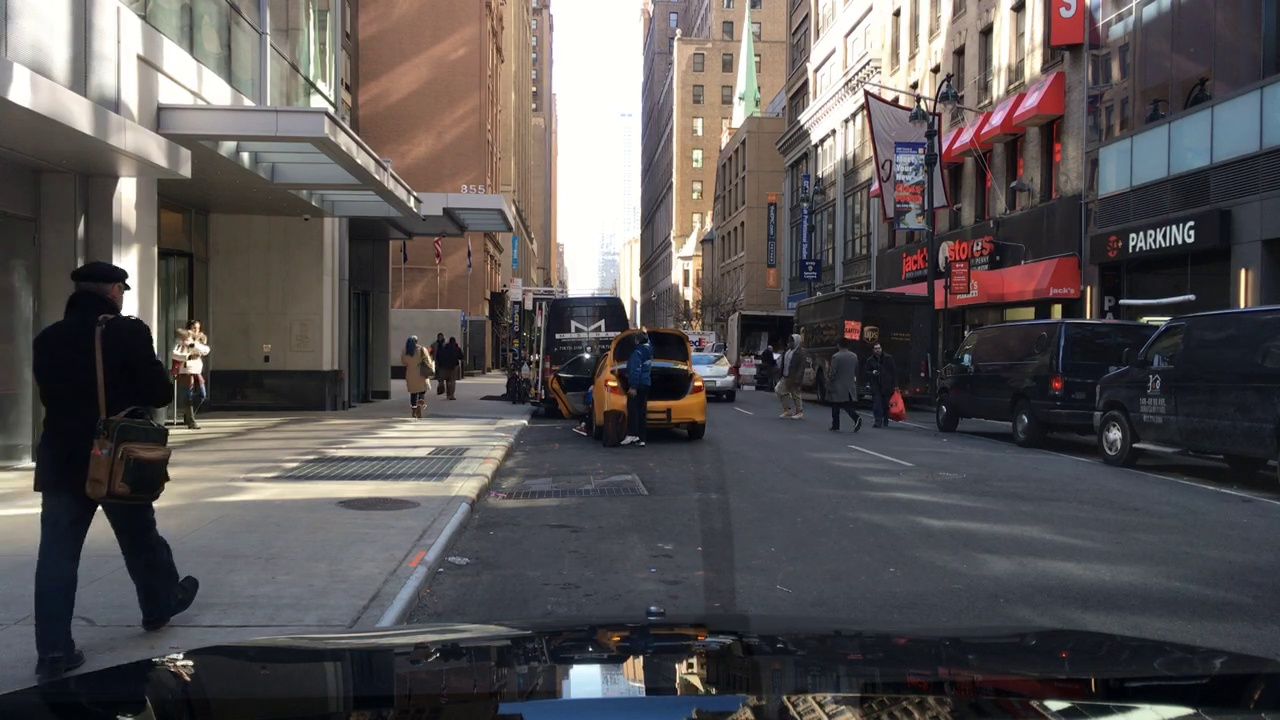} &
\includegraphics[width=\linewidth]{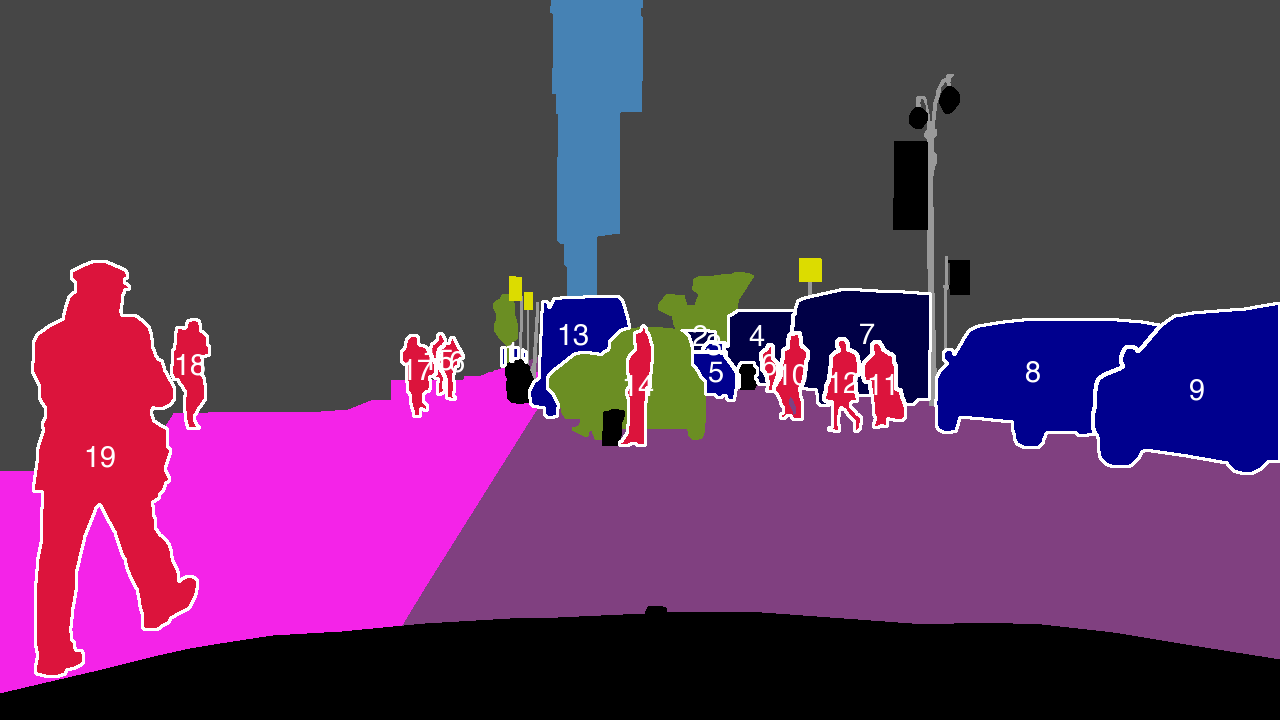} &
\includegraphics[width=\linewidth]{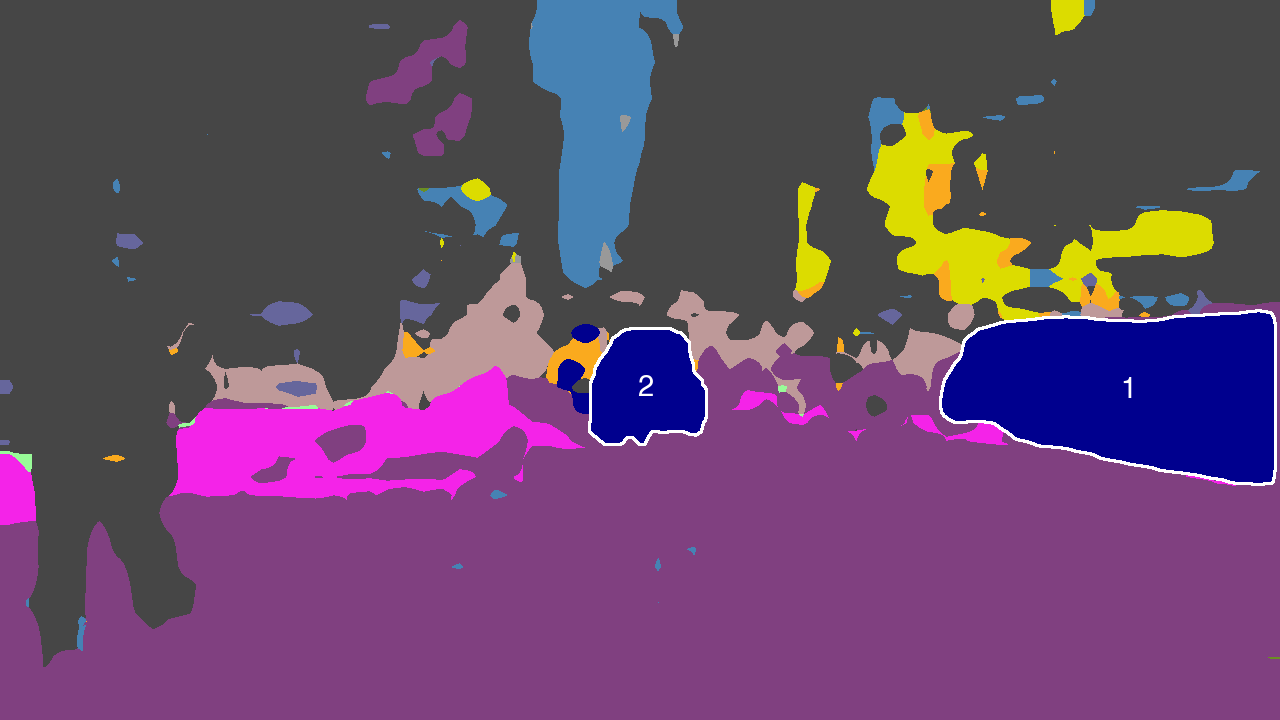} &
\includegraphics[width=\linewidth]{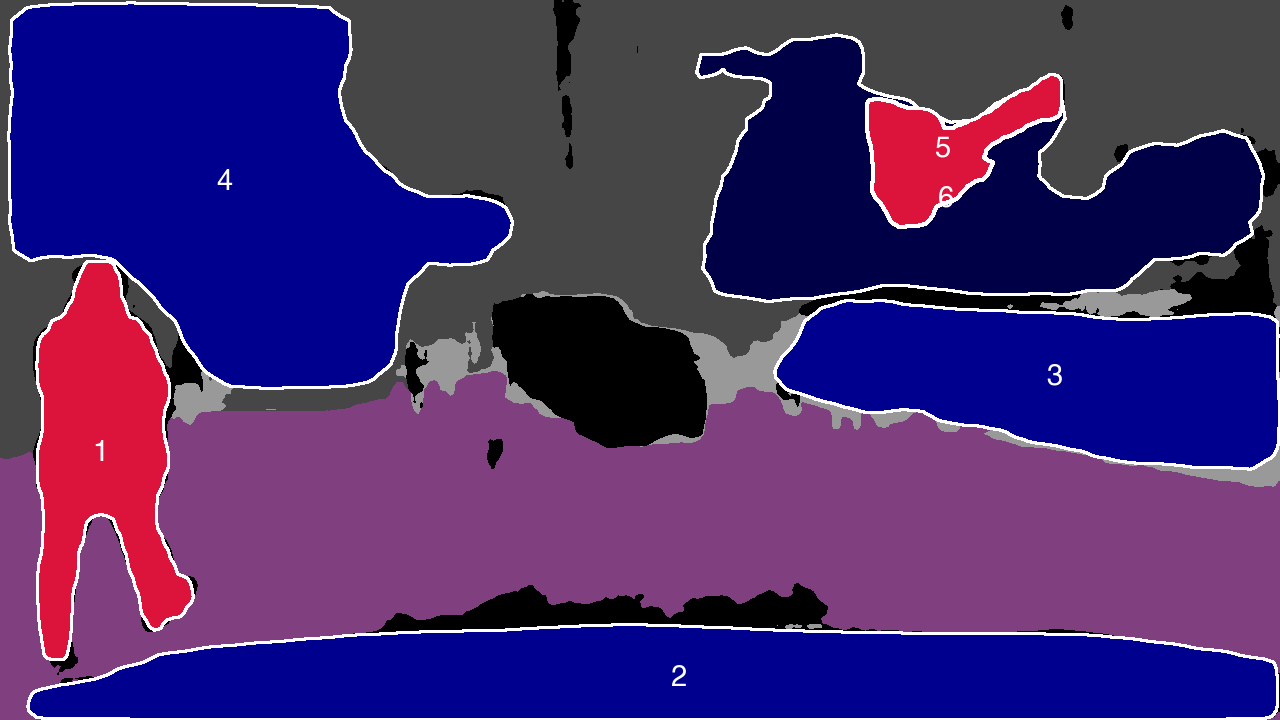} &
\includegraphics[width=\linewidth]{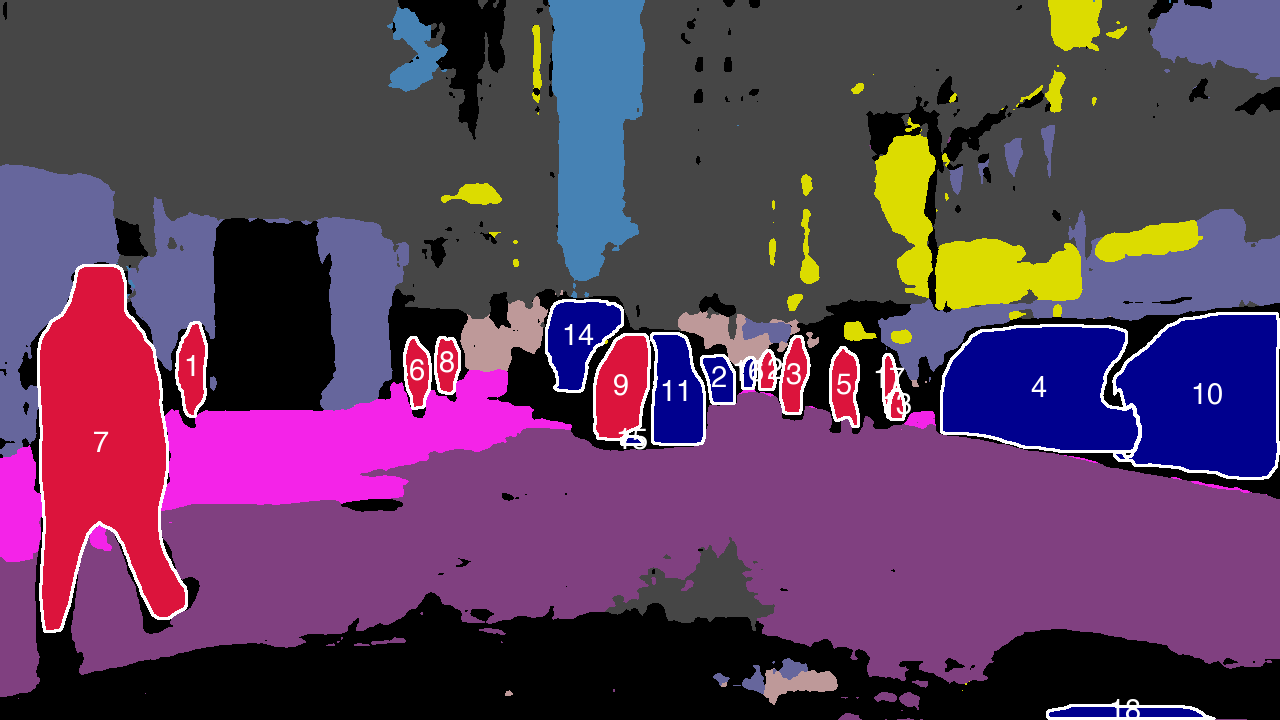} \\

\includegraphics[width=\linewidth]{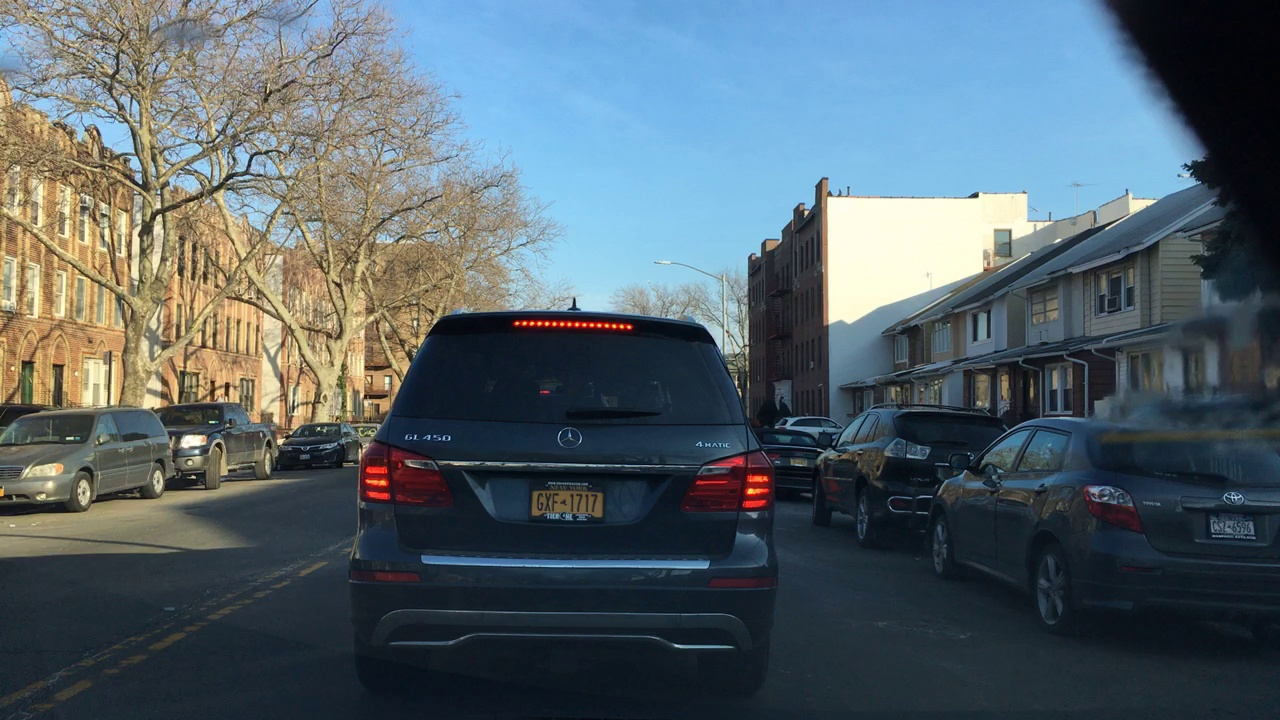} &
\includegraphics[width=\linewidth]{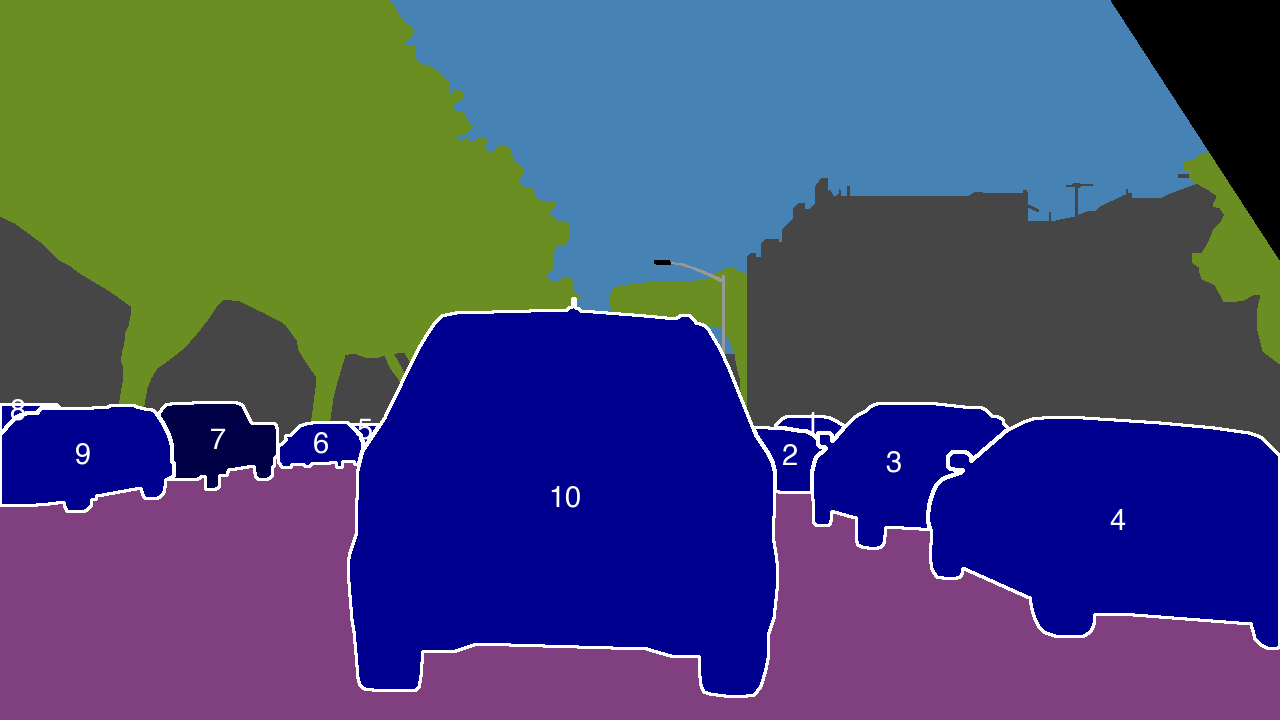} &
\includegraphics[width=\linewidth]{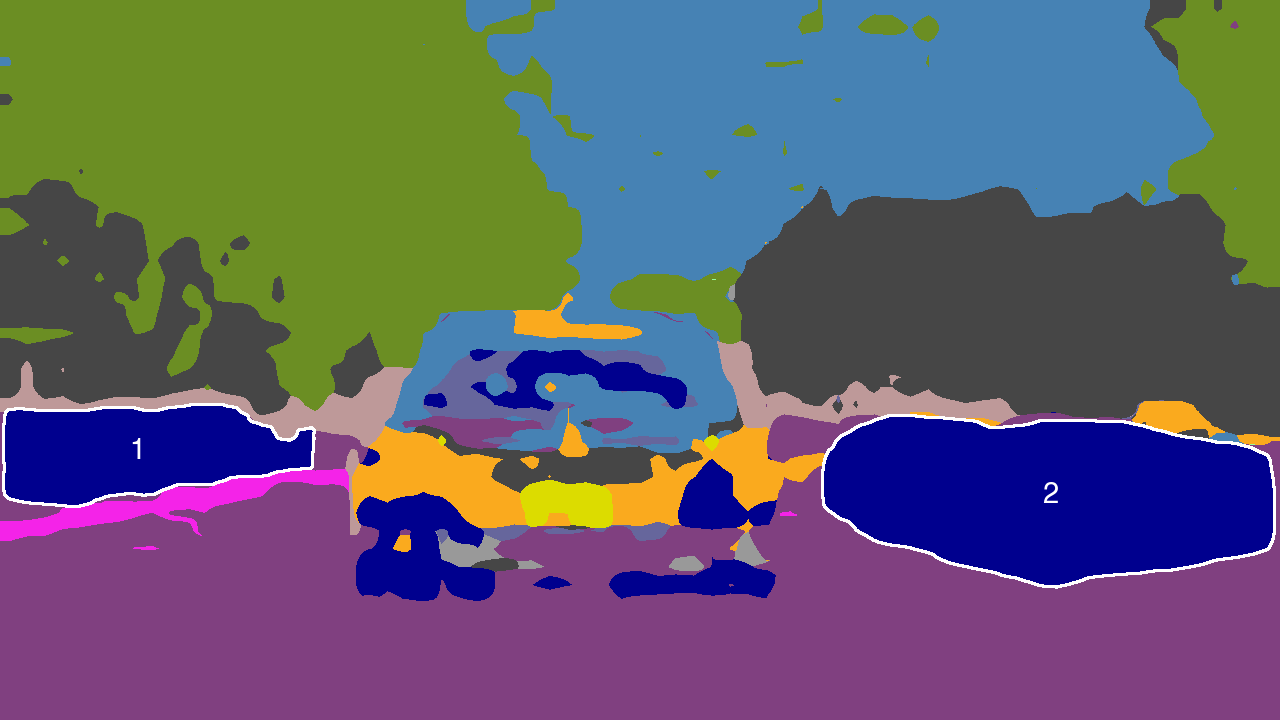} &
\includegraphics[width=\linewidth]{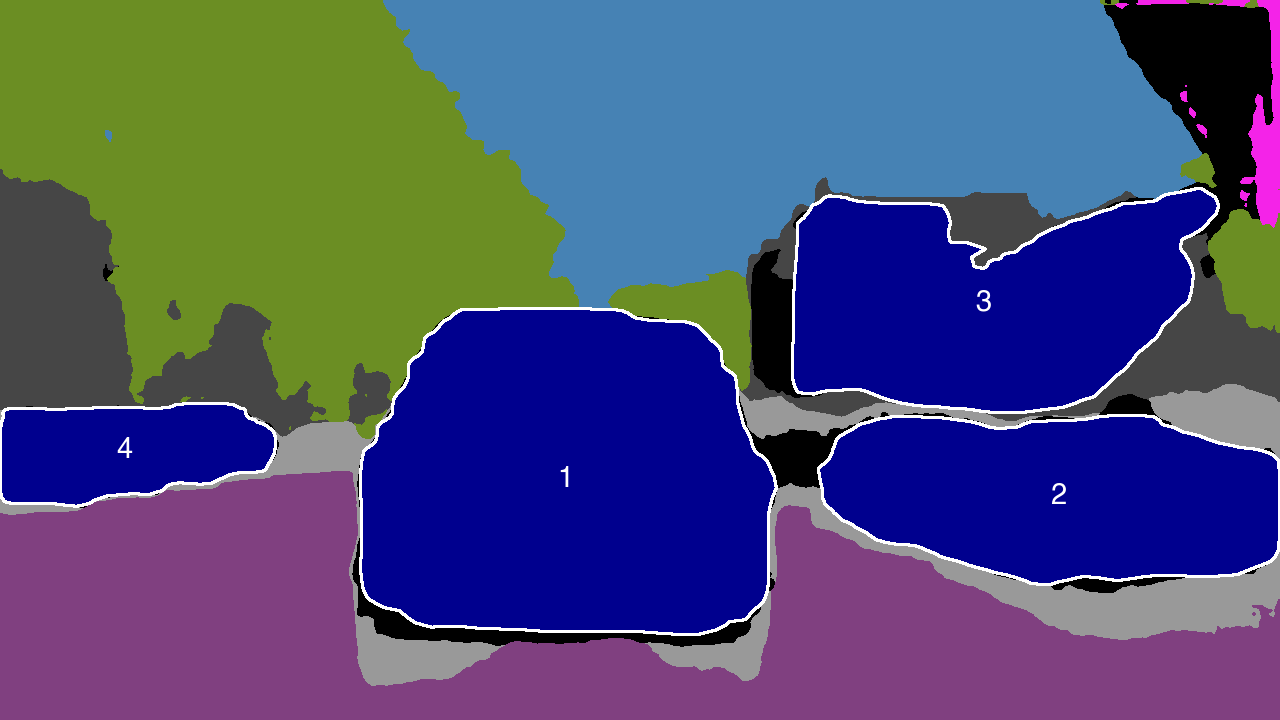} &
\includegraphics[width=\linewidth]{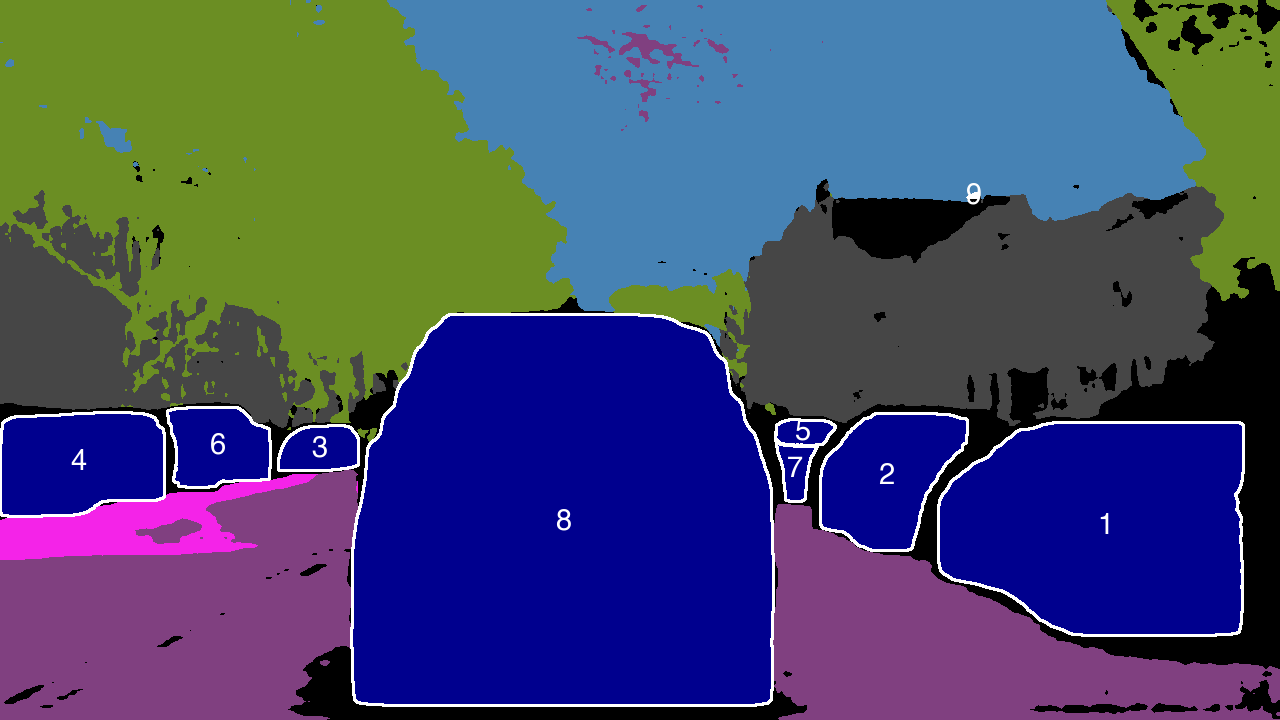} \\[-1pt]

\end{tabular}

\tiny
\renewcommand{\arraystretch}{1.3}
\begin{tabularx}{\textwidth}{*{19}{>{\centering\arraybackslash}X}}  
    \cellcolor{road}\textcolor{white}{Road}
    & \cellcolor{sidewalk}\textcolor{white}{Sidewalk}
    & \cellcolor{building}\textcolor{white}{Building}
    & \cellcolor{wall}\textcolor{white}{Wall}
    & \cellcolor{fence}\textcolor{white}{Fence}
    & \cellcolor{pole}\textcolor{white}{Pole}
    & \cellcolor{trafficlight}\textcolor{white}{Traffic~Light}
    & \cellcolor{trafficsign}\textcolor{white}{Traffic~Sign}
    & \cellcolor{vegetation}\textcolor{white}{Vegetation}
    & \cellcolor{terrain}\textcolor{white}{Terrain}
    & \cellcolor{sky}\textcolor{white}{Sky}
    & \cellcolor{person}\textcolor{white}{Person}
    & \cellcolor{rider}\textcolor{white}{Rider}
    & \cellcolor{car}\textcolor{white}{Car}
    & \cellcolor{truck}\textcolor{white}{Truck}
    & \cellcolor{bus}\textcolor{white}{Bus}
    & \cellcolor{train}\textcolor{white}{Train}
    & \cellcolor{motorcycle}\textcolor{white}{Motorcycle}
    & \cellcolor{bicycle}\textcolor{white}{Bicycle}
\end{tabularx}

        \vspace{-0.5em}
        \caption{\textbf{BDD} --- Qualitative unsupervised panoptic segmentation examples.\label{fig:qualitative_bdd}}
    \end{subfigure}\\

\end{figure*}
\begin{figure*}
\ContinuedFloat
    \vspace{-0.5em}
    \begin{subfigure}[t]{\textwidth}
        \centering
        \small
\sffamily
\setlength{\tabcolsep}{0pt}
\renewcommand{\arraystretch}{0.0}
\begin{tabular}{>{\centering\arraybackslash} m{0.2\textwidth} 
                >{\centering\arraybackslash} m{0.2\textwidth} 
                >{\centering\arraybackslash} m{0.2\textwidth}
                >{\centering\arraybackslash} m{0.2\textwidth}
                >{\centering\arraybackslash} m{0.2\textwidth}}

{Image} & {Ground Truth} & {Baseline} & {U2Seg~\cite{Niu:2024:UUI}} & {\MethodName \textit{(Ours)}} \\[4pt]

\includegraphics[width=\linewidth]{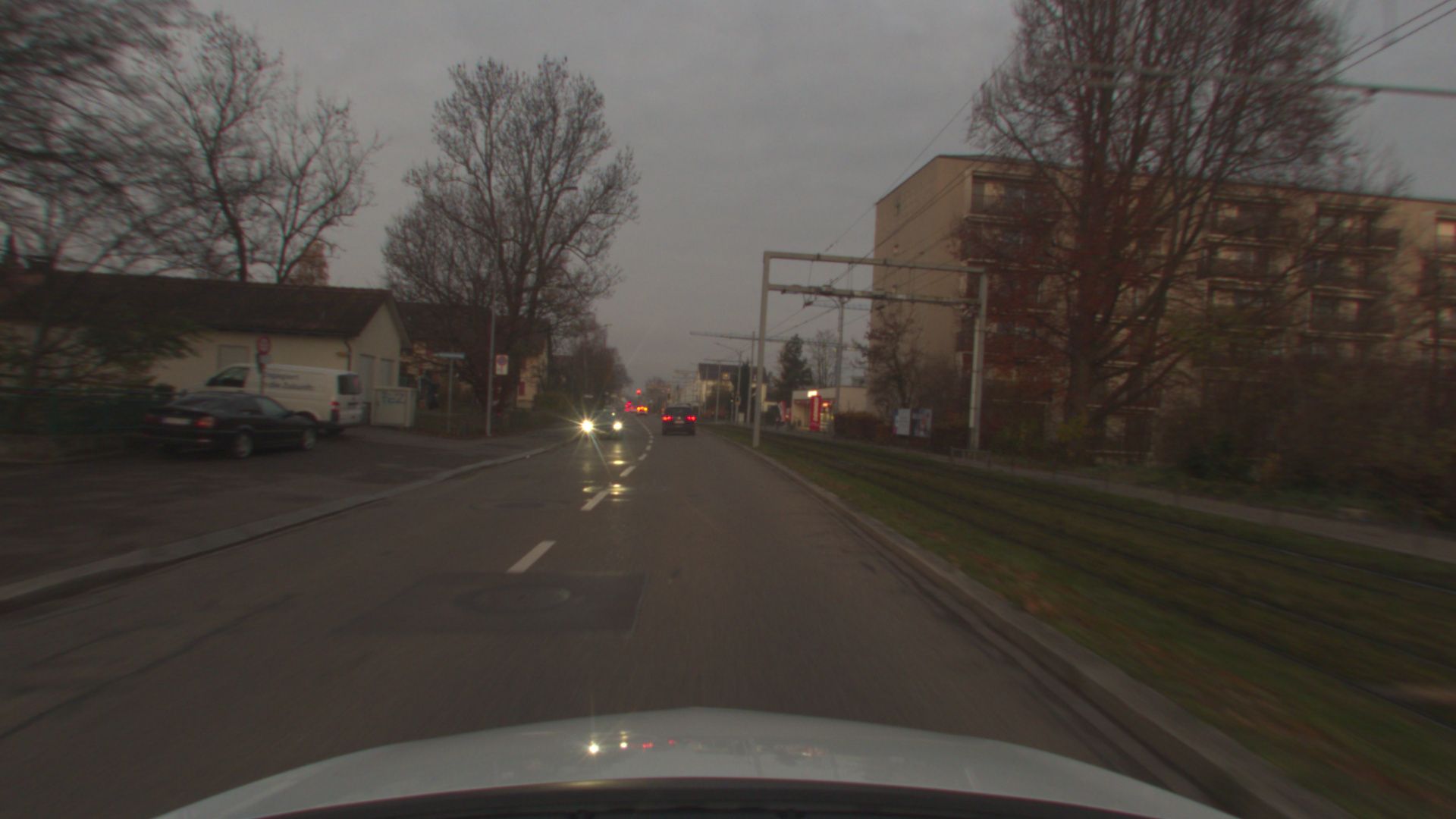} &
\includegraphics[width=\linewidth]{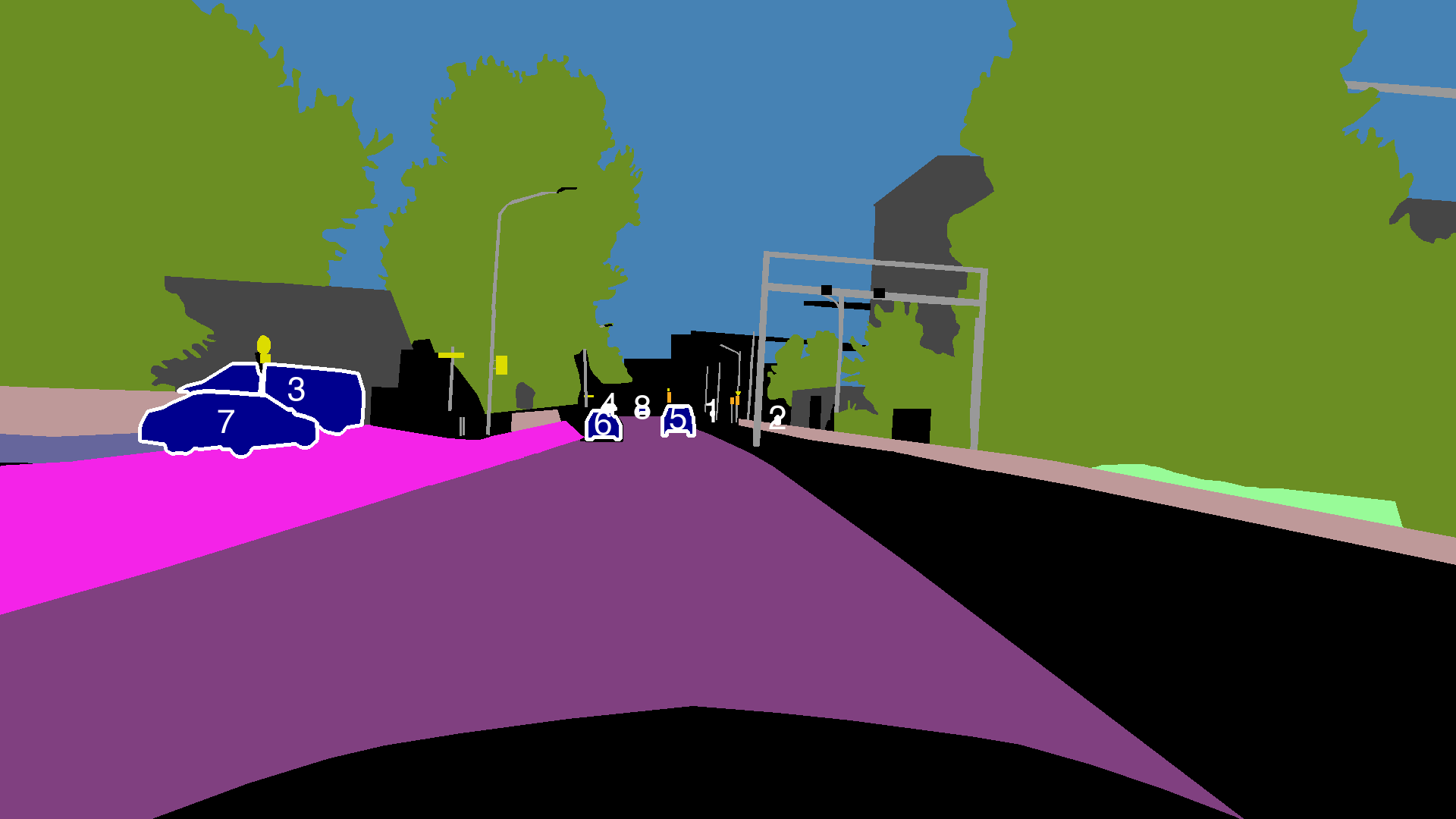} &
\includegraphics[width=\linewidth]{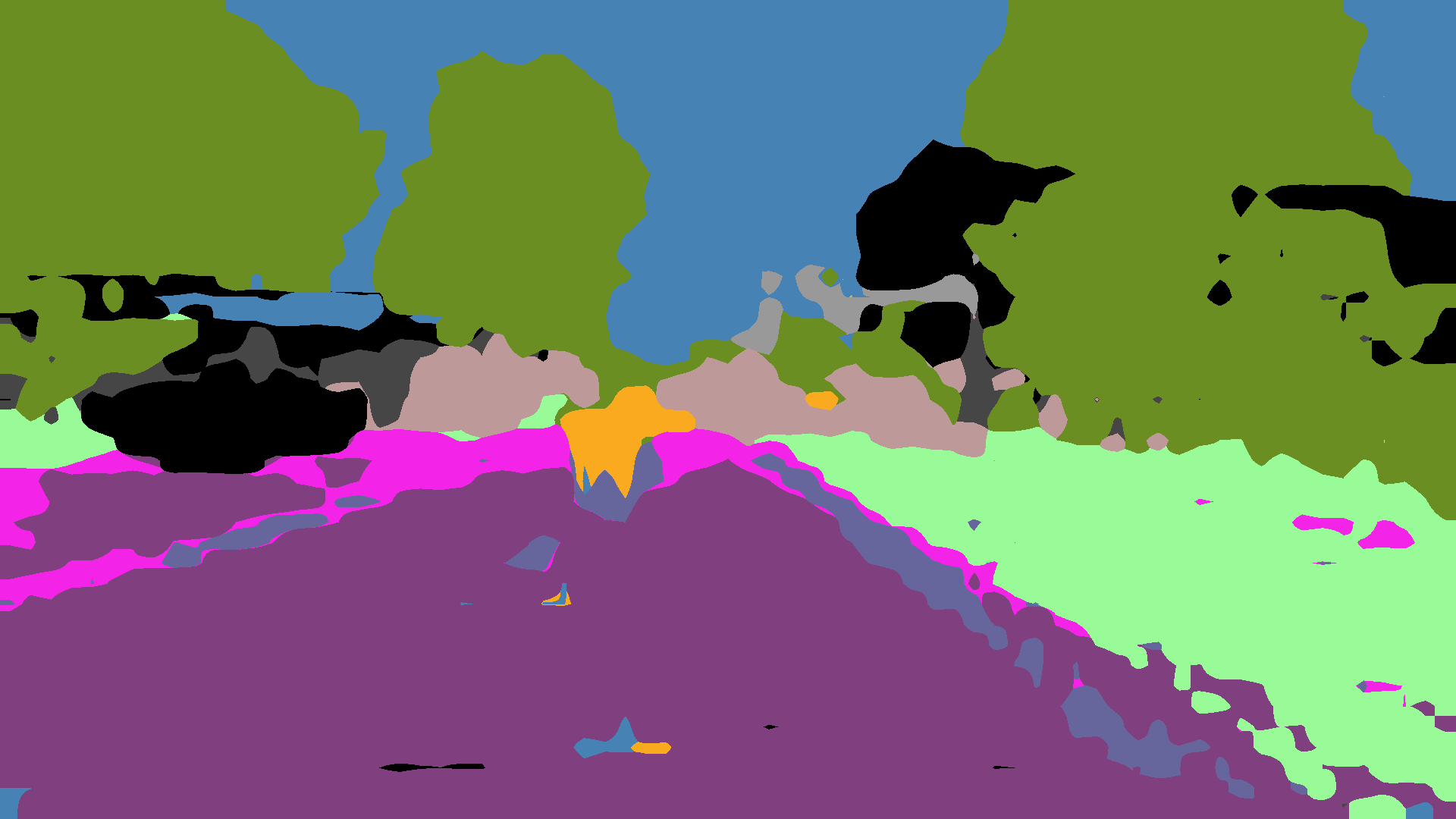} &
\includegraphics[width=\linewidth]{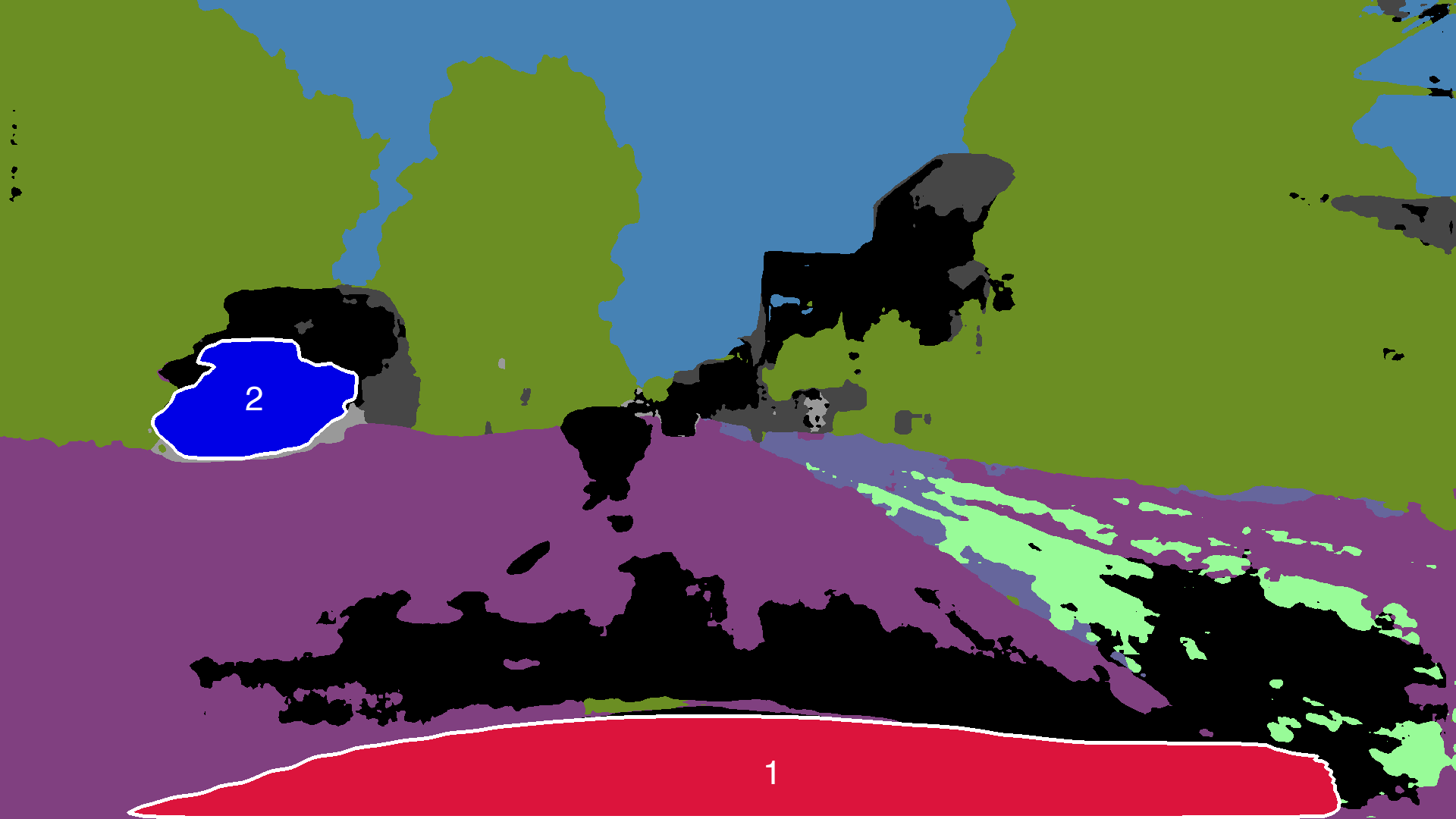} &
\includegraphics[width=\linewidth]{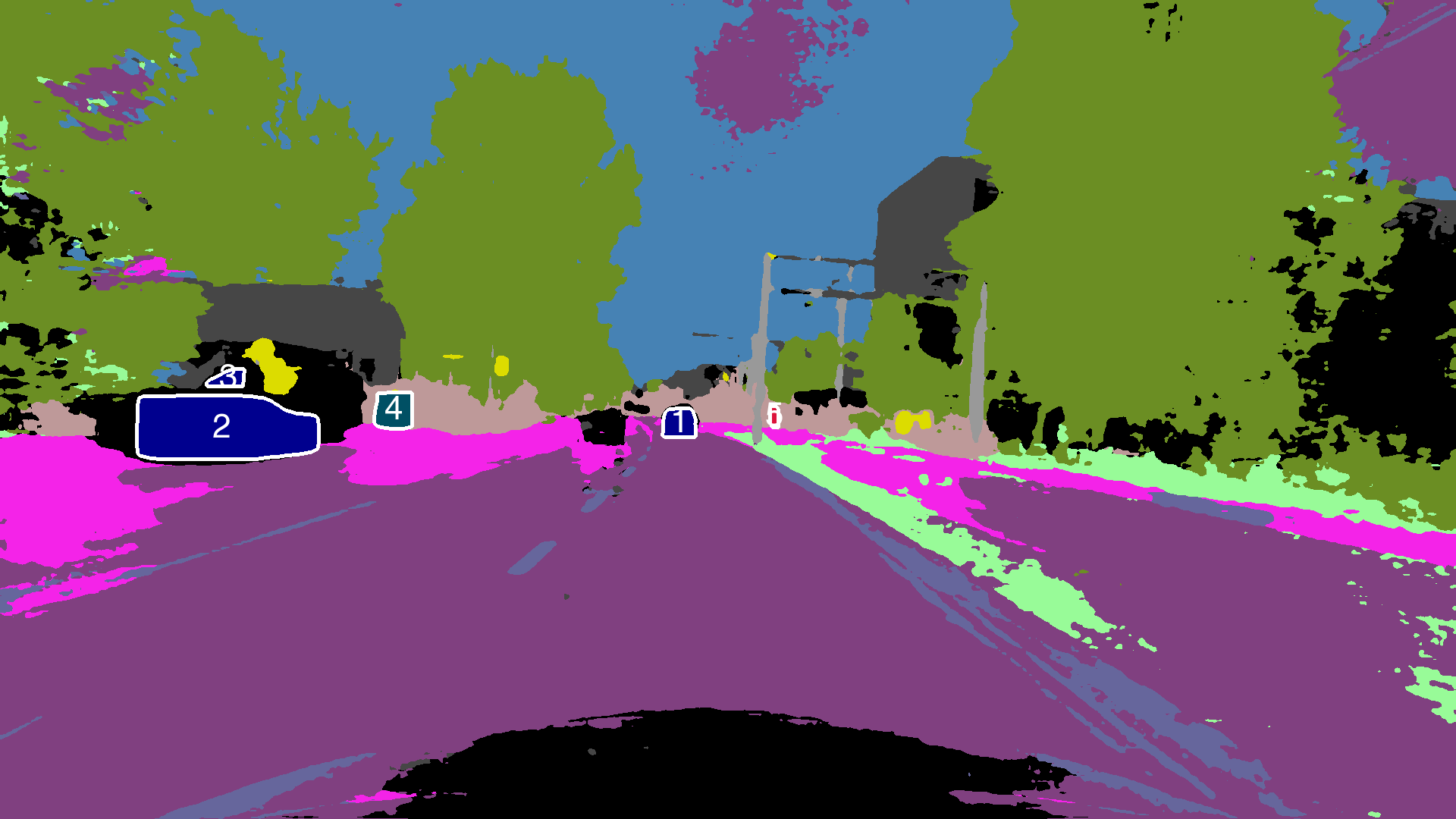} \\

\includegraphics[width=\linewidth]{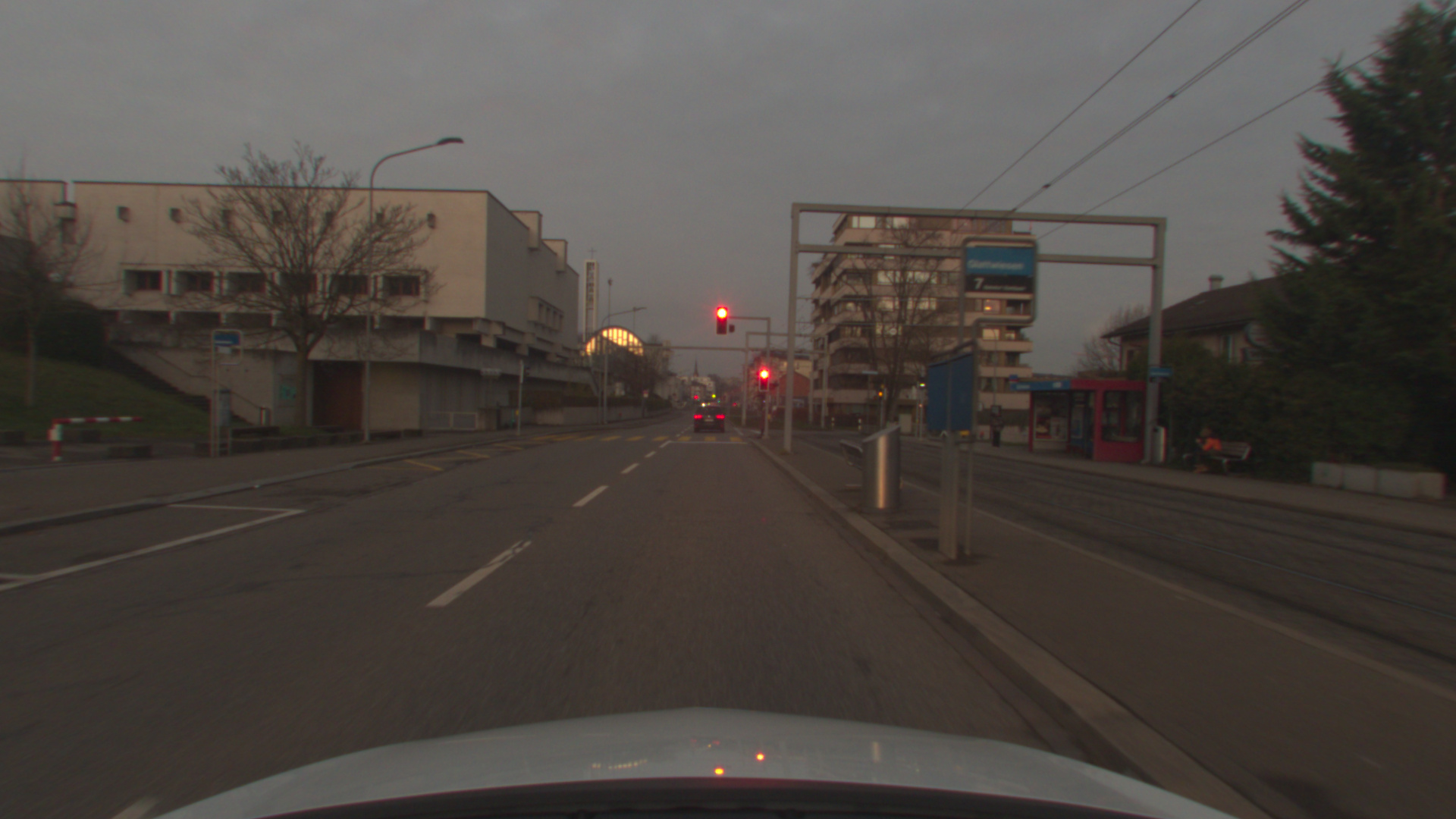} &
\includegraphics[width=\linewidth]{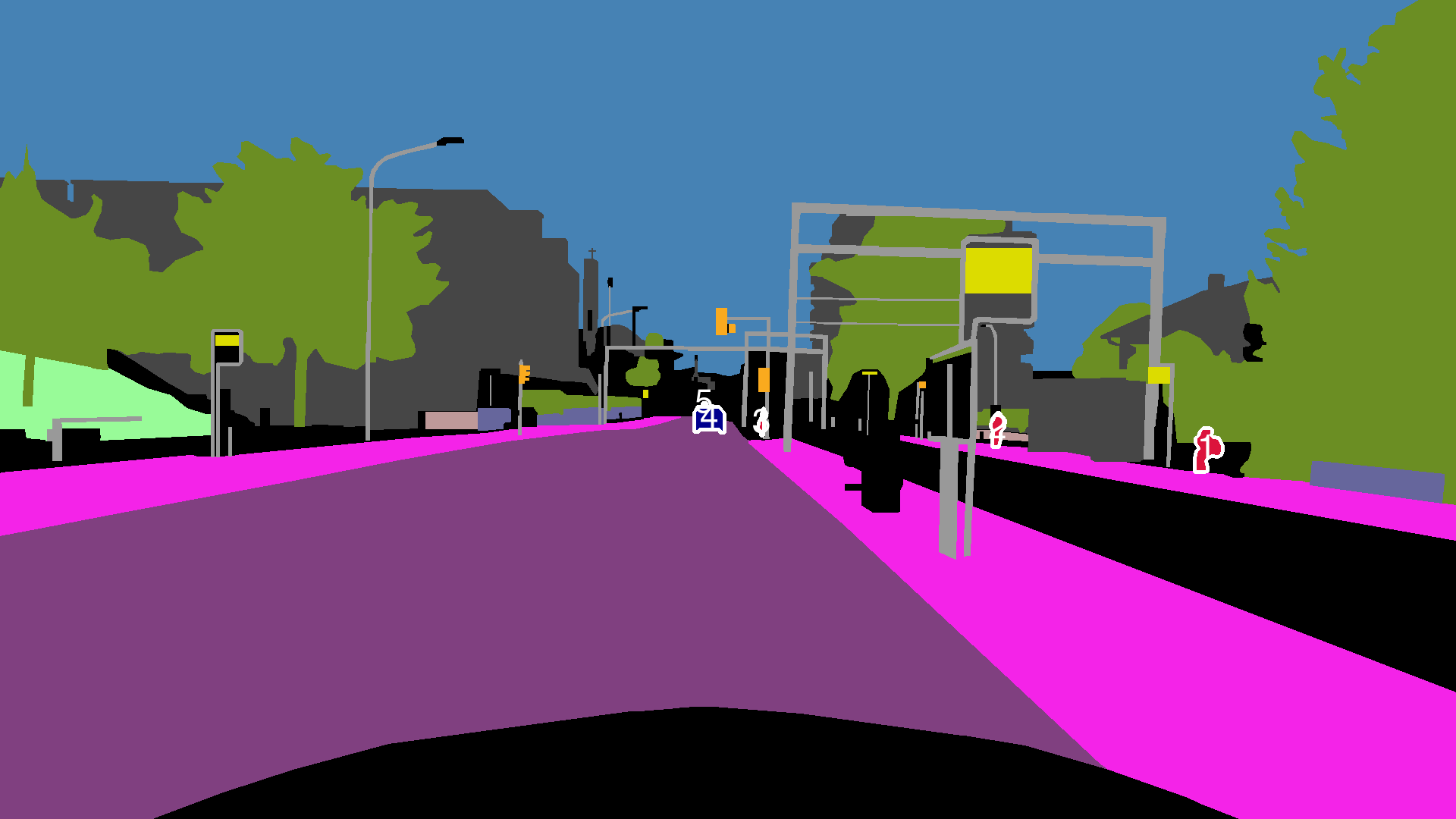} &
\includegraphics[width=\linewidth]{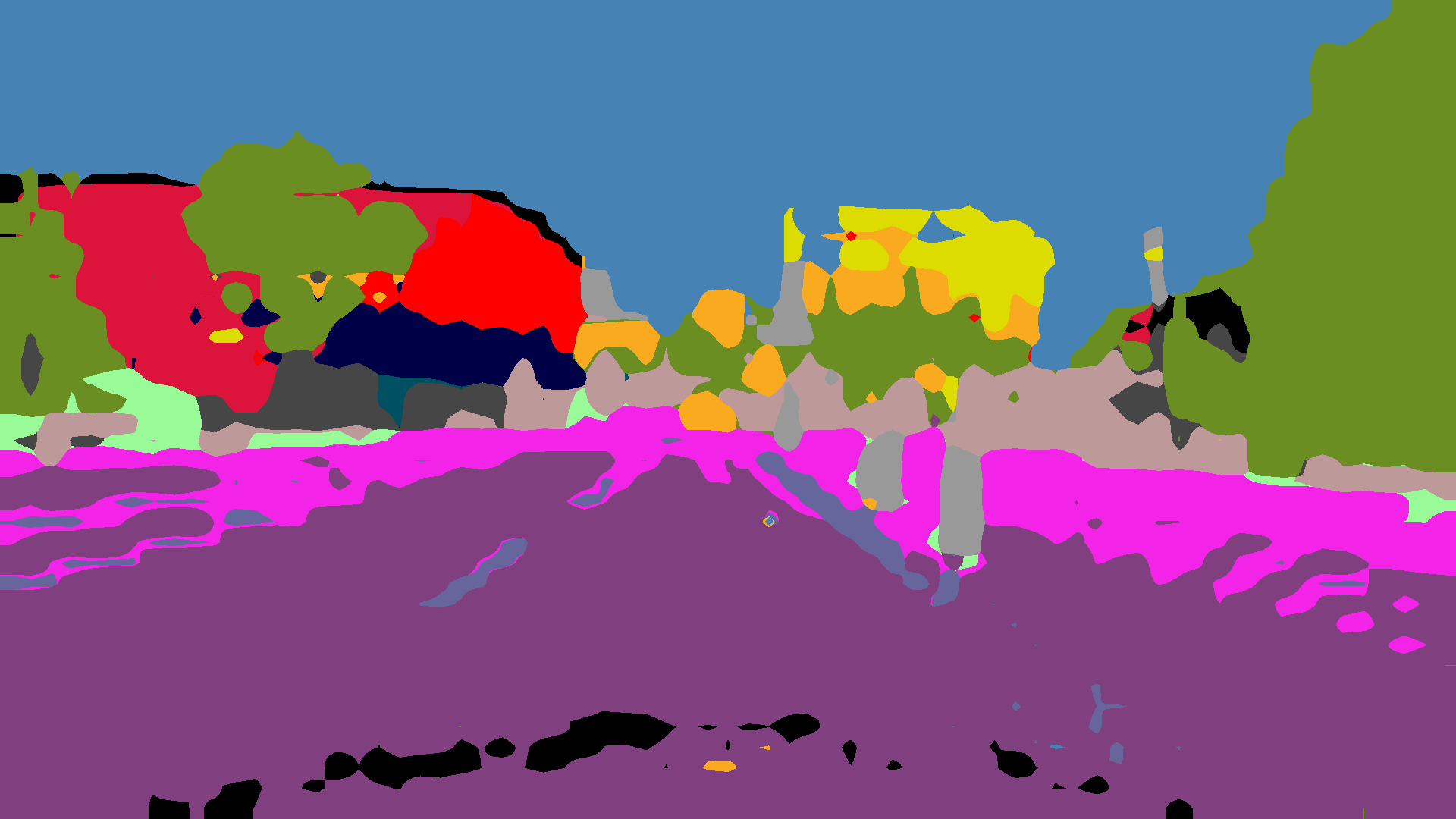} &
\includegraphics[width=\linewidth]{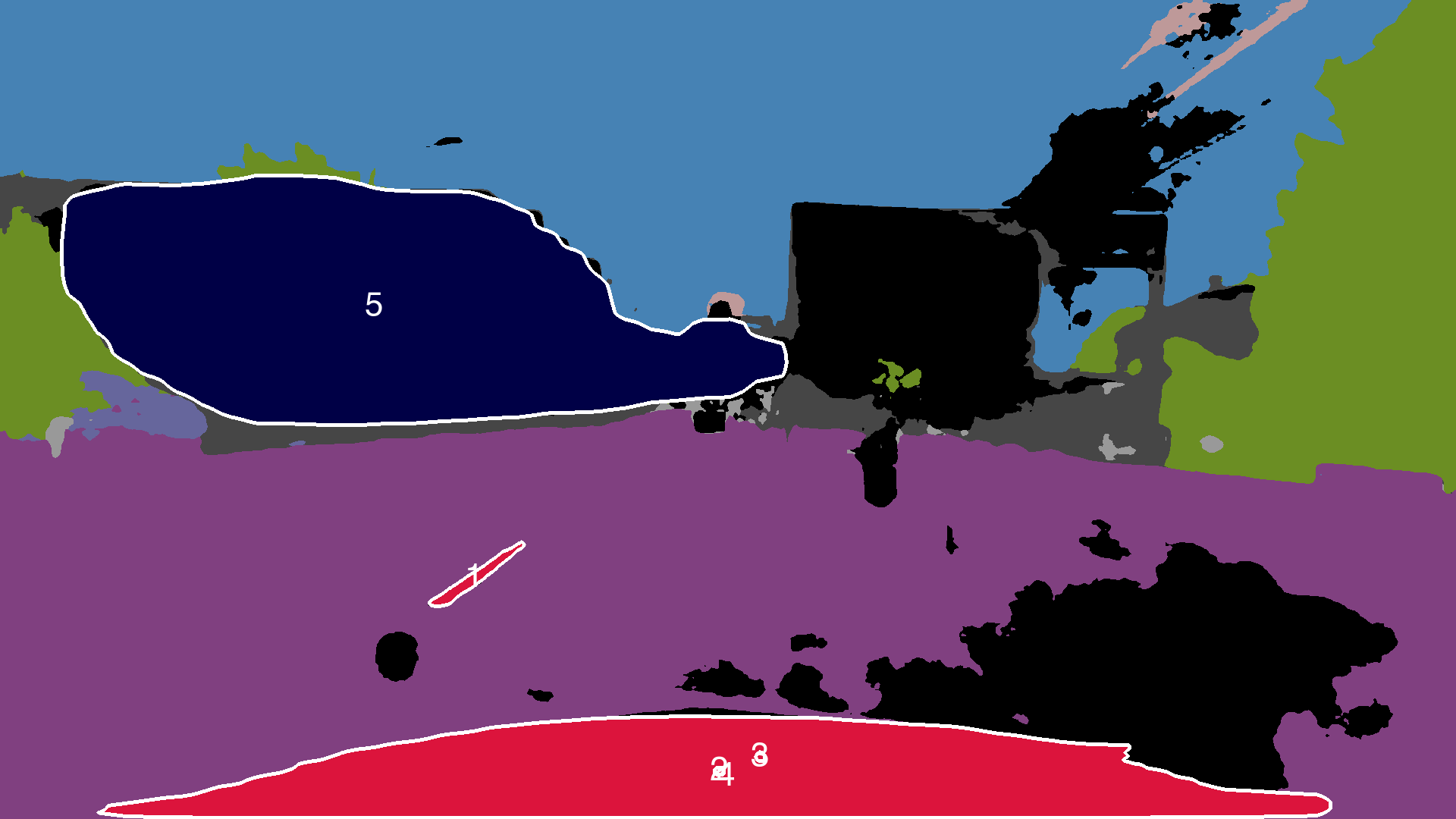} &
\includegraphics[width=\linewidth]{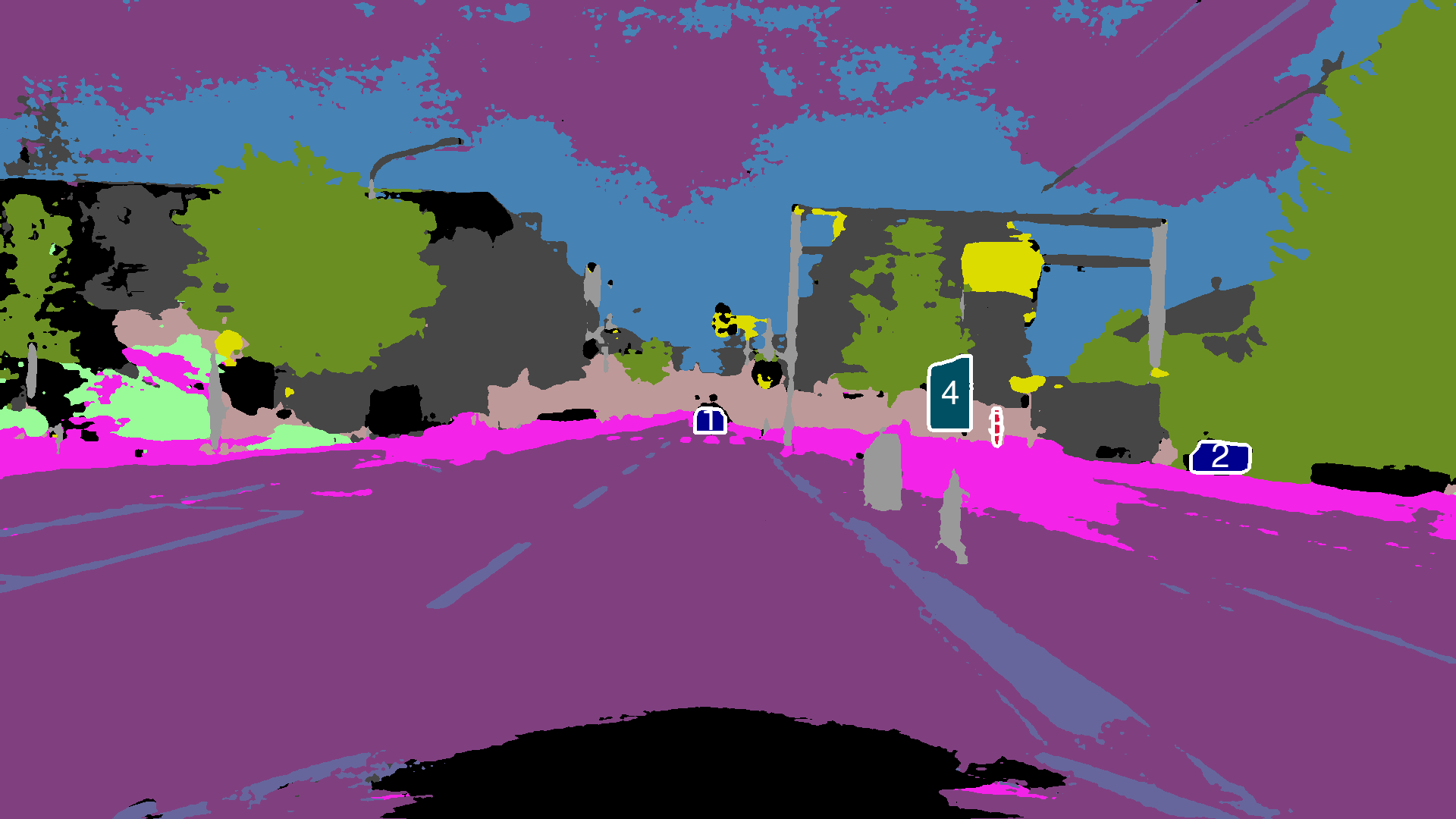} \\

\includegraphics[width=\linewidth]{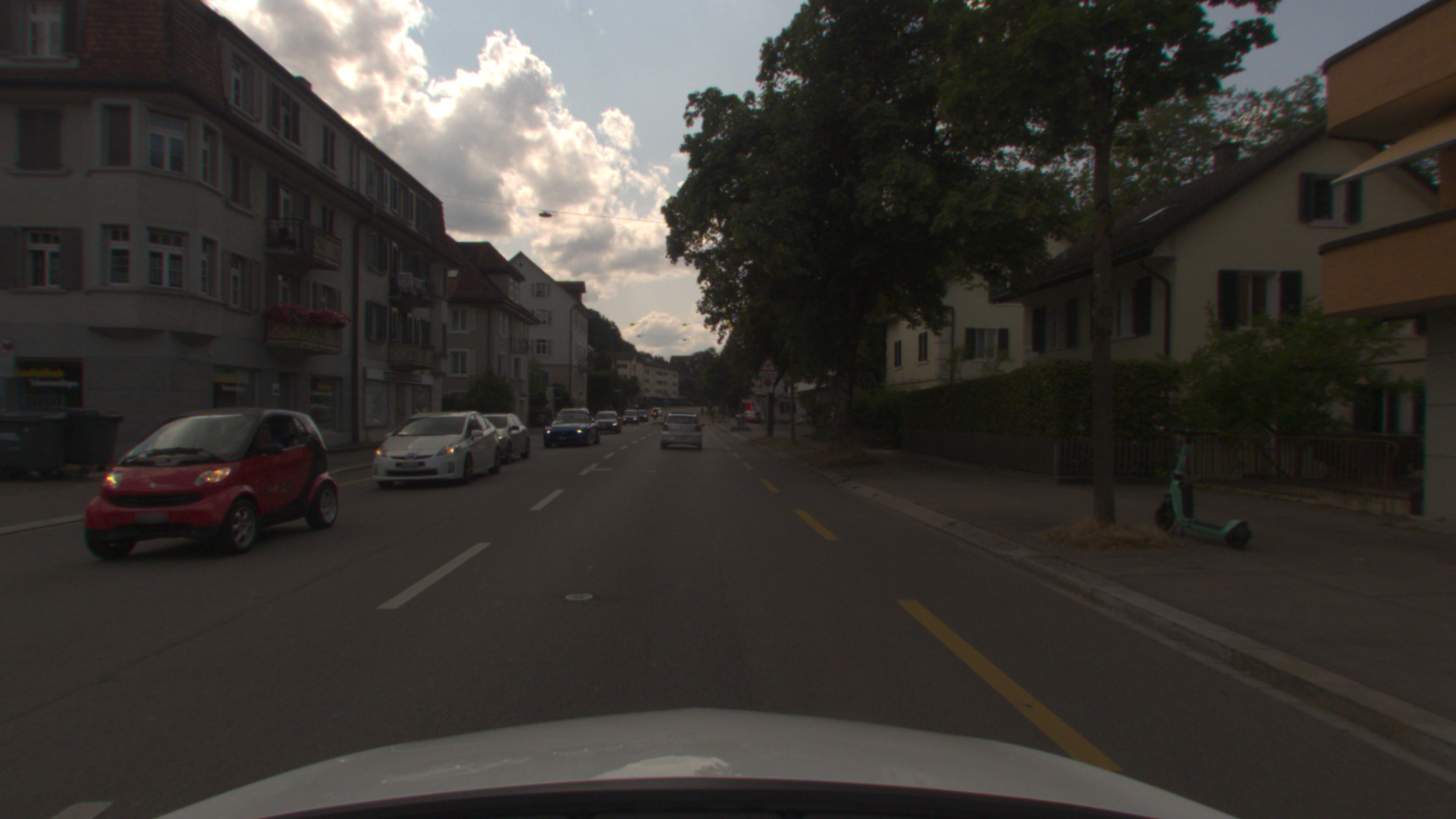} &
\includegraphics[width=\linewidth]{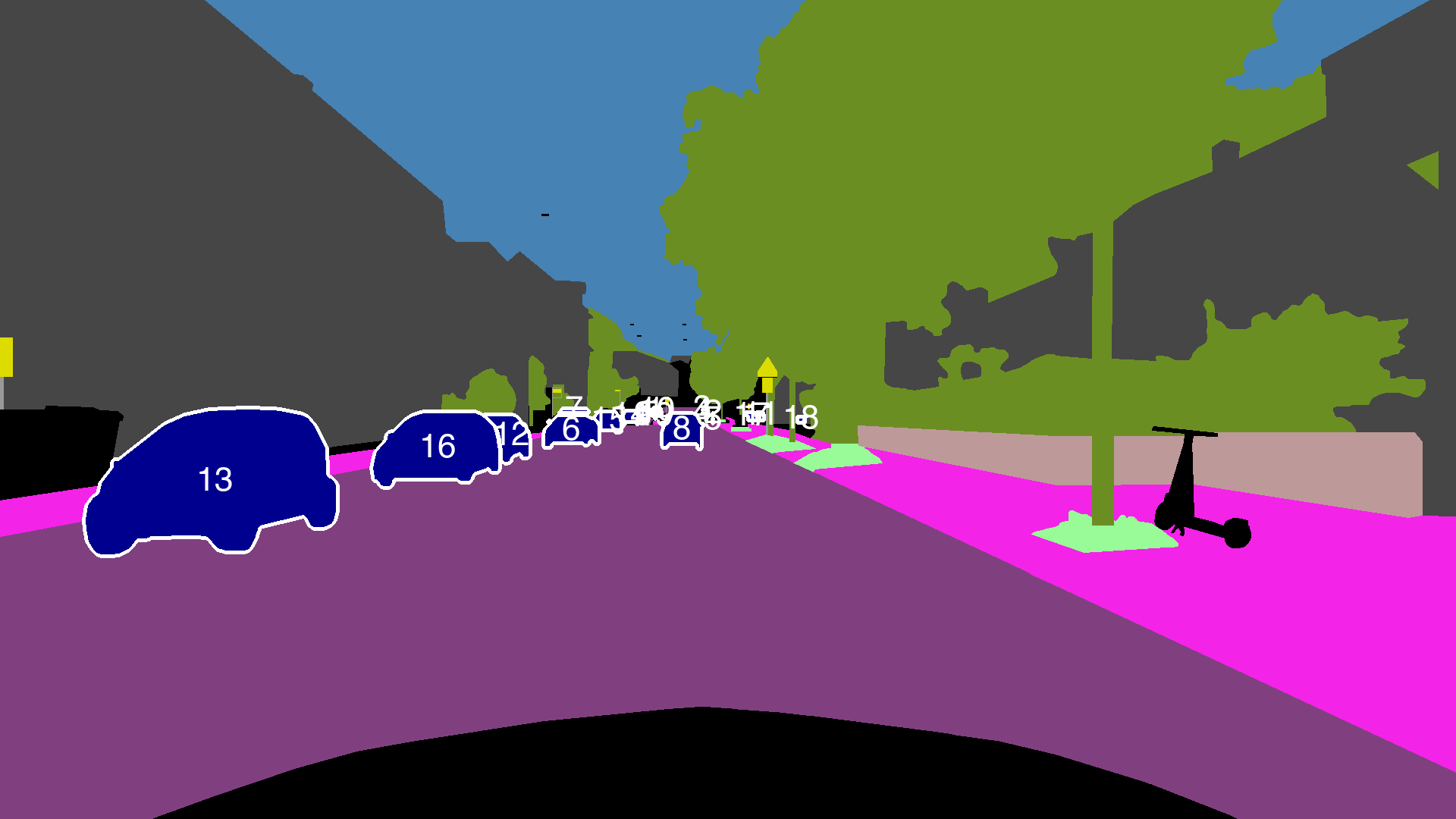} &
\includegraphics[width=\linewidth]{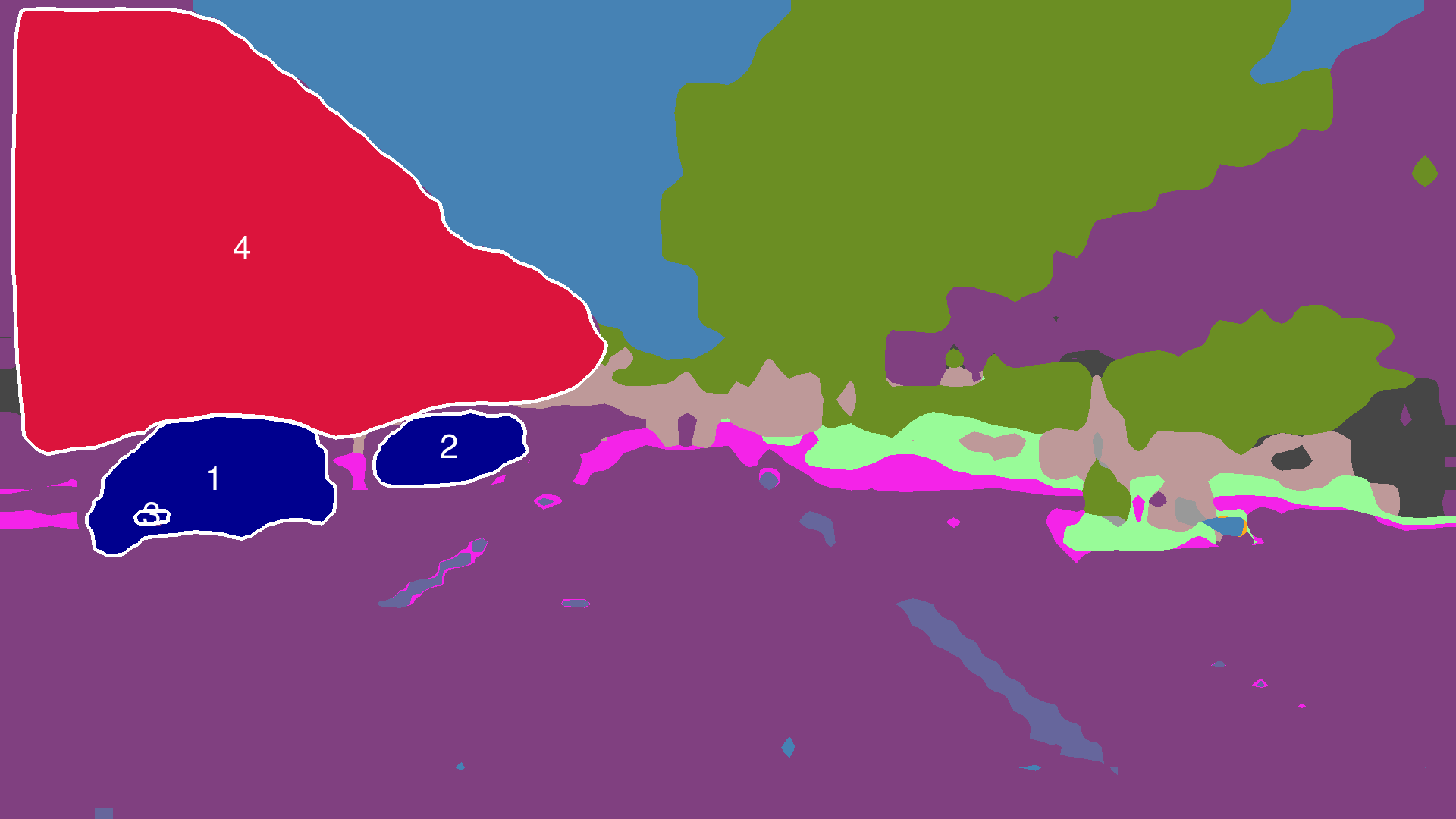} &
\includegraphics[width=\linewidth]{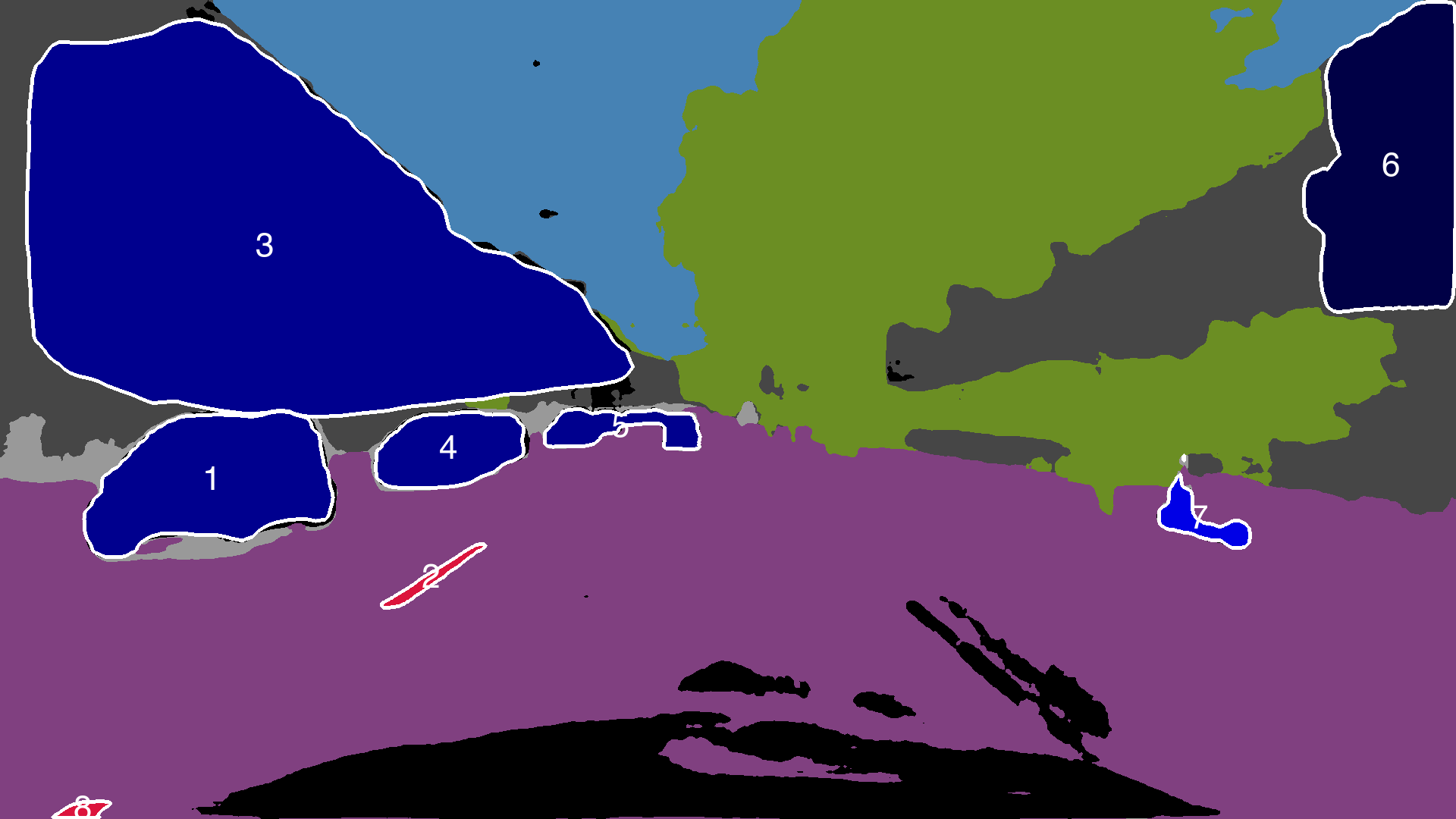} &
\includegraphics[width=\linewidth]{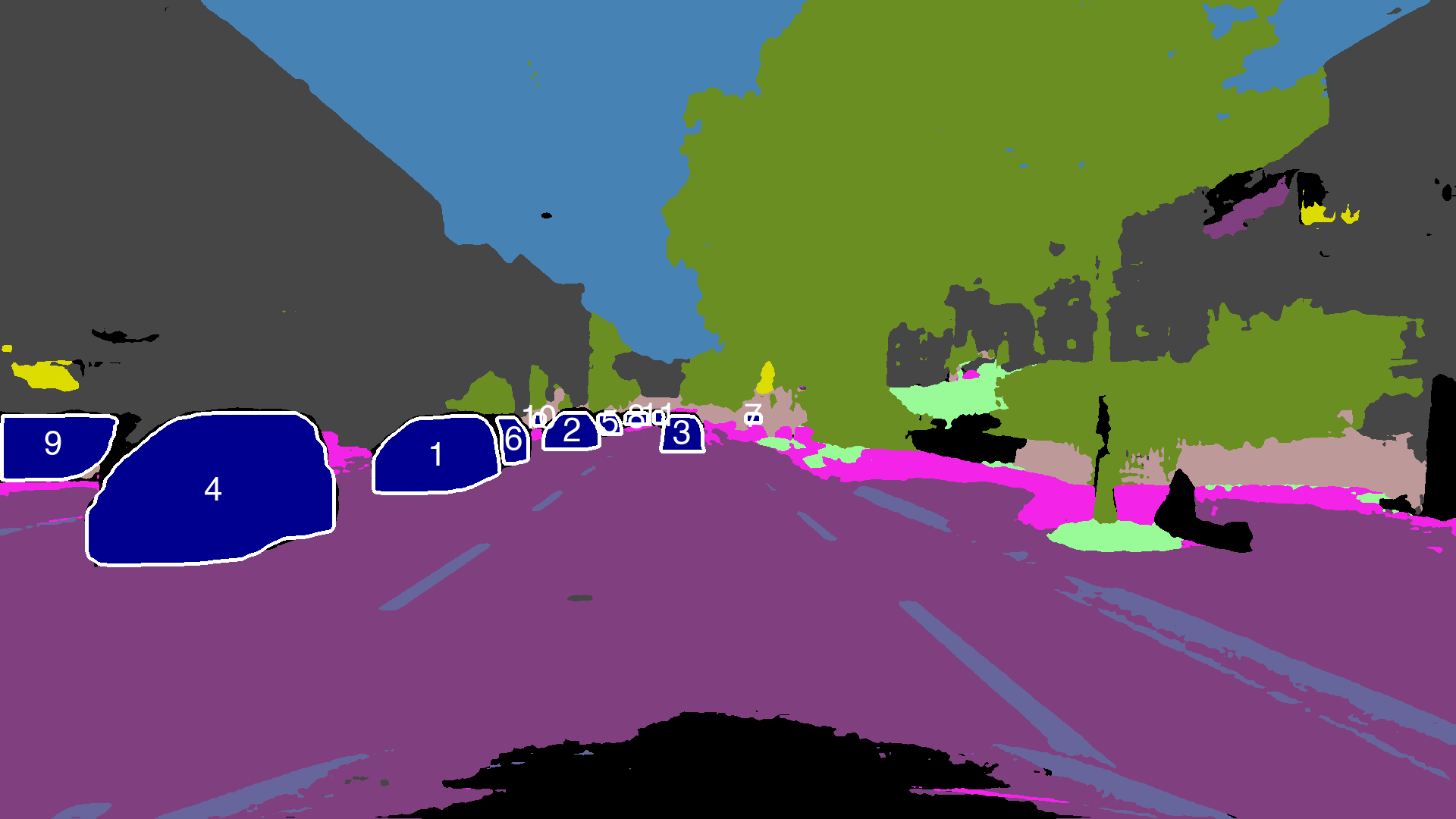} \\[-1pt]

\end{tabular}

\tiny
\renewcommand{\arraystretch}{1.3}
\begin{tabularx}{\textwidth}{*{19}{>{\centering\arraybackslash}X}}  
    \cellcolor{road}\textcolor{white}{Road}
    & \cellcolor{sidewalk}\textcolor{white}{Sidewalk}
    & \cellcolor{building}\textcolor{white}{Building}
    & \cellcolor{wall}\textcolor{white}{Wall}
    & \cellcolor{fence}\textcolor{white}{Fence}
    & \cellcolor{pole}\textcolor{white}{Pole}
    & \cellcolor{trafficlight}\textcolor{white}{Traffic~Light}
    & \cellcolor{trafficsign}\textcolor{white}{Traffic~Sign}
    & \cellcolor{vegetation}\textcolor{white}{Vegetation}
    & \cellcolor{terrain}\textcolor{white}{Terrain}
    & \cellcolor{sky}\textcolor{white}{Sky}
    & \cellcolor{person}\textcolor{white}{Person}
    & \cellcolor{rider}\textcolor{white}{Rider}
    & \cellcolor{car}\textcolor{white}{Car}
    & \cellcolor{truck}\textcolor{white}{Truck}
    & \cellcolor{bus}\textcolor{white}{Bus}
    & \cellcolor{train}\textcolor{white}{Train}
    & \cellcolor{motorcycle}\textcolor{white}{Motorcycle}
    & \cellcolor{bicycle}\textcolor{white}{Bicycle}
\end{tabularx}

        \vspace{-0.55em}
        \caption{\textbf{MUSES} --- Qualitative unsupervised panoptic segmentation examples.\label{fig:qualitative_muses}}
    \end{subfigure}\vspace{1.25em}\\
       
    \begin{subfigure}[t]{\textwidth}
        \centering
        \small
\sffamily
\setlength{\tabcolsep}{0pt}
\renewcommand{\arraystretch}{0.0}
\begin{tabular}{>{\centering\arraybackslash} m{0.2\textwidth} 
                >{\centering\arraybackslash} m{0.2\textwidth} 
                >{\centering\arraybackslash} m{0.2\textwidth}
                >{\centering\arraybackslash} m{0.2\textwidth}
                >{\centering\arraybackslash} m{0.2\textwidth}}

{Image} & {Ground Truth} & {Baseline} & {U2Seg~\cite{Niu:2024:UUI}} & {\MethodName \textit{(Ours)}} \\[4pt]

\includegraphics[width=\linewidth, height=7em]{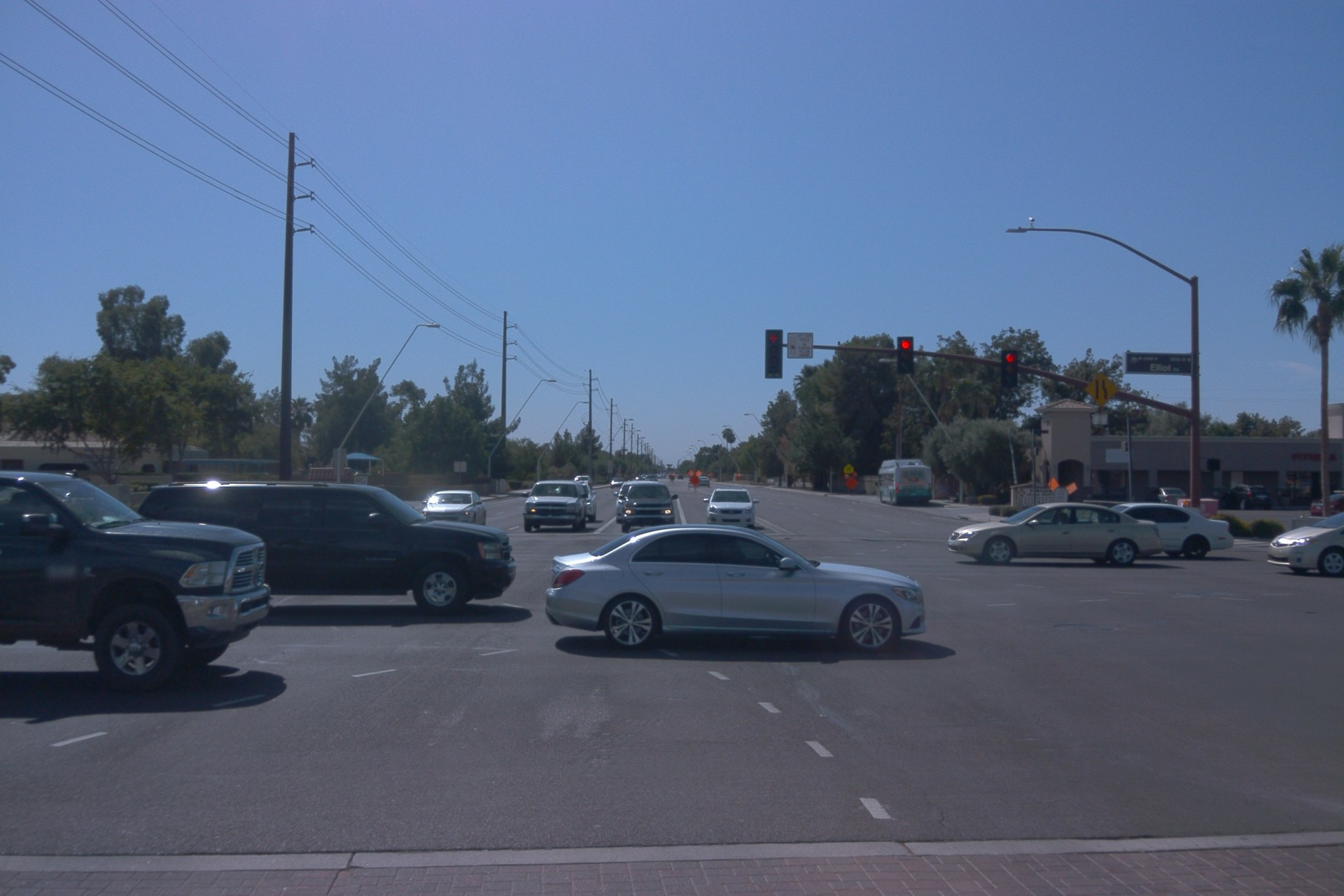} &
\includegraphics[width=\linewidth, height=7em]{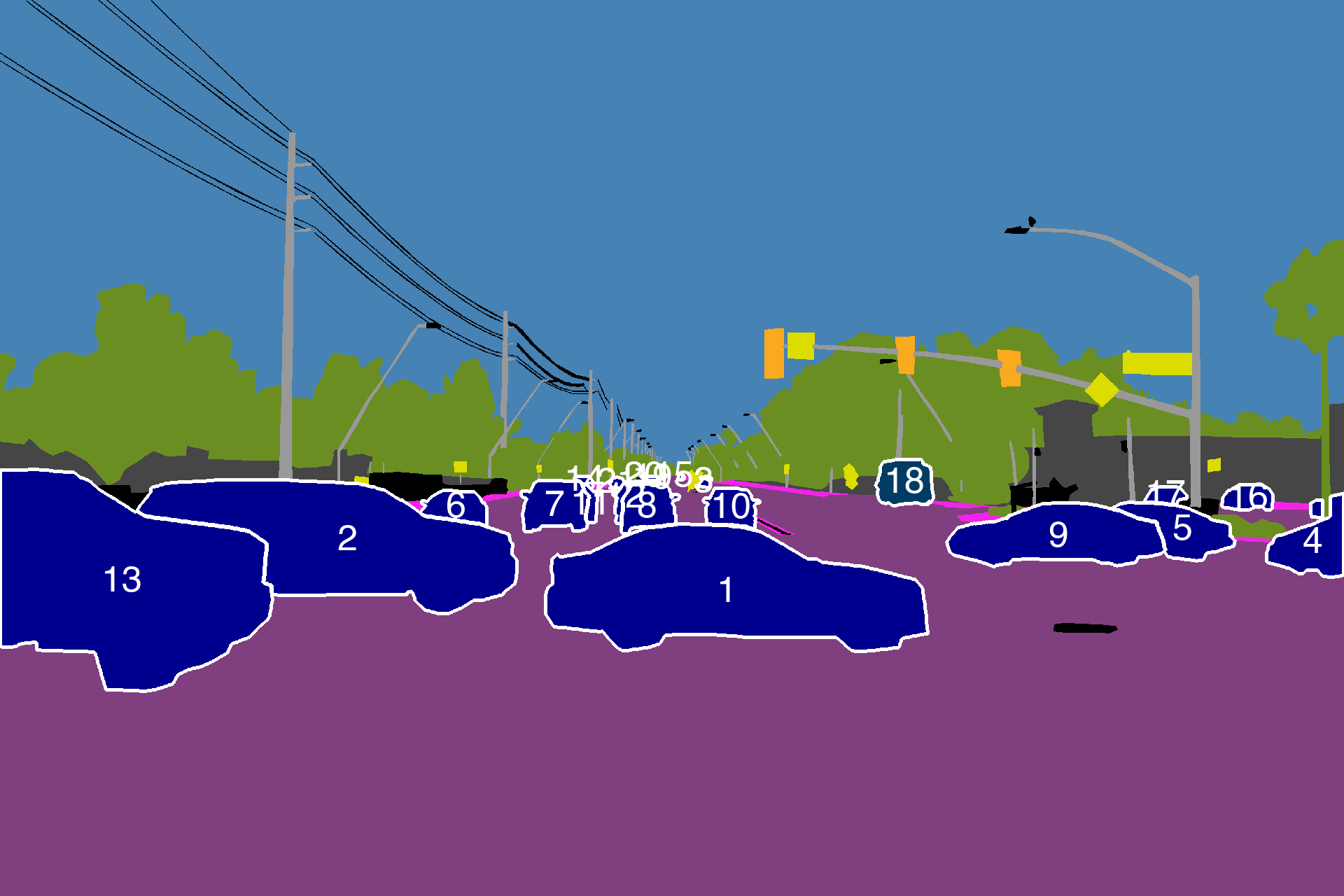} &
\includegraphics[width=\linewidth, height=7em]{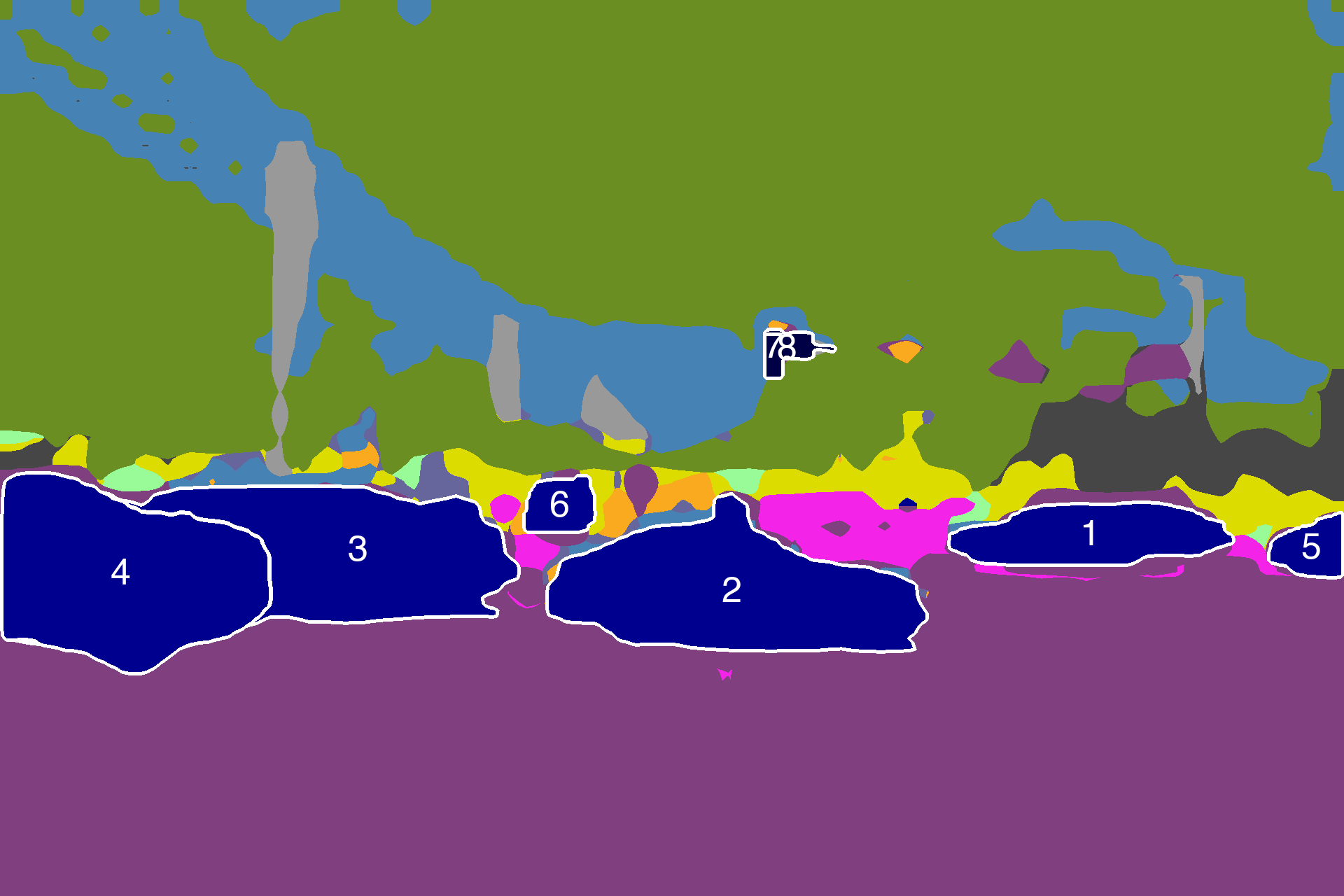} &
\includegraphics[width=\linewidth, height=7em]{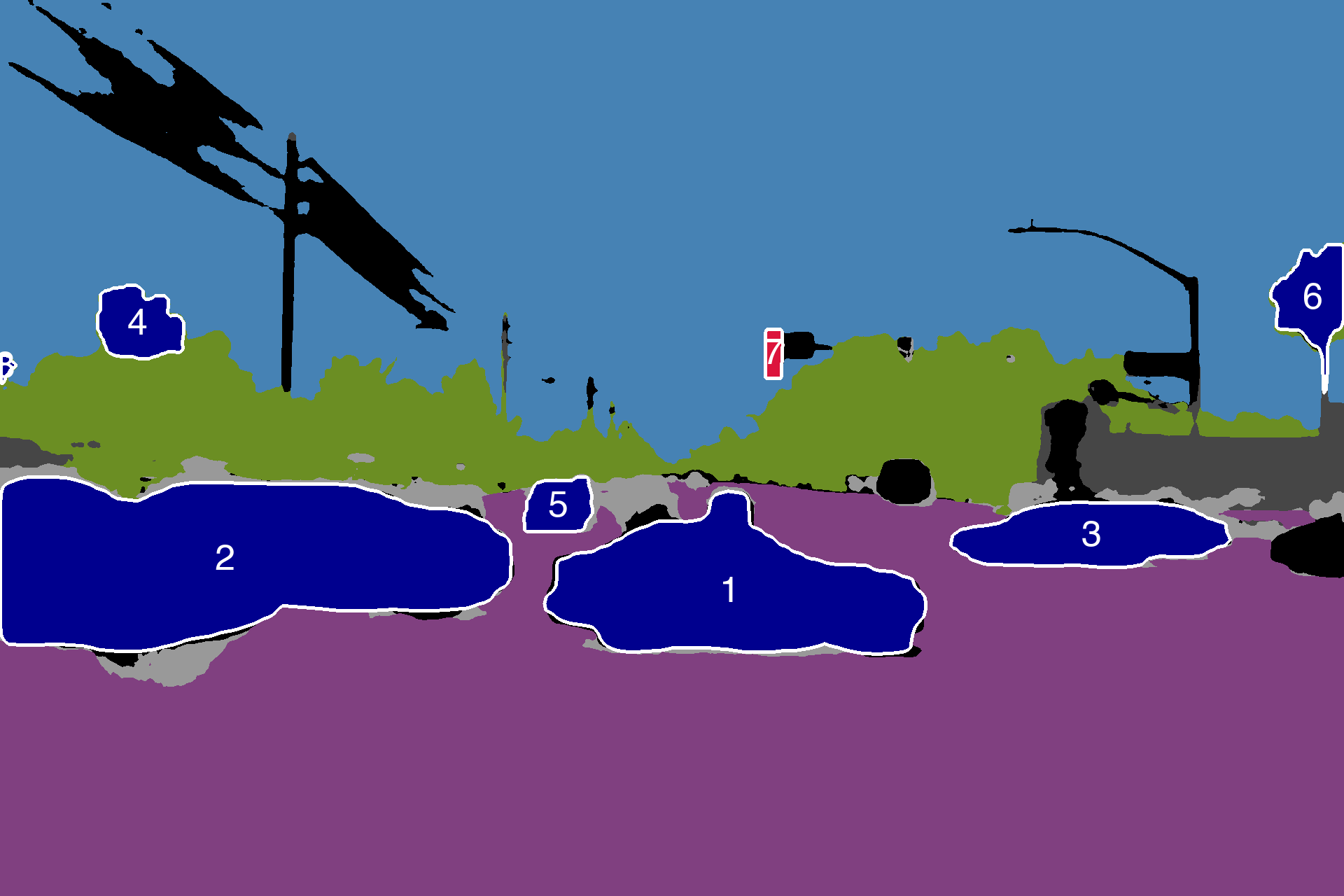} &
\includegraphics[width=\linewidth, height=7em]{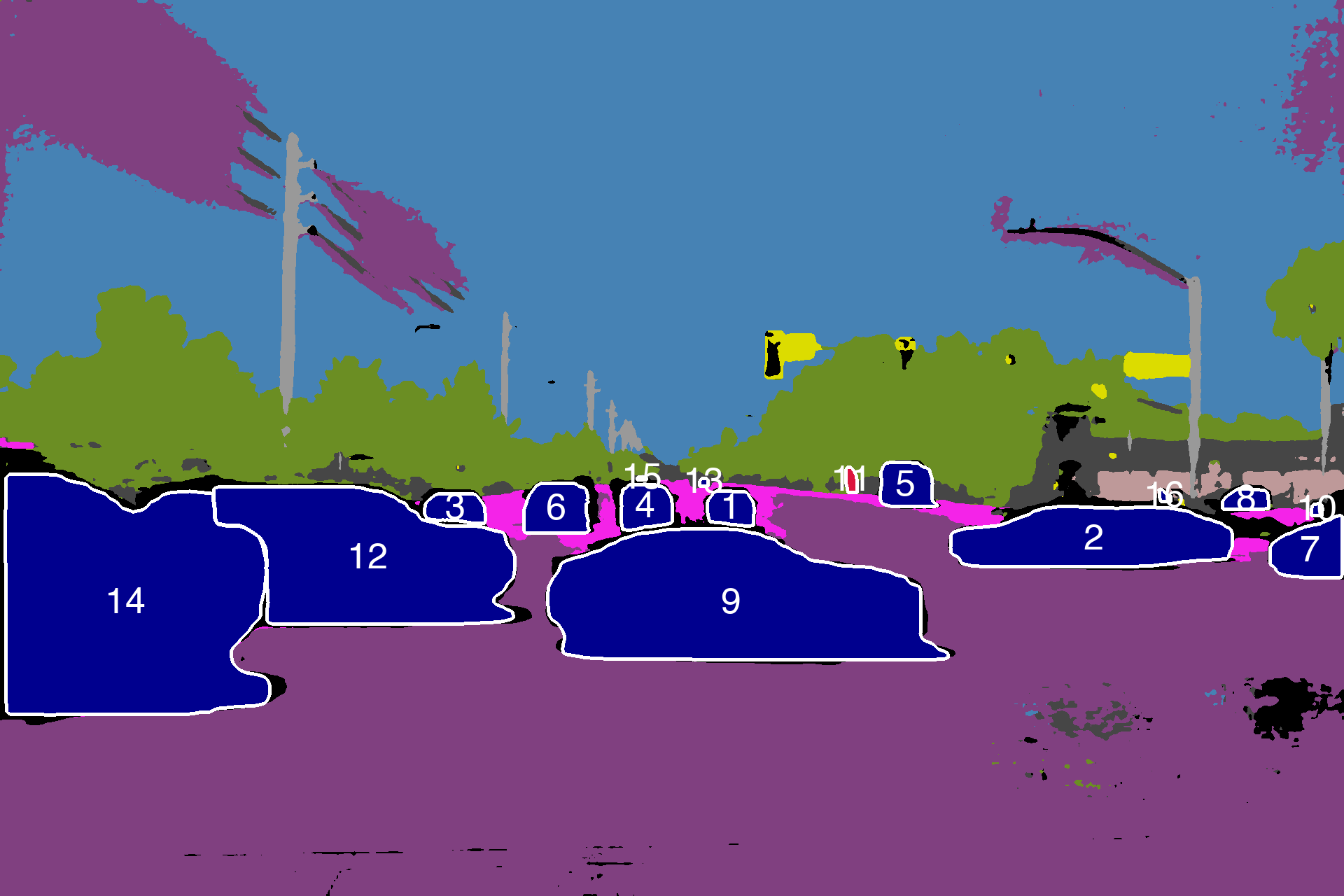}\\

\includegraphics[width=\linewidth, height=7em]{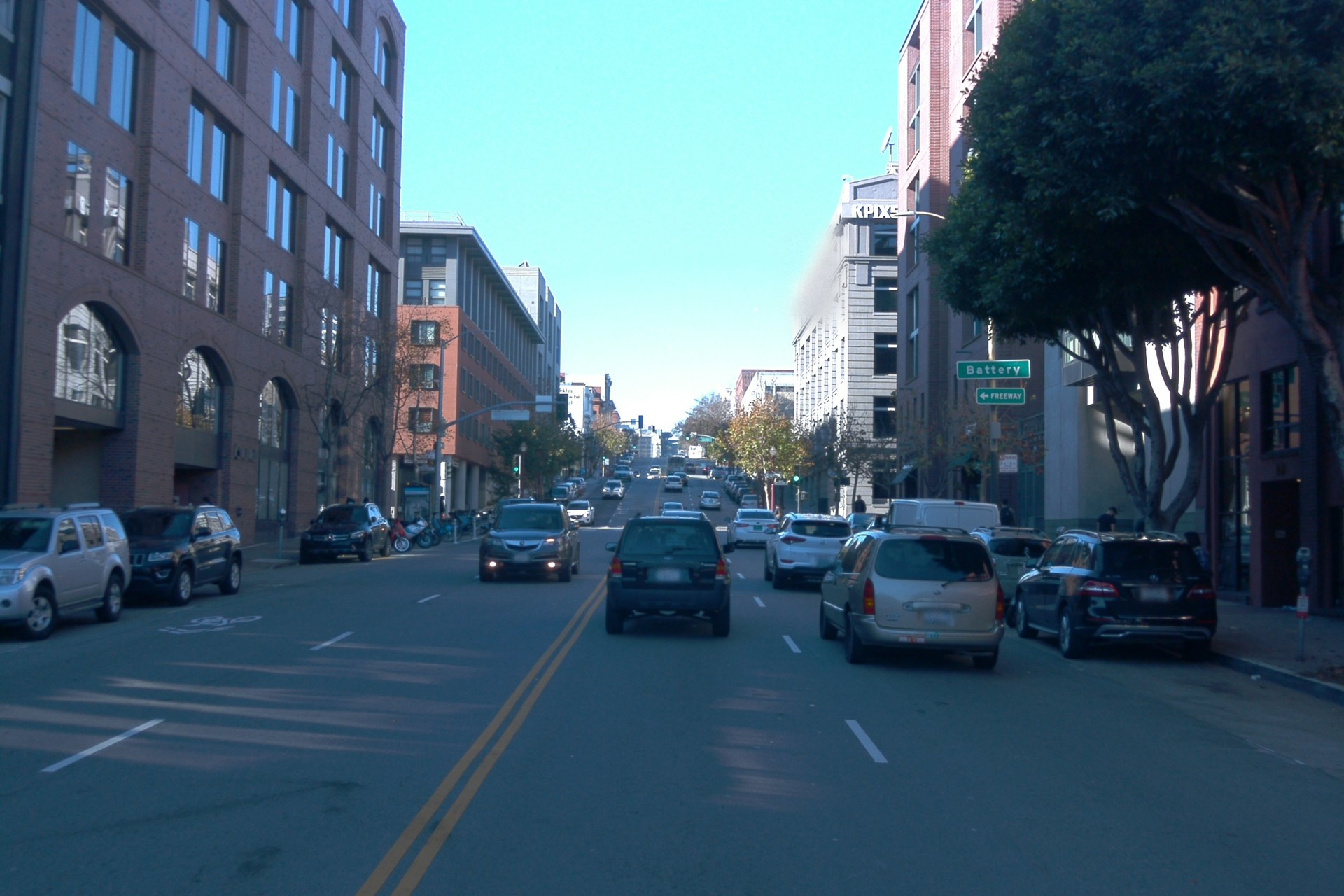} &
\includegraphics[width=\linewidth, height=7em]{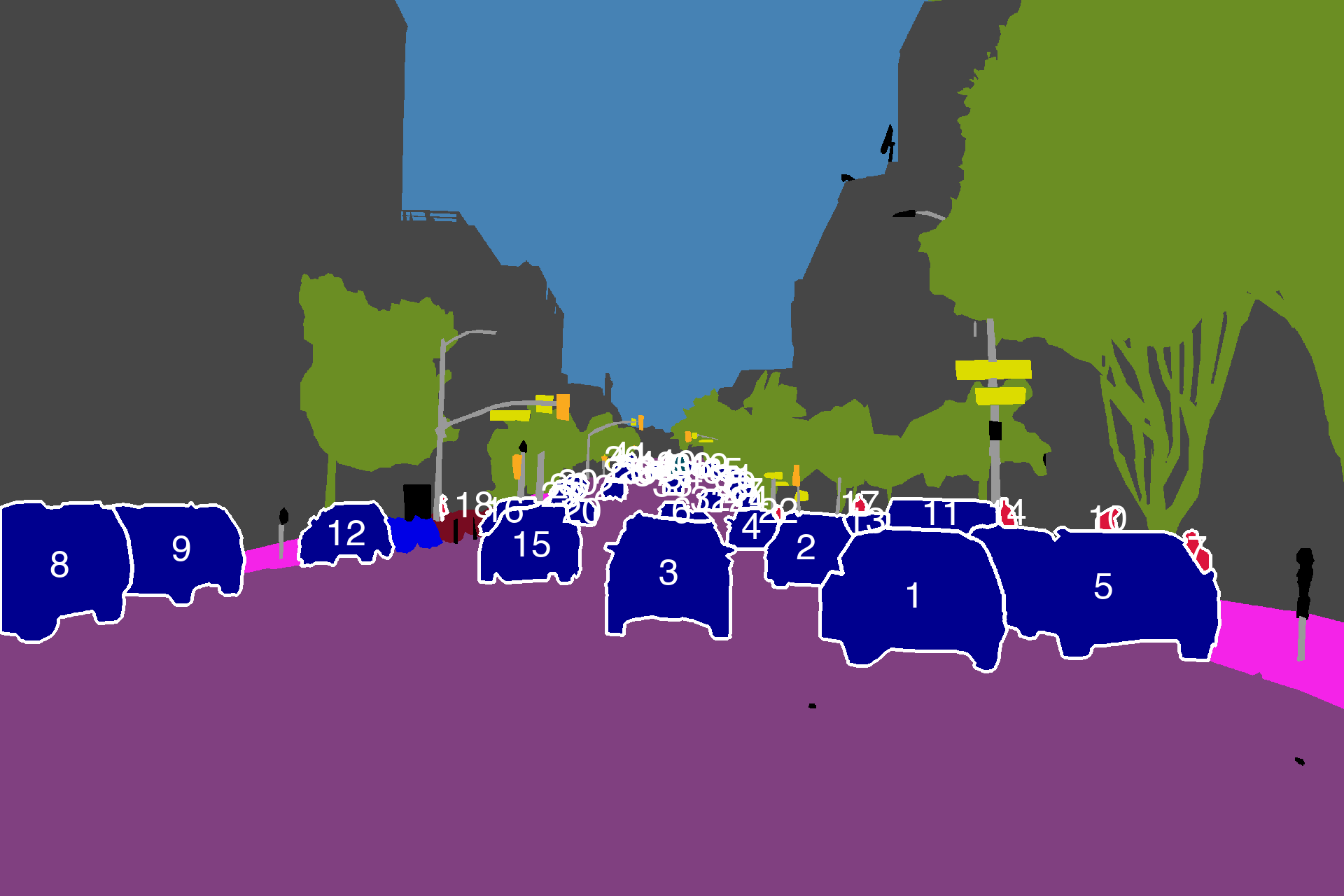} &
\includegraphics[width=\linewidth, height=7em]{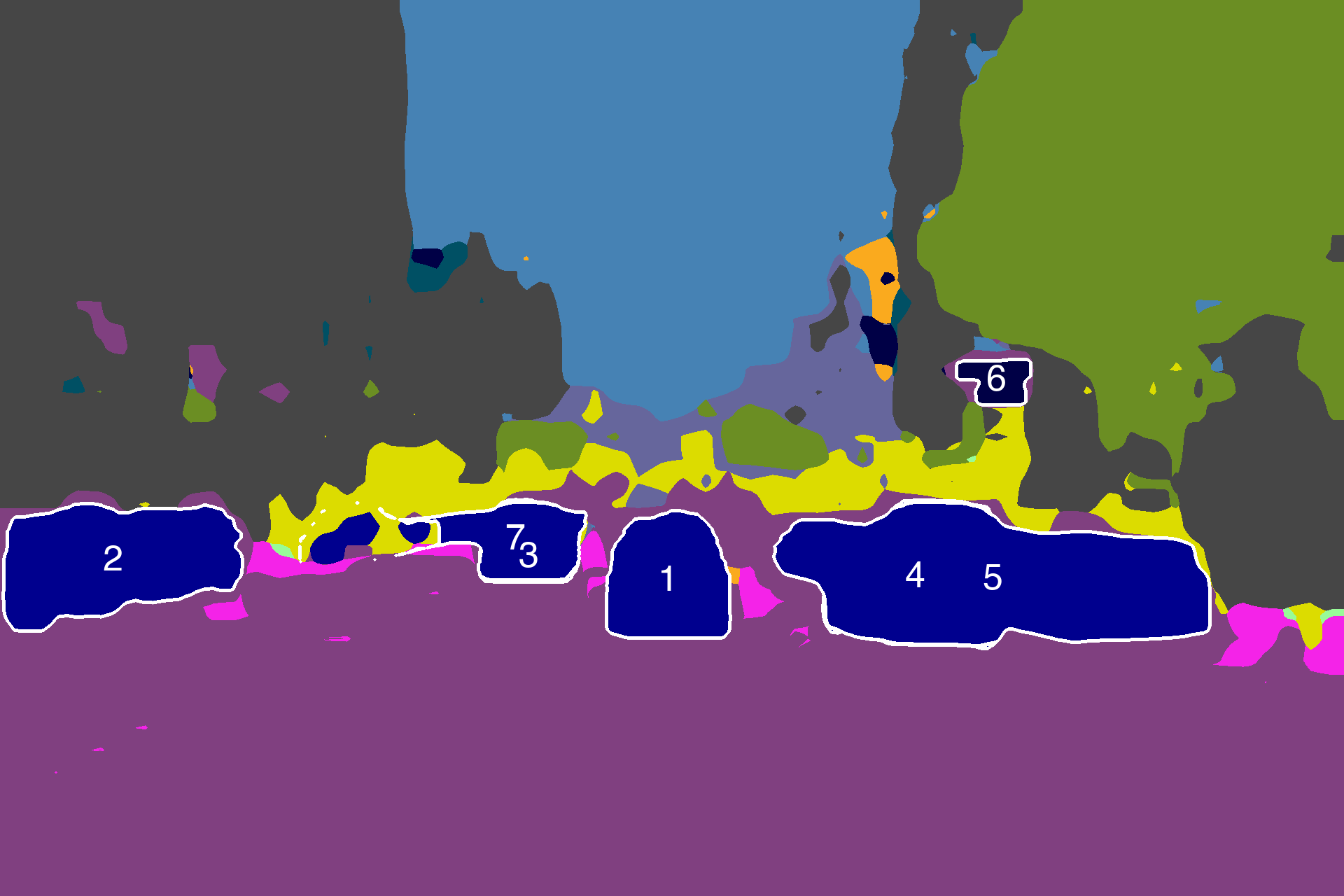} &
\includegraphics[width=\linewidth, height=7em]{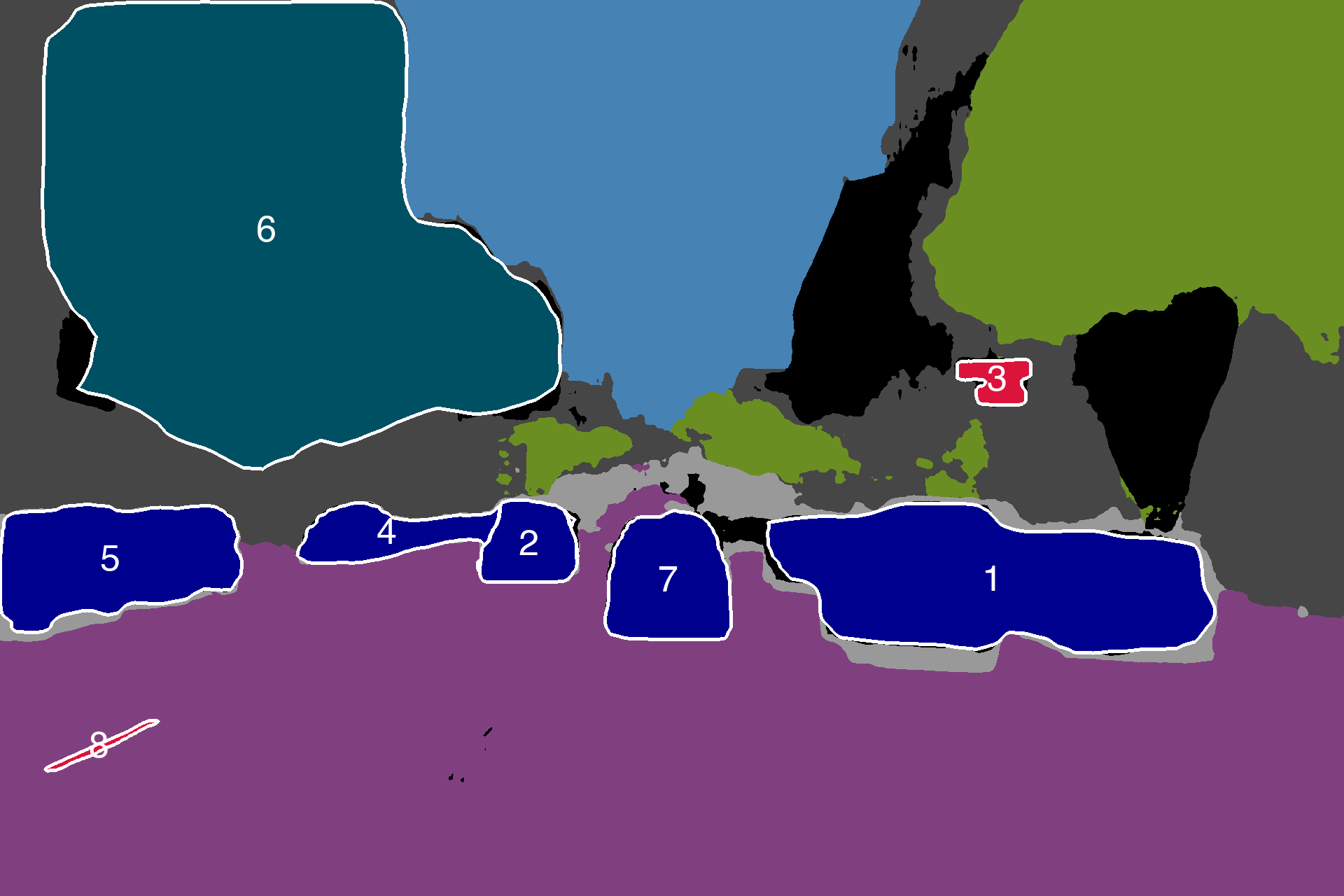} &
\includegraphics[width=\linewidth, height=7em]{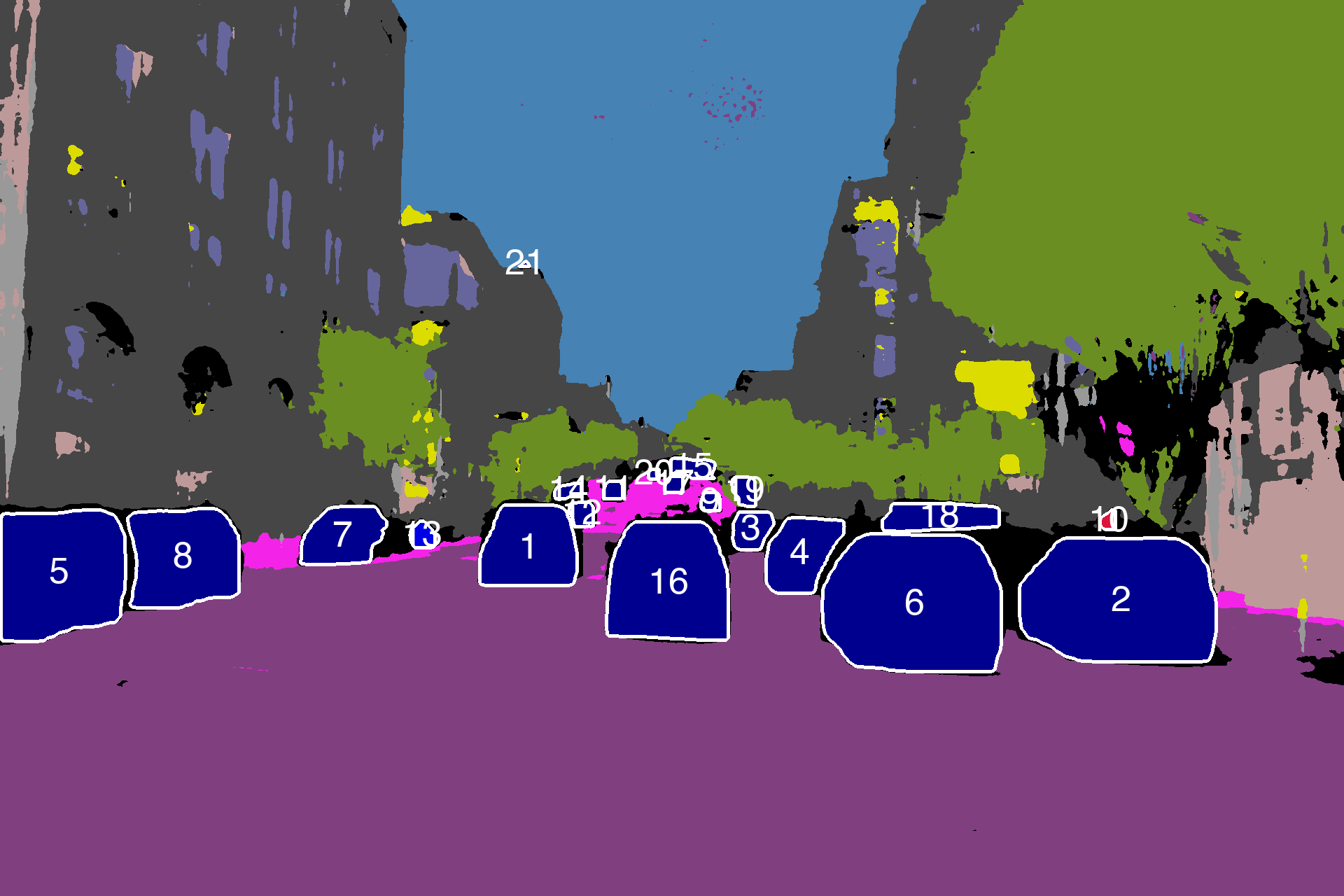}\\

\includegraphics[width=\linewidth, height=7em]{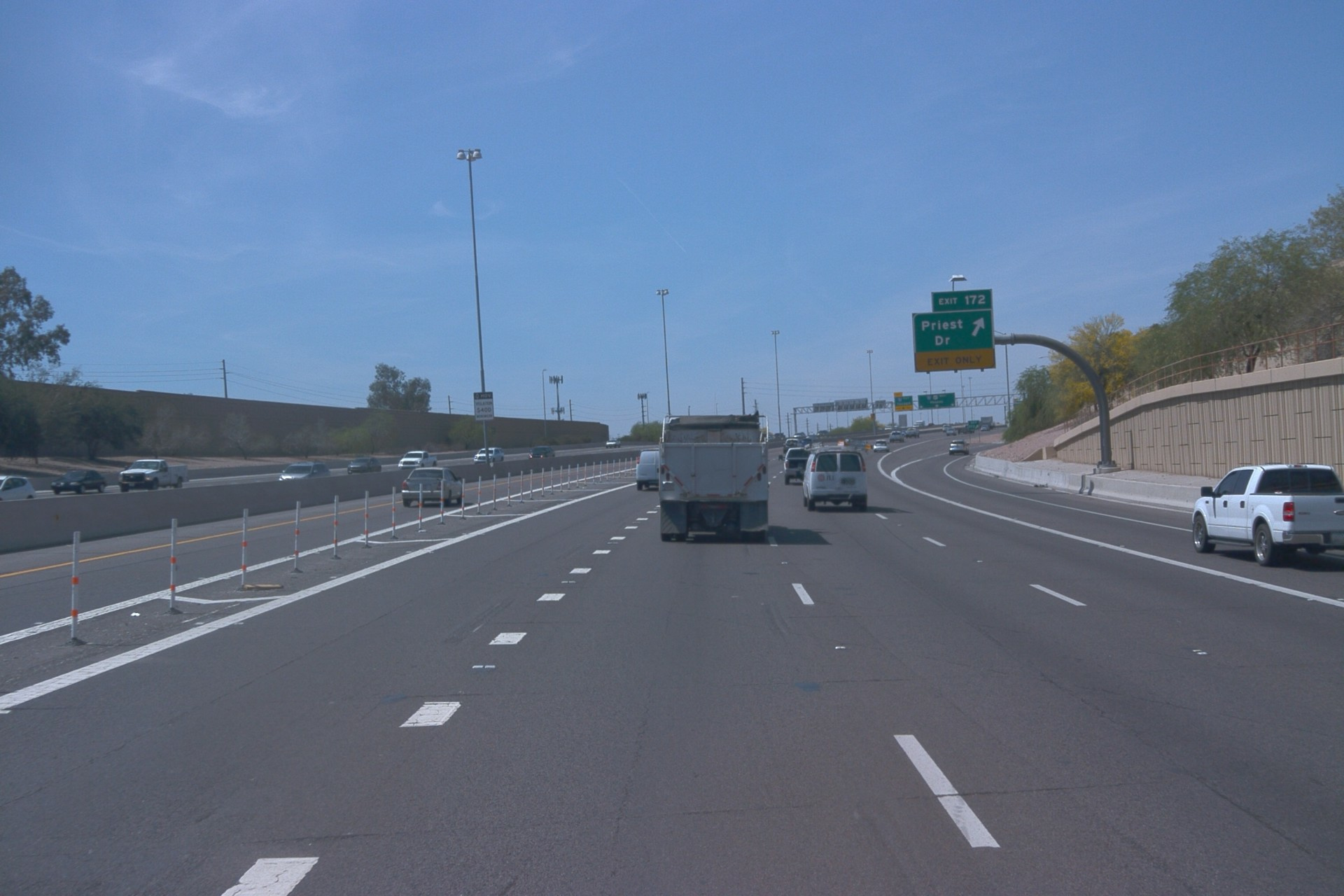} &
\includegraphics[width=\linewidth, height=7em]{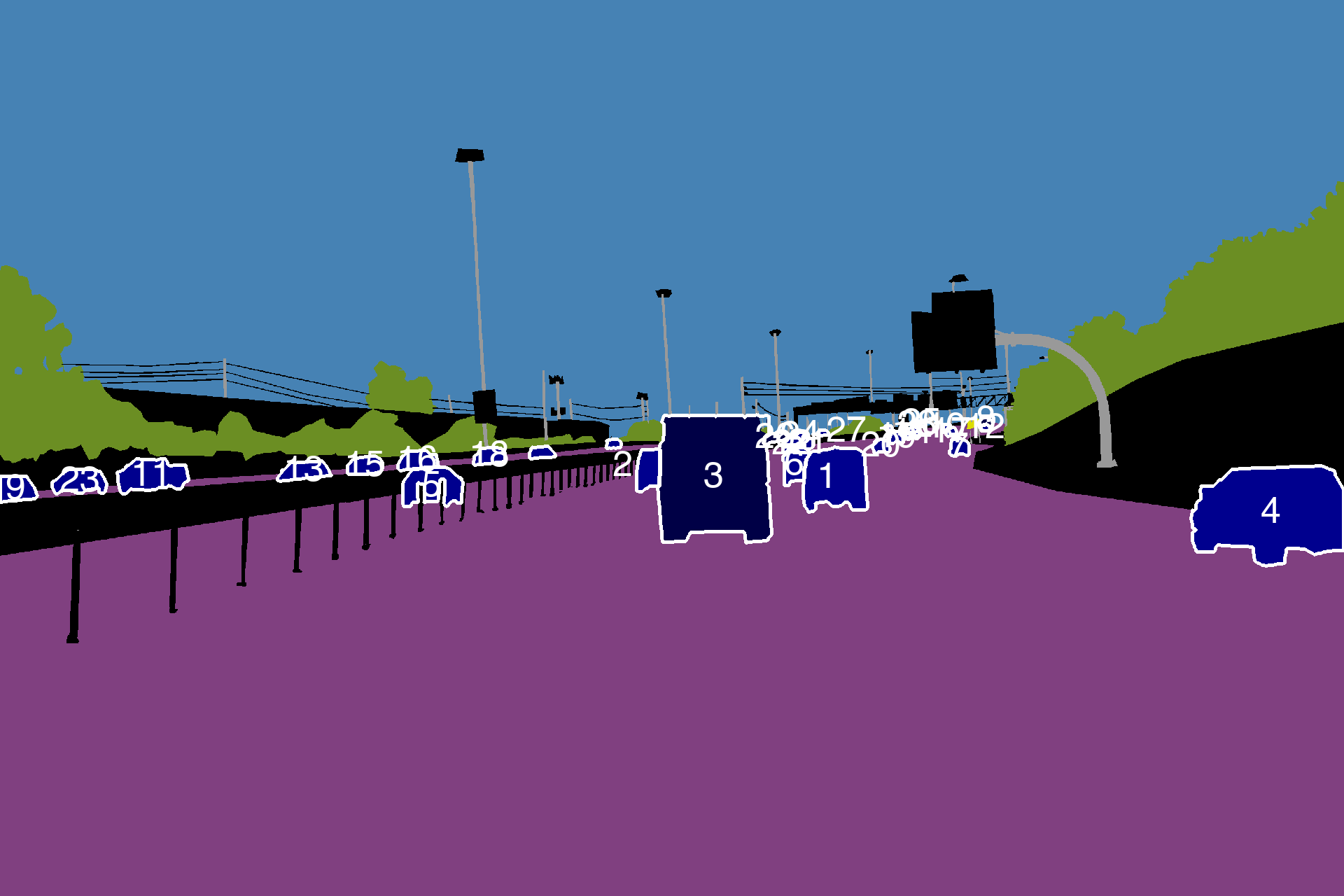} &
\includegraphics[width=\linewidth, height=7em]{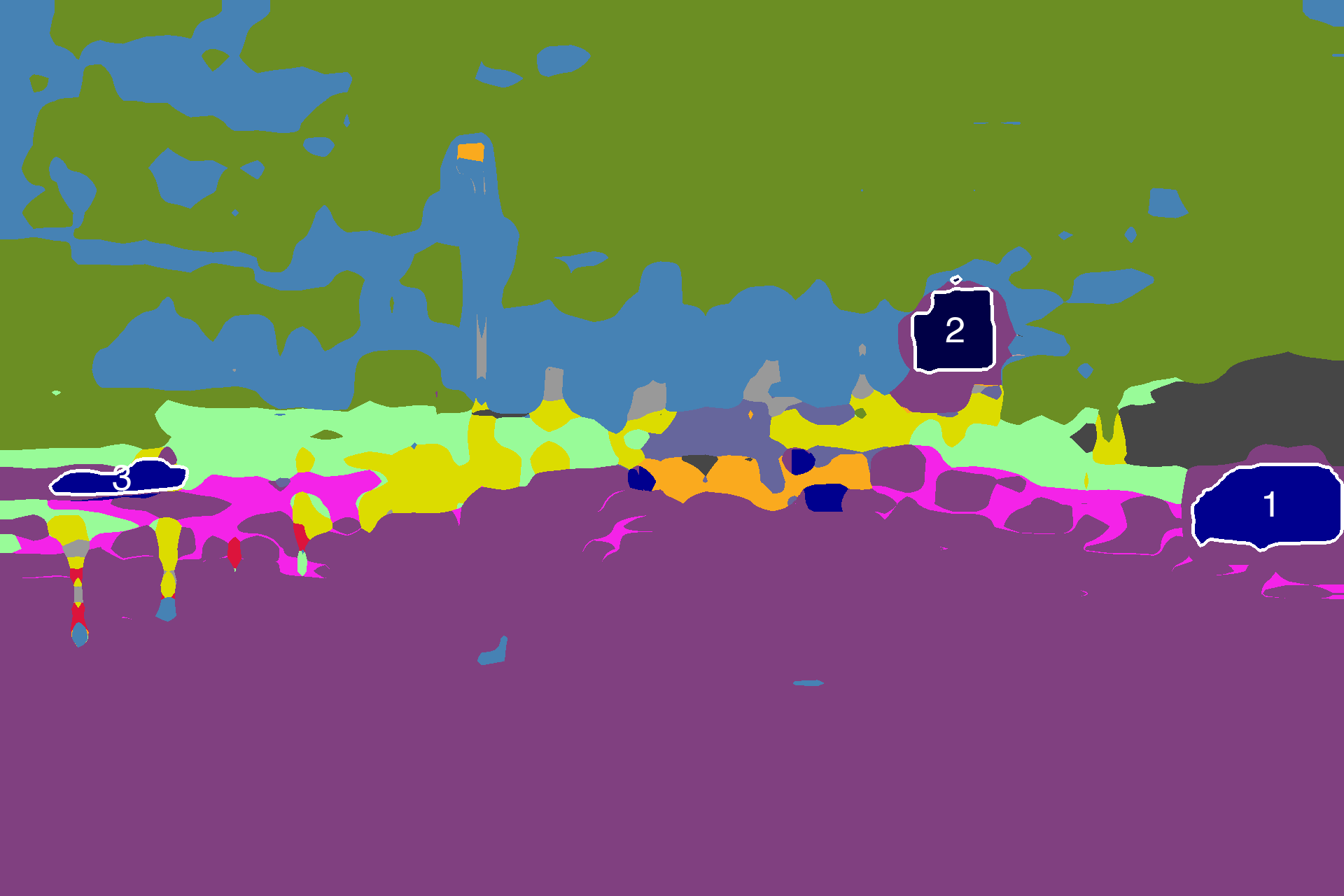} &
\includegraphics[width=\linewidth, height=7em]{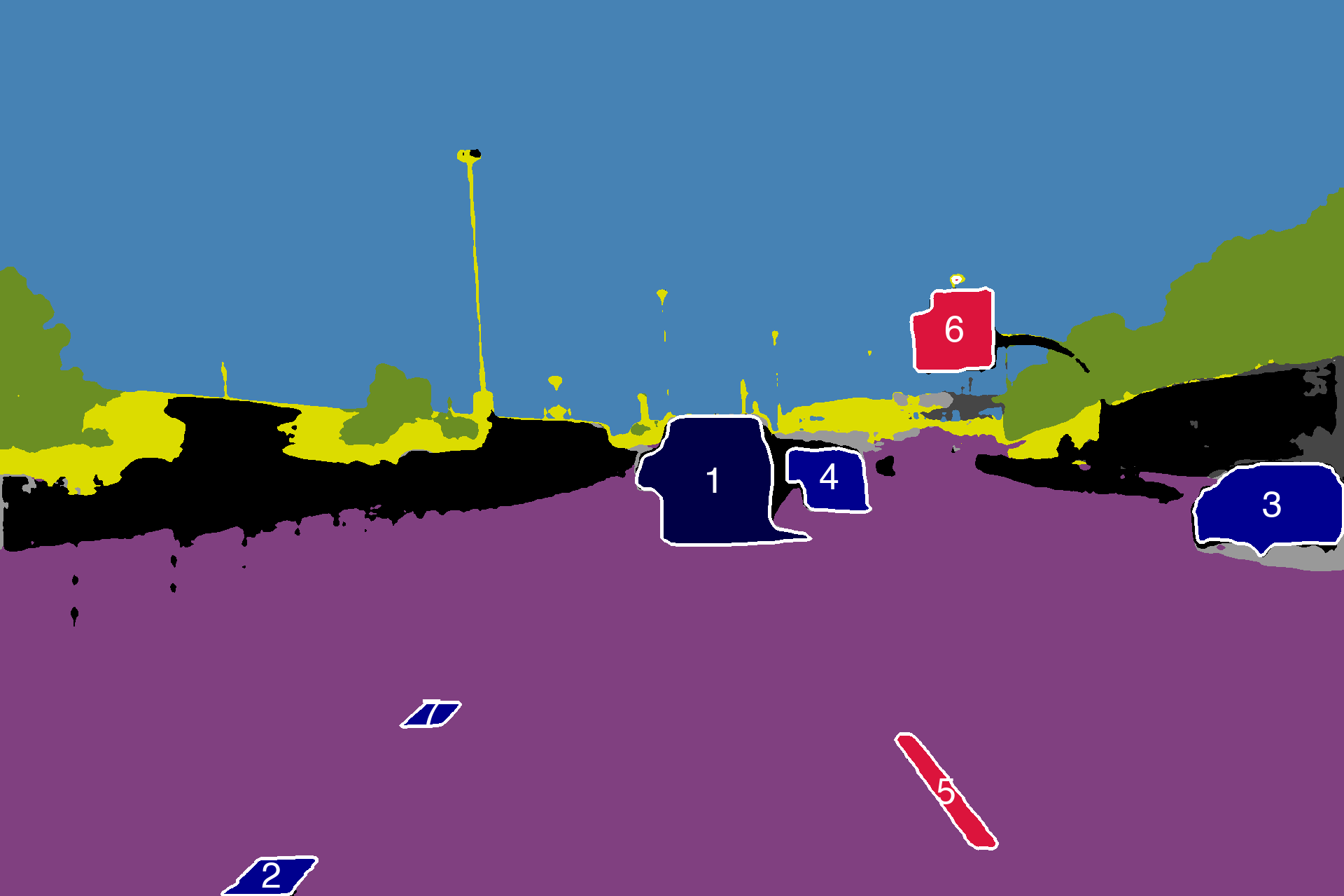} &
\includegraphics[width=\linewidth, height=7em]{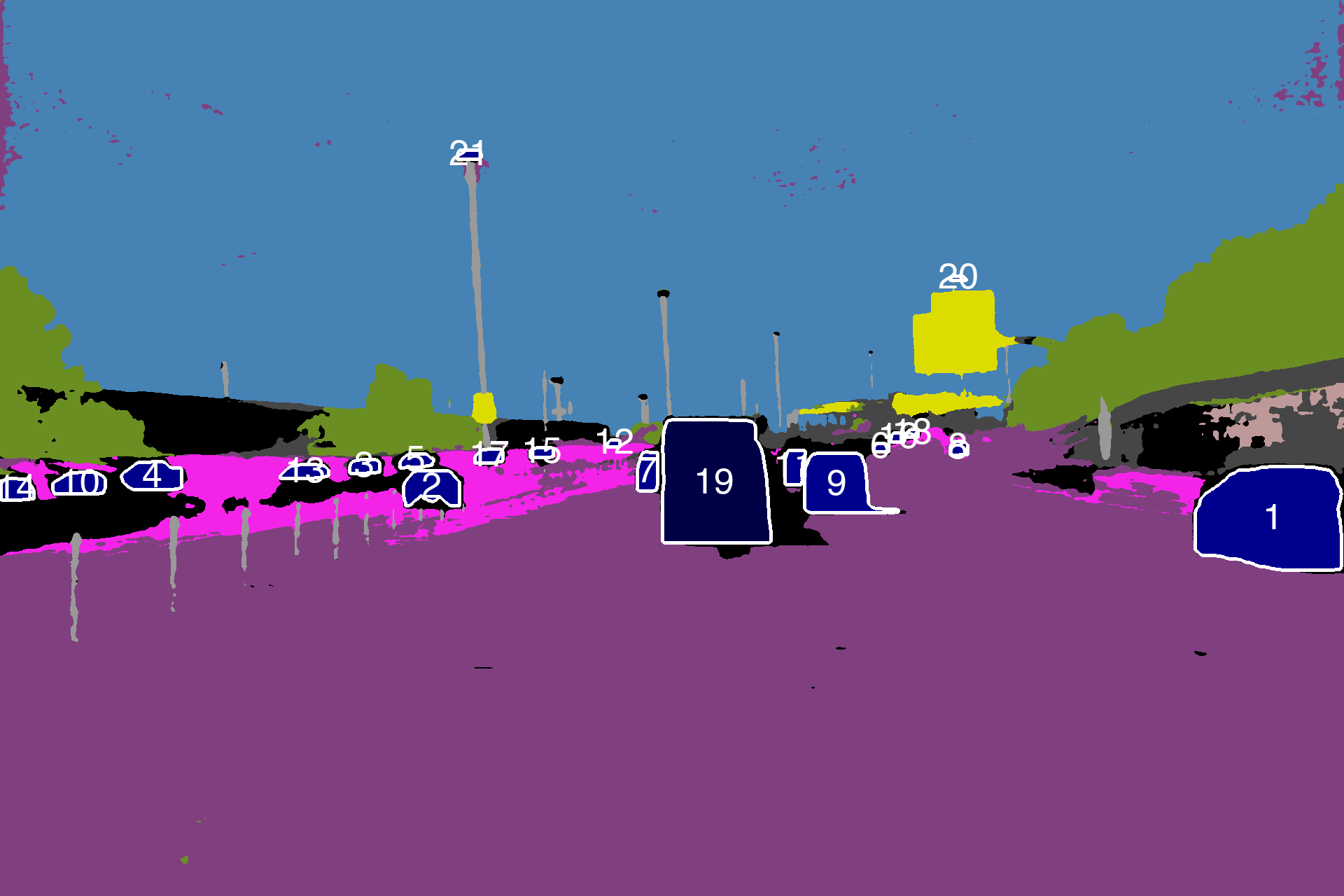}\\[-1pt]

\end{tabular}

\tiny
\renewcommand{\arraystretch}{1.3}
\begin{tabularx}{\textwidth}{*{19}{>{\centering\arraybackslash}X}}  
    \cellcolor{road}\textcolor{white}{Road}
    & \cellcolor{sidewalk}\textcolor{white}{Sidewalk}
    & \cellcolor{building}\textcolor{white}{Building}
    & \cellcolor{wall}\textcolor{white}{Wall}
    & \cellcolor{fence}\textcolor{white}{Fence}
    & \cellcolor{pole}\textcolor{white}{Pole}
    & \cellcolor{trafficlight}\textcolor{white}{Traffic~Light}
    & \cellcolor{trafficsign}\textcolor{white}{Traffic~Sign}
    & \cellcolor{vegetation}\textcolor{white}{Vegetation}
    & \cellcolor{terrain}\textcolor{white}{Terrain}
    & \cellcolor{sky}\textcolor{white}{Sky}
    & \cellcolor{person}\textcolor{white}{Person}
    & \cellcolor{rider}\textcolor{white}{Rider}
    & \cellcolor{car}\textcolor{white}{Car}
    & \cellcolor{truck}\textcolor{white}{Truck}
    & \cellcolor{bus}\textcolor{white}{Bus}
    & \cellcolor{train}\textcolor{white}{Train}
    & \cellcolor{motorcycle}\textcolor{white}{Motorcycle}
    & \cellcolor{bicycle}\textcolor{white}{Bicycle}
\end{tabularx}

        \vspace{-0.55em}
        \caption{\textbf{Waymo} --- Qualitative unsupervised panoptic segmentation examples.\label{fig:qualitative_waymo}}
    \end{subfigure}\vspace{1.25em}\\

    \begin{subfigure}[t]{\textwidth}
        \centering
        \small
\sffamily
\setlength{\tabcolsep}{0pt}
\renewcommand{\arraystretch}{0.0}
\begin{tabular}{>{\centering\arraybackslash} m{0.2\textwidth} 
                >{\centering\arraybackslash} m{0.2\textwidth} 
                >{\centering\arraybackslash} m{0.2\textwidth}
                >{\centering\arraybackslash} m{0.2\textwidth}
                >{\centering\arraybackslash} m{0.2\textwidth}}

{Image} & {Ground Truth} & {Baseline} & {U2Seg~\cite{Niu:2024:UUI}} & {\MethodName \textit{(Ours)}} \\[4pt]

\includegraphics[width=\linewidth, height=7.8em]{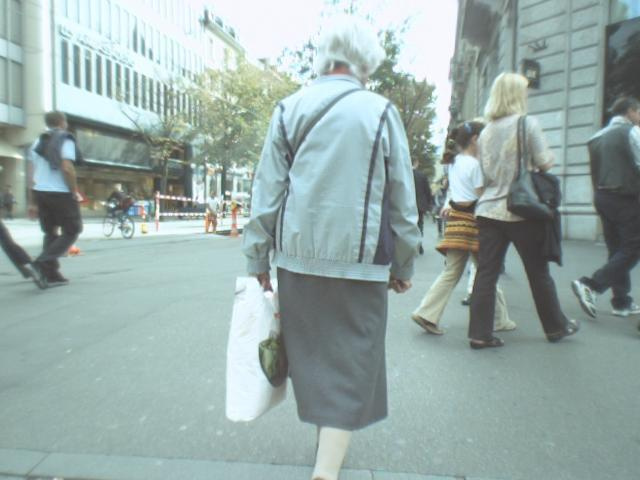} &
\includegraphics[width=\linewidth, height=7.8em]{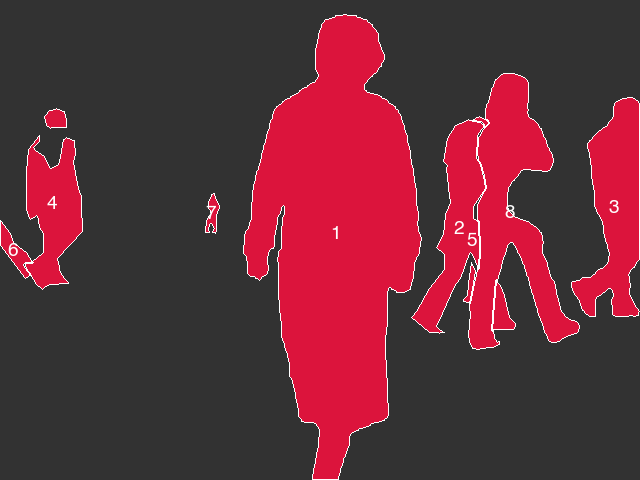} &
\includegraphics[width=\linewidth, height=7.8em]{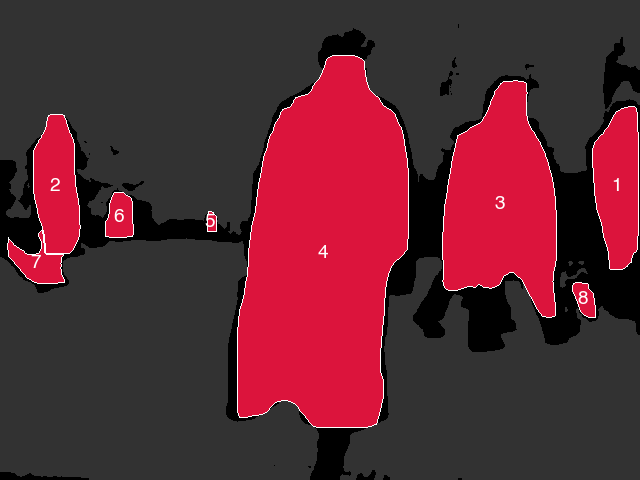} &
\includegraphics[width=\linewidth, height=7.8em]{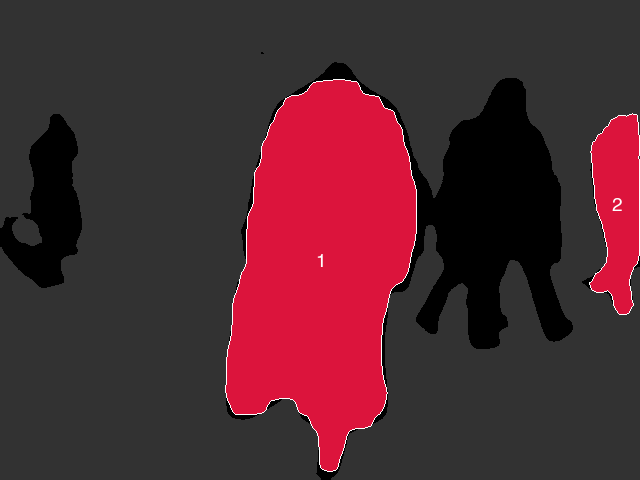} &
\includegraphics[width=\linewidth, height=7.8em]{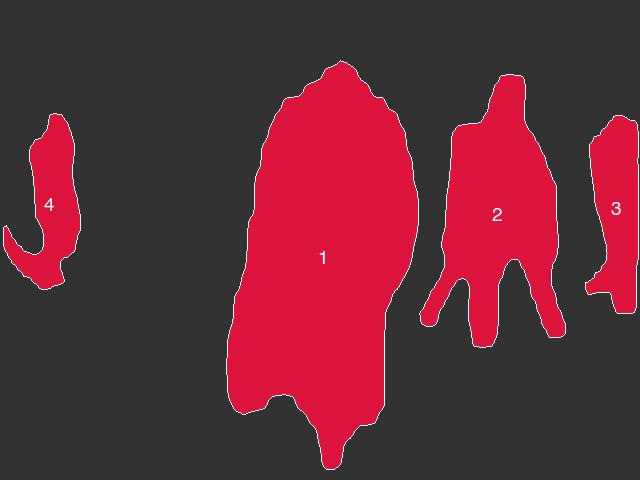}\\

\includegraphics[width=\linewidth]{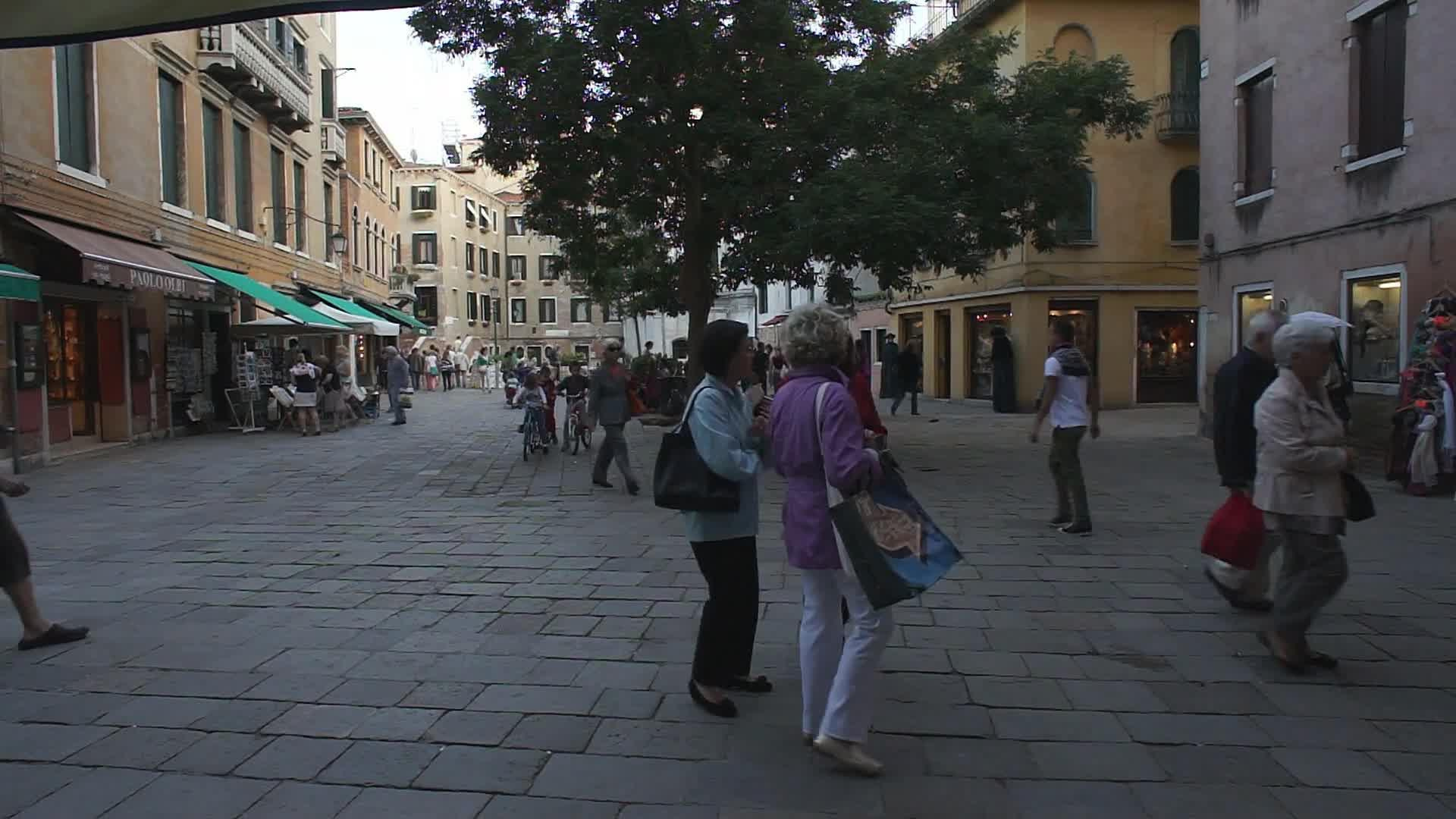} &
\includegraphics[width=\linewidth]{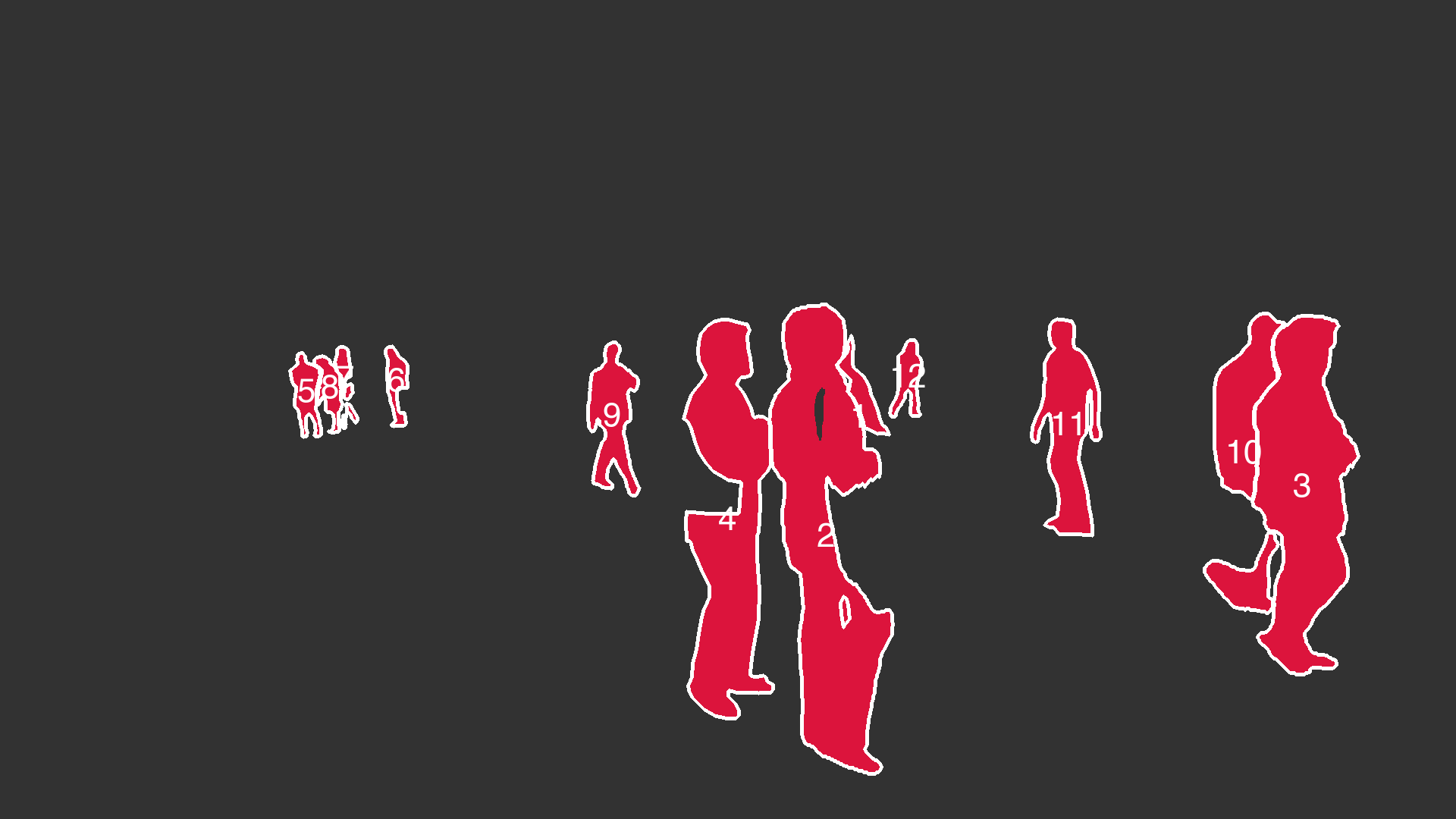} &
\includegraphics[width=\linewidth]{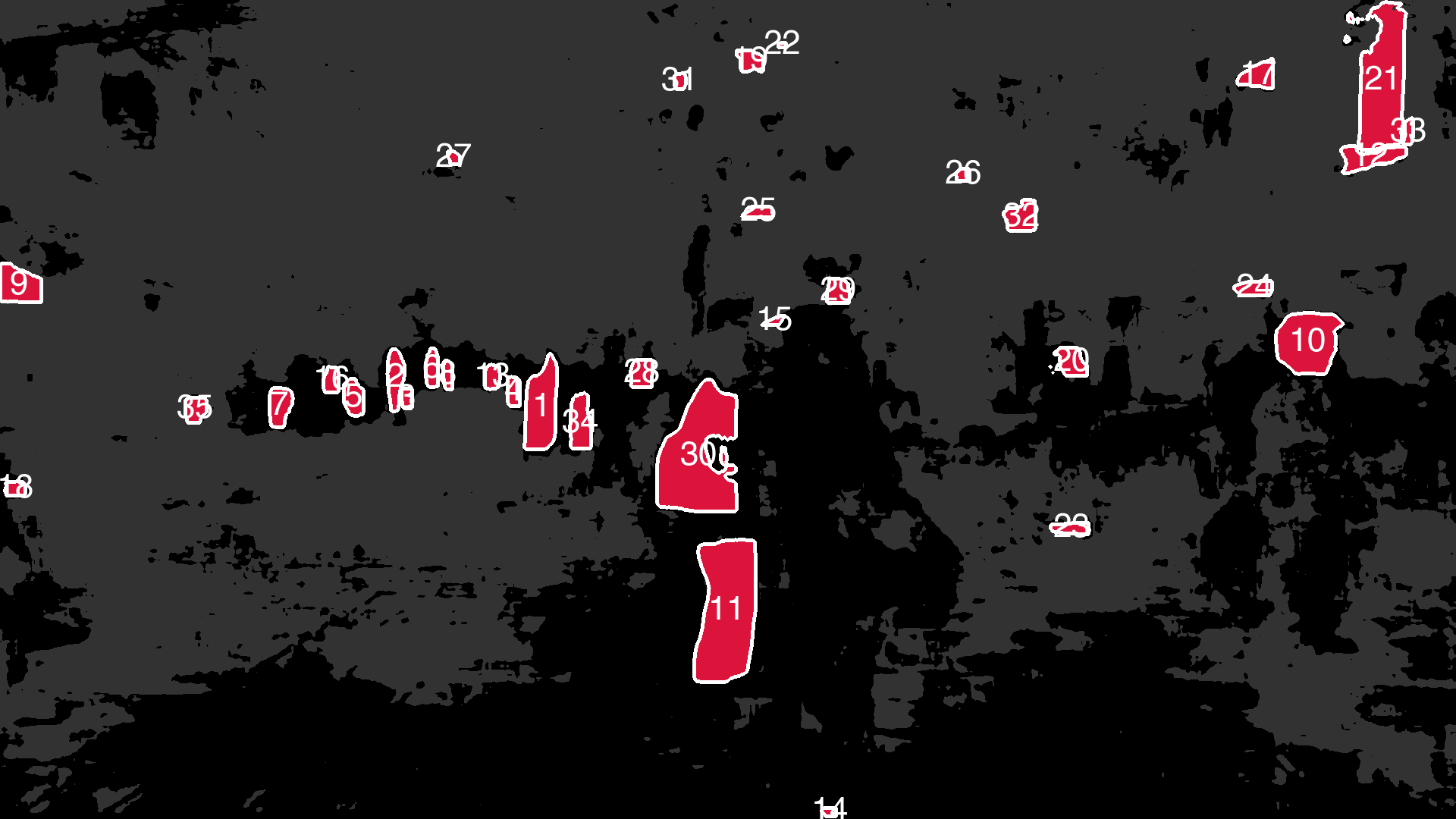} &
\includegraphics[width=\linewidth]{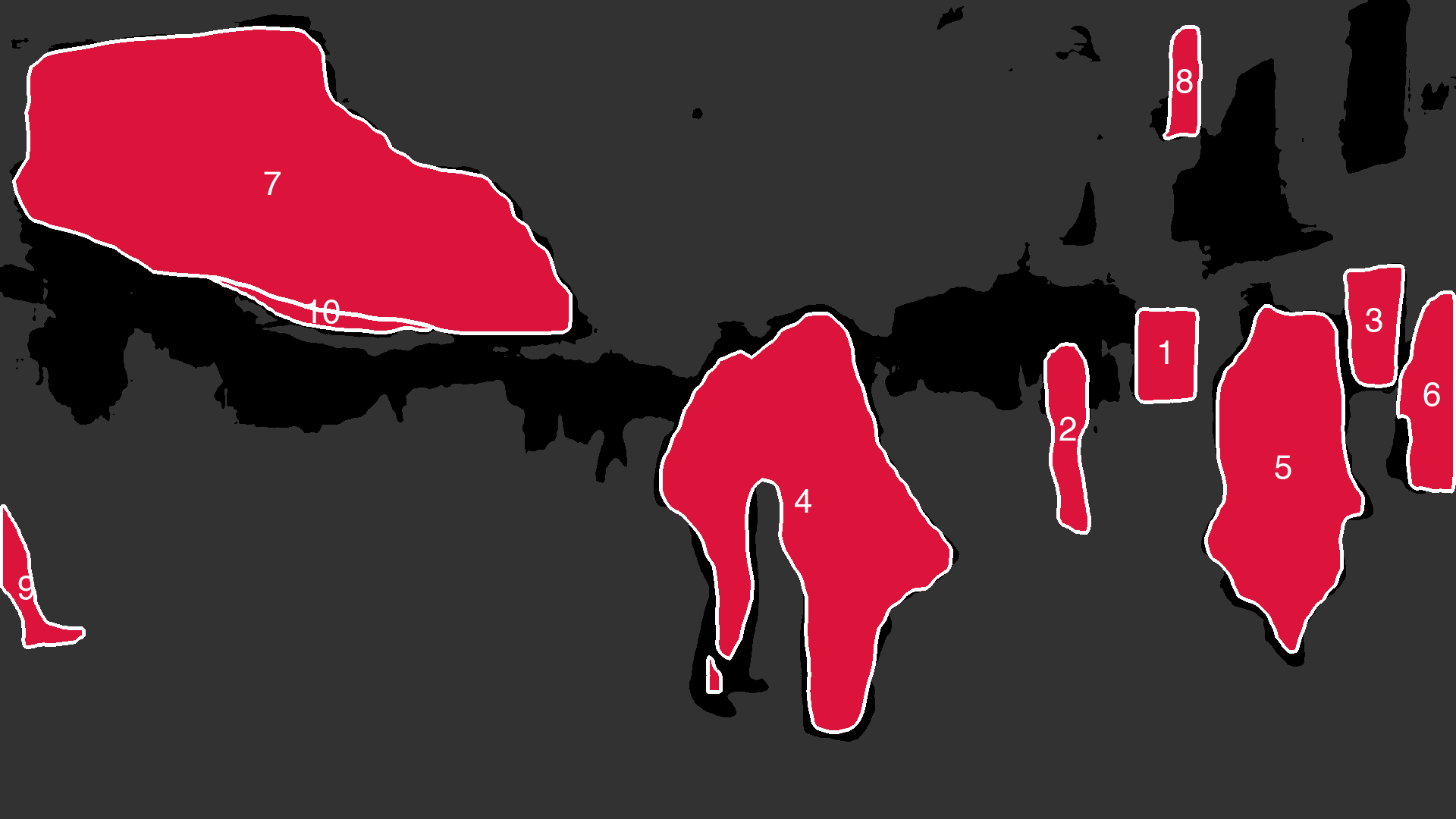} &
\includegraphics[width=\linewidth]{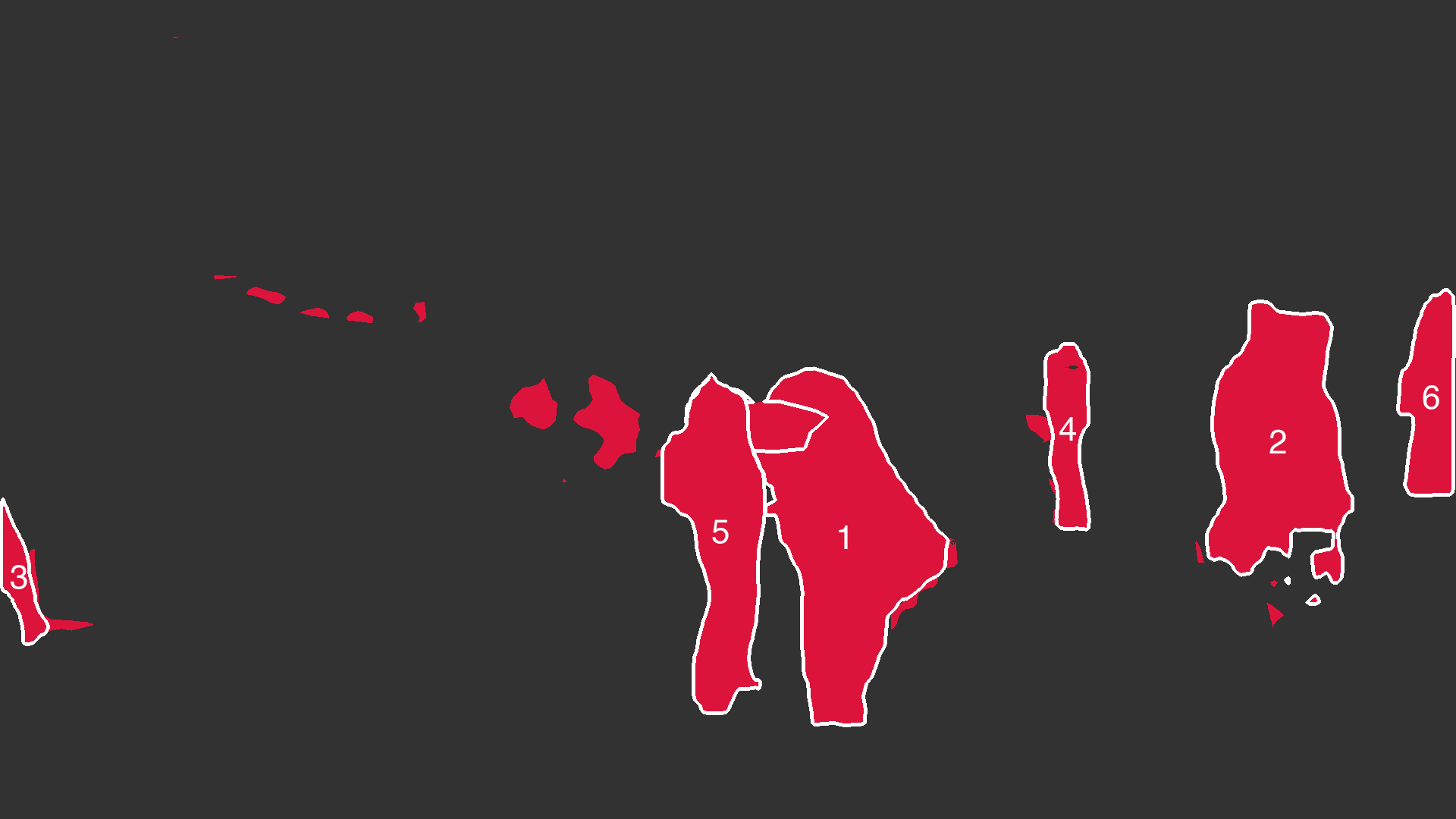}\\

\includegraphics[width=\linewidth]{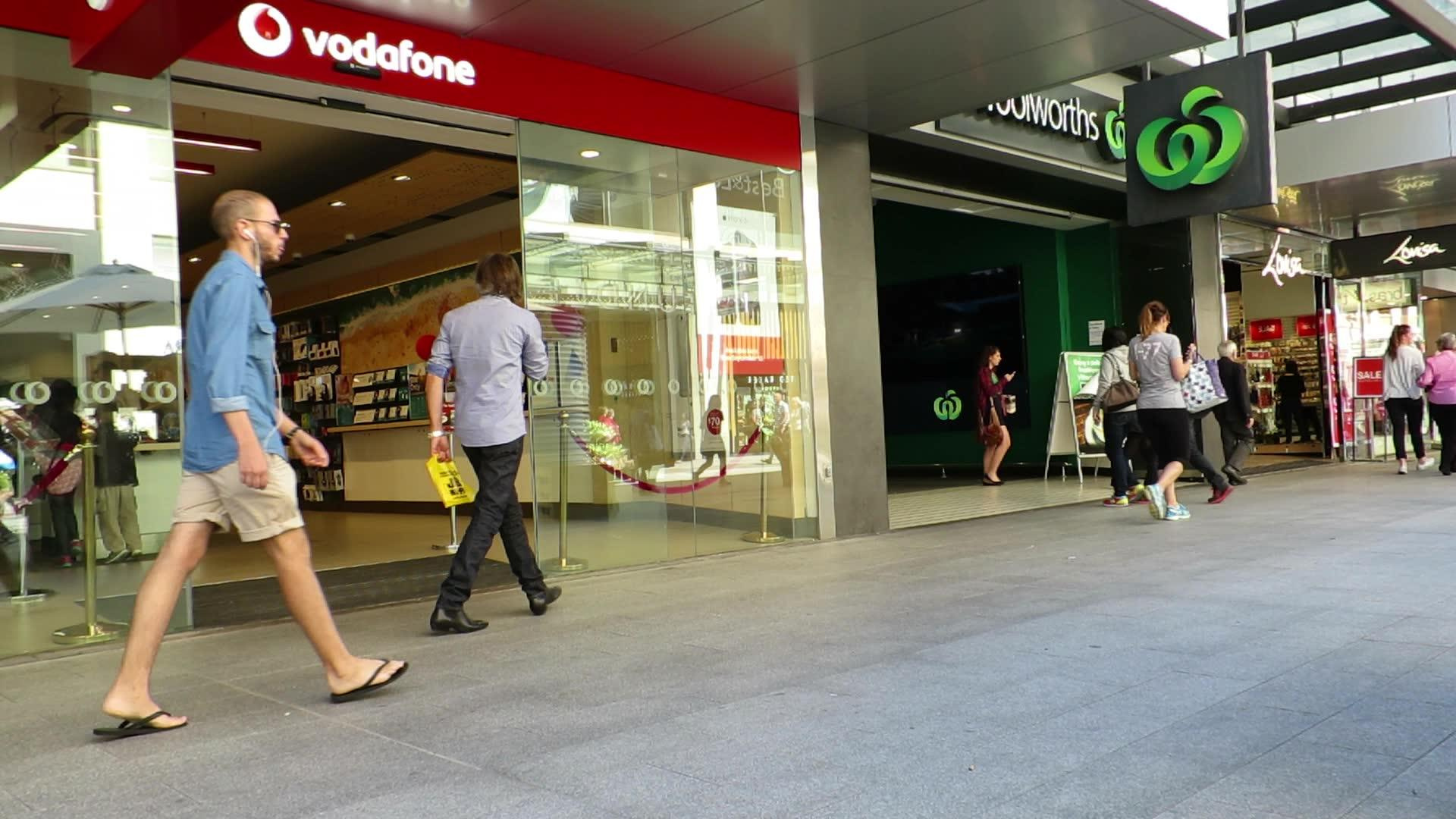} &
\includegraphics[width=\linewidth]{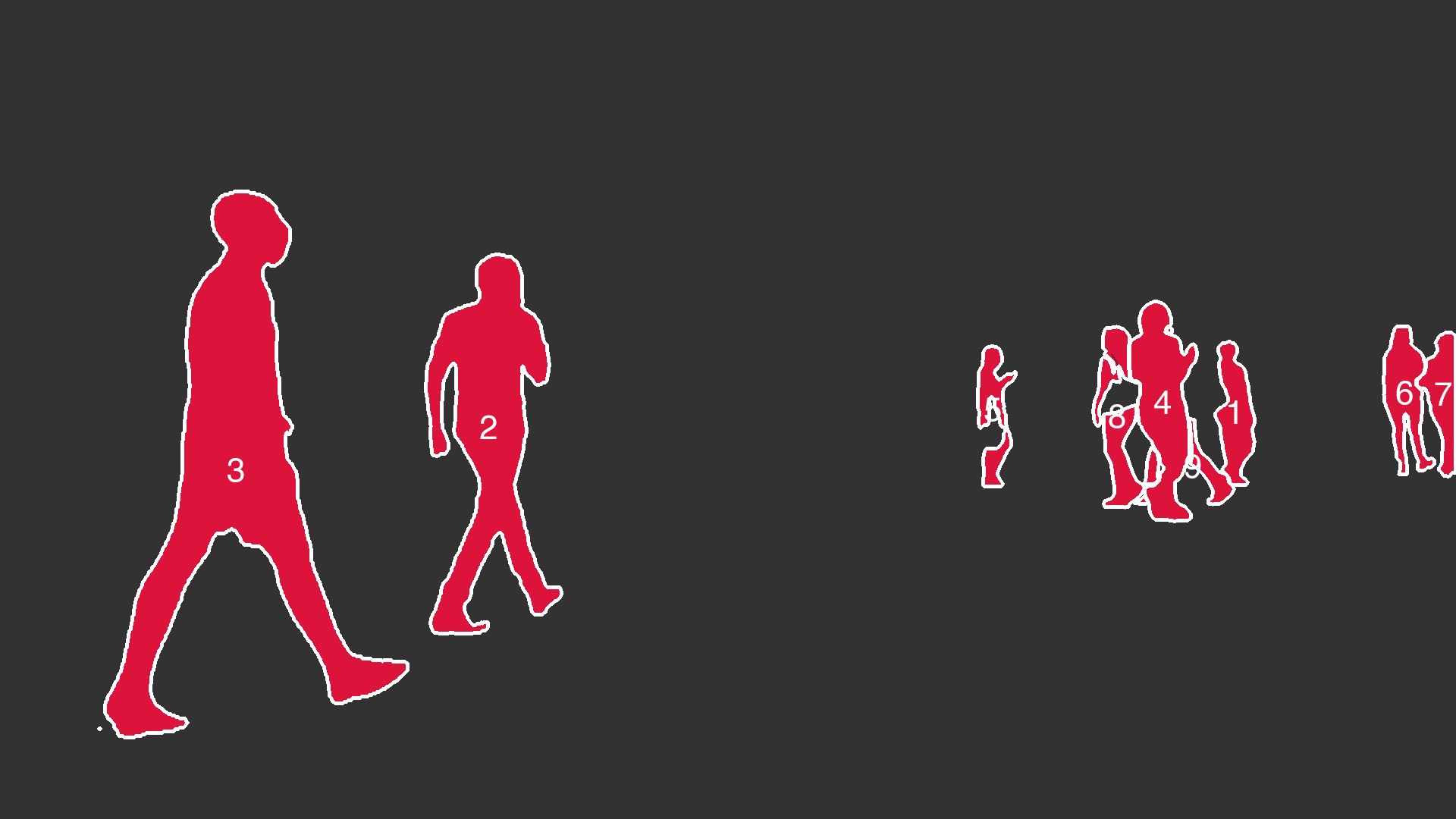} &
\includegraphics[width=\linewidth]{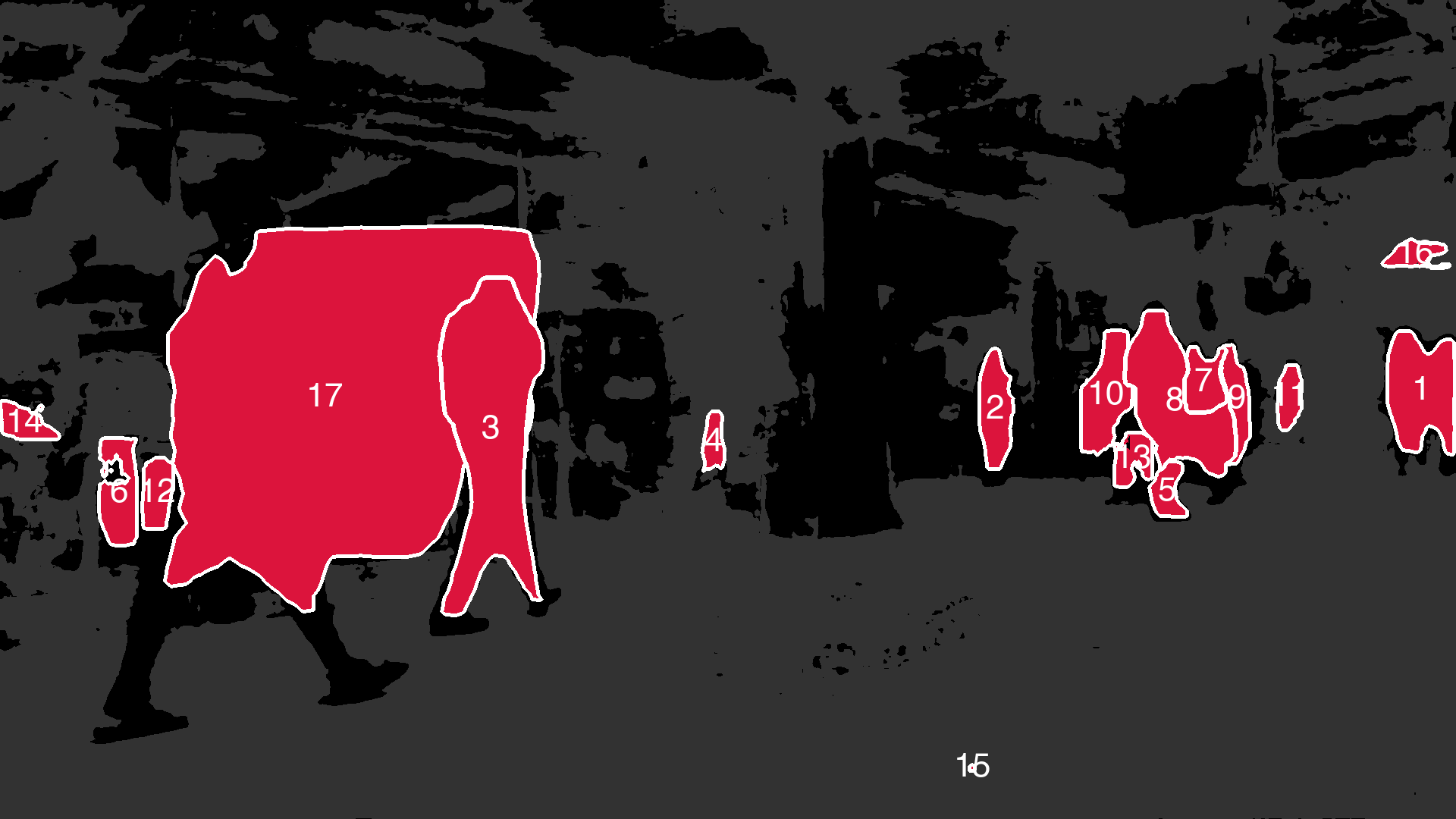} &
\includegraphics[width=\linewidth]{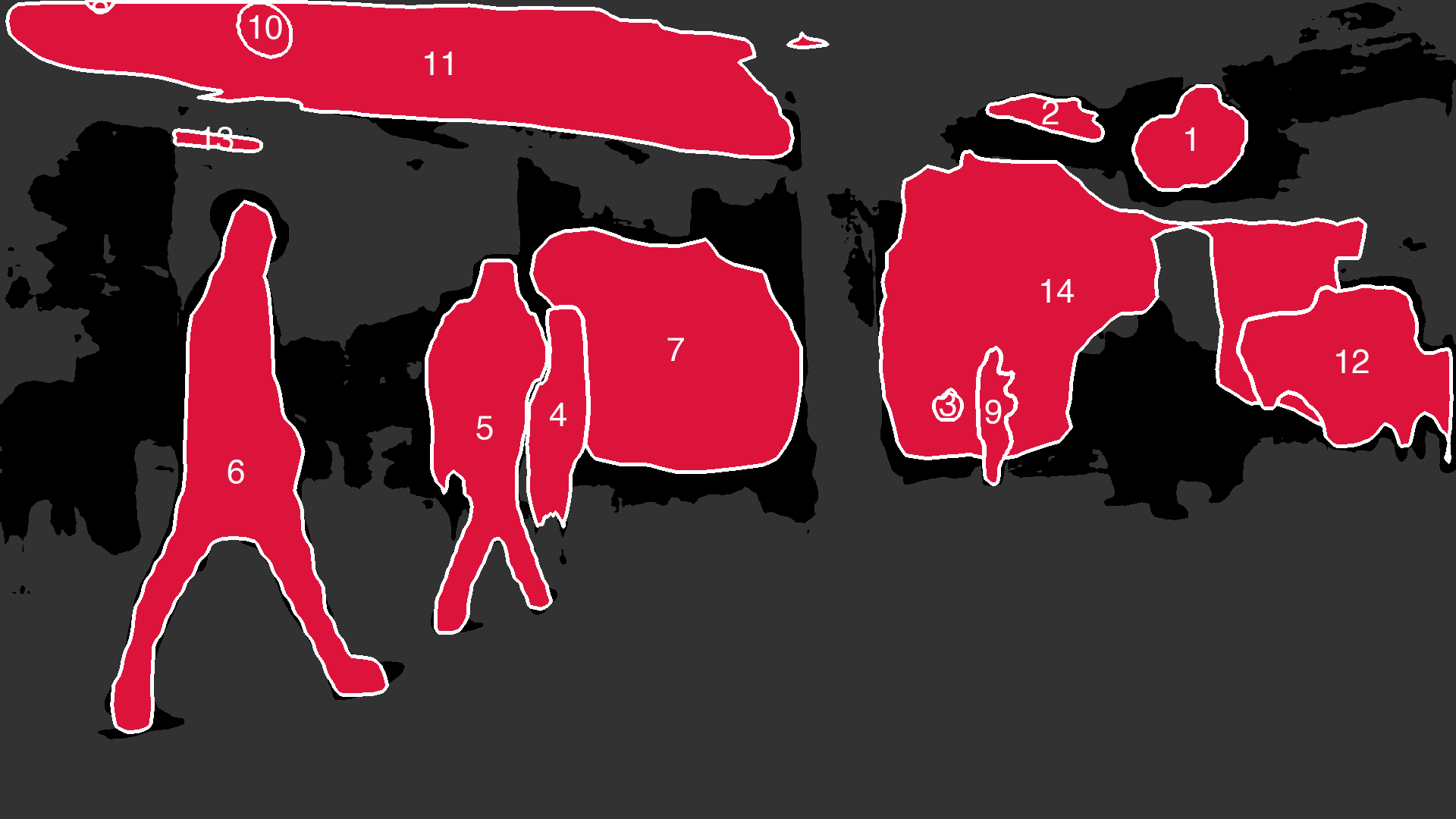} &
\includegraphics[width=\linewidth]{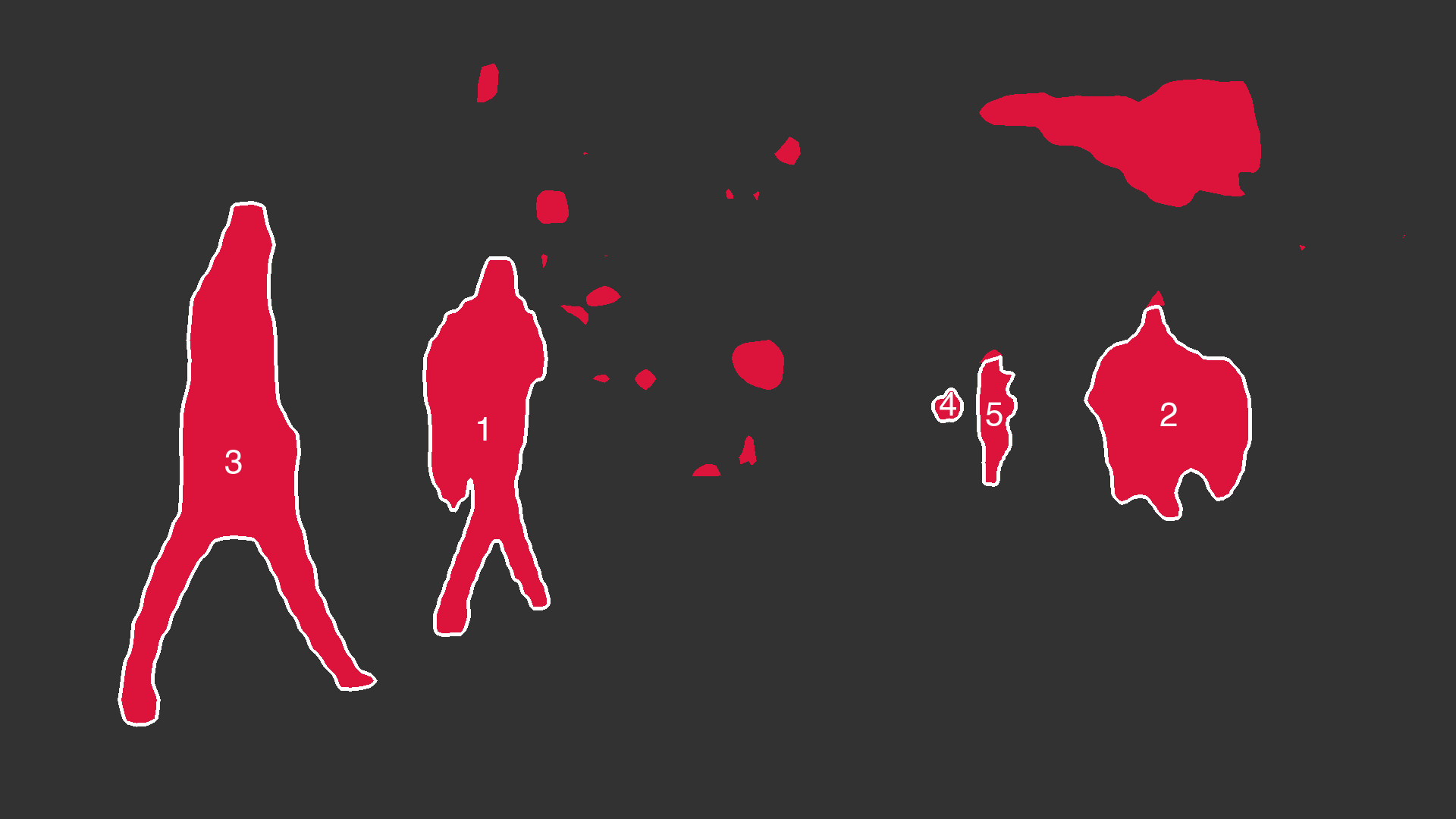}\\[-1pt]

\end{tabular}

\tiny
\renewcommand{\arraystretch}{1.3}
\begin{tabularx}{\textwidth}{*{2}{>{\centering\arraybackslash}X}}  
    \cellcolor{building}\textcolor{white}{Background}
    & \cellcolor{person}\textcolor{white}{Person}
\end{tabularx}

        \vspace{-0.55em}
        \caption{\textbf{MOTS} --- Qualitative unsupervised panoptic segmentation examples.\label{fig:qualitative_mots}}
    \end{subfigure}\\

\end{figure*}

\begin{figure*}
    \centering
    \small
\sffamily
\setlength{\tabcolsep}{0pt}
\renewcommand{\arraystretch}{0.1}
\def\inheight{8.17em}
\def\inheighttwo{7.77em}

\begin{tabularx}{\textwidth}{cccccc}  
\includegraphics[height=\inheight]{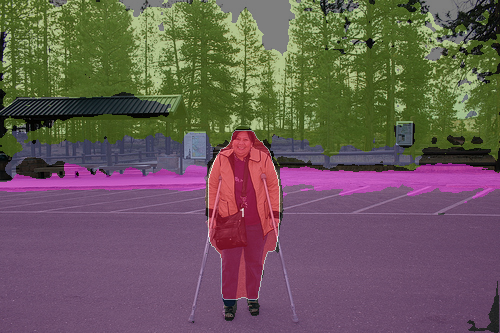} &
\includegraphics[height=\inheight]{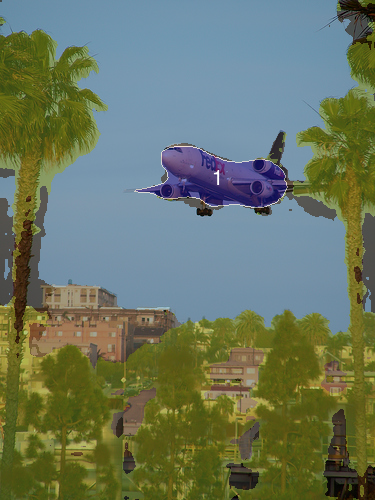} &
\includegraphics[height=\inheight]{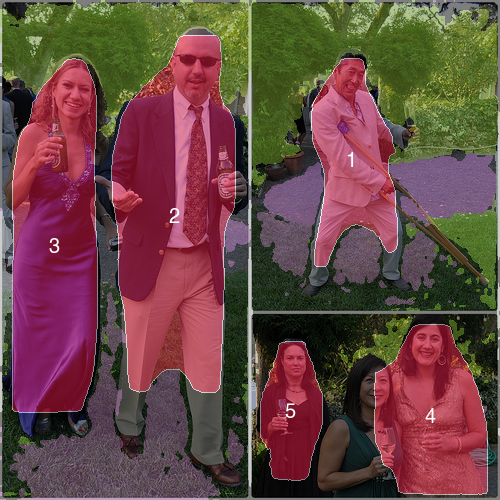} &
\includegraphics[height=\inheight]{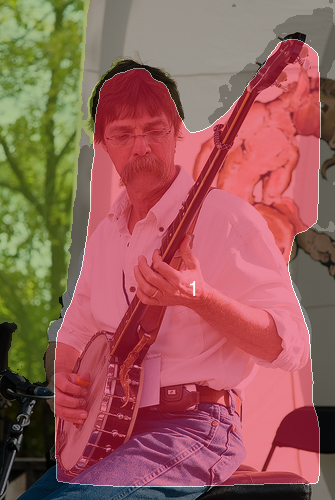} &
\includegraphics[height=\inheight]{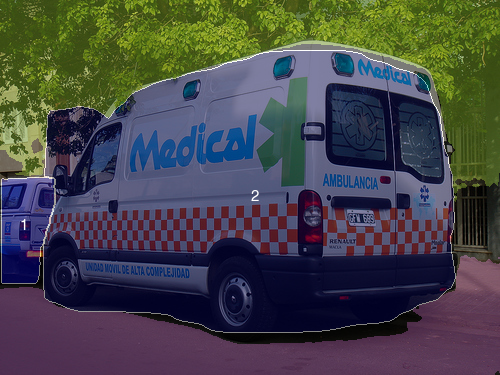} &
\includegraphics[height=\inheight]{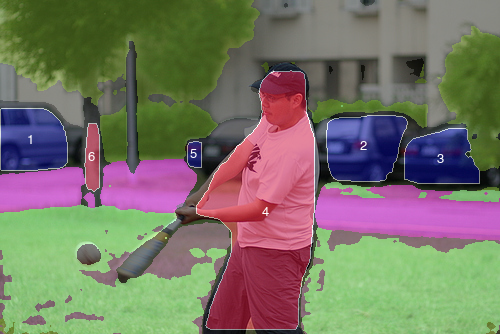} \\[-1.5pt]
\end{tabularx}

\begin{tabularx}{\textwidth}{cccccc}  
\includegraphics[height=\inheighttwo]{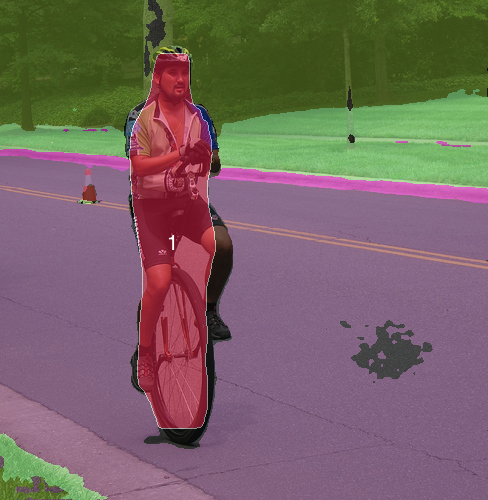} &
\includegraphics[height=\inheighttwo]{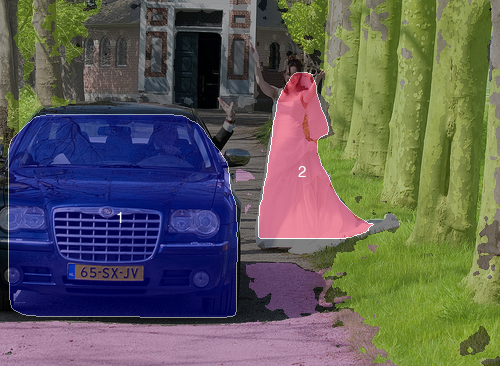} &
\includegraphics[height=\inheighttwo]{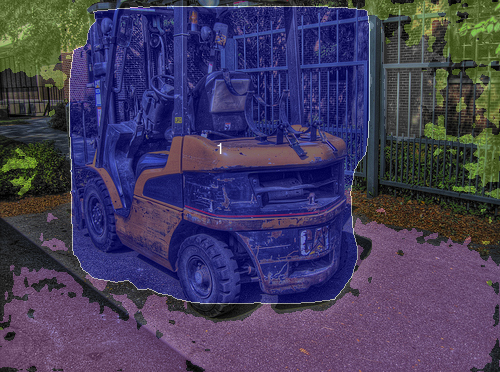} &
\includegraphics[height=\inheighttwo]{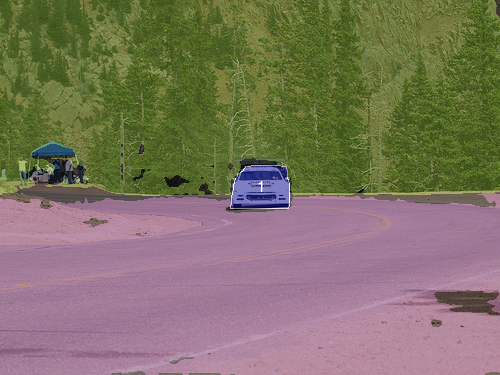} &
\includegraphics[height=\inheighttwo]{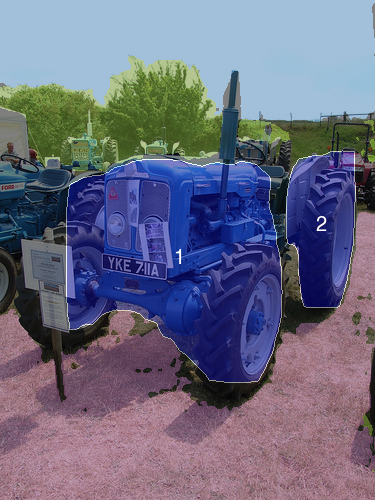} &
\includegraphics[height=\inheighttwo]{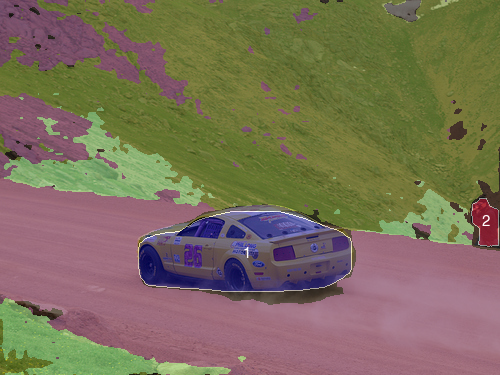} \\[-1.5pt]
\end{tabularx}

\tiny
\renewcommand{\arraystretch}{1.3}
\begin{tabularx}{\textwidth}{*{19}{>{\centering\arraybackslash}X}}  
    \cellcolor{road}\textcolor{white}{Road}
    & \cellcolor{sidewalk}\textcolor{white}{Sidewalk}
    & \cellcolor{building}\textcolor{white}{Building}
    & \cellcolor{wall}\textcolor{white}{Wall}
    & \cellcolor{fence}\textcolor{white}{Fence}
    & \cellcolor{pole}\textcolor{white}{Pole}
    & \cellcolor{trafficlight}\textcolor{white}{Traffic~Light}
    & \cellcolor{trafficsign}\textcolor{white}{Traffic~Sign}
    & \cellcolor{vegetation}\textcolor{white}{Vegetation}
    & \cellcolor{terrain}\textcolor{white}{Terrain}
    & \cellcolor{sky}\textcolor{white}{Sky}
    & \cellcolor{person}\textcolor{white}{Person}
    & \cellcolor{rider}\textcolor{white}{Rider}
    & \cellcolor{car}\textcolor{white}{Car}
    & \cellcolor{truck}\textcolor{white}{Truck}
    & \cellcolor{bus}\textcolor{white}{Bus}
    & \cellcolor{train}\textcolor{white}{Train}
    & \cellcolor{motorcycle}\textcolor{white}{Motorcycle}
    & \cellcolor{bicycle}\textcolor{white}{Bicycle}
\end{tabularx}

    \vspace{-0.5em}
    \caption{\textbf{Qualitative OOD examples for \MethodName on ImageNet val.} Applying the pseudo class to ground-truth matching of \MethodName from Cityscapes for visualization purposes. \label{fig:qual_imagenet}}
    \vspace{-0.5em}
\end{figure*}

\inparagraph{Analysis of panoptic segmentation architecture.} Our method does not make particular assumptions regarding the downstream panoptic segmentation model. In principle, CUPS can be applied to various panoptic segmentation architectures without significant changes; hyperparameter tuning may be required for optimal accuracy.
As a preliminary experiment, we perform stage-1 training (\ie, only pseudo-label training) of CUPS using the Mask2Former~\cite{Cheng:2022:M2F} architecture and observe comparable panoptic segmentation accuracy relative to the Panoptic Cascade Mask R-CNN baseline. Specifically, Mask2Former achieves slightly inferior RQ but marginally superior SQ, resulting in an overall lower PQ. 
We attribute this weaker recognition performance to architectural differences: Mask2Former jointly predicts semantic and instance labels per mask, whereas Panoptic Mask R-CNN separates these tasks into two branches, facilitating a more effective application of the drop loss. In particular, Mask2Former applies the drop loss to both ``thing'' and ``stuff'' predictions, while Panoptic Cascade Mask R-CNN only drops ``thing'' masks.
Our findings indicate that prior work \cite{Wang:2024:USA} has only partially addressed the application of the drop loss to Mask2Former, thus limiting the effectiveness of the drop loss. While initial results appear promising, further investigation is necessary.

\section{Qualitative Results}

We show a qualitative comparison of \MethodName to the DepthG+CutLER baseline and U2Seg \cite{Niu:2024:UUI} across all datasets in \cref{fig:panotic_qualitative}. 

On Cityscapes (\cf \cref{fig:qualitative_cs}), we observe a significant qualitative difference to U2Seg. We attribute this improvement to the quality of our pseudo labels, which enable predicting small instances in the background. Despite some errors, such as the ``Fence'' being incorrectly predicted in small regions of the building in the upper image, \MethodName identifies substantially more classes and provides more precise instance segmentation compared to both the baseline and U2Seg.
On KITTI (\cf \cref{fig:qualitative_kitti}), we observe a similar trend. \MethodName detects and segments more objects, offering a finer-grained panoptic segmentation compared to U2Seg, which tends to merge overlapping objects. For instance, in the upper example, the parked cars are incorrectly merged into a single mask by U2Seg, while \MethodName successfully separates them.
On the BDD dataset (\cf \cref{fig:qualitative_bdd}), the impact of the domain shift is evident across all methods. \MethodName exhibits minor artifacts, such as predictions related to parts of the ego vehicle or dirt on the windshield. Additionally, signs on buildings are occasionally misclassified as traffic signs. In contrast, U2Seg often produces large, erroneous masks that span across the image, resembling the MaskCut artifacts in \cref{fig:maskcut}.
Similarly, for MUSES and Waymo (\cf \cref{fig:qualitative_muses,fig:qualitative_waymo}), all methods are somewhat affected by the domain shift and challenging viewing conditions. However, \MethodName consistently detects instances compared to both other approaches. For the upper Waymo example, one can observe an occasional artifact for \MethodName. For example, it incorrectly classifies the shadows forming underneath the vehicles in sunny weather conditions. %
This is a result of the instance cue being derived from unsupervised flow, which can introduce artifacts due to the apparent motion.
MOTS (\cf \cref{fig:qualitative_mots}) is challenging for all approaches. Nonetheless, \MethodName produces accurate predictions with fewer artifacts compared to both the baseline and U2Seg, showcasing its robustness even in complex scenarios.

Overall, \MethodName predicts less noisy and more accurate semantics, aligning well with the image while predicting significantly more and better instance masks. This observation is in line with our quantitative experiments (\cf \cref{sec:experiments}).

Additionally, we run \MethodName and U2Seg on a demo (validation) video sequence from Cityscapes (\cf \url{https://visinf.github.io/cups}).
For this analysis, we process each individual frame independently using the respective method and concatenate the outputs into a video, as both methods are designed for per-frame processing. 
On this sequence, \MethodName is qualitatively superior to U2Seg.

\inparagraph{Results on object-centric images.} To further evaluate the generalization capabilities of our approach, we tested \MethodName on randomly selected out-of-domain images, sampled from ImageNet~\cite{imagenet}. %
Qualitative results, shown in \cref{fig:qual_imagenet}, demonstrate that \MethodName effectively generalizes to novel domains, viewpoints, and object categories. 
We find that objects such as tractors, forklifts, and airplanes are classified as cars, which is reasonable given the classes available in Cityscapes.
Additionally, objects and surroundings in diverse scenarios are accurately segmented. For instance, despite never encountering a racing car on a mountain road during training, \MethodName provides contextually appropriate and coherent segmentation, further highlighting the robustness of our method.

\section{Limitations and Future Work}

\MethodName utilizes stereo videos to extract depth cues for pseudo labeling of complex scenes. Although stereo videos are widely available and are inexpensive to record, overcoming the need for the stereo setup could further broaden the application spectrum. %
Future work could explore replacing the stereo input with a state-of-the-art self-supervised monocular depth estimation method, such as ProDepth~\cite{prodepth}.

The evaluation of \MethodName has been also largely constrained to driving datasets.
This is due to the wide availability of panoptic annotation specifically for this domain.
Nevertheless, we believe that \MethodName has the potential for applications beyond traffic scenarios, as it relies on domain-agnostic cues, such as depth and motion as well as general-purpose visual representations.

U2Seg and \MethodName approach the task of unsupervised panoptic segmentation from two distinct perspectives: object-centric and scene-centric training data. Combining the strengths of both methods could open a promising avenue for future research, offering a more comprehensive solution to the challenges of unsupervised panoptic scene understanding.

An additional direction for future work could scale such an approach by exploring more advanced panoptic segmentation networks, such as Mask2Former~\cite{Cheng:2022:M2F}, and increasing the amount of training data.

\end{document}